\documentclass{article} 
\usepackage{colm2025_conference}

\usepackage{microtype}
\usepackage{hyperref}
\usepackage{url}
\usepackage{booktabs}

\usepackage{lineno}

\usepackage{multicol}
\definecolor{darkblue}{rgb}{0, 0, 0.5}
\hypersetup{colorlinks=true, citecolor=darkblue, linkcolor=darkblue, urlcolor=darkblue}

\usepackage{amsmath,amsfonts,bm}

\def\eqref#1{equation~\ref{#1}}

\def\1{\bm{1}}

\DeclareMathAlphabet{\mathsfit}{\encodingdefault}{\sfdefault}{m}{sl}
\SetMathAlphabet{\mathsfit}{bold}{\encodingdefault}{\sfdefault}{bx}{n}

\usepackage{xspace}
\usepackage{hyperref}
\usepackage{url}
\usepackage{enumitem}
\usepackage{marvosym} 
\usepackage[utf8]{inputenc} 
\usepackage[T1]{fontenc} 
\usepackage{hyperref} 
\usepackage{url} 
\usepackage{booktabs} 
\usepackage{amsfonts} 
\usepackage{microtype} 
\usepackage{xcolor} 
\usepackage{epigraph}
\usepackage[export]{adjustbox}
\usepackage{multirow}
\usepackage{graphicx}
\usepackage{threeparttable}
\usepackage{listings}
\usepackage{colortbl}
\usepackage{color}
\usepackage{pifont}
\usepackage{fontawesome}
\usepackage{wrapfig}
\usepackage{subcaption}
\usepackage{caption}
\usepackage{float}
\usepackage{enumitem}
\usepackage{tabularx}
\usepackage{makecell}
\usepackage{cellspace}
\usepackage{minitoc} 
\usepackage{amssymb}
\usepackage{algorithm}
\usepackage{algorithmic}
\usepackage{array}

\title{\magi: Autoregressive Video Generation at Scale}

\newcommand{\magi}{\textsc{Magi-1}\xspace}

\begin{document}

\maketitle

\begin{abstract}
We present \magi, a world model that generates videos by \textit{autoregressively} predicting a sequence of video chunks, defined as fixed-length segments of consecutive frames. Trained to denoise per-chunk noise that increases monotonically over time, \magi enables causal temporal modeling and naturally supports streaming generation. It achieves strong performance on image-to-video (I2V) tasks conditioned on text instructions, providing high temporal consistency and scalability, which are made possible by several algorithmic innovations and a dedicated infrastructure stack. \magi\ facilitates controllable generation via chunk-wise prompting and supports real-time, memory-efficient deployment by maintaining constant peak inference cost, regardless of video length. The largest variant of \magi comprises 24 billion parameters and supports context lengths of up to 4 million tokens, demonstrating the scalability and robustness of our approach. The code and models are available at \href{https://github.com/SandAI-org/Magi-1}{\texttt{magi-source}} and \href{{https://github.com/SandAI-org/MagiAttention}}{\texttt{magi-attention}}. The product can be accessed at \href{https://sand.ai}{\texttt{magi-product}}.
\end{abstract}

\section{Introduction}

World modeling and video generation have emerged as central challenges in artificial intelligence, requiring the synthesis of temporally coherent and photorealistic sequences conditioned on semantically rich inputs such as natural language, static imagery, or short video clips. This task resides at the intersection of spatial perception and temporal reasoning, with profound implications for fields including robotics, embodied artificial intelligence, interactive media, and scientific simulation. As video becomes a dominant modality for both human communication and machine understanding, the demand for generative models that are not only high-fidelity and computationally efficient, but also causally consistent and compatible with streaming applications, has become increasingly urgent.

Building on the remarkable success of diffusion~\citep{sohl2015deep, ho2020denoising, song2020score} and flow-matching frameworks~\citep{lipman2022flow, liu2022flow} in image generation, recent research has increasingly focused on extending these approaches to video synthesis. However, most large-scale video diffusion models continue to rely on globally conditioned denoising architectures that process the entire temporal sequence simultaneously. These models typically employ uniform noise levels and require full-sequence access during inference. Such designs disregard the causal structure inherent to temporal data, rendering them suboptimal for scenarios requiring streaming, real-time interaction, or autoregressive generation.

To overcome these limitations, we present \magi: a large-scale diffusion-based generative model that produces video through the \textit{autoregressive} generation of temporally segmented chunks, each consisting of a fixed-length sequence of consecutive frames. This chunk-wise approach offers a principled trade-off between causal modeling and temporal abstraction, enabling the model to capture mid-range temporal dependencies while maintaining strict left-to-right temporal consistency. Training is conducted at the chunk level with temporally \textit{progressive} noise levels, resulting in a model that is both autoregressively structured and adaptable in its conditional generation capacity.

\magi adheres strictly to causal constraints and facilitates real-time, streaming-compatible video synthesis that approximates multi-step diffusion trajectories with reduced-step, chunk-level predictions. This is enabled by a Transformer~\citep{transformer} backbone specifically designed for bidirectional spatial and causal temporal denoising, supported by a carefully engineered training infrastructure. Central to this infrastructure is a novel distributed attention mechanism (MagiAttention) tailored for ultra-long autoregressive contexts, along with a scalable execution framework optimized for low-latency, parallelized inference. These core components are further augmented by a robust data curation pipeline that supports multi-stage training and dynamically adapts the data distribution based on ongoing model evaluation. Together, these architectural and algorithmic advances empower \magi to deliver efficient, scalable, and controllable video generation. Notably, the inference-time peak resource usage of \magi is independent of the total video length, as each chunk is processed with a fixed computational and memory footprint. This makes \magi particularly suitable for low-latency, memory-efficient applications. The largest variant of the model comprises 24 billion parameters and supports context lengths of up to 4 million tokens, demonstrating the scalability and robustness of the framework.

We evaluate \magi using both internal metrics and publicly available benchmarks, with a particular focus on the image-to-video (I2V) generation task. Our evaluation protocol assesses prompt fidelity, temporal coherence, and subject integrity. On VBench-I2V~\citep{huang2024vbenchcomprehensiveversatilebenchmark} and Physics-IQ Benchmark~\citep{physicsiqbenchmark}, \magi achieves substantial improvements over previous models, especially in its ability to synthesize complex motion, preserve semantic alignment, and model physically plausible interactions.

In summary, \magi establishes a scalable and autoregressive foundation for diffusion-based video synthesis. By integrating architectural innovations, high-throughput inference techniques, and a comprehensive data processing framework, \magi bridges the gap between high-quality generative performance and real-time applicability. The complete inference codebase and pre-trained models are publicly accessible at \href{https://github.com/sandai-org/magi-1}{\texttt{magi-source}}, the distributed attention available at \href{https://github.com/sandai-org/magiattention}{\texttt{magi-attention}}, and a live demonstration available at \href{https://sand.ai}{\texttt{magi-product}}.

\section{\magi}
\begin{figure}[htb!]
\begin{center}
\includegraphics[width=0.98\textwidth]{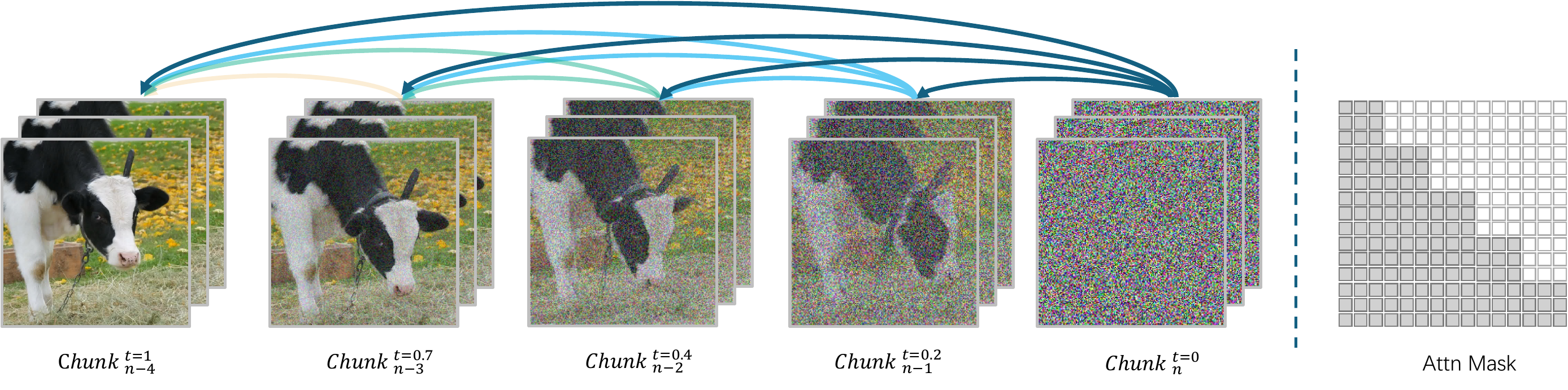}
\end{center}
\caption{(Left) \magi performs chunk-wise autoregressive denoising. The video is generated in chunks of 24 frames, where each chunk attends to all previously denoised chunks. Once a chunk reaches a certain denoising level, the next chunk begins generation. (Right) A block-causal attention mask enforces temporal causality across chunks, enabling pipelined and parallel generation.}
\label{fig:main_algorithm}
\end{figure}

\magi is an autoregressive denoising video generation model operating in latent space. The generation process is illustrated in Fig.~\ref{fig:main_algorithm}. Unlike other bi-directional denoising models (\emph{e.g.,} Sora~\citep{openaisora2024}) that generates the video as a whole, \magi generates the video chunk-by-chunk in a pipeline manner. Specifically, each chunk consists of multiple frames that are denoised holistically. As a chunk is denoised to a certain extent (not necessary completely clean), the next chunk begins generation, conditioned to all preceding chunks. This design allows multiple chunks to be processed concurrently. In our implementation, each chunk contains 24 raw frames (equivalent to one second video clip at 24 FPS), and up to four chunks can be inferred simultaneously.

Compared to fully denoising one chunk before starting subsequent chunks, our method leverages parallelism to better utilize computation, reducing the latency of obtaining subsequent clean chunks and enabling real-time streaming video generation. Moreover, the auto-regressive design naturally supports video continuation without additional specific designs, and extends seamlessly to image-to-video generation. This unified framework enables us to cover text-to-video generation, video continuation, and image-to-video generation within a single pre-training process, eliminating the need for task-specific fine-tuning required by other methods. By maintaining consistency between pre-training and downstream tasks, our approach achieves superior performance in both video continuation and image-to-video generation. 

In this section, we will systematically introduce the training, distillation, and inference of \magi in detail.

\subsection{Transformer-based Variational Auto-Encoder}
To improve the efficiency of both training and inference, \magi employs a variational autoencoder (VAE) to obtain a compressed latent space, over which denoising is performed. While most open-source VAEs are built upon convolutional architectures (\emph{e.g.}, U-Net~\citep{ronneberger2015u}), they are considerably slower than the transformer-based counterparts (\emph{e.g.}, ViT~\citep{dosovitskiy2020image}) of comparable model size on modern GPUs. To address this, we design our VAE architecture based on transformers.

\begin{figure}[htb!]
\begin{center}
\includegraphics[width=0.7\textwidth]{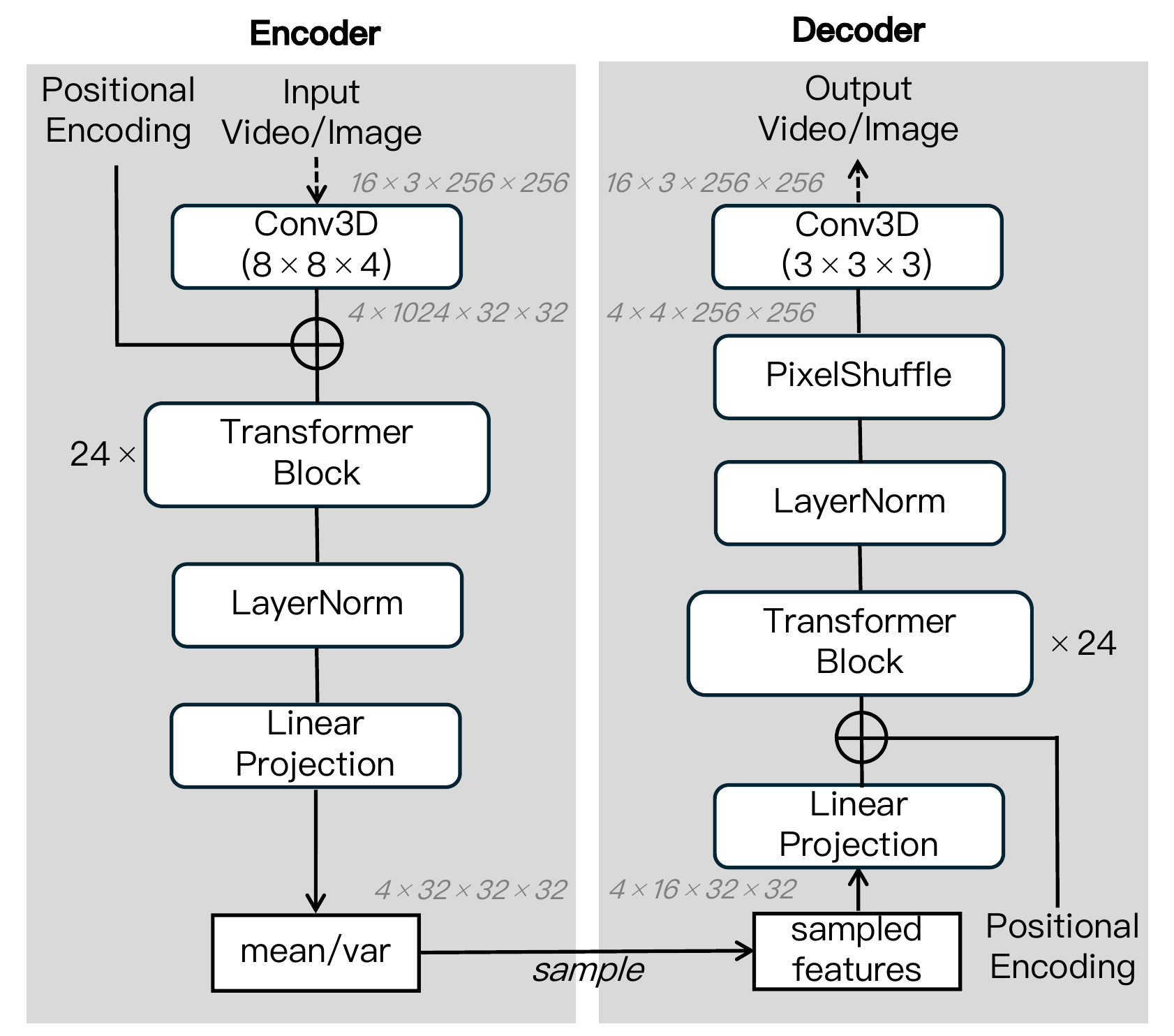}
\end{center}
\caption{Model Architecture of Transformer-based VAE.}
\label{fig:vae_arch}
\end{figure}

The architecture of our VAE is illustrated in Fig.~\ref{fig:vae_arch}. In the encoder, the input is first processed by an embedding module based on a 3D convolution with a kernel size of $8 \times 8 \times 4$\footnote{The kernel size is specified in the order of height, width, and temporal dimensions.} and a stride of $8 \times 8 \times 4$, producing an output with 1024 channels. Absolute positional embeddings are then added to enrich spatial and temporal representations. Building on this, we stack 24 transformer blocks, where self-attention is stabilized through query, key and value normalization to improve training stability. The output of the final transformer block is normalized by a LayerNorm and then projected via a linear layer to 32 channels: the first 16 channels represent the predicted mean, and the remaining 16 channels represent the predicted log-variance. Compared to the raw video input, the encoded features are downsampled by a factor of 8 in the spatial dimensions and by a factor of 4 in the temporal dimension.

The decoder adopts a symmetric architecture to the encoder. To restore the original spatial and temporal resolution, we first apply a pixel shuffle operation to the output of the final transformer block, followed by a 3D convolution with a kernel size of $3 \times 3 \times 3$ and 3 output channels to generate the final output in pixel space. For image inputs consisting of a single frame, we replicate the frame four times along the temporal dimension, which yields better performance compared to padding with three empty frames.

\begin{table}[htb!]
\centering
\small
\begin{tabular}{c|c|c|c|c}
\toprule
\multirow{2}{*}{\makecell[l]{\textbf{VAE}}}     & \multirow{2}{*}{\makecell[l]{\textbf{PSNR}}} & \multirow{2}{*}{\makecell[c]{\textbf{Params}\\\textbf{(M)}}} &  \multirow{2}{*}{\makecell[c]{\textbf{Avg Encode Time}\\\textbf{(ms)}}} & \multirow{2}{*}{\makecell[c]{\textbf{Avg Decode Time}\\\textbf{(ms)}}} \\ & & & &
\\
\midrule
OpenSoraPlan-1.2~\citep{lin2024open} & 28.39         & 239       & 51.08           & 17.48                        \\ 
CogVideoX~\citep{cogvideox}        & 35.99         & 216        &  40.19        & 142.96                        \\ 
HunyuanVideo~\citep{kong2024hunyuanvideo}     & 37.27         & 246        &  124.39        & 47.11                        \\ 
StepVideo~\citep{ma2025step}        & 33.75         & 499          &  30.47      & 18.12                         \\ 
Wan2.1~\citep{wang2025wan}           & 35.95         & 127        &   51.91       & 79.43                        \\ 
Ours             & 36.55         & 614         &  36.68       & 12.28                        \\
\bottomrule
\end{tabular}
\vspace{0.5em}
\caption{Comprehensive comparison of our VAE with other open-source approaches. Thanks to the optimized inference support of transformers, our VAE achieves the fastest decode speed under identical hardware conditions, despite having the largest model size.}
\label{tab:vae_compare}
\end{table}

The training process of the VAE consists of two stages.
In the first stage, we use a fixed input resolution during training: 16-frame short clips with a spatial resolution of $256\times256$ pixels, to maximize training efficiency by avoiding unnecessary padding. In the second stage, two key modifications are introduced. First, both image data (single frame) and video data (16-frame clip) are jointly used during training. Second, we adopt variable spatial resolutions and aspect ratios by randomly sampling at each training step, enabling the VAE to generalize across different resolutions. Specifically, we constrain the total number of pixels (height $\times$ width) is approximately $256^2$ or $384^2$, while sampling the aspect ratio uniformly from the range [0.25, 4.0].
In both stages, we apply a combination of L1 loss, KL divergence loss, LPIPS loss, and GAN loss, following common practice.

During inference, we use sliding window approach to support arbitrary resolutions. In the spatial dimension, we adopt a window size of $256\times256$ pixels with a stride of 192 pixels, resulting in a $25\%$ overlap between adjacent patches in spatial. In the temporal dimension, no overlap is applied.

Tab.~\ref{tab:vae_compare} shows the comparison with other open-source VAEs. All models were evaluated on a single NVIDIA H800 GPU. To eliminate potential biases from varying slicing strategies at higher resolutions, we report the average processing speed measured across 169 test videos, each containing 25 frames with a spatial resolution of 256$\times$256 pixels. 
Despite having the largest model size, our transformer-based VAE achieves the fastest average decoding time among all models and significantly outperforms most baselines in encoding speed. In terms of reconstruction quality (measured by PSNR), it remains highly competitive, ranking second overall.

\subsection{Auto-Regressive Denoising Model}
\subsubsection{Training objective}
\magi employes flow-matching~\citep{albergo2022building,liu2022flow,lipman2022flow} as its training objective. Given a training video clip contains $n$ chunks, we sample independent Gaussian noises for each chunk. The linear interpolation with respect to the denoising timestep $t$ between the sampled noise and the clean latent of the $i$-th chunk is defined as:
\begin{equation}
    x_i^t = (1-t) x_i^0 + t x_i^1,
\end{equation}
where $x_i^1$ denotes the latent of $i$-th chunk and $x_i^0$ is the corresponding sampled Gaussian noise. The ground-truth velocity for each chunk is given by:
\begin{equation}
    v^*(x_i^t) = \frac{d x_i^t}{dt} = x_i^1 - x_i^0.
\end{equation}

In the auto-regressive model, earlier chunks are cleaner than later ones. For convenience, we define the noise timestep sampled assigned to each chunk as $t_i$, and impose the constraint $t_i < t_j$ whenever $i < j$.\footnote{In completely denoising cases, $t_i=t_j=0$, but we use the strict inequality here for simplicity.} The interpolation of the entire video clip is then defined as: $X_T=\{x_0^{t_0}, x_1^{t_1}, ..., x_n^{t_n}\}$. The model is trained to minimize the following objective:
\begin{equation}
\mathbb{E}_{c, X_T}\parallel v(x_i^{t_i} | t_i, c, \{x_{j<i}^{t_j}\}; \theta) - v^*(x_i^{t_i})\parallel^2.
\end{equation}
where $v(\cdot; \theta)$ is the denoising model parameterized by $\theta$, and $c$ denotes the conditioning text inputs. Note that the prediction of velocity for $x_i$ explicitly conditioned on all its preceding chunks $x_j$ where $j<i$.

In contrast, typical bi-directional denoising video models do not enforce monotonicity of the noise timestep. Instead, they apply the equality constraint, where all chunks share the same noise timestep. Accordingly, their training objective is formulated as:
\begin{equation}
\mathbb{E}_{c, X_T}\parallel v(x_i^{t_i} | t_i, c, X_T; \theta) - v^*(x_i^{t_i})\parallel^2.
\end{equation}
where the velocity prediction for $x_i$ is conditioned on all chunks, regardless their temporal order.

\begin{figure}[htb!]
\begin{center}
\includegraphics[width=0.75\textwidth]{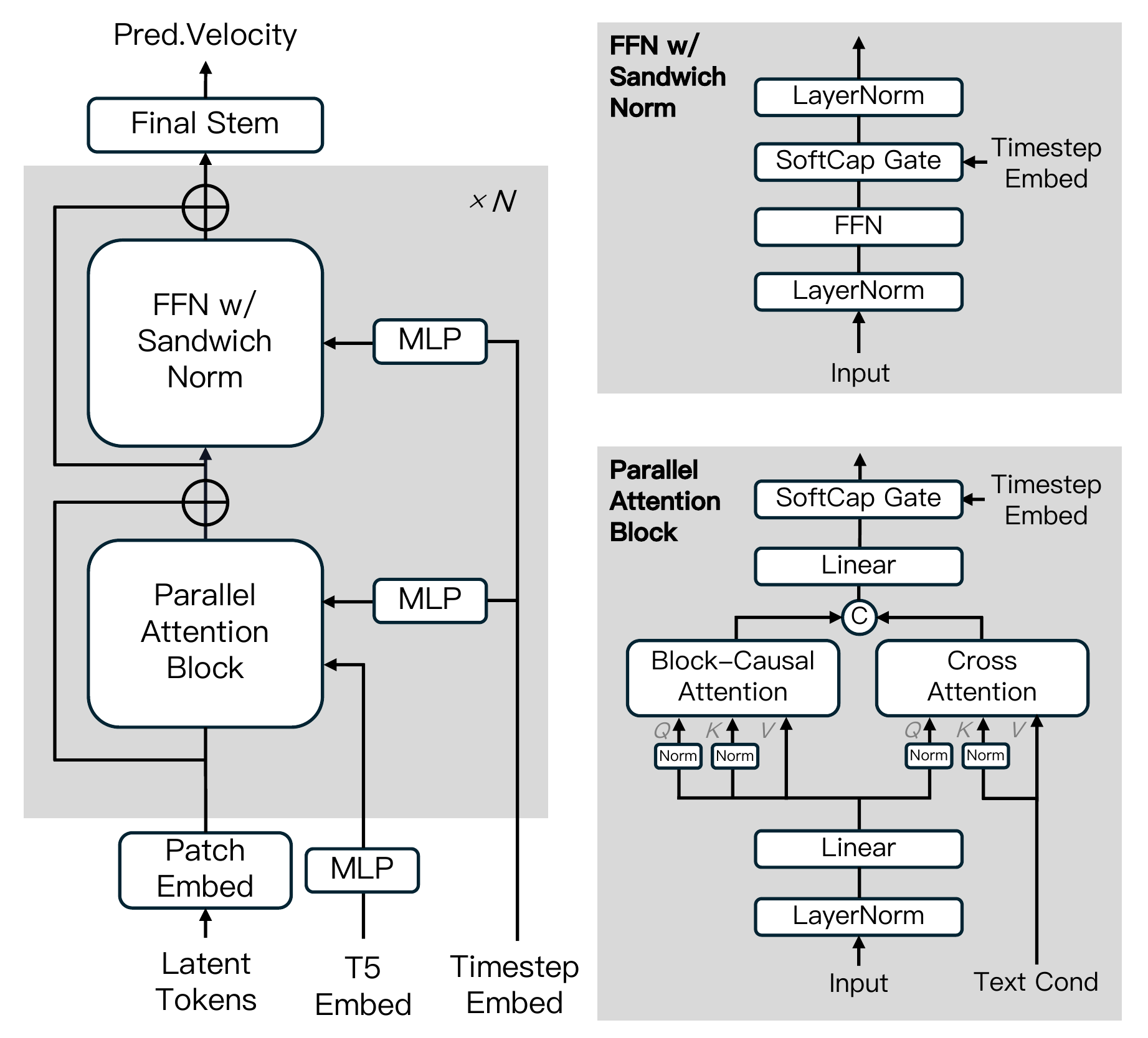}
\end{center}
\caption{Model Architecture of Auto-Regressive Denoising Model.}
\label{fig:ar_model_architecuture}
\end{figure}

\subsubsection{Model Architecture}
\magi is built upon the Diffusion Transformer (DiT) architecture. However, to better meet the requirements of auto-regressive modeling and to improve training efficiency and stability at scale, we introduce several key modifications. As shown in Fig.~\ref{fig:ar_model_architecuture}(a), \magi follows a high-level architecture similar to the standard DiT, consisting of four main components: patch embedding, attention, feed-forward network (FFN), and final stem. We employ T5~\citep{raffel2020exploring} to extract text embeddings, while the timestep information is encoded using sinusoidal positional embeddings.
Our primary modifications target the attention and FFN modules, which are illustrated in Fig.~\ref{fig:ar_model_architecuture}(b) and Fig.~\ref{fig:ar_model_architecuture}(c), respectively. In the following, we provide a detailed description of these modifications.

\paragraph{Block-Causal Attention}
\magi employs full attention within each chunk and causal attention across chunks. Spatial and temporal positional information is encoded using a learnable 3D RoPE~\citep{su2024roformer}, in which the base frequency is learnable. However, existing attention implementations~\citep{dao2022flashattention,dao2023flashattention} do not efficiently support block-causal attention, therefore, we implemented a new kernel called Flexible-Flash-Attention on top of FlashAttention-3. Further details can be found in Sec.~\ref{sec:magiattn}.
 
\paragraph{Parallel Attention Block}
\magi adopts a parallel design for spatial-temporal self-attention and cross-attention with external conditioning input, offering improved computational efficiency over the serial attention architecture. In the serial setup, each attention module independently computes query projections and incurs a separate round of Tensor Parallel (TP) communication. In contrast, the parallel block computes  query projections once and applies them to both attention types concurrently, reducing TP communication from two rounds to one per block. This optimization lowers inter-GPU synchronization overhead and enhances scalability in large-scale models.

\paragraph{QK-Norm and GQA} 
Earlier studies on vision transformers~\citep{liu2022swin,dehghani2023scaling} have shown that normalizing the queries and keys of attention can significantly improve training stability. Moreover, inspired by recent advances in large language models (LLMs), we replace the standard multi-head attention (MHA) with grouped-query attention (GQA)~\citep{ainslie2023gqa} to reduce memory consumption. Both techniques are applied to the spatial-temporal attention and cross-attention modules in our design. 

\paragraph{Sandwich Normalization in FFN} In practice, we have noticed that the numerical problems are more likely to appears in FFN modules as the model size increase. Therefore, we have added LayerNorm before and after the FFN input and output to alleviate the challenge.

\paragraph{SwiGLU} SwiGLU~\citep{shazeer2020glu} has been widely adopted in large language models and has been shown to consistently improve performance than ReLU. Therefore, we employ SwiGLU in the feed-forward network (FFN) of our 24B model.

\paragraph{Softcap Modulation} The standard DiT incorporates timestep information via \emph{adaLN}, where the denoising timestep is used to compute a scaling factor that modulate both the input and output activations of the attention and FFN. While this design works well for small models, we observed that in large models it tends to amplify activation magnitudes and exacerbate numerical instability. To address this issue, we apply a Softcap to the scaling factor, constraining its values within the range of $[-1,1]$. Furthermore, since we adopt QK-Norm in attention modules, we remove the input modulation of \emph{adaLN}.

\begin{table}
    \centering
    \small
    \begin{tabular}{c|c|c}
    \toprule
     & 4.5B & 24B \\
    \midrule
    Layers & 34 & 48 \\
    Model Dimension  & 3072  & 6144 \\
    FFN Activation & GLU & SwiGLU \\
    FFN Dimension & 12288 &  16384 \\
    Attention Type & GQA + QK-Norm &  GQA + QK-Norm  \\
    Block-Casual Attention Head & 128 & 128 \\
    Block-Casual Attention Group & 8 & 8\\
    Cross Attention Head & 128 &128\\
    Cross Attention Group & 8 & 8\\
    Positional Embedding & Learnable 3d RoPE & Learnable 3d RoPE  \\
    \midrule
    Optimizer & AdamW & AdamW \\
    Weight Decay & $1 \times 10^{-1}$ & $1 \times 10^{-1}$ \\
    Peak LR & $1 \times 10^{-4}$ & $1 \times 10^{-4}$\\
    Warm-up & 1000 & 10000 \\
    $  \beta_{1}$ & 0.9 & 0.9 \\
    $  \beta_{2}$ & 0.95 & 0.95 \\
    \bottomrule
    \end{tabular}
    \vspace{0.5em}
    \caption{Model Specification of \magi. }
    \label{table:model_specification}
\end{table}

\subsubsection{Training Recipes}
\label{sec:training_recipes}
\paragraph{Training Configurations}
We train a 4.5B and 24B \magi models and their configurations is shown in Tab.~\ref{table:model_specification}. The training is organized into three stages. Take the 4.5B model as an example. In the first two stages, the resolution of training data is set to 360p and 480p, respectively, with video length is up to 8 seconds. In the third stage, the resolution is further increased to 720p, and the video length is extended up to 16 seconds. Throughout all three stages, the image and video are trained jointly. At the beginning of training, we apply a learning rate warmup, gradually increasing the learning rate to $1\mathrm{e-}{4}$ in $1000$ steps. Then, we adopt a stepwise learning rate scheduling strategy. In the first two stages, the learning rate remains constant, and the stage is switched when the visual assessment of generated video does not significantly improve. In the third stage, we gradually reduce the learning rate once the validation loss reaches a plateau, eventually reducing to $1\mathrm{e-}{5}$. 

For the 24B model, we reduce the resolution in the first training stage from 360p to 256p, as this stage primarily serves to making the model learn global motion dynamics and semantic concepts. Lowering the resolution allows for more training iterations within the same computational budget, thereby improving training efficiency. In addition, we extend the learning rate warmup phase to 10,000 steps to enhance stability during the early training phase. Furthermore, since larger models typically require longer training to reach performance saturation, we proportionally increase the number of training steps at each stage, guided by empirical visual assessment on the validation set.

\begin{figure}[htb!]
\begin{center}
\includegraphics[width=0.85\textwidth]{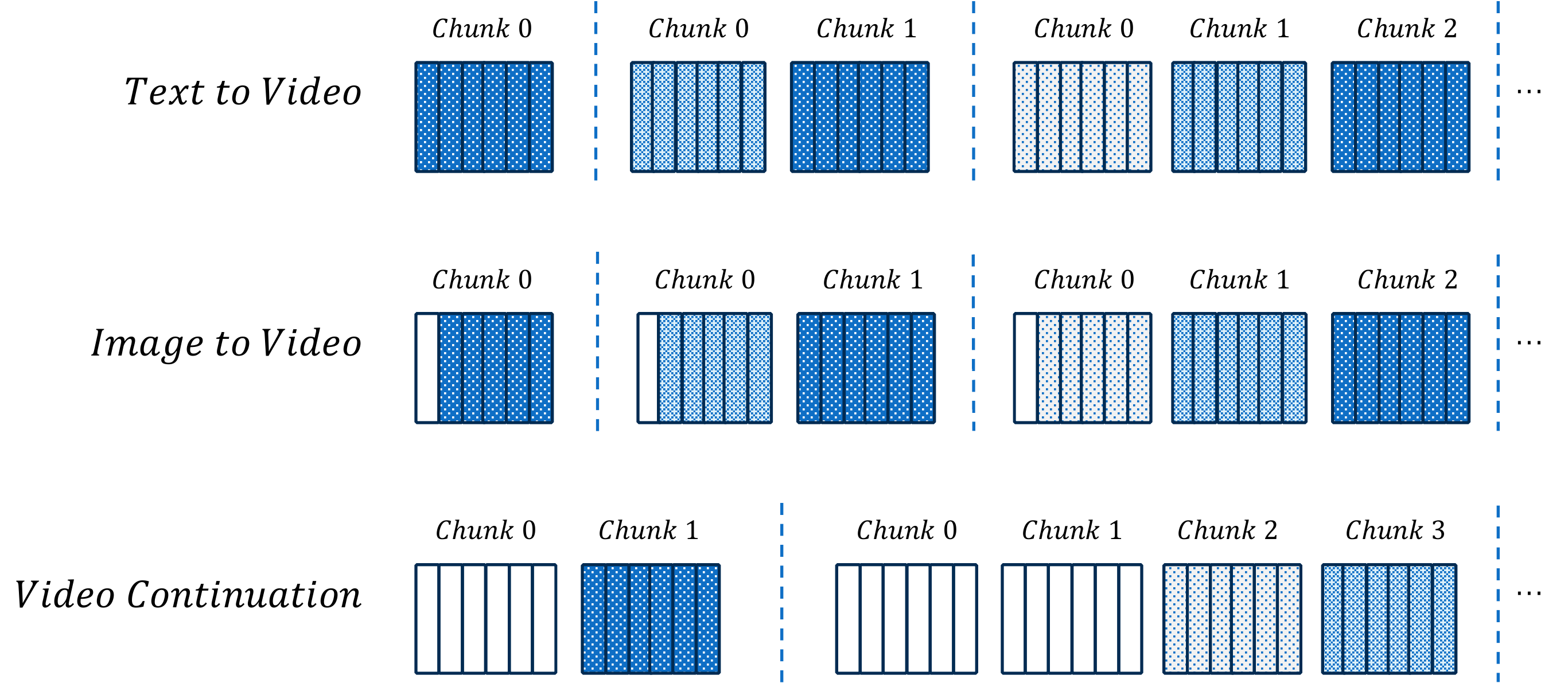}
\end{center}
\caption{The figure shows how different tasks can be unified by varying the proportion of clean chunks. Each vertical bar represents a latent frame in a chunk, with darker bars indicating higher noise levels and the \emph{white} bars denoting clean frames. The first row illustrates the early inference stage of T2V generation, starting from a single fully noisy chunk and progressing to multiple noisy chunks, before any clean chunk has been produced. The middle row depicts the case of I2V generation, treated as a special case of continuation in which only the first frame of the first chunk is clean. The last row describes a general stage where clean chunks are already available, applicable to video continuation and other scenarios involving prior denoised content.}
\label{fig:clean_chunk}
\end{figure}

\paragraph{Multi-Task Training via Data Configurations}
Bi-directional denoising models typically support only text-to-video generation during pretraining, while tasks such as image-to-video generation often require dedicated architectural designs or additional finetuning. In contrast, within the auto-regressive framework, text-to-video, image-to-video, and video continuation tasks differ solely in the proportion of clean chunks present in the training data. As illustrated in Fig.~\ref{fig:clean_chunk}, the early stage of text-to-video generation corresponds to the case of all chunks are noisy, while the inclusion of some clean chunks represents to video continuation. Image-to-video generation is a special case of video continuation, with only the first frame of the first chunk being clean.

Thanks to this property, our auto-regressive model enables unification of various generation tasks under a single training objective without additional task specific fine-tuning and requiring only adjustment of the proportion of clean chunks in the training data.

Furthermore, unlike bi-directional denoising models — where the text condition must be predefined for the entire video and remains fixed throughout generation — \magi allows for different text conditions to be provided for each chunk, enabling fine-grained, chunk-wise text control. To better support this capability, we design a dedicated auto-regressive captioning strategy (Data details are described in Sec.~\ref{sec:caption}) that adapts training accordingly. Additional examples of this fine-grained control are provided in Sec.~\ref{sec:model_capability_study}.

\begin{figure}[htb!]
\begin{center}
\includegraphics[width=0.55\textwidth]{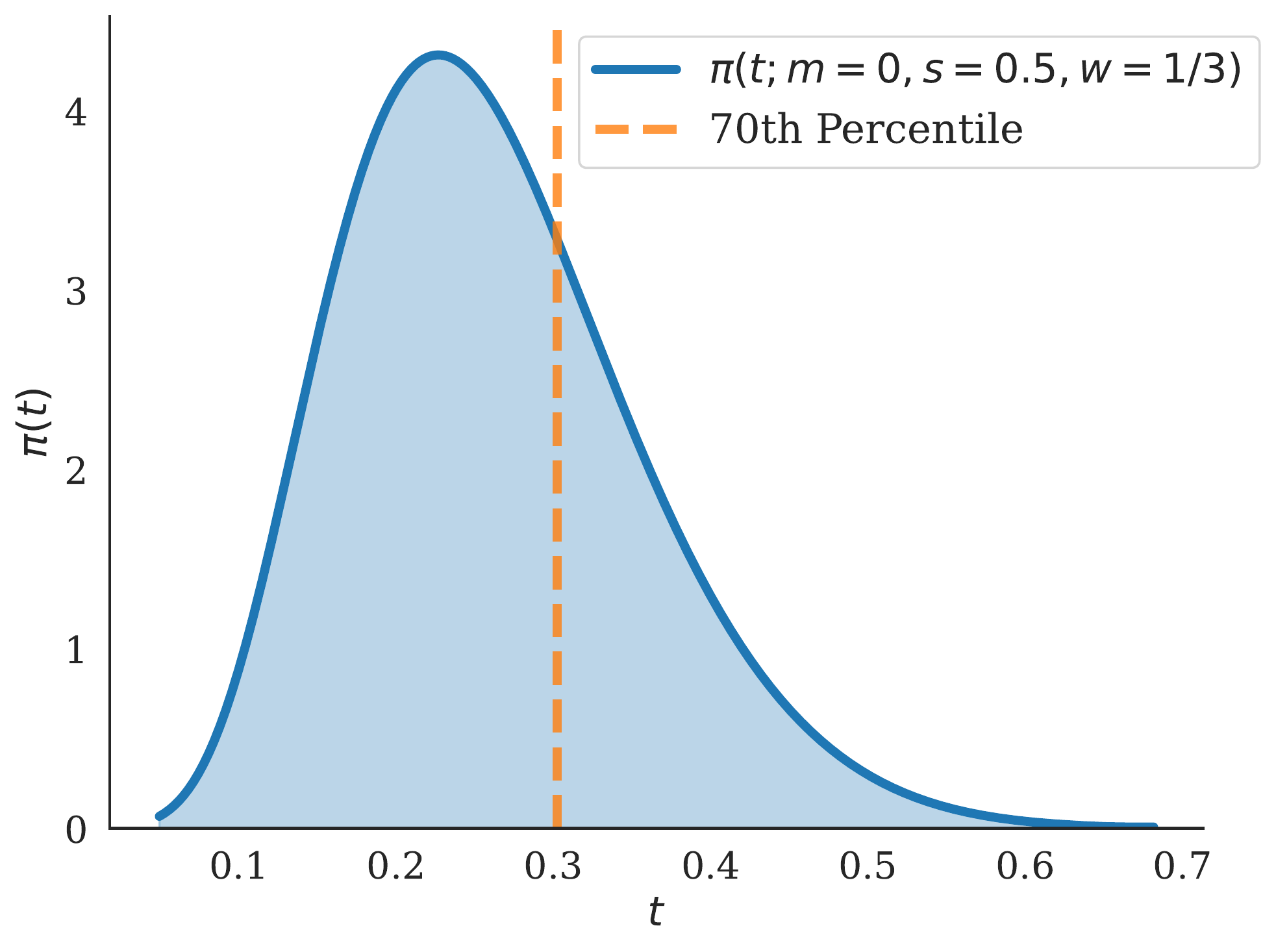}
\end{center}
\caption{The probability density of training timestep. We generally aim to allocate 70\% of training computation when t < 0.3.}
\label{fig:training_timestep}
\end{figure}

\paragraph{Timestep Sampler in Training}
Early studies have demonstrated that improving the design of timestep sampler (commonly known as SNR sampler) can facilitate training efficiency by better allocating computations across different noise levels. \citep{esser2024scaling} introduce the Logit-Normal sampling strategy, which provides a flexible framework for controlling the distribution of sampled timestep, the transformed timestep density $\pi(t)$ is:
\begin{equation}
\pi(t; m, s) = \frac{1}{s\sqrt{2\pi}}\frac{1}{t(1-t)}\exp(-\frac{(\text{logit}(t)-m)^2}{2s^2}),
\end{equation}
where $\text{logit}(t)=\log \frac{t}{1-t}$. In addition, \citep{esser2024scaling} further introduce a timestep shift strategy to handle the resolution increasing:
\begin{equation}
\label{eq:t_scale_transform}
    t' = \frac{wt}{1 - (1-w)t}
\end{equation}

\begin{figure}[htb!]
\begin{center}
\includegraphics[width=0.95\textwidth]{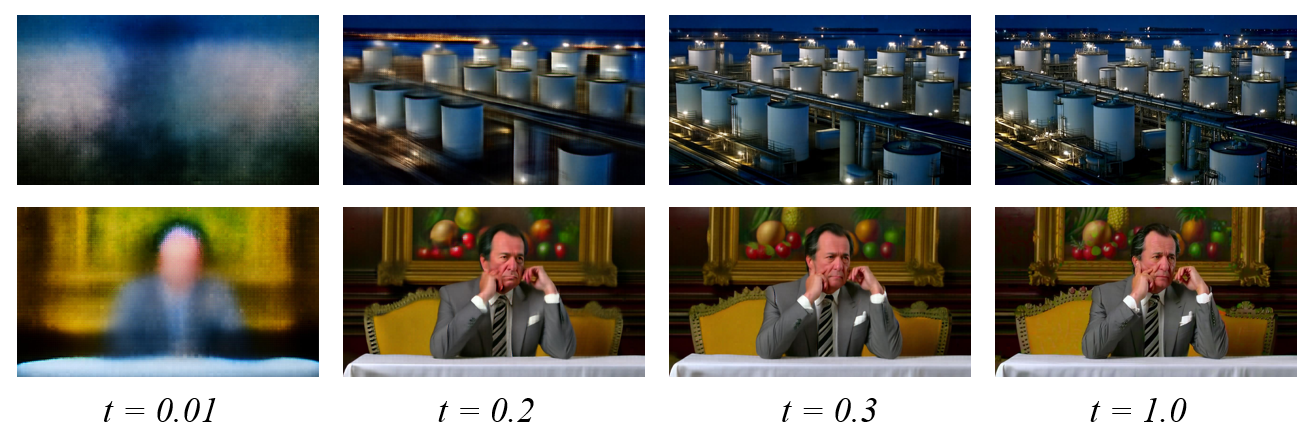}
\end{center}
\caption{The generation results at the given timestep $t$. Through empirical experiments, we found that model is capable of generating quite clear video outputs at $t=0.3$.}
\label{fig:stop_t}
\end{figure}

In \magi, we draw inspiration from these two sampling strategies but make adjustment for video data. Since videos typically contain more redundant information than images, we aim to shift the overall sampling distribution further towards the noise side compared to images. In our preliminary experiments, as shown in Fig.~\ref{fig:stop_t}, we observed that the model is capable of generating reasonably clear video outputs at $t = 0.3$. Based on this observation, we heuristically allocate approximately $70\%$ of the training computation budget to the region where $t < 0.3$. Following this principle, we set the $m = 0$, $s = 0.5$, and $w = 1/3$ for all cases during training.

\paragraph{Design Choices for Clean Chunks}
There are two types of chunks in the training of \magi: noisy chunks and clean chunks, and we adopt three key designs to handle clean chunks:

First, in practical video continuation scenarios, users typically upload an initial video clip and provide follow-up text descriptions or dynamically update the prompt during the continuation process. Considering this usage, we argue that clean chunks should not be conditioned on text inputs.

Second, exposure bias is a well-recognized challenge in training auto-regressive models~\citep{bengio2015scheduled,wiseman2016sequence}. A common mitigation strategy is to inject a small amount of noise into clean data during training. Following this practice, we inject up to $5\%$ noise into clean chunks to alleviate exposure bias.

Finally, since clean chunks are relatively abundant in the pre-training data, they risk dominating the training signal. To address this, we apply the loss function exclusively to noisy chunks. Nevertheless, clean chunks still participate in training through the attention mechanism and continue to receive gradient updates. Empirically, we observe that blocking the gradients of clean chunks leads to a significant degradation in model performance.

\subsection{Distillation Using Shortcut Model}
Flow-matching formulates the generative process as an ODE that maps noise to data along high-dimensional, curved trajectories. Sampling from such models is computationally intensive, typically requiring dozens of function evaluations with sufficiently small step sizes to incrementally transform noise into data. This inefficiency motivates the development of diffusion distillation methods~\citep{luhman2021knowledge, salimans2022progressive} that can reduce the required number of inference steps without sacrificing sample quality.

This work adopts shortcut model~\citep{frans2024one} as the distillation target. Given a noise-data interpolation defined by $x_i^t = (1 - t)x_i^0 + t x_i^1$, where \( x_i^1 \) is the clean data point at the \( i \)-th chunk and \( x_i^0 \) denotes Gaussian noise, the shortcut model uses a \textit{single} neural network to predict a velocity field \( v(x_i^t \mid t, s) \)\footnote{For clarity, we omit irrelevant variables and denote \( v(x_i^t \mid t, s) \) as a shorthand for the full expression \( v(x_i^{t_i} \mid t_i, s, c, \{x_{j<i}^{t_j}\}; \theta) \).}, conditioned not only on the current timestep \( t \), but also on the \textit{desired step size} \( s \propto \Delta t \).
Here, \( \Delta t \) denotes the interval between adjacent timesteps, while the reciprocal \( 1/s \in 2^{\mathbb{N}} \) specifies the number of function evaluations required to complete the denoising process\footnote{In practice, note that \( s \) and \( \Delta t \) may differ due to nonlinearities in the denoising schedule.}.

The generation process of the shortcut model closely resembles the flow-matching formulation and can be expressed as $\hat{x}_i^{t + \Delta t} = x_i^t + \Delta t \cdot v(x_i^t, t, s)$, where $\hat{x}_i^{t + \Delta t}$ denotes the model-predicted next point in the denoising trajectory, explicitly indicated by the hat symbol over ${x}_i^{t + \Delta t}$. As $\Delta t \to 0$, this formulation recovers the standard flow-matching scenario, where the shortcut model approximates the instantaneous velocity.

During training, the shortcut model constructs distillation targets using a bootstrap procedure, leveraging the principle that a single shortcut step is equivalent to two consecutive steps of half the step size. Formally, the update rule $\hat{x}_i^{t + \Delta t_1 + \Delta t_2} = x_i^t + (\Delta t_1 + \Delta t_2) \cdot v(x_i^t, t, 2s)$ can also be written as $\hat{x}_i^{t + \Delta t_1 + \Delta t_2} = \hat{x}_i^{t + \Delta t_1} + \Delta t_2 \cdot v(\hat{x}_i^{t + \Delta t_1}, t + \Delta t_1, s) = x_i^t + \Delta t_1 \cdot v(x_i^t, t, s) + \Delta t_2 \cdot v(\hat{x}_i^{t + \Delta t_1}, t + \Delta t_1, s)$, leading to the relationship:
\begin{equation} 
v(x_i^t, t, 2s) = \frac{\Delta t_1}{\Delta t_1 + \Delta t_2} v(x_i^t, t, s) + \frac{\Delta t_2}{\Delta t_1 + \Delta t_2} v(\hat{x}_i^{t + \Delta t_1}, t + \Delta t_1, s) 
\label{eq:shortcut_velocity} 
\end{equation}
In practice, the smallest $s$ utilized is $1/64$, corresponding to the standard flow-matching inference setting that requires 64 function evaluations. When training with this minimal step size, we incorporate classifier-free guidance (CFG) distillation~\citep{meng2023distillation} (see Sec.~\ref{CFG Design} for details). The step size $s$ for distillation is cyclically sampled from the set $[1/64] \times 8 \cup [1/32, 1/16, 1/8]$. This sampling strategy enables a single distilled model to perform denoising with different computational budgets (64, 32, 16, or 8 steps), thus providing flexibility to dynamically balance generation quality and inference efficiency at test time.

\subsection{Inference Approach}
\subsubsection{Diffusion Guidance}
\label{CFG Design}
\begin{figure}[!htb]
    \centering
    \begin{minipage}{0.98\textwidth}
    \begin{subfigure}[b]{1.0\textwidth}
        \centering
        \includegraphics[width=1.0\textwidth]{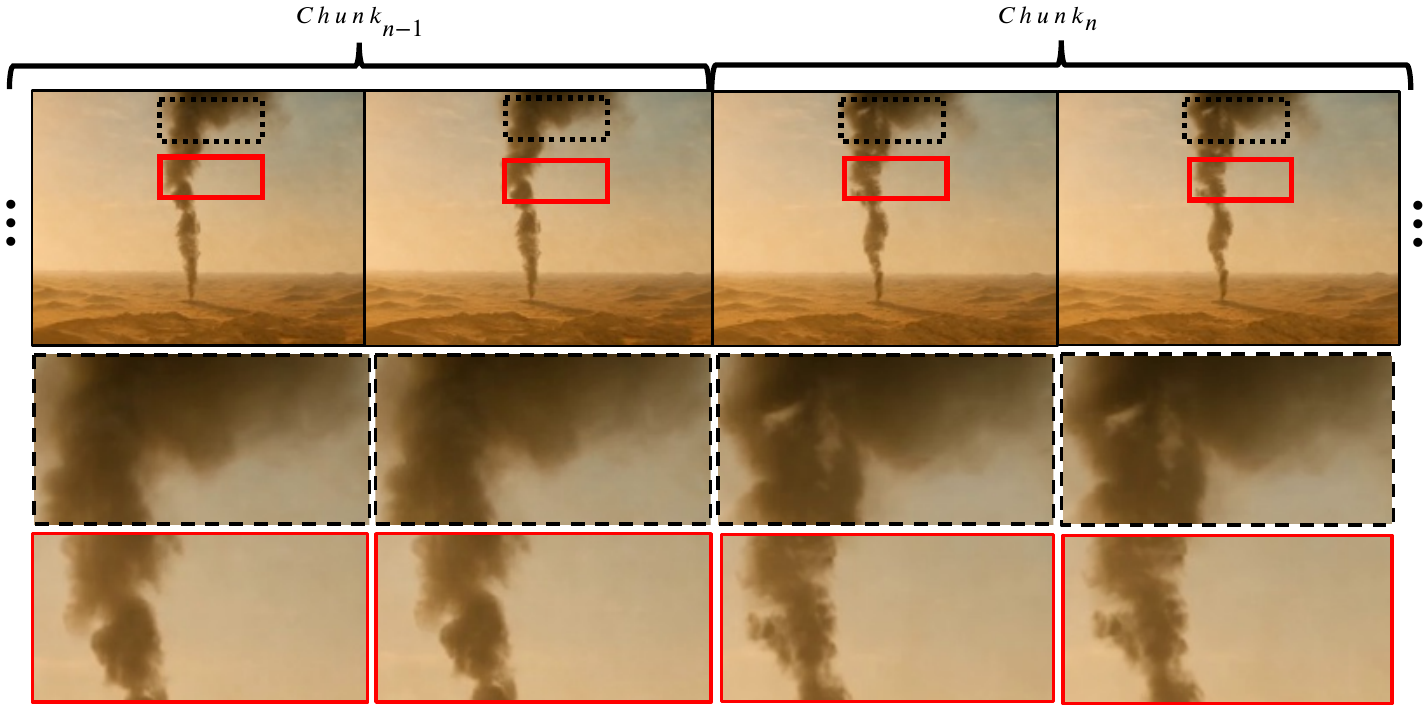}
        \caption{\(w_{\text{prev}} = 1.0\)}
        \label{fig:prev_cfg_a}
    \end{subfigure}
    \hfill
    \begin{subfigure}[b]{1.0\textwidth}
        \centering
        \includegraphics[width=1.0\textwidth]{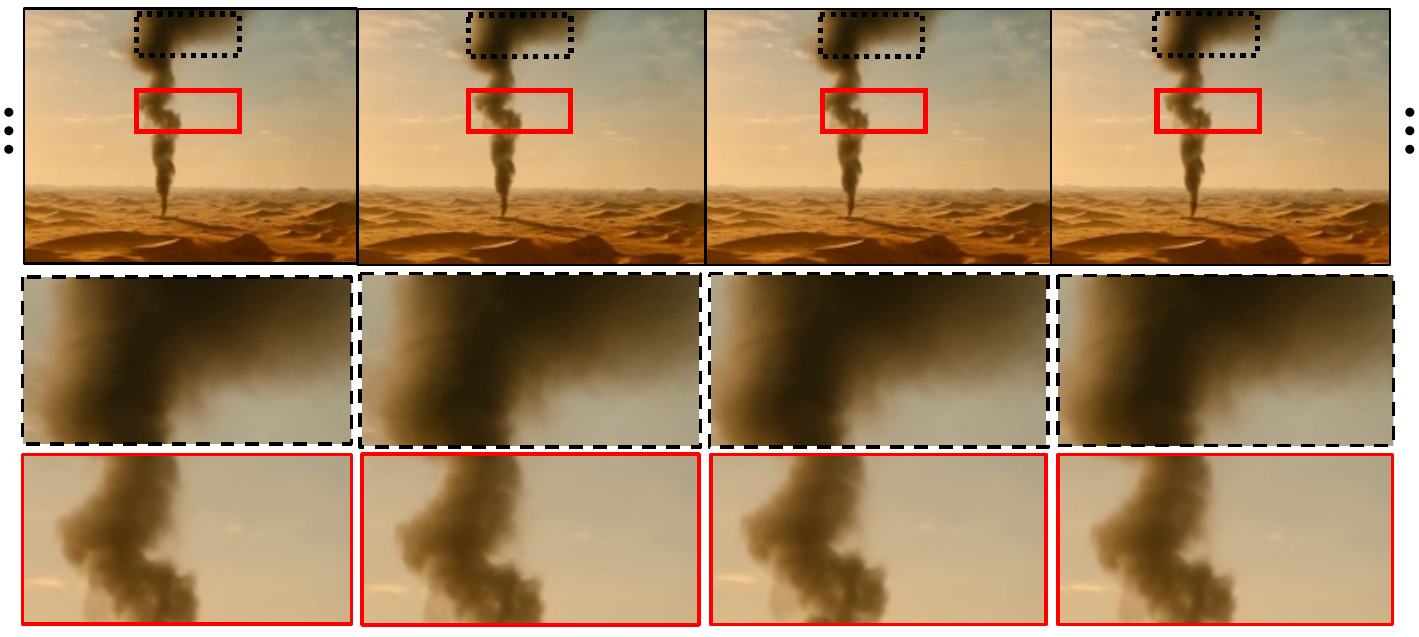}
        \caption{\(w_{\text{prev}} = 1.5\)}
        \label{fig:prev_cfg_b}
    \end{subfigure}
    \end{minipage}
    \caption{This figure demonstrates the impact of \(w_{\text{prev}}\) on the generation results. (a) When \(w_{\text{prev}} = 1.0\), there are perceptible misalignments between adjacent chunks (\emph{e.g.}, the shape of the smoke). (b) When \(w_{\text{prev}} = 1.5\), this phenomenon is significantly alleviated.}
    \label{fig:prev_cfg}
\end{figure}

Classifier-free guidance~\citep{ho2022classifier}, a widely adopted low-temperature sampling technique in diffusion models, offers a principled approach to mediating the inherent trade-off between sample fidelity and diversity in generative modeling. This technique is particularly effective in text-to-video generation, where the objective is to synthesize temporally coherent video frames that conform to given textual prompts.

To improve clarity, we omit irrelevant variables and express the \textit{guided} posterior distribution of the latent variable \( x_t \) given condition \( c \) using Bayes’ rule as \( p_\text{guided}(x_t \mid c) \propto p(x_t) \cdot p(c \mid x_t)^w \), where the exponent \( w \geq 1 \) serves as an inverse temperature parameter. Exponentiating the conditional likelihood \( p(c \mid x_t) \) concentrates the distribution around modes better aligned with the conditioning signal, thereby reducing entropy and improving sample fidelity.

In our setting, the generation of the \(i\)-th video chunk \(x_i\) is conditioned not only on the textual prompt \(c_{\text{text}}\), but also on a sequence of preceding chunks \(x_{<i}\), which may include partially denoised or fully noised representations. The guided conditional distribution is thus formulated as:
\begin{equation}
p_{\text{guided}}(x_i \mid x_{<i}, c_{\text{text}}) \propto p(x_i) \cdot p(x_{<i} \mid x_i)^{w_{\text{prev}}} \cdot p(c_{\text{text}} \mid x_{<i}, x_i)^{w_{\text{text}}},
\label{eq:multi-guidance-1}
\end{equation}
where \(w_{\text{prev}}\) and \(w_{\text{text}}\) are scalar weights modulating the influence of temporal and semantic signals, respectively.
Taking the logarithm of both sides of Eq.~\ref{eq:multi-guidance-1}, we obtain the guided score:  
\(\nabla_{x_i} \log p_{\text{guided}}(x_i \mid x_{<i}, c_{\text{text}}) = \nabla_{x_i} \log p(x_i) + w_{\text{prev}} \cdot \nabla_{x_i} \log p(x_{<i} \mid x_i) + w_{\text{text}} \cdot \nabla_{x_i} \log p(c_{\text{text}} \mid x_{<i}, x_i)\).
Applying Bayes' rule, we rewrite the gradients as  
\(\nabla_{x_i} \log p(x_{<i} \mid x_i) = \nabla_{x_i} \log p(x_i \mid x_{<i}) - \nabla_{x_i} \log p(x_i) \),  
and  
\(\nabla_{x_i} \log p(c_{\text{text}} \mid x_{<i}, x_i) = \nabla_{x_i} \log p(x_i \mid x_{<i}, c_{\text{text}}) - \nabla_{x_i} \log p(x_i \mid x_{<i}) \).
Substituting and regrouping, we arrive at the final guided score:
\begin{equation}
\begin{aligned}
\nabla_{x_i} \log p_{\text{guided}}(x_i \mid x_{<i}, c_{\text{text}}) 
&= \ (1 - w_{\text{prev}}) \cdot \nabla_{x_i} \log p(x_i) \\
&+\ (w_{\text{prev}} - w_{\text{text}}) \cdot \nabla_{x_i} \log p(x_i \mid x_{<i}) \\
&+\ w_{\text{text}} \cdot \nabla_{x_i} \log p(x_i \mid x_{<i}, c_{\text{text}}).
\end{aligned}
\label{eq:final-score}
\end{equation}
This decomposition cleanly separates the contributions of the unconditional prior, temporal context, and prompt conditioning. It enables controllable trade-offs between coherence and semantic fidelity in autoregressive generation.

As a special case, when \( w_{\text{prev}}\) is 1, \emph{i.e.}, the guidance from previous chunks is disabled, the score function in Eq.~\ref{eq:final-score} simplifies to \( \nabla_{x_i} \log p_{\text{guided}}(x_i \mid x_{<i}, c_{\text{text}}) = (1 - w_{\text{text}}) \cdot \nabla_{x_i} \log p(x_i \mid x_{<i}) + w_{\text{text}} \cdot \nabla_{x_i} \log p(x_i \mid x_{<i}, c_{\text{text}}) \). This form recovers the standard classifier-free guidance formulation widely adopted in bidirectional text-to-video diffusion models, which interpolates between unconditional and prompt-conditioned signals.

However, during our chunk-wise generation process, we observed subtle yet perceptible misalignments between adjacent chunks, resulting in temporal artifacts. This observation underscores the necessity of reinforcing temporal guidance to maintain chunk-to-chunk coherence. To this end, we increase \( w_{\text{prev}} \) to 1.5, thereby amplifying the influence of preceding content. As shown in Fig.~\ref{fig:prev_cfg}, this adjustment significantly enhances inter-chunk alignment and mitigates flickering artifacts, resulting in smoother and more temporally consistent video synthesis. Nevertheless, it should be noted that further increasing \( w_{\text{prev}} \) beyond this optimal range may lead to saturation artifacts or even cause the video to become static (\emph{i.e.}, still frames) as playback progresses. We follow standard practice by setting \( w_{\text{text}} \) to 7.5.

\begin{figure}[htb!]
\begin{center}
\includegraphics[width=0.6\textwidth]{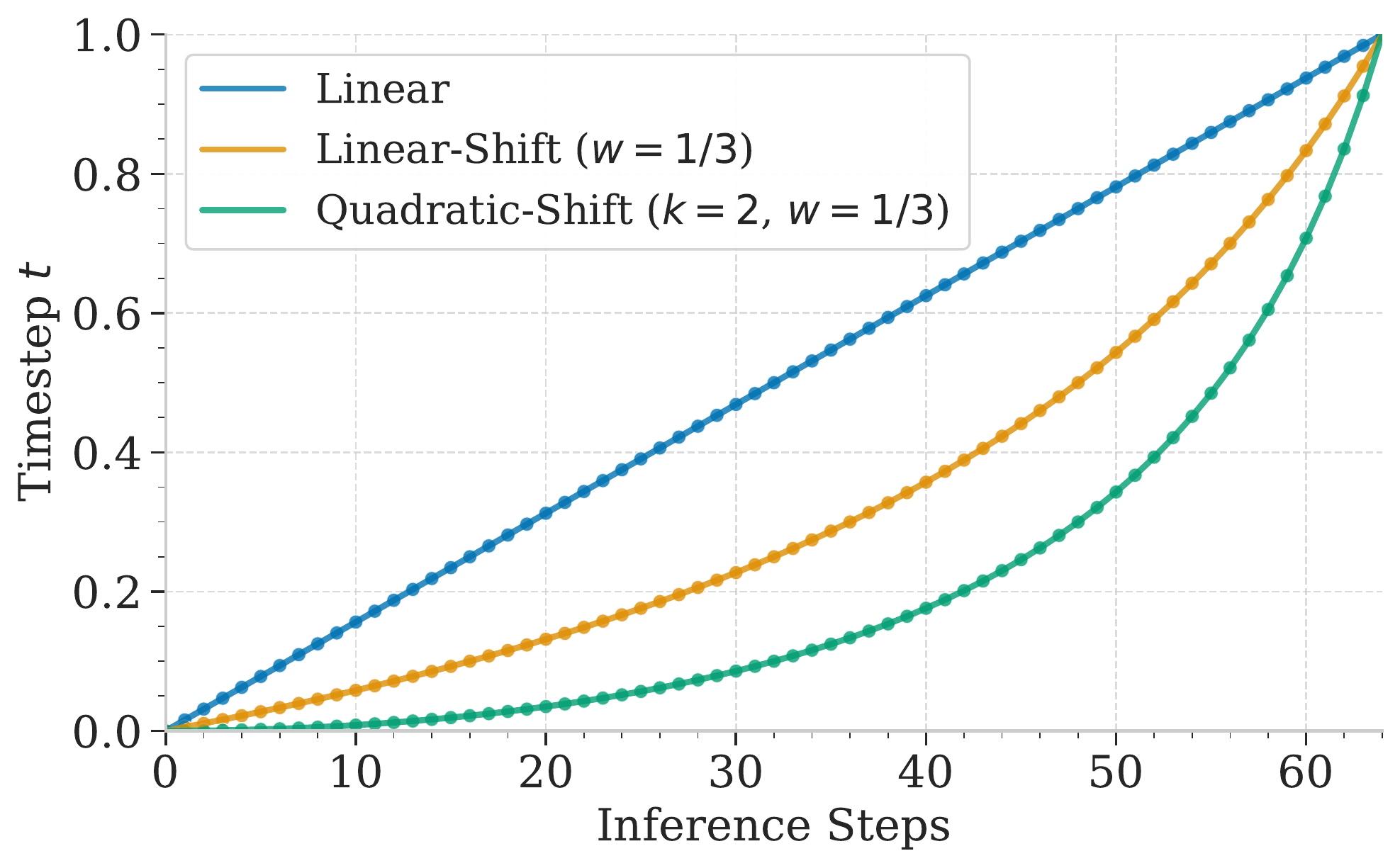}
\end{center}
\caption{Inference timestep sampling of non-distilled model. }
\label{fig:inference_timestep}
\end{figure}

\subsubsection{Inference Timestep Sampler}
Previous video generation studies have demonstrated that applying targeted timestep sampling strategies during inference can significantly improve generation quality. In our work, we observed similar behavior in \magi. To enable finer-grained control over the sampling process, we introduce an additional tunable power transformation based on the scaling formula (Eq.~\ref{eq:t_scale_transform}) $t' = \frac{wt^k}{1-(1-w)t^k}$. Through extensive experiments, we found that setting $w = 1/3$ and $k = 2$ yields the best visual quality, and visualization of the sampler is shown in Fig.~\ref{fig:inference_timestep}.

\begin{figure}[h]
    \centering
    \begin{minipage}{1.0\textwidth}
    \begin{subfigure}[b]{1.0\textwidth}
        \centering
        \includegraphics[width=1.0\textwidth]{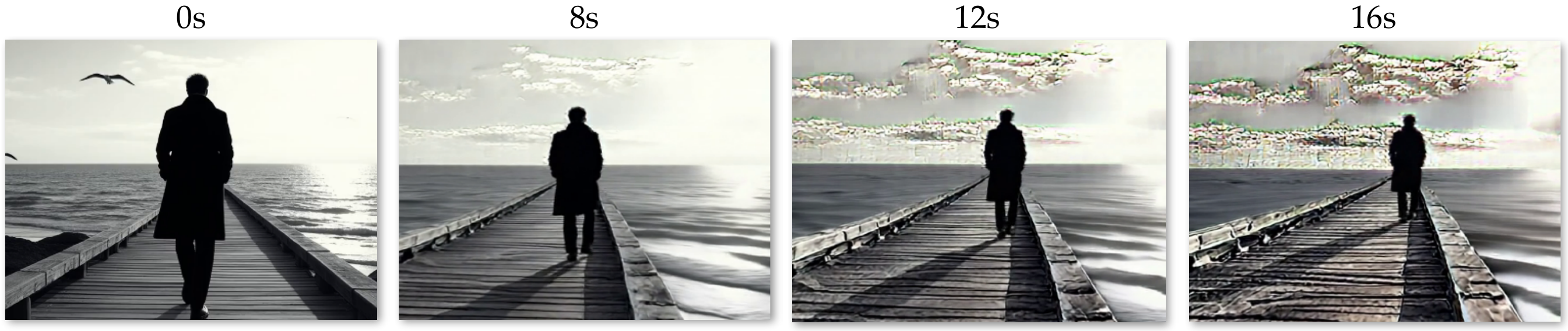}
        \caption{w{\!/}o Fine-Grained Control}
        \label{fig:fgc_a}
    \end{subfigure}
    \hfill
    \begin{subfigure}[b]{1.0\textwidth}
        \centering
        \includegraphics[width=1.0\textwidth]{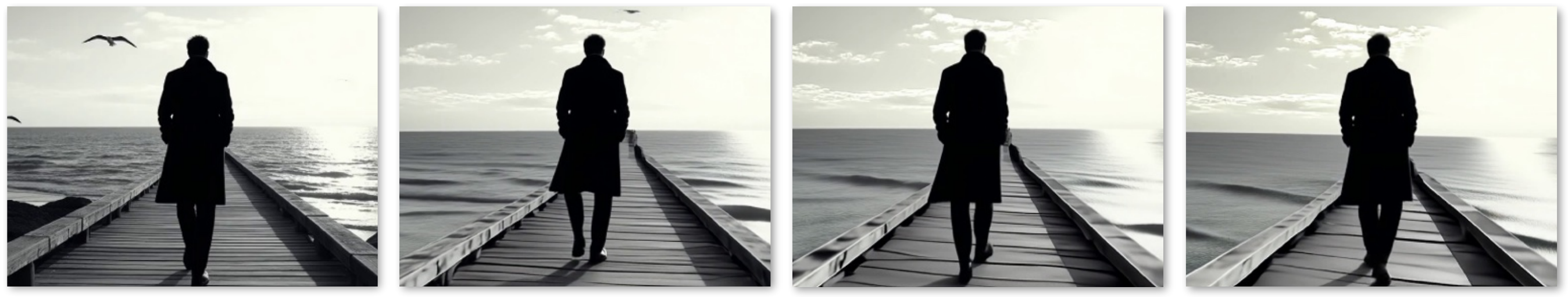}
        \caption{w{\!/} Fine-Grained Control}
        \label{fig:fgc_b}
    \end{subfigure}
    \end{minipage}
    \caption{(a) When the generation length exceeds $5$ seconds, severe artifacts emerge and intensify over time. (b) By adjusting the guidance strength (\emph{i.e.}, \(w_{\text{prev}}=1.0\) and \(w_{\text{text}}=0.0\) when \(t > 0.3\)), there are no serious artifacts in the entire generation.}
    \label{fig:fine_grained_control}
\end{figure}

\subsubsection{Fine-Grained Control of Guidance Strength}
\paragraph{Non-distilled Model.}
In the case of the non-distilled model, as described in Sec.~\ref{CFG Design}, we set \( w_{\text{prev}} = 1.5 \) and \( w_{\text{text}} = 7.5 \) during generation. In practice, when synthesizing longer videos (typically exceeding 5 seconds), we observe noticeable saturation and checkerboard artifacts progressively emerging during playback. These artifacts are primarily attributed to excessively strong guidance. However, uniformly reducing the strength of \(w_{\text{prev}}\) and \(w_{\text{text}}\) often results in degraded content quality and increased flickering artifacts. This motivates a more fine-grained strategy in which the guidance scales are dynamically adjusted throughout the denoising process, that is, varying \( w_{\text{prev}} \) and \( w_{\text{text}} \) as the denoising timestep \( t \) progresses from 0 to 1.

To investigate when strong guidance is necessary, we analyze the evolution of latent representations throughout the denoising process (Fig.~\ref{fig:stop_t}). As \( t \) approaches 0.3, just before the final denoising stage begins, we observe that the decoded latent representations already exhibit coherent video content, with both structural and semantic elements largely established. The remaining denoising steps, from \( t = 0.3 \) to \( t = 1 \), primarily serve to refine local details, resembling a super-resolution process. Based on this observation, we hypothesize that strong guidance from either the text or previous chunks is no longer necessary during this stage. Accordingly, for \( t > 0.3 \), we reduce the guidance scales to \( w_{\text{prev}} = 1.0 \) and \( w_{\text{text}} = 0.0 \), such that only the \( \nabla_{x_i} \log p(x_i \mid x_{<i}) \) term remains active. As illustrated in Fig.~\ref{fig:fine_grained_control}, this simple yet effective adjustment significantly alleviates temporal artifacts and improves the overall coherence of longer video generations.

\paragraph{Distilled Model.}
A similar observation holds for the distilled model. Saturation artifacts progressively intensify as the video plays, motivating a comparable mitigation strategy. In the first three stages, we directly use the distilled model's output score, $\nabla_{x_i} \log p_{\text{distilled}}(x_i \mid x_{<i}, c_{\text{text}})$. In the final denoising range, we incorporate additional \textit{guidance} to reduce the influence of the previous chunk, even though the model has already undergone classifier-free guidance distillation. Specifically, we adopt the following guided score in the final stage:
\begin{equation}
\begin{aligned}
\nabla_{x_i} \log p_{\text{guided, distilled}}(x_i \mid c_{\text{text}}, x_{<i}) 
&= (1 - w_{\text{prev}}) \cdot \nabla_{x_i} \log p_{\text{distilled}}(x_i \mid c_{\text{text}}) \\
&\quad + w_{\text{prev}} \cdot \nabla_{x_i} \log p_{\text{distilled}}(x_i \mid c_{\text{text}}, x_{<i}) .
\end{aligned}
\label{eq:distilled-guidance}
\end{equation}
This formulation is derived by switching the positions of $x_{<i}$ and $c_{\text{text}}$ in Eq.~\ref{eq:multi-guidance-1} and Eq.~\ref{eq:final-score}, resulting in the form $\nabla_{x_i} \log p_{\text{guided}}(x_i \mid c_{\text{text}}, x_{<i}) = (1 - w_{\text{text}}) \cdot \nabla_{x_i} \log p(x_i) + (w_{\text{text}} - w_{\text{prev}}) \cdot \nabla_{x_i} \log p(x_i \mid c_{\text{text}}) + w_{\text{prev}} \cdot \nabla_{x_i} \log p(x_i \mid c_{\text{text}}, x_{<i})$. By setting $w_{\text{text}} = 1$, thereby disabling the text guidance term, the first component vanishes and the expression simplifies to Eq.~\ref{eq:distilled-guidance}.
The rationale behind this modification is that we do not introduce a null text token during distillation, and therefore do not explicitly model $p_{\text{distilled}}(x_i)$ or $p_{\text{distilled}}(x_i \mid x_{<i})$. In our experiments, we set $w_{\text{prev}} = 0.7$, which effectively attenuates the influence of previous chunk guidance in the final denoising stage and helps mitigate temporal saturation artifacts.

\subsubsection{KV Cache}\label{subsubsec:kv_cache}
Thanks to its auto-regressive nature, \magi can leverage the KV cache mechanism during inference, which is a widely adopted technique in language models to avoid redundant computations. Specifically, once a chunk has been sufficiently denoised, its features can be cached and reused by subsequent denoising chunks without the need for recomputation.

Furthermore, by constraining the KV range, \magi can easily support long video generation. For example, by setting the KV range to 8 for all chunks, each newly generated chunk depends only on the preceding 8 seconds of video content. This design ensures that the computational cost of generating long videos scales linearly with their duration.

In addition, many KV compression~\citep{hooper2024kvquant,xiao2023efficient,sheng2023flexgen} techniques have recently been developed to reduce the computational overhead of auto-regressive model while preserving the ability to reference the full history as much as possible. \magi is theoretically compatible with these advancements, although we leave their exploration in \magi for future work.

\magi also benefits from the unique characteristics of denoising models: at higher noise levels, the model focuses on capturing global structural information, whereas at lower noise levels, it produces fine details and textures. By dynamically adjusting the KV range at different denoising stages, we can unlock new capabilities that were previously challenging to achieve, such as enabling temporally controllable shot transitions while preserving subject identities, or allowing changes in object identities while maintaining consistent global layouts. More details and experimental results are provided in Sec.~\ref{sec:model_capability_study}.

\subsection{Prompt-Enhancement Strategy}
\magi is trained with highly descriptive captions that follow a specific structure as text conditions. However, in real-world scenarios, user inputs vary widely: ranging from very brief prompts to overly elaborate descriptions. This mismatch between the training distribution and real user inputs often leads to suboptimal inference performance. To address this gap, we propose a Prompt Enhancement (PE) strategy during inference.
We take the image-to-video (I2V) task as an example to illustrate our PE approach. In this setting, users typically provide an image along with an optional textual prompt. To enhance the user input, we employ a state-of-the-art multi-modal large language model (MLLM) to perform prompt refinement. Our PE pipeline consists of two parallel sub-processes:
\begin{itemize}
    \item The first sub-process analyzes and describes the content of the uploaded image.
    \item The second sub-process predicts the temporal evolution of the scene or objects in the first frame, such as actions, motion trajectories, and object transitions.
\end{itemize}
This structured enhancement strategy significantly improves generation quality. However, due to the large size of the state-of-the-art MLLM, it incurs high computational cost and latency, limiting its feasibility in real applications.
To enable lightweight deployment, we distill the enhanced prompts generated by the large MLLM into a smaller, more efficient model (\textasciitilde7B). We construct a training corpus of approximately 2 million examples, filtering out samples with excessively long target texts to ensure controlled output length. Based on human evaluation, the distilled model achieves comparable video generation quality to its larger counterpart, while greatly reducing inference latency and computational resource usage.

\begin{figure}[htbp]
    \centering
    \begin{minipage}{1.0\textwidth}
    \begin{subfigure}[b]{1.0\textwidth}
        \centering
        \includegraphics[width=1.0\textwidth]{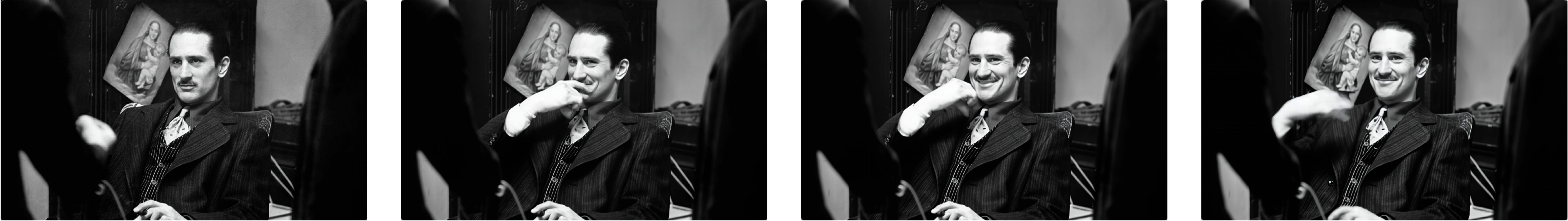}
        \caption{A man smiles while resting his chin on his hand.}
        \label{fig:chunk_control_a}
    \end{subfigure}
    \hfill
    \begin{subfigure}[b]{1.0\textwidth}
        \centering
        \includegraphics[width=1.0\textwidth]{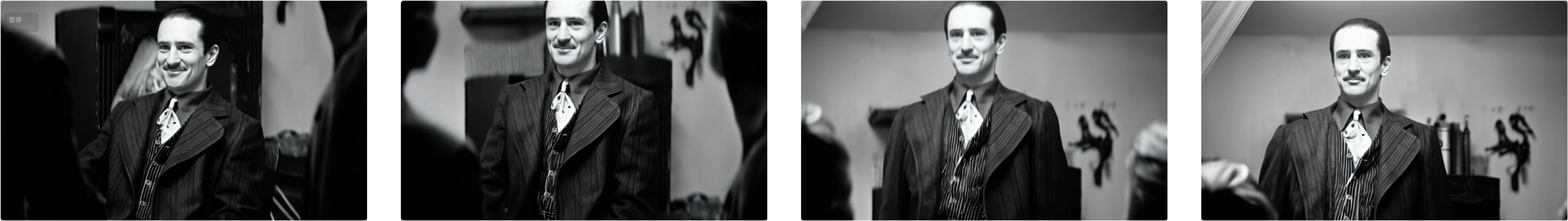}
        \caption{He slowly rises from his seat.}
        \label{fig:chunk_control_b}
    \end{subfigure}
    \hfill
    \begin{subfigure}[b]{1.0\textwidth}
        \centering
        \includegraphics[width=1.0\textwidth]{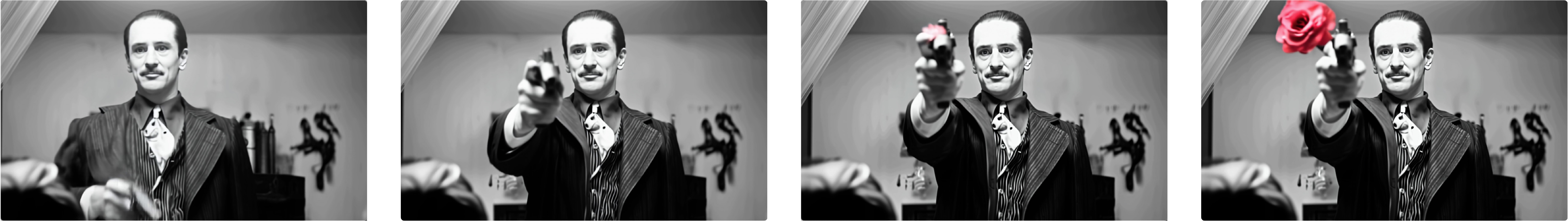}
        \caption{Draws a pistol, from which a red rose is fired.}
        \label{fig:chunk_control_c}
    \end{subfigure}
    \begin{subfigure}[b]{1.0\textwidth}
        \centering
        \includegraphics[width=1.0\textwidth]{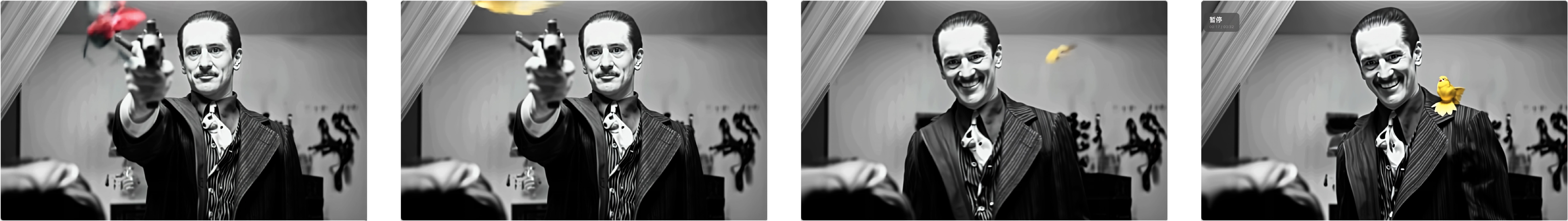}
        \caption{The rose transforms into a yellow bird that lands on his shoulder as he makes a playful expression.}
        \label{fig:chunk_control_d}
    \end{subfigure}
    \begin{subfigure}[b]{1.0\textwidth}
        \centering
        \includegraphics[width=1.0\textwidth]{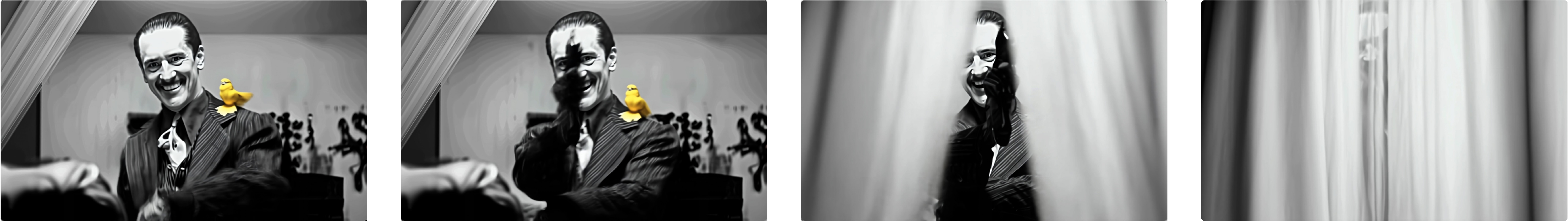}
        \caption{He performs a juggling gesture as curtains on both sides gradually close, concealing him completely.}
        \label{fig:chunk_control_e}
    \end{subfigure}
    \begin{subfigure}[b]{1.0\textwidth}
        \centering
        \includegraphics[width=1.0\textwidth]{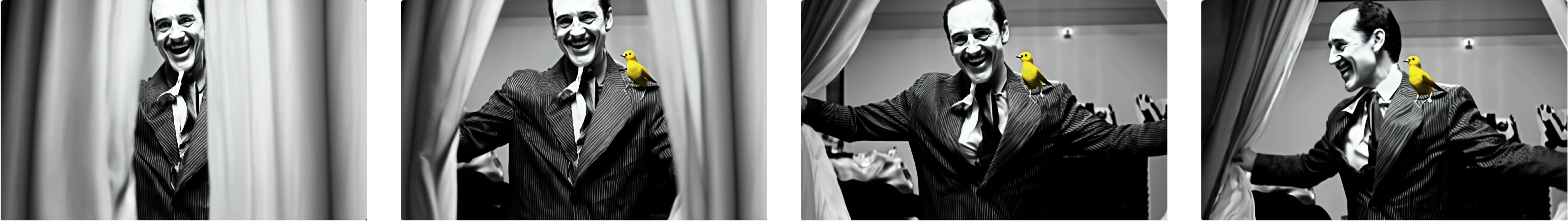}
        \caption{The curtains reopen by the man, then he turns and walks away.}
        \label{fig:chunk_control_f}
    \end{subfigure}
    \begin{subfigure}[b]{1.0\textwidth}
        \centering
        \includegraphics[width=1.0\textwidth]{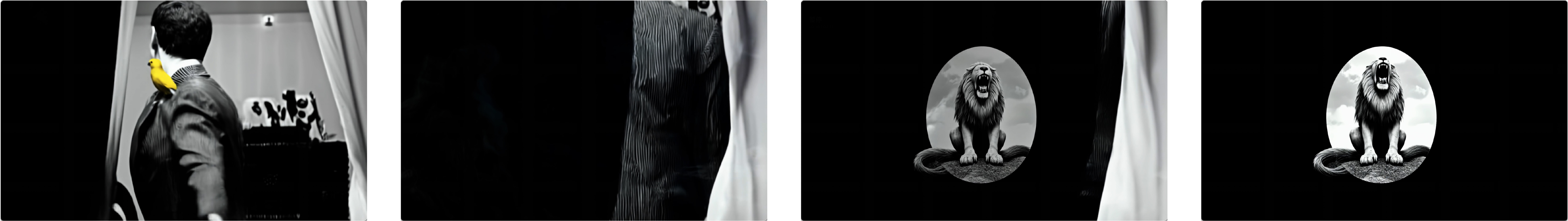}
        \caption{As he departs, a roaring lion logo slowly fades into view on the screen.}
        \label{fig:chunk_control_g}
    \end{subfigure}
    \end{minipage}
    \caption{This figure presents a near 30-second video generation example that demonstrates the capability of our model for complex actions and narrative structures through chunk-wise controllability and long-video generation. The sequence progresses from (a) to (g), with each sub-caption corresponding to the text prompt used during generation.}
    \label{fig:chunk_control}
\end{figure}

\subsection{Model Capability Study}
\label{sec:model_capability_study}
\paragraph{Real-time Streaming Video Generation}
The chunk-by-chunk pipelined inference of \magi offers two key advantages: (1) the time to display the first clear chunk is independent of the total generated video length; and (2) the generation latency between consecutive chunks is significantly reduced. Combined with a high-performance inference infrastructure, \magi enables real-time streaming video generation, unlocking new applications in interactive content and live streaming. More implementation details are in Sec.~\ref{sec:real_time_infra}.

\paragraph{Chunk-wise Text Controllability}
Chunk-wise text controllability is one of the key features of \magi, enabling us to decompose complex actions into simpler, shorter segments and significantly enhancing the model's ability to generate intricate action sequences. Furthermore, when combined with the capability of \magi for long video generation, this makes it possible to create videos with complex narrative structures, as illustrated in Fig.~\ref{fig:chunk_control}

\paragraph{Video Continuations}

\begin{figure}[ht]
    \centering
    \begin{minipage}{1.0\textwidth}
        \begin{subfigure}[b]{1.0\textwidth}
            \centering
            \includegraphics[width=1.0\textwidth]{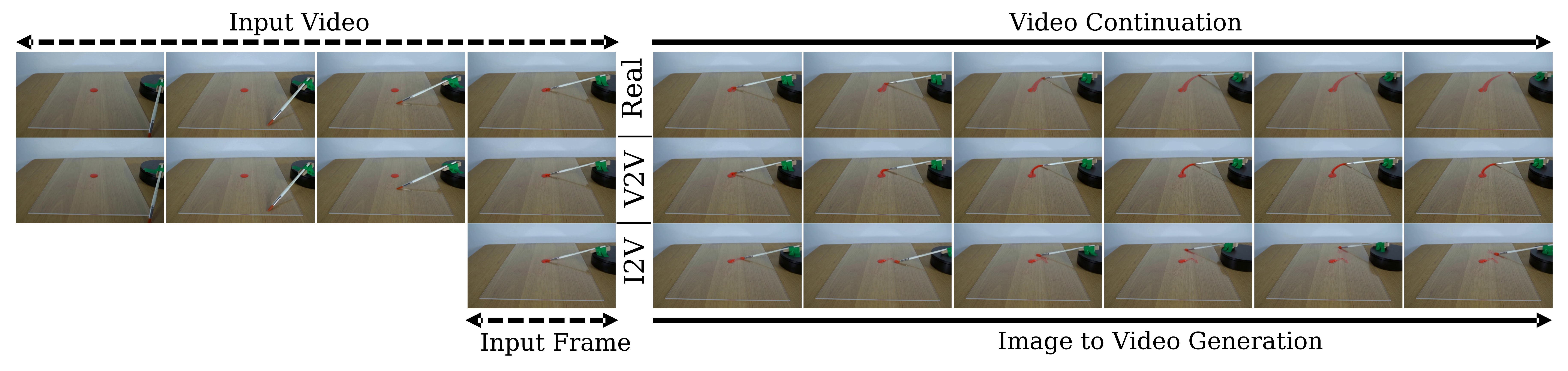}
            \caption{Text Guidance: A clear acrylic sheet placed on a wooden table with a small dollop of red paint. A rotating paintbrush attached to a rotating platform rotates clockwise and goes through the paint. Static shot with no camera movement.}
            \label{fig:video_continues_demo_01}
        \end{subfigure}
        \begin{subfigure}[b]{1.0\textwidth}
            \centering
            \includegraphics[width=1.0\textwidth]{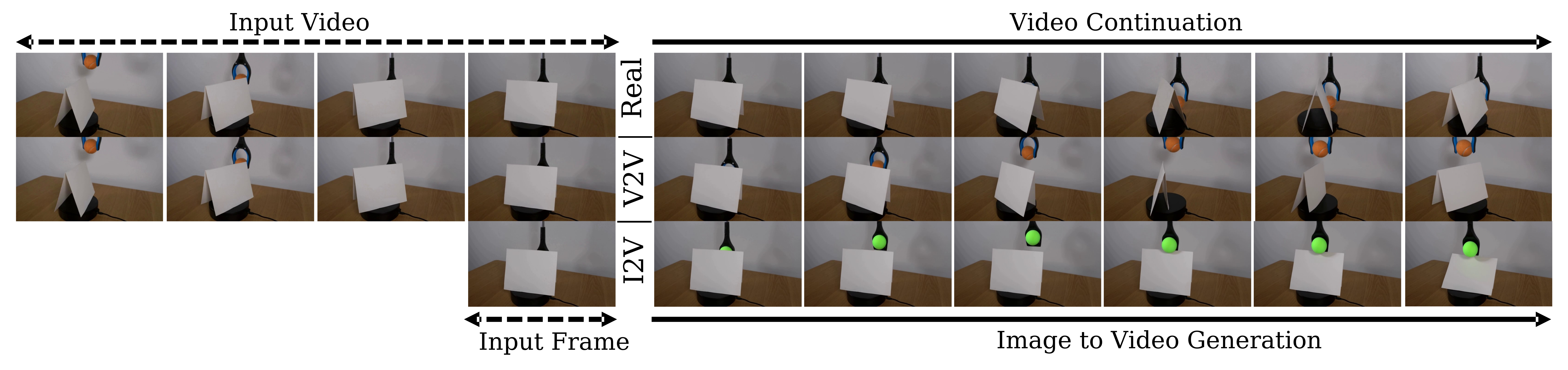}
            \caption{Text Guidance: A grabber arm is holding a tennis ball above a piece of cardstock propped up on a rotating platform sitting on a table that rotates clockwise. The grabber lowers the ball and places is on the table as the cardstock rotates. Static shot with no camera movement.
    }
            \label{fig:video_continues_demo_02}
        \end{subfigure}
    \end{minipage}
    \caption{Comparison between video-conditioned (V2V) and image-conditioned (I2V) video continuation. (a) \magi (V2V) accurately captures the pen’s rotational trajectory by leveraging historical motion information, while I2V fails to reproduce the correct motion due to the absence of temporal context. (b) In an occlusion scenario, V2V successfully predicts post-occlusion behavior by utilizing information before the occlusion, whereas I2V shows poor temporal consistency. Each example presents the real-world scene (top row), \magi (V2V) generation (middle row), and \magi (I2V) generation (bottom row).}
\label{fig:video_continues_demo}
\end{figure}

Video continuation is a task that \magi natively supports. In the community, an alternative approach to video continuation relies on image-to-video generation (I2V), where the last frame of the given prefix video is used as the starting frame for the extended video. However, this approach often struggles to maintain consistent motion trajectories between the generated continuation and the prefix video, leading to motion discontinuities or generating implausible predictions due to the loss of essential historical information. Fig.~\ref{fig:video_continues_demo} shows such cases. In the pen rotation example, I2V fails to capture the correct rotational velocity because it lacks access to preceding motion dynamics. Similarly, in the occlusion scenario, I2V cannot accurately predict the object's reappearance after occlusion due to missing temporal information. In contrast, conditioning on the full prefix video allows \magi to naturally preserve motion continuity by leveraging historical patterns and temporal dependencies, enabling seamless video continuation.

\begin{figure}[htbp]
    \centering
    \begin{minipage}{1.0\textwidth}
    \begin{subfigure}[b]{1.0\textwidth}
        \centering
        \includegraphics[width=1.0\textwidth]{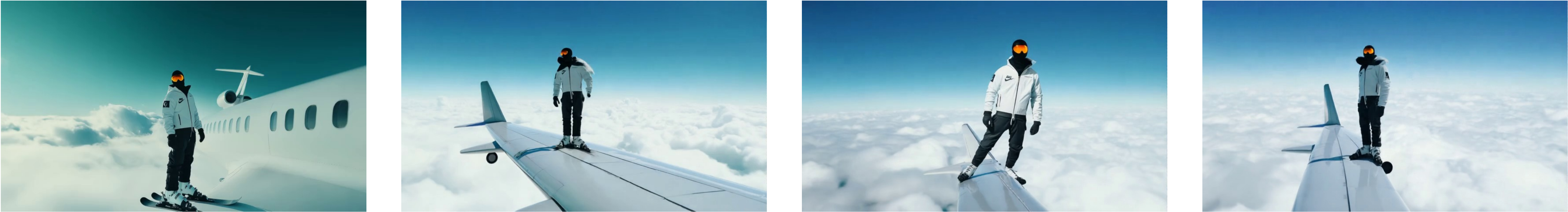}
        \caption{Shot transition with preserved identity.}
        \label{fig:shot_transition_layout}
    \end{subfigure}
    \hfill
    \begin{subfigure}[b]{1.0\textwidth}
        \centering
        \includegraphics[width=1.0\textwidth]{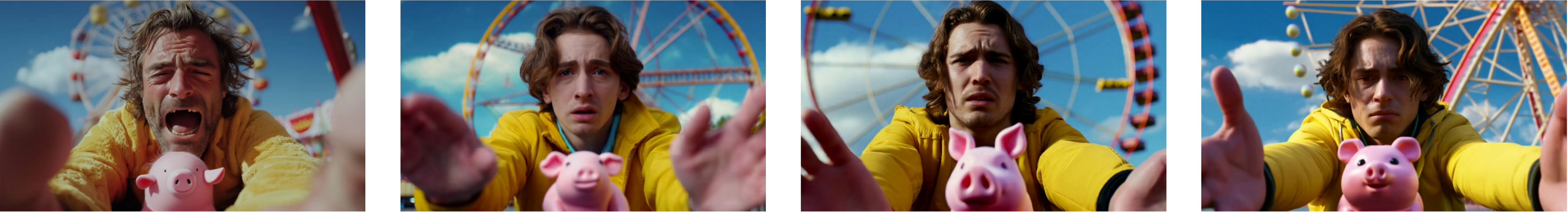}
        \caption{Transition with consistent scene layout but changing object details.}
        \label{fig:shot_transition_finegrain}
    \end{subfigure}
    \end{minipage}
    \caption{This figure illustrates two examples of realizing distinct shot transitions by modulating the KV range at different denoising stages. (a) demonstrates a case where the KV range is set to 1 only at the high-noise denoising stages, whereas (b) applies it at the low-noise denoising stages.}
    \label{fig:controlable_shot_transition}
\end{figure}

\paragraph{Controllable Shot Transition}
Another exciting feature of \magi is its ability to enable diverse and controllable transitions at any designated chunk by adjusting the KV range across different denoising stages. Specifically, by setting the KV range to 1 only at high-noise denoising stages (meaning the model cannot access the preceding video content) while keeping a normal KV range (\emph{e.g.}, 8) at other stages, we can achieve shot transitions while preserve object identities unchanged, as shown in Fig.~\ref{fig:shot_transition_layout}.
Conversely, by setting the KV range to 1 only at low-noise stages, we can produce transitions where the overall layout of the scene remains consistent, but the fine details of the objects change, as illustrated in Fig.~\ref{fig:shot_transition_finegrain}.

We believe the above capabilities can offer an entirely new level of creative control for video content creation.

\section{DATA}
Training a high-performance video generation model demands massive, high-quality, and diverse data. To this end, we have developed a scalable data processing system that constructs the training dataset for \magi from tens of petabytes of raw videos and images collected from a wide range of sources.

\begin{figure}[htb!]
\begin{center}
\includegraphics[width=0.9\textwidth]{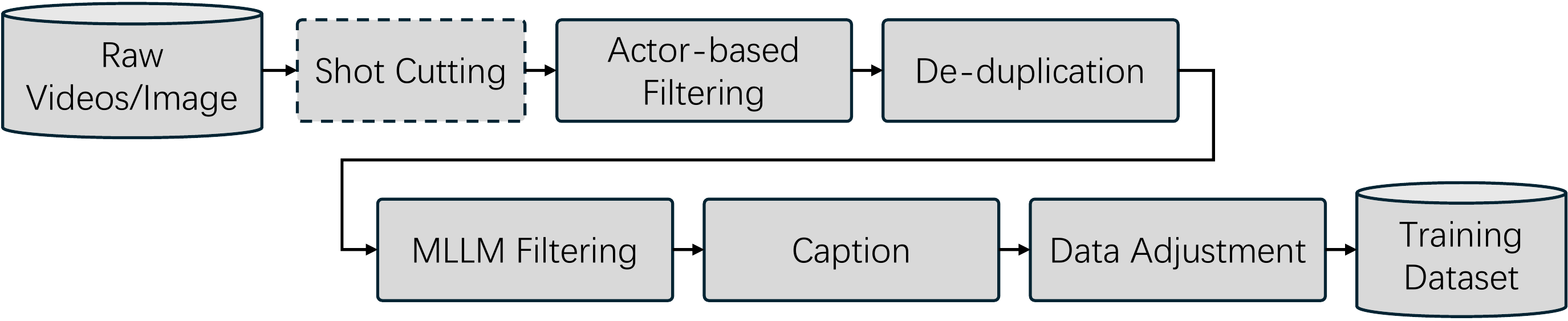}
\end{center}
\caption{Overview of the our data processing pipeline. The shot cutting module is only applied for video data. }
\label{fig:data_pipeline_overview}
\end{figure}

An overview of the data processing pipeline is shown in Fig.~\ref{fig:data_pipeline_overview}. We utilize PySceneDetect\footnote{PySceneDetect: \url{https://github.com/Breakthrough/PySceneDetect}} to cut long videos into short clips, ensuring that each clip contains only a single shot. Next, we apply a series of filters to remove low-quality data and eliminate duplicates. While this initial filtering stage effectively discards most of the low-quality data, some problematic cases still persist. To further improve data quality, we incorporate a multi-modal large language model (MLLM) as a stronger filter. Data that passes this filter is then captioned by the MLLM to provide accurate and detailed descriptions.

Through this process, we curate training data with satisfactory visual and motion quality. However, the distribution of the data — particularly in terms of semantic concept — still requires consideration. Specifically, we observed that the modeling difficulty varies significantly across different concepts. To address this, we use a dynamic distribution adjustment strategy based on evaluation results obtained during training. Additionally, we tailor the data distribution to accommodate the multi-stage training strategy.

In the sections that follow, we provide a detailed description of each component in our data processing pipeline.

\subsection{Filter Actors}
We have developed a set of filtering actors to ensure the quality of the training data. These actors are described below in details:

\paragraph{Video Quality Assessment}
We adopt DOVER~\citep{wu2023exploring} to assess the visual quality of each video clip. DOVER provides three distinct quality scores: overall score, aesthetic score, and technical score. Through empirical evaluation, we found that the technical score alone is the most effective indicator for our use case.

\paragraph{Aesthetics}
We employ the LAION aesthetic model~\citep{schuhmann2022laion} to predict aesthetic score for each image and video. Since the LAION aesthetic model is originally designed for images, we use the aesthetic score of the first frame to represent the quality of the entire video clip.

\paragraph{Overexposed and Underexposed}
Some videos suffer from overexposure or underexposure, which we have found to adversely affect training stability. To remove such data, we convert every frame of the video to the HSI color space and compute the average brightness across the entire video. Videos identified as either overexposed or underexposed, based on their average brightness, are excluded from the training set.

\paragraph{Motion Strength}
To quantify the motion strength of each video, we employ the RAFT optical flow model~\citep{teed2020raft}. To reduce computational overhead, all videos are first downsampled to 8 FPS before computing the optical flow between adjacent frames. The optical flow is calculated at the pixel level, and the overall motion strength is obtained by averaging the flow magnitudes across all pixels in the clip.

However, this approach tends to underestimate motion in cases where the background remains static while the foreground exhibits significant movement. To mitigate this issue, we additionally apply a saliency detection model~\citep{zhao2019pyramid} to each frame. The resulting saliency maps enable us to distinguish between foreground and background regions, allowing us to compute the average optical flow separately for both.

As a result, we derive three motion statistics: overall motion strength, foreground motion strength, and background motion strength. To balance data quality and training difficulty, we prioritize video clips with moderate motion strength, avoiding both overly static and excessively dynamic videos. Specifically, we define lower and upper thresholds for all three motion statistics to guide data selection.

\paragraph{Camera Movement Stability}
A significant portion of collected videos is captured with handheld devices, which often results in erratic camera movements that are challenging for the model to learn. Since such cases are not effectively filtered by motion strength alone, we estimate camera stability by evaluating the consistency of optical flow between adjacent frames, filtering out clips with unstable camera motion.

\paragraph{Slides Movement}
Slide movements, such as floating photos or banners commonly found in screen recordings or slideshow presentations, are another undesirable case. To detect these, we analyze the divergence of the optical flow across all pixels in each frame. If the divergence remains consistently low over time, the clip is identified as containing slide movements and is removed.

\paragraph{Border Detection}
We perform edge detection on each frame and apply the Hough transform to identify persistent vertical and horizontal lines across frames. These lines are treated as potential borders, and the proportion of frames containing such borders serves as a confidence score for filtering.

\paragraph{Text Detection}
We perform text detection on video frames to identify and exclude clips containing excessive textual content. Specifically, if any frame within a clip contains an overly large number of characters or if the detected text regions occupy a substantial portion of the frame, the corresponding clip is discarded.

A notable exception is subtitles, which typically consist of fewer characters and occupy relatively limited spatial regions, rendering them less likely to be filtered out by the aforementioned criteria. Nevertheless, subtitles exhibit distinctive spatiotemporal patterns: they consistently appear in fixed locations where most commonly at the top or bottom of the frame, and persist across multiple consecutive frames. By leveraging these characteristics, we are able to reliably detect and exclude video clips containing subtitles from the training data.

\paragraph{Logo Detection}
Many videos contain logos in the corners, which is an undesirable pattern for model training. To address this, we employ the Florence-2 model~\citep{xiao2024florence}, which supports open-vocabulary object detection. By providing a predefined set of keywords, Florence-2 accurately detects and localizes logos within video frames and providing confidence scores for filtering.

\paragraph{Corner Face Detection}
In commentary videos, narrators typically appear in a fixed corner of the screen, and we aim to exclude such patterns from our training data. To achieve this, we employ a face detection model, leveraging both face location and detection confidence to identify potential narrators. Specifically, we average the detection confidence of faces located in fixed corners across all frames to estimate the likelihood of a narrator's presence.

\paragraph{Transition Detection}
While PySceneDetect can segment raw videos into clips based on shot boundaries, it struggles to handle complex transitions, and as a result, the resulting clips may still contain multiple shots. To address this issue, we sparsely sample keyframes from each video and use CLIP~\citep{radford2021learning} to compute the semantic similarity between adjacent keyframes. If the similarity falls below a predefined threshold, the clip is considered to contain multiple shots and is subsequently removed.

\subsection{De-duplication}
Recent studies on large language models~\citep{lee2021deduplicating,hernandez2022scaling} have shown that even small amounts of duplicate data can significantly degrade performance. Motivated by this, we conduct rigorous de-duplication. We compute pairwise similarity scores using both CLIP~\citep{radford2021learning} and DINOv2~\citep{oquab2023dinov2}, and treat any clip exceeding the threshold in either similarity as a duplicate to be removed.

\subsection{MLLM as Advanced Filter}
After the above filtering and de-duplication processes, most of the undesired data have been effectively removed. However, due to the limitations of the current filtering actors, a small portion of low-quality data still remains. As the remaining data size has been significantly reduced and to further improve data quality, we leverage a multi-modal large language model (MLLM) to perform an additional round of filtering. This enables us to detect more complex bad cases. Notably, this step can be seamlessly integrated into the subsequent caption procedure, thereby reducing overall costs and improving efficiency.

\subsection{Caption}
\label{sec:caption}
\paragraph{Highly Descriptive Caption}
Recent advances~\citep{betker2023improving} have demonstrated that using MLLMs to generate highly descriptive captions is crucial for improving image generation quality, and we adopt this approach for captioning our data. Compared to images, videos have richer temporal information, including actions, camera movements, and scene changes. However, most mainstream MLLMs are primarily designed for images. To address this, we process each video by extracting a set of key-frames to form an image sequence. Through empirical analysis, we find that using 4 to 12 frames per video clip (depending on its duration) reaches the best trade-off between descriptive accuracy and computational efficiency.  
For video data, the captioning prompt is structured into two stages. In the first stage, the model is guided through a series of targeted questions aimed at eliciting responses on predefined attributes of the video clip (as summarized in Tab.~\ref{table:caption_attributes}). This step encourages the model to perform a structured analysis of the content. In the second stage, the model generates the final descriptive caption, which can incorporate salient observations identified in the preceding analysis of first stage. In contrast, for image data, we directly prompt the model to generate a caption without the attribute-based pre-analysis. Example captions are provided in Tab.~\ref{table:caption_examples}.

\begin{table}
    \centering
    \small
    \begin{tabular}{c|c}
    \toprule
    Attribute & Instruction  \\
    \midrule
    Scene Count & Identify the number of distinct scenes in the video. \\
    Camera Transitions & Note any noticeable transitions between shots. \\
    Camera Shot Type & Specify the type of camera shot used. \\ 
    Camera Movement & Describe any camera movements.  \\
    Main Subject Identification & Determine who or what is the central focus of the video. \\
    Subject Attributes & Describe the main subject's appearance. \\
    Subject Position & Indicate where the main subject is within the frame.  \\
    Subject Action  & Explain what the main subject is doing.  \\
    \bottomrule
    \end{tabular}
    \vspace{0.5em}
    \caption{Predefined attributes used in caption instruction.}
    \label{table:caption_attributes}
\end{table}

\begin{table}
\centering
\small
\begin{tabular}{m{3cm}|p{10cm}}
\toprule
Caption Type & Example \\
\midrule
Video Detail Caption & Medium shot of a hotel reception desk with two staff members. A woman stands on the left, and a man in a suit and red tie stands on the right. White orchids are in vases on either side of the desk. A painting hangs on the wall behind the desk. The man on the right picks up a telephone receiver and begins a phone conversation.\newline\newline The man is now more prominently featured in the frame, his upper body taking up a larger portion of the screen. The woman on the left is still visible, but less prominent. The man continues his phone conversation, his expression becoming more serious.\newline\newline The new arrival is now standing at the reception desk, slightly behind the man on the phone. The woman is still visible on the left. A man in a dark suit approaches the reception desk from the right side of the frame.\\
\midrule
Image Detail Caption & A young woman with long dark hair stands on a rocky beach. She is wearing a light beige, strapless top and matching wide-legged pants. Her arms are crossed, and her hands are near her chest. She is barefoot. The rocks are various shades of brown and tan, some smooth and some rough. The rocks are wet in places. The ocean is visible in the background. The sky is light blue and mostly clear. A small child is partially visible in the lower left corner of the frame, seemingly playing near the water's edge. The woman is positioned in the center of the frame, slightly off-center towards the right. She is facing the camera directly. The rocks behind her are large and form a backdrop to her figure. The rocks in the foreground are smaller and scattered around her feet. The child is in the lower left corner, facing towards the center of the frame, and is partially obscured by rocks. The ocean is in the far background, above the rocks, and extends across the entire width of the frame. The sky is visible above the rocks and the ocean, occupying the upper portion of the frame. The lighting is natural, with sunlight illuminating the scene. The overall composition is balanced, with the woman as the focal point, surrounded by the natural elements of the beach. \\
\midrule
AR Caption & \textbf{1st second:} A woman holds a lipstick tube, her expression changes subtly. The background is a simple, light brown wooden wall. The woman in the frames is wearing a beige lace sleeveless top and gold necklaces. She holds a gold lipstick tube in her right hand. Her makeup is subtle, and her expression changes slightly throughout the two frames. Her hair is dark brown and styled in a shoulder-length cut. The lighting is soft and even, creating a neutral mood. There are no other objects or people visible in the frames. \newline\newline \textbf{2nd second:} The woman's head tilts slightly, her expression shifts from a neutral to a slight smile. The lipstick remains in her hand. The camera remains static, focusing on the woman. \newline\newline\textbf{3rd second:} The woman's head is slightly turned to the left, her expression is more serious. The lipstick is still in her hand. The camera remains static, focusing on the woman. \newline\newline\textbf{4th second:} The woman's head is turned slightly to the right, her expression is neutral. The lipstick is still in her hand. The camera remains static, focusing on the woman. \\
\bottomrule
\end{tabular}
\vspace{0.5em}
\caption{Caption examples used in \magi.}
\label{table:caption_examples}
\end{table}

\paragraph{Auto-Regressive Caption}
Unlike typical bi-directional denoising video generation models that produce an entire video as a whole, our model generates videos in an auto-regressive manner. This design allows our model to condition different parts of the video on distinct text prompts, offering greater controllability.  
To enable this capability, we provide fine-grained, second-by-second descriptions for each video clip. Tab.~\ref{table:caption_examples} shows example. Specifically, the caption of the first second is instruct to generates a detailed description. For caption of subsequent seconds, they focus on describing changes relative to the previous one. 

\begin{table}
    \centering
    \small
    \begin{tabular}{c|c|c|c}
    \toprule
    & stage-1 & stage-2 & stage-3 \\
    \midrule
    Resolutions & 256p/360p & 480p & 720p\\
    Video Duration & $\leq8$s & $\leq8$s & $\leq16$s \\
    Image-Video Ratio & 4:1 & 4:1 & 4:1 \\
    AR Caption Ratio & 0\% & 10\% & 10\% \\
    \bottomrule
    \end{tabular}
    \vspace{0.5em}
    \caption{Data configuration of different stages.}
    \label{table:data_configuration}
\end{table}

\subsection{Data Adjustment in Training}
\label{sec:data_adjustment}
We have two different data adjustment scenarios during training. First, we use a multi-stage training strategy, with later stages having higher data quality; Second, we dynamically adjust the data distribution during training based on the evaluation results. 

\paragraph{Multi-stage Adjustment}
\magi is trained in three stages, with the data resolution gradually increasing from 256p to 480p and ultimately to 720p. Alongside the resolution improvements, the data volume is progressively reduced, and more rigorous filtering strategies are employed to ensure higher data quality. Furthermore, in the final stage, the video duration is extended from a maximum of 8 seconds to a maximum of 16 seconds, allowing the model to capture richer temporal dynamics. The data specifications for each stage are summarized in Tab.~\ref{table:data_configuration}.

\paragraph{Dynamic Distribution Adjustment}
An appropriate data distribution is crucial for training high-performance models. However, identifying the optimal distribution in advance is challenging. For instance, during training, we observed that landscape scenes are relatively easy for the model to learn, while human expressions are significantly more difficult. These insights are hard to predict beforehand. To address this, we adopt a dynamic distribution adjustment strategy. By continuously monitoring model performance throughout the training process, we can adaptively adjust the proportion of specific data subsets to strengthen the underperforming aspects of the model, thereby enabling a more effective learning process.

\section{Infrastructure}
In this section, we introduce our training infrastructure and inference infrastructure.  
\subsection{Training Infrastructure}

Efficient training of large-scale autoregressive denoising models like \magi requires carefully tailored distributed training infrastructure. Existing distributed training frameworks, such as Megatron~\citep{shoeybi2020megatronlm} and DeepSpeed~\citep{rajbhandari2020zeromemoryoptimizationstraining} are primarily designed for large language models (LLMs). However, \magi differs significantly from LLMs in both algorithmic side and data side.

On the algorithmic side, \magi integrates both autoregressive and denoising modeling paradigms, resulting in a model architecture that is notably more complex than that of typical LLMs. It incorporates components such as gating, cross-attention, and block-causal attention that are rarely used in language models.

On the data side, a single video training example typically contains tens to hundreds of times more tokens than a text example. Furthermore, ensuring the temporal and semantic integrity of video content imposes strict constraints, making it infeasible to directly apply common data processing strategies from LLMs, such as arbitrary sequence truncation or concatenation of multiple samples into a single training sequence offline, in the context of video generation.

These fundamental differences introduce unique challenges, necessitating a new, purpose-built distributed system design. In this section, we propose novel solutions to address these challenges to enable efficient and scalable training of \magi.

Specifically, the training of \magi leverages a combination of data parallelism (DP), context parallelism (CP), and tensor parallelism (TP). To address the DP load imbalance caused by variable-length video sequence and the insufficient GPU utilization on short token sequences, we introduce a distributed Packing and Padding (PnP) during training, that performs online batching of video data in each training iteration. This strategy mitigates GPU bubbles thereby significantly improving overall training efficiency (Sec.~\ref{sec:train_infra_pnp}).

Due to the use of PnP and the inherent demands of block-causal attention in \magi, we require an efficient attention implementation capable of supporting highly flexible attention masks. Additionally, given the extremely long token sequences typical in video training data, native support for context parallelism is essential. To address these requirements, we propose MagiAttention: a scalable distributed attention mechanism that efficiently handles diverse attention masks and is optimized for ultra-long sequences (Sec.~\ref{sec:magiattn}).

Through the above innovations, we enable the efficient training of \magi. However, while developing the \magi training system, we identified several limitations in existing large-scale training frameworks. For instance, most current frameworks (including ours) do not treat verifiable numerical accuracy in distributed environments as a first-class design concern. Moreover, the tight coupling between algorithm development and infrastructure implementation often creates friction between algorithm researchers and infrastructure engineers, hindering efficient collaboration. To address these challenges, we discuss potential directions and design principles for next-generation training infrastructure in Sec.~\ref{sec:train_infra_odin}, with the goal of providing insights and practical guidance for the broader research and engineering community.

\subsubsection{Distributed Packing and Padding}\label{sec:train_infra_pnp}

Due to the integrity constraints of video data and the variability in video lengths and resolutions, we adopt a Packing and Padding (PnP) strategy~\citep{sirluk2024efficient, kundu2024enhancingtrainingefficiencyusing} to batch video samples in a way that minimizes excessive padding and reduces unnecessary computational overhead in distributed training scenarios. Moreover, the data composition is frequently adjusted during the training of \magi (See Sec.~\ref{sec:data_adjustment}), and to accommodate such flexibility, we employ an online PnP strategy instead of a offline approach.

The core idea of PnP is to efficiently utilize GPU resources by concatenating multiple short sequences into a batch while minimizing redundant filling. The offline formulation of this problem aligns with the classic bin-packing problem: given a set of input samples, the goal is to pack them into a set of bins, each with a fixed capacity \texttt{max\_length}, while minimizing overall unused space. Although this problem is NP-complete, it can be efficiently approximated in practice using the First-Fit Decreasing (FFD)~\citep{dosa2007tight} greedy algorithm.

In our online setting, we must process streaming data inputs while ensuring compatibility with the 3D parallelism strategy employed during training. To this end, we reformulate the problem as follows: given M candidate samples, we aim to pack them into N bins of size \texttt{max\_length}, minimizing overall space waste. Here, M denotes the size of the candidate pool with M $\gg$ N; N must be divisible by the \texttt{DP\_SIZE}; and \texttt{max\_length} must be divisible by \texttt{TP\_SIZE}$\times$\texttt{CP\_SIZE}.

In practice, we extend the FFD algorithm with custom heuristics to support efficient online packing under these constraints. This approach enables us to achieve a $99\%$ capacity utilization rate and the differences between different DP groups can be neglected, thus substantially reducing computational overhead during training.

\subsubsection{MagiAttention: Towards Linear Scalability for Ultra-Long and Heterogeneous Mask Training.}\label{sec:magiattn}

\begin{figure}[!htbp]
    \centering
    \includegraphics[width=\linewidth]{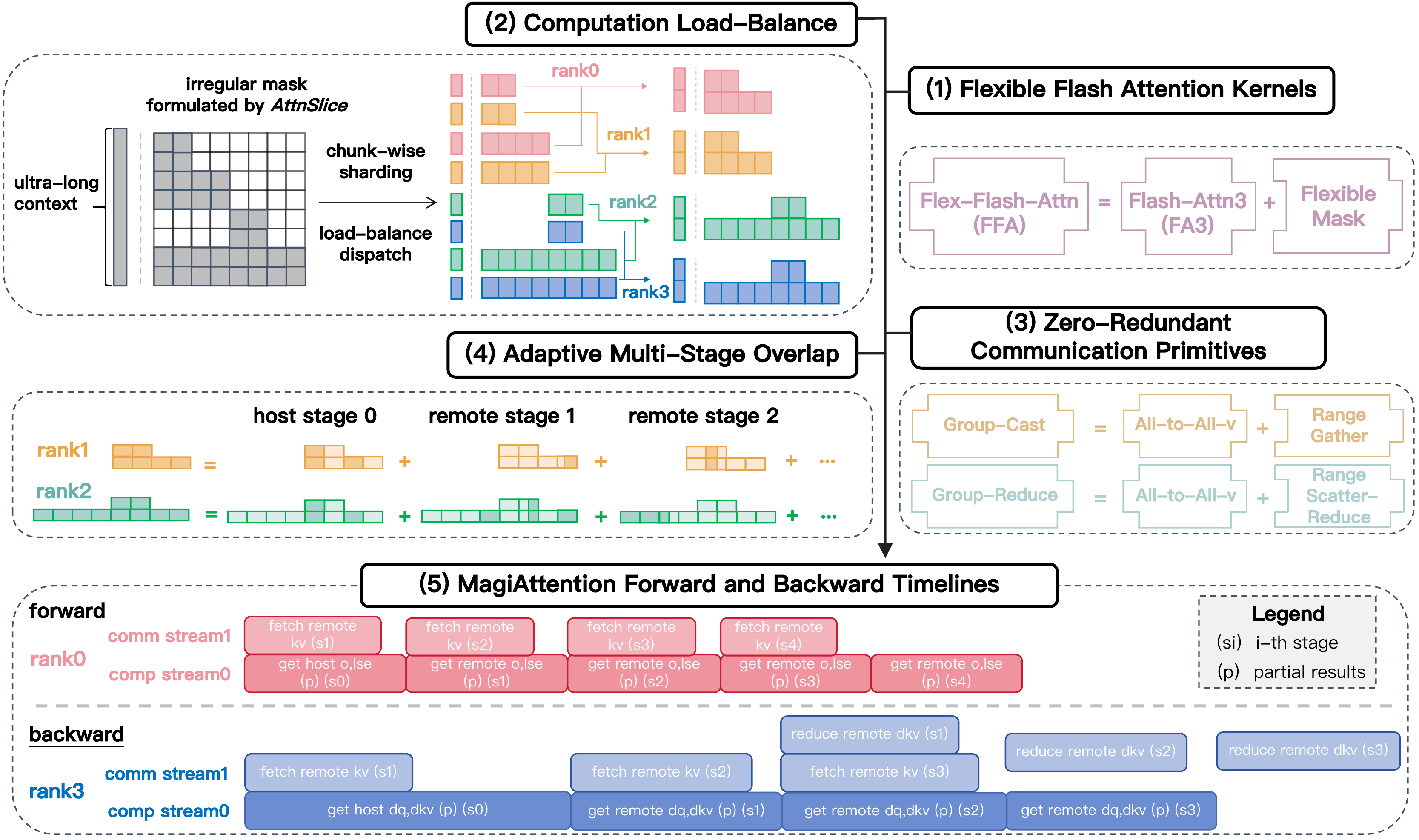}
    \caption{Overview of MagiAttention: (1) Flex-Flash-Attention(FFA), an efficient attention supports flexible mask patterns and native considers distribution requirements; (2) The \textit{dispatch solver} shards and dispatches packed data with ultra-long contexts and heterogeneous masks, ensuring load-balanced computation; (3) Group-Cast and Group-Reduce primitives eliminate redundant communication; (4) The \textit{adaptive multi-stage overlap} strategy effectively hides communication latency; (5) Forward and backward timelines of MagiAttention. With all techniques together, MagiAttention reach linear scalability under diverse scenarios.}

    \label{fig:magiattn_overview}
\end{figure}

Training large-scale autoregressive diffusion models like \magi for video generation presents two major challenges: 

\begin{itemize}
    \item The extremely long context length of video tokens, which reaching up to 4 million during training, results in prohibitive computational and memory overhead. Context-Parallelism (CP) is designed for dealing such long context challenge, but existing state-of-the-art CP methods~\citep{jacobs2023deepspeed, liu2023ringattentionblockwisetransformers, fang2024uspunifiedsequenceparallelism, gu2024loongtrainefficienttraininglongsequence, chen2024longvilascalinglongcontextvisual} face scalability limitations that face scalability limitations due to size constraints or the high communication overhead inherent in inefficient ring-style point-to-point (P2P) patterns. While recent efforts~\citep{wang2024datacentricheterogeneityadaptivesequenceparallelism, zhang2024dcp, ge2025bytescaleefficientscalingllm} dynamically adjust CP sizes to avoid unnecessary sharding and redundant communication for shorter sequences, they still incur extra memory overhead for multiple NCCL process groups and involve complex scheduling to balance loads and synchronize across different subsets of ranks.
    \item The combination of block-causal attention and Packing-and-Padding introduces highly complex attention mask patterns (Sec.\ref{sec:train_infra_pnp}), which cannot be efficiently handled by existing attention implementations.
\end{itemize}

To address the aforementioned challenges, we propose MagiAttention, which aims to support a wide variety of attention mask types (\emph{i.e.}, kernel flexibility) while achieving linear scalability with respect to context-parallel (CP) size across a broad range of scenarios. Achieving this goal depends on meeting the following fundamental conditions:

\begin{itemize}
    \item \emph{Linearly Scalable Attention Kernel}: The performance of the attention kernel should not degradate as CP size increases. To this end, we introduce \emph{Flex-Flash-Attention}, an extension of FlashAttention-3 (FA3), which native considers the efficiency impact of attention mask partitioning in distributed environments. It supports distributable mask representations with a tailored kernel implementation to ensure scalability while accommodating a broader range of attention mask types.
    \item \emph{Balanced Computational Workloads}: Imbalances in the computational load across CP ranks lead to unavoidable idle bubbles that hinder scalability. MagiAttention is natively designed to ensure \emph{Computation Load Balancing}, mitigating such inefficiencies.
    \item \emph{Full Overlap of Communication and Computation}: Without sufficient overlap, increasing CP size results in communication-induced idle time on GPUs, impairing scalability. MagiAttention introduces novel \emph{Zero-Redundant Communication Primitives} to minimize communication overhead, along with an \emph{Adaptive Multi-Stage Overlap} strategy that enables effective communication-computation overlap.
\end{itemize}

The overview of MagiAttention is shown in Fig.~\ref{fig:magiattn_overview}, and we will introduce key designs in the following, with comprehensive experimental results presented in Appendix~\ref{appendix:magiattn_exps}.

\paragraph{Flex-Flash-Attention.}\label{sec:magiattn_ffa}
\begin{figure}[htbp]
    \centering
    \includegraphics[width=\linewidth]{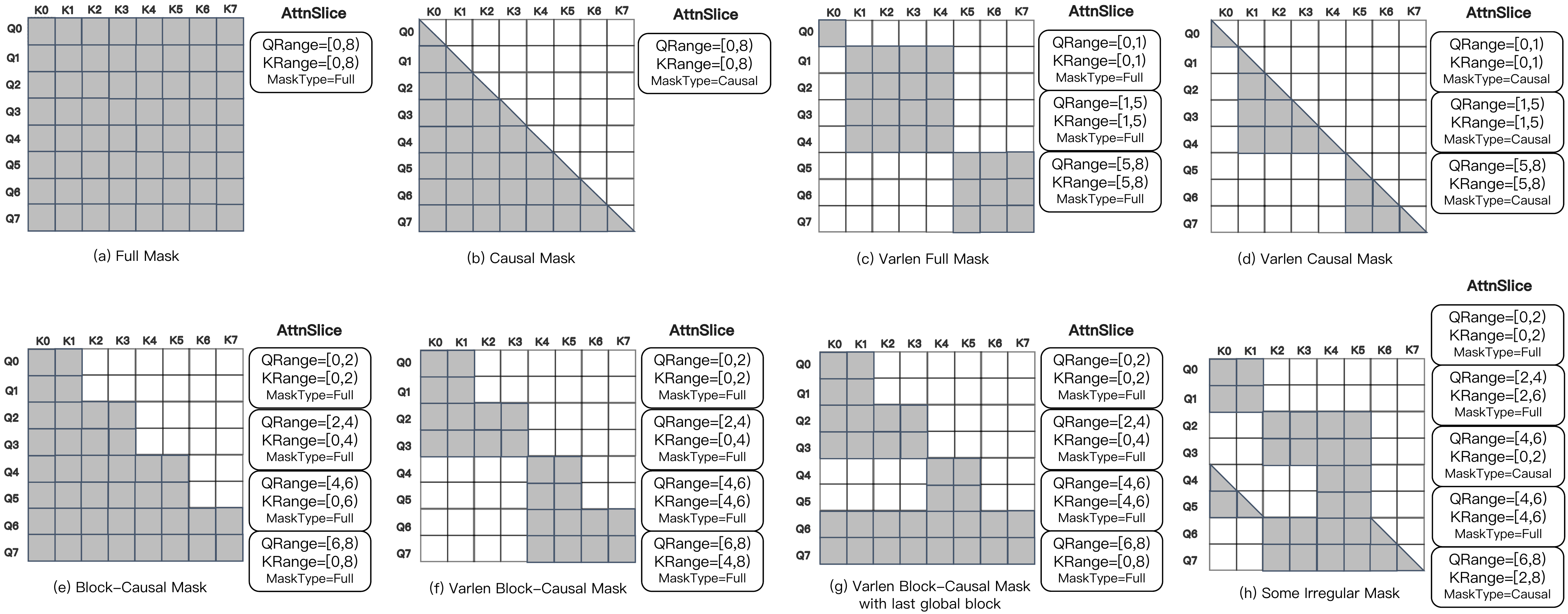}
    \caption{Examples of mask patterns formulated by \texttt{AttnSlice}. (a)-(d) Standard FA3-compatible patterns; (e)-(h) Irregular masks beyond FA3's capabilities, including our novel \textit{varlen block-causal} design, which FFA supports seamlessly while maintaining performance comparable to FA3.}
    \label{fig:mask_with_attn_slice}
\end{figure}

FlashAttention~\citep{dao2022flashattention, dao2023flashattention, shah2024flashattention3fastaccurateattention} is a foundational technique in large-scale model training for its superior performance and support for varlen-packed data with causal attention masks. However, it offers limited support for irregular attention masks, particularly when such patterns are distributed across CP ranks, resulting in increased complexity and underscoring the need for a more flexible attention kernel~\citep{pytorch_sdpa, dong2024flexattentionprogrammingmodel,wang2025flashmaskefficientrichmask} without compromising performance.

Therefore, we introduce Flex-Flash-Attention (FFA), which is natively designed for distribution scenarios and provides greater flexibility in handling diverse attention mask types. The core idea behind FFA is to generalize a distributable formulation for irregular attention masks by decomposing the entire mask into multiple computational units, each referred to as an \texttt{AttnSlice}. Each \texttt{AttnSlice} is defined by a triplet \texttt{\(QRange, KRange, MaskType\)}, which specifies a submask with a basic shape bounded by a contiguous 2D query-key region (see Fig.~\ref{fig:attnslice_interpret}). Using this formulation, a wide variety of commonly used attention masks (Fig.~\ref{fig:mask_with_attn_slice}) (including our varlen block-causal mask) can be expressed as a composition of multiple such triplets, making FFA highly suitable for distributed attention computation.

Built on FA3 kernels, Flex-Flash-Attention leverages NVIDIA Hopper GPUs' TMA feature~\citep{nvidia2024accelerating} and introduces slice-level parallelism with atomic operations for correctness (Fig~\ref{fig:ffa_slice_parallel}), achieving comparable MFU to FA3 while supporting the flexible \texttt{AttnSlice} formulation~\footnote{Redundant computation from padding tokens is excluded by easily passing empty \texttt{QRange} or \texttt{KRange}.} (see Appendix~\ref{appendix:magiattn_exps} for benchmarks).

\paragraph{Computation Load-Balance.}\label{sec:magiattn_comp}

In context-parallelism (CP) settings, different CP ranks may be assigned heterogeneous attention masks, resulting in imbalanced computational workloads across ranks. Ring-Attention~\citep{ring_flash_attention_issue2} employs a specialized partitioning strategy designed specifically for causal attention, which limits its applicability to more general attention patterns. To overcome this limitation, we propose a generic and efficient \texttt{dispatch solver} that enables balanced workload distribution across CP ranks for a broad range of attention types.

First, to enable finer-grained control, we propose a chunk-wise permutable sharding strategy (Fig~\ref{fig:magiattn_overview} (2)). Specifically, the entire mask is evenly partitioned along the query-dimension into dispatch chunks, each associated with a submask area: $\{(C_i, \mathrm{Area}(C_i))\}_{i=1}^n$, where $C_i$ indicates i-th dispatch chunk, $\mathrm{Area}(C_i)$ is the mask area of $C_i$, $n$ is $\frac{seqlen}{dispatch\_chunk\_size}$, and $dispatch\_chunk\_size$ is a hyperparameter controlling granularity. 
These dispatch chunks are then equally assigned to $cp\_size$ buckets, with each bucket containing the exact same number of dispatch chunks to ensure token-level load balance in non-attention modules, attaching with a summed submask area, denoted as $\{(B_j, \mathrm{SumArea}(B_j))\}_{j=1}^{cp\_size}$.

With above strategy, we could fine-grained control the computational workloads of each CP rank, and the load-balancing dispatch becomes a combinatorial optimization problem, defined as finding an optimal mapping function $f^*: \{C_i\}_{i=1}^n \rightarrow \{B_j\}_{j=1}^{cp\_size}$ as follows

\begin{align}
    &f^* = \arg \min\limits_{f}\max\limits_{j}\left\{\mathrm{SumArea}(B_j)\right\} \label{eq:comp_load_balance}\\
    &\text{s.t.}\;\;|B_j| = \frac{n}{cp\_size}, \;\; seqlen \;\%\; (cp\_size \times dispatch\_chunk\_size) = 0\nonumber
\end{align}

However, this optimization is a known NP-hard problem, making it impractical to find an optimal solution on-the-fly during each training iteration, especially given the varying mask patterns across micro-batches. Thus, we propose an efficient greedy algorithm (as shown in Alg.~\ref{alg:minhp}) that provides a suboptimal yet effective solution within $O(n\log n)$ complexity.

\paragraph{Zero-Redundant Communication Primitives.}\label{sec:magiattn_comm}

The existing ring-style implementation uses point-to-point \texttt{send}/\texttt{recv} communication primitives, which cannot provide sufficient communication granularity, resulting in redundant communication. Take causal mask as an example, we analyze the redundant communication by recording the distribution of remote key-value ($\mathrm{KV}$) requests and their gradients ($\mathrm{dKV}$) under sparse attention masks. As shown in Fig~\ref{fig:ring-p2p-redundancy}, $\mathrm{KV}_0$ is required by all queries and should be sent to all devices via Broad-Cast in the forward pass, with $\mathrm{dKV}_0$ reduced via All-Reduce in the backward pass. In contrast, $\mathrm{KV}_7$ is only needed by its host device but still circulates through all devices, and this redundancy intensifies in varlen scenarios.

To address this, we introduce two communication primitives: \texttt{group-cast} and \texttt{group-reduce}, which model the communication patterns of low-demand $\mathrm{KV}$ and $\mathrm{dKV}$ (Fig~\ref{fig:group-gather-reduce-all2allv}). For example, in the causal mask, $\mathrm{KV}_5$ on $\mathrm{rank}_2$ is required only by $\{\mathrm{Q}_6,\mathrm{Q}_7\}$ and should be sent exclusively to the target ranks $\{\mathrm{rank}_0, \mathrm{rank}_1\}$ via \texttt{group-cast}, while the partial $\mathrm{dKV}_5$ is collected and reduced back to $\mathrm{rank}_2$ via \texttt{group-reduce} accordingly.

As no existing communication kernels support these primitives, we prototype them using \texttt{all-to-all-v} (Fig~\ref{fig:group-gather-reduce-all2allv}), achieving zero-redundant communication in both forward and backward passes. However, this approach introduces extra pre-/post-processing overhead, similar to (un)permutation in expert parallelism (EP)~\citep{gale2022megablocks}. While kernel fusion mitigates the overhead, a dedicated implementation of \texttt{group-cast} and \texttt{group-reduce} remains a key direction for future work.

\paragraph{Adaptive Multi-Stage Overlap.}\label{sec:magiattn_overlap}

Leveraging previous optimizations, we achieve high-performance computation through an efficient kernel and balanced workload dispatch, while minimizing communication overhead with our new primitives. To drive true linear scalability, we further improve end-to-end performance by introducing a multi-stage compute-communication overlap strategy, that effectively hides communication latency and adaptively optimizes overlap through manual or automatic tuning.

Similar to prior works~\citep{liu2023ringattentionblockwisetransformers,zhao2023pytorch,async_tensor_parallelism_in_pytorch}, we schedule pipeline stages to overlap computation with communication for both forward and backward passes (Fig~\ref{fig:multi_stage_overlap_fwd_bwd}). Each $\mathrm{rank}_i$ first partitions its remote $\mathrm{KV}$/$\mathrm{dKV}$ communication into stages. 
In the forward pass, the scheduler first launches the \texttt{group-cast} kernel to prefetch the next remote $\mathrm{KV}$, then asynchronously executes the FFA kernel for partial attention computation, hiding all communication behind computation~\footnote{To prevent all SMs from being occupied by the attention kernel, we ensure the communication kernel picked first by setting \texttt{CUDA\_DEVICE\_MAX\_CONNECTIONS}=1~\citep{cuda_device_max_connections_issue}.}. 
In the backward pass, besides prefetching the next $\mathrm{KV}$, the \texttt{group-reduce} kernel reduces the last $\mathrm{dKV}$ in a separate CUDA stream before launching the FFA kernel for the current stage, ensuring communication is overlapped across all stages except the final $\mathrm{dKV}$ reduction~\footnote{Due to PyTorch's one-to-one mapping for process groups and collective communication streams including \texttt{all-to-all-v}~\citep{collectives_nccl_stream_issue}, we internally use an additional CP group for \texttt{group-reduce} to enable full overlap between communication kernels in the backward pass.}.

To adaptively control overlap granularity, we further introduce a tunable hyperparameter, $num\_stages$, accounting for varying compute-to-communication ratios across training setups, microbatches, or between forward and backward passes. This parameter can be manually configured or automatically determined by our \textit{overlap solver}, with a simple dynamic search algorithm (See Alg.~\ref{alg:dynamic-overlap-search} for more details).

\subsubsection{Rethinking System Design for Robust Distributed Training Frameworks with DTensor}\label{sec:train_infra_odin}

As large-scale models continue to evolve, the growing  complexity of training procedures has exposed fundamental limitations in existing distributed training frameworks~\citep{shoeybi2020megatronlm, rajbhandari2020zeromemoryoptimizationstraining}. Two major bottlenecks are particularly prominent:
\begin{itemize}
    \item Lack of testability by design. Most frameworks were not initially built with testability as a first-class feature, resulting in fragile infrastructure with limited maintainability and reliability;
    \item Tight coupling between model implementation and parallelization strategy. This entanglement prevents algorithm researchers and system engineers from working independently, hindering collaboration and modular development
\end{itemize}
We argue that next-generation distributed training frameworks must directly address these two pain points to support large-scale model research and deployment.

Inspired by early explorations~\citep{xu2021gspmdgeneralscalableparallelization, yuan2022oneflowredesigndistributeddeep} and PyTorch’s pioneering implementations~\citep{zhao2023pytorch, pytorchdtensor2024, liang2024torchtitanonestoppytorchnative}, we propose a blueprint for redesigning robust distributed training frameworks based on Pytorch Distributed Tensor (DTensor)~\citep{pytorchdtensor2024} and Parallel Plan:

\paragraph{DTensor} PyTorch DTensor introduces three parallel placements: \texttt{Replicated}, \texttt{Shard}, and \texttt{Partial}, alongside a distributed initialization strategy to maintain placement semantics~\citep{pytorch_offset_rngtracker_2025}, and a propagation mechanism that deduces output placements from input ones for supported ops, triggering communication as needed~\footnote{In practice, DTensor selects communication patterns based on estimated redistribution cost, but these estimates are often inaccurate.}~\citep{pytorch_sharding_prop_2025}. While it supports basic ops including naive distributed matmul, its current implementations lack the generality to handle more complex yet commonly scenarios in modern training workflows, as shown in Tab.~\ref{tab:dtensor_placements}.

\paragraph{Parallel Plan} 
Parallel Plan provides a declarative interface for specifying parallelization strategies across model submodules. It works in conjunction with the \texttt{parallelize\_module} function and is built on top of DTensor. However, its current capabilities are mostly limited to tensor parallelism (TP) and do not generalize well to other parallelism.

In our architecture design, we extend both DTensor and Parallel Plan to support a broader range of usages. These extensions enable the following key features:

\paragraph{Decoupling Modeling from Parallelization.} This feature allows model researchers to concentrate on model design and algorithm development without needing to manage low-level parallelism details. At the same time, infrastructure engineers can independently optimize parallelization strategies without modifying model implementation. This clear separation of concerns enables more efficient collaboration and improved training throughput.

\paragraph{High-Precision Alignment with Non-Distributed Oracles.} 
By disabling all parallel plans, we can seamlessly revert to non-distributed configurations, yielding "pure" model code that serves as a baseline or oracle for evaluating distributed correctness. To ensure alignment within a relative error of $10^{-8}$, we upcast tensors to higher precision~\footnote{In our experiments, \texttt{float32} is insufficient for fully alignment; \texttt{float64} suffices in most cases.}, enforce deterministic algorithms~\citep{pytorch2024reproducibility}, and control randomness using consistent seed management. This design enables precise infrastructure testing, ultimately improving reliability and debuggability.

\subsection{Inference Infrastructure}
As an innovative large-scale autoregressive denoising model, \magi introduces two pivotal architectural innovations: multi-chunk parallel inference and KV cache, which unlock new possibilities for user experiences, such as real-time streaming video generation, and enables cost-effective deployment. However, these advancements also introduce new challenges to the inference infrastructure. In this section, we present our infrastructure design tailored to two major scenarios: real-time streaming inference on H100/H800, and cost-efficient deployment on RTX 4090 GPU.

\subsubsection{Real-Time Streaming Video Generation}
\label{sec:real_time_infra}
Our model adopts an auto-regressive architecture that supports real-time streaming video generation. To ensure a seamless user experience, we optimize for two key latency metrics: Time to First Chunk (TTFC), which measures the delay between task submission and starting to see the video, and Time Per Output Chunk (TPOC), which reflects the time required to generate each subsequent chunk. Maintaining a low TTFC enhances responsiveness, while keeping TPOC below 1 second is essential for uninterrupted playback.

We encountered three major challenges when designing the infrastructure:

\begin{itemize} \item \magi consists of multiple sub-models: T5 for text embedding extraction, a VAE encoder for processing user-uploaded images and prefix videos, a VAE decoder for decode the denoised output, and a core auto-regressive denoising model. These components exhibit distinct computational characteristics: T5 and VAE are memory-bound, while the denoising model is compute-bound. Efficiently handling this heterogeneity is essential.

\item To meet the TPOC target of under 1 second, \magi demands approximately 9 PFLOPS of compute per second of video, which far exceeds the capabilities of a single H100/H800 GPU. Achieving this requires serving models on multiple H100/H800 GPUs and a highly optimized parallelism strategy.

\item First-chunk inference differs significantly from subsequent chunks. It is not compute-bound but CPU-bound, due to limited token workloads per GPU, resulting in a long TTFC. 
\end{itemize}

To address these challenges, we propose a systematically optimized framework, enabling real-time streaming video generation for our largest 24B \magi model on 3-node, 24 H100 GPUs. Here, we briefly introduce our solutions.

\paragraph{Multi-Model Heterogeneous Serving Pipeline}
We designed a heterogeneous serving architecture that co-locates T5 and \magi on high-performance GPUs, while deploying the VAE to cost-efficient hardware. This approach enables concurrent execution of \magi inference and VAE decoding, minimizing idle time and improving overall throughput. Profiling-driven resource allocation strategies further enhance utilization efficiency. With this design, we could efficiently handling the heterogeneity of different models and achieve the best performance. 

\paragraph{TPOC Optimization}
Given that the denoising model of \magi is compute-bound, we prioritized aggressive quantization and distributed inference optimizations:

\begin{itemize} \item \emph{Quantization.} We adopted W8A8 SmoothQuant~\cite{xiao2023smoothquant} to quantize both weights and activations to \texttt{FP8} precision, except the first and last layers. The quantization delivered a $30\%$ speedup without compromising generation quality. 

\item \emph{Multi-Node Parallel Inference.} We adopt a Ulysses-based multi-node parallel inference strategy with sufficiently computation and communication overlapping (less than $3\%$ of communication time remaining unoverlapped in the execution timeline). As a result, the TPOC is optimized to be within 1 second when we generating 480p (3:4 aspect ratio) videos using 16 denoising steps and KV range of 5 on 24 H100/H800 GPUs.
\end{itemize}

\paragraph{TTFC Optimization}

For first-chunk inference, only a few hundred tokens need to be processed. In this scenario, the GPU workload is relatively light, and CPU-side bottlenecks become the primary constraint. To address this issue, we employ CUDA Graphs to minimize kernel launch overhead, reducing 30.4\% latency. Additionally, we accelerate VAE decoding through a tile-based parallel mechanism and \texttt{torch.compile}, bringing latency down from 1 second to around 70 milliseconds. Collectively, these optimizations reduced TTFC to 2.3 seconds, ensuring a smooth real-time streaming experience. Tab. ~\ref{tab:opt_and_gain} summarizes the key optimizations and their corresponding latency gains\footnote{While TPOC latency is expected to be the sum of autoregressive diffusion model and VAE decoder latencies—similar to TTFC—in our serving pipeline where autoregressive diffusion model and VAE run on separate machines, TPOC is instead computed as the maximum of the two.}.

\begin{table}[h]
    \centering

    \begin{tabular}{l|l|c|c|c|c}
        \toprule
        \textbf{Model} & \textbf{Optimization} & \textbf{TTFC(s)} & \textbf{Gain} & \textbf{TPOC(s)} & \textbf{Gain} \\
        \midrule
        \multirow{5}{*}{\makecell[l]{Autoregressive \\ Denoising Model}} 
            & Baseline     & 73.34 & -      & 45.49 & -      \\
            & KV Cache     & 73.34 & -      & 23.94 & 1.90X  \\
            & Ulysses      & 3.86  & 18.0X  & 1.26  & 18.0X  \\
            & Smooth Quant & 3.00  & 1.29X  & 0.98  & 1.29X  \\
            & Cuda Graph   & 2.30  & 1.30X  & 0.98  & -      \\
        \midrule
        \multirow{3}{*}{\makecell[l]{Vae Decoder}}  
            & Baseline      & 1.00  & -     & 1.00  & -     \\
            & Tile Parallel & 0.20  & 5.00X & 0.20  & 5.00X \\
            & torch.compile & 0.07  & 2.86X & 0.07  & 2.86X \\
        \midrule
        End-to-End & - & 2.37 & - & 0.98 & - \\
        \bottomrule
    \end{tabular}
    \caption{Inference Optimization and Latency Gain}
    \label{tab:opt_and_gain}
\end{table}

\subsubsection{Cost-effective Inference on RTX 4090} 
The NVIDIA GeForce RTX 4090 is a highly cost-effective GPU with 24G memory. However, through in-depth memory profiling and analysis, we identified memory insufficiency as the primary bottleneck to serve our model on it. To address this challenge, we developed a highly memory-efficient inference architecture and performs systematically optimizations. As a result, we successfully deployed and ran our 4.5B-parameter model on a single RTX 4090 GPU, and also support our largest 24B model on an 8$\times$RTX 4090 GPUs. In the following section, we briefly introduce the key optimization techniques.

\paragraph{Memory Optimization} 
To address the memory constraints of the RTX 4090, we used a variety of techniques to do systematically optimization:
 
\begin{itemize}
\item \emph{Quantization:} We adopt the same quantization strategy (WA8A SmoothQuant) as for streaming video generation.
\item \emph{KV-offload:} KV-offload is a technique that stores the KV cache in CPU memory by default and dynamically re-load it back to the GPU as needed. This approach significantly reduces peak GPU memory usage and is widely adopted in long-sequence processing for large language models (LLMs). In \magi, we also adopt this technique to effectively address memory constraints.

\item \emph{Hybrid Parallelism and Communication Optimization:} The above two optimizations only enable 4.5B model deployment on a single RTX 4090 GPU. However, the largest 24B model further requires multi-GPU parallelism. Unlike the streaming setting where we primarily adopt a Ulysses-based context-parallelism (CP) approach, deployment on RTX 4090 employs a hybrid strategy combining pipeline-parallelism (PP) and context-parallelism.

Specifically, pipeline-parallelism is used to partition model weights, while context-parallelism is used to partition activations. However, since the RTX 4090 utilizes PCIe for inter-GPU communication, both PP and CP suffer from communication-induced bubbles that degrade compute utilization, as measured by Model FLOPs Utilization (MFU). For PP, we mitigate this by interleaving tasks to overlap GPU idle. For context-parallelism, we initially adopted the Ulysses approach, but found that communication could not be fully overlapped with computation under PCIe constraints.

Therefore, we propose an enhancement to Ulysses called Context Shuffle Overlap (CSO)(Details in Sec.~\ref{sec:appendix_context_shuffle_overlap}), which scatters each chunk evenly across all GPUs, enabling finer-grained overlap between computation and communication than plain Ulysses. This strategy significantly improves MFU under the limited interconnect bandwidth of the RTX 4090. 
\end{itemize}

With the above optimizations, we constrained peak memory usage to 19.07GB for the 4.5B model on a single RTX 4090 GPU, and 19.29 GB for the 24B model on 8$\times$RTX 4090 GPUs. For the 24B model, the maximum MFU reached 66\%.

\section{Evaluation}
\newcommand{\bbf}[1]{\textbf{ #1}}
\newcommand{\abf}[1]{#1}
\newcommand{\und}[1]{\underline{#1}}
\newcommand{\addword}[1]{\textcolor{blue}{#1}}
Evaluation methods for video generation models in the research community are typically categorized into two complementary types: the first focuses on the perceptual quality of the generated videos, while the second evaluates the model’s ability to faithfully capture underlying physics, which is often regarded as essential for modeling a realistic world. In \magi, we adopt both evaluation types to obtain a comprehensive understanding of the model’s strengths and limitations.

For perceptual quality evaluation, the inherently subjective nature of human preference, combined with the high-dimensional and diverse characteristics of video content (\emph{e.g.}, motion continuity, aesthetic, and identity consistency), makes it challenging to rely solely on objective metrics. As a result, the community typically employs a hybrid evaluation protocol that integrates human subjective assessments with standardized automated metrics, ensuring a more robust and comprehensive evaluation. 

There is currently no universally accepted human evaluation protocol or human evaluation platform within the community for perceptual quality evaluation. To address this, we design our own in-house evaluation benchmark based on a comprehensive review of existing human evaluation methodologies, combined with our understanding of both evaluation criteria and model capabilities. Human experts serve as evaluator in this system, comparing our model against other competitors under strict double-blind conditions, and providing assessments across multiple perceptual dimensions.
For objective evaluation, we adopt VBench~\citep{huang2024vbenchcomprehensiveversatilebenchmark}, which is currently the most widely used benchmark in the community. VBench consists of two evaluation tracks: text-to-video (T2V) and image-to-video (I2V). We primarily focus on the I2V track, as it more closely reflects real-world usage patterns: users typically generate videos from images rather than from text. For the same reason, we also allocate a larger proportion of I2V tasks during the training of \magi, aiming to better align the model's capabilities with practical deployment scenarios.

Physics-IQ~\citep{physicsiqbenchmark} is one of the most representative benchmarks for evaluating a model’s ability to capture physical dynamics in video. It presents a short video clip depicting real-world physical motion and asks the model to predict future frames. The predictions are then compared against ground-truth sequences to assess the model’s understanding of physical rules.

The evaluation framework and the corresponding benchmark metrics are summarized in Tab.~\ref{tab:evaluation_framework}. The following sections present our evaluations in detail, and if not specified, we evaluate our 24B model by default.

\begin{table}[h]
    \centering
    \begin{tabular}{l|l|l}
        \toprule 
        \textbf{Evaluation Category}                          & \textbf{Benchmark}                                       & \textbf{Metrics}     \\\midrule 
        \multirow{6}{*}{\makecell[l]{Perceptual\\Evaluation}} & \multirow{4}{*}{\makecell[l]{In-house Human Evaluation}} & Overall              \\
                                                              &                                                          & Motion Quality \\
                                                              &                                                          & Instruction Following \\
                                                              &                                                          & Visual Quality        \\ \cline{2-3}
                                                              & \multirow{2}{*}{\makecell[l]{VBench-I2V}}                & \multirow{2}{*}{\makecell[l]{Automated Quality Metrics}} \\            
                                                              &                                                          &                             \\                                                                                                                                                                                                   
        \midrule \makecell[l]{Physical\\Evaluation}           & Physics-IQ-Benchmark                                     & Physics-IQ-Score \\
        \bottomrule
    \end{tabular}
    \caption{Evaluation Benchmark Overview}\label{tab:evaluation_framework}
    
\end{table}

\subsection{Perceptual Evaluation}
\subsubsection{In-house Human Evaluation Benchmark}
\label{sec:i2v_human_evaluation} 
Our in-house evaluation benchmark is primarily designed for I2V task, and integrates three complementary components to ensure comprehensive and unbiased assessment.
First, we design a hierarchical metric system that prioritizes completeness over simplicity, while enforcing orthogonality among metrics to enable fine-grained evaluation across multiple quality dimensions without redundancy.
Second, we construct a benchmark dataset of 100 diverse image-prompt pairs through systematic selection. These pairs span a broad spectrum of scenarios, from simple object motions to complex human activities, and each curated to probe specific aspects of video generation capability.
Third, we implement a double-blind comparison protocol with standardized output normalization, ensuring that each model operates under fair conditions for a meaningful comparison.

\paragraph{Evaluation Metrics.}

To ensure a comprehensive and reliable evaluation while avoiding unnecessary complexity, we adhere to three guiding principles in our metric design: comprehensiveness first, simplicity second, and orthogonality third.
Unlike T2V, where both visual content and motion are generated from scratch, I2V starts with a fixed visual input provided by the user’s uploaded image, while the subsequent dynamics are guided by the input text condition. This distinction shifts the evaluation focus toward assessing the motion and temporal quality of generated video while ensuring faithful preservation of the original visual elements.

Through preliminary analysis, we identified several common failure modes in I2V generation, including distortion, clipping, and temporal jittering. These typical issues guided the design of our evaluation framework, which emphasizes motion quality, temporal coherence, and the trade-off between source image fidelity and natural animation.
Therefore, our evaluation framework organizes metrics into four primarily dimensions:
\emph{Overall}, \emph{Motion Quality}, \emph{Instruction Following}, and \emph{Visual Quality}.
Each dimension is further broken down into specific sub-metrics designed to
capture particular aspects of video generation quality as shown in Tab.~\ref{tab:evaluation_metrics}.

\begin{table}[h]
    \centering
    \begin{tabular}{l|l|l}
        \toprule 
        \multirow{2}{*}{\makecell[l]{Main\\Metric}}          & \multirow{2}{*}{\makecell[l]{ Sub \\Metric}}  & \multirow{2}{*}{\makecell[l]{ Description}}    \\
                                                             &                                               &                                                \\\midrule 
        \multirow{1}{*}{\makecell[l]{Overall}}               & -                                             & General preference    \\   \midrule                                                   
        \multirow{4}{*}{\makecell[l]{Motion\\Quality}}       & Motion Speed                                  & Appropriate timing of movements                \\
                                                             & Motion Amplitude                              & Natural range of movement                      \\
                                                             & Motion Smoothness                             & Continuous movement without jitter             \\
                                                             & Movement Direction                            & Logical and consistent direction               \\\midrule    
        \multirow{3}{*}{\makecell[l]{Instruction\\Following}}& Subject Adherence                             & Following behavioral instructions              \\
                                                             & Environment Adherence                         & Meeting contextual requirements                \\
                                                             & Camera Adherence                              & Following camera movement requests             \\\midrule 
        \multirow{4}{*}{\makecell[l]{Visual\\Quality}}       & Subject Features                              & Consistency of main subject                    \\
                                                             & Scene Features                                & Consistency of environment                     \\
                                                             & Lighting Changes                              & Quality of lighting transitions                \\
                                                             & Texture Changes                               & Consistency of surface appearances             \\
        \bottomrule
    \end{tabular}
    \caption{Hierarchical Evaluation Framework}\label{tab:evaluation_metrics}
\end{table}

\paragraph{Dataset Construction.}

We construct a benchmark dataset consisting of 100 high-quality image-prompt pairs, each carefully selected to challenge different aspects of I2V generation. To ensure diversity and representativeness, we source data from four sources: 1) user-submitted inputs from existing video generation platforms, 2) synthetic images generated by FLUX~\citep{flux2024}, 3) authentic photographs from public repositories, and 4) professional cinematographic materials. Each sample is annotated with specific evaluation targets defined by our metric framework, enabling broad coverage of assessment dimensions while avoiding redundancy.

The dataset construction process follows a systematic multi-stage pipeline. We first establish a set of selection criteria focused on key challenges in I2V generation, including complex object deformation, multi-object interaction, dynamic camera motion, and lighting transitions. Based on these criteria, experts nominate candidate samples, which are then finalized through a collaborative voting procedure. This curated process ensures the resulting benchmark presents a diverse yet focused set of evaluation cases for rigorously testing I2V models.

\paragraph{Results and Analysis.}
\label{para:i2v_results_and_analysis}
Our evaluation methodology employs a paired comparison approach designed to directly measure relative model performance. Specifically, for each test case, we generate two videos (one from our model and one from a comparative model) using identical prompts and input images. Expert evaluators with strong aesthetic training then indicate their preference between each pair (Win/Tie/Lose) across multiple evaluation dimensions without knowledge of which model produced which video.

\magi's autoregressive design enables generation of arbitrary-length videos. For fair comparison, we adapt our generation length to match each comparison model: for example, 5 seconds for Kling and 6 seconds for Hailuo. To avoid potential manipulation of visual quality, we maintain each model's native output without post-processing like resolution normalization. In addition, all models are evaluated using raw user inputs without any manual refinement from our side, relying solely on their built-in prompt enhancement (PE) mechanisms.

\begin{figure}[h]
    \centering
    \includegraphics[width=.8\textwidth]{
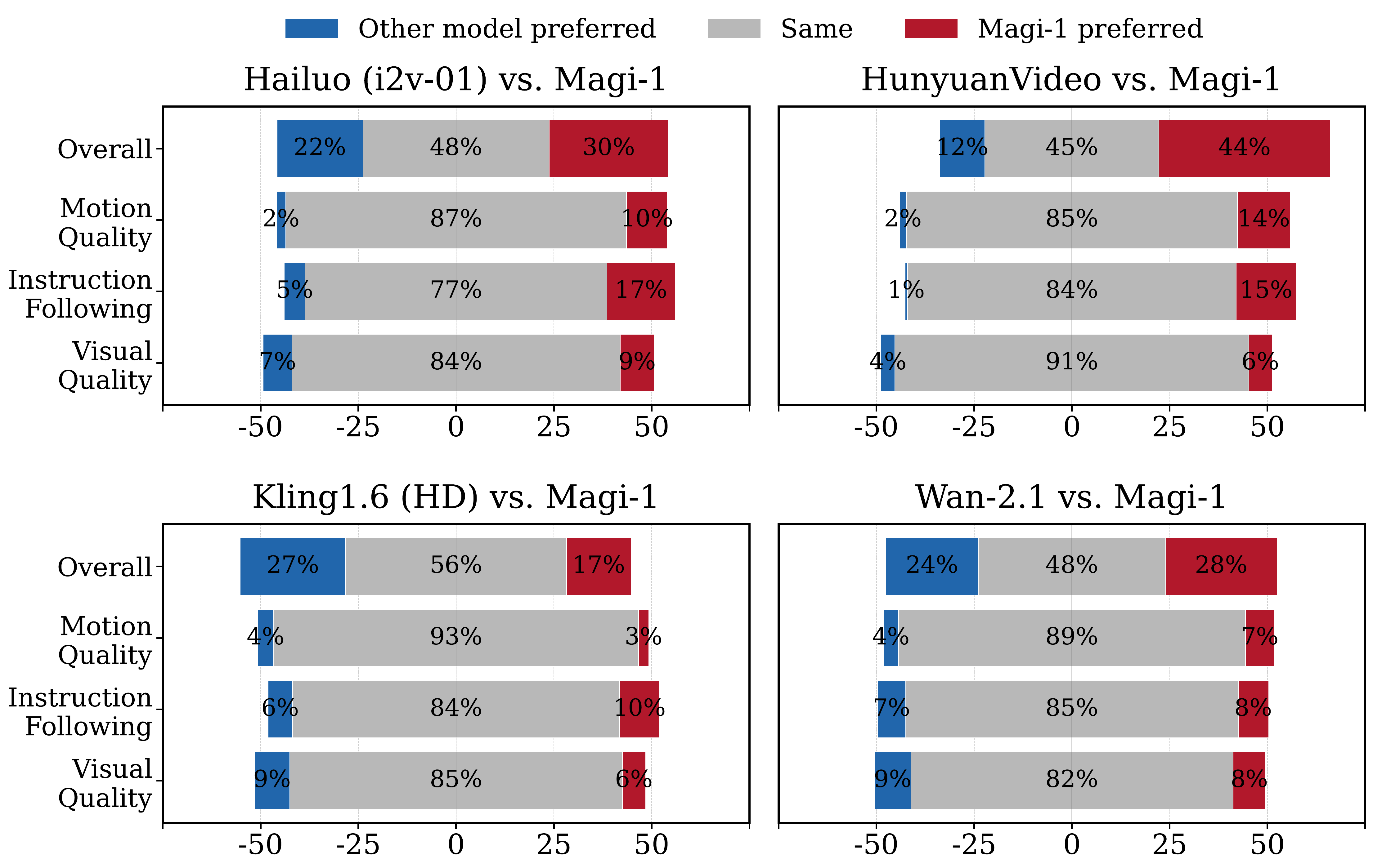
    }    \vspace{0.5em}
    \caption{Comparative evaluation of our model against leading open-source and proprietary video generation models across multiple metrics. Each bar is divided into three sections: red, gray, and blue, representing Win-Tie-Loss percentages for each comparison. Blue sections indicate where users preferred the competitor model, gray sections represent ties, and red sections show where users preferred our model. The evaluation includes both API-based assessments like Kling1.6 (HD)~\citep{kuaishou2024kling} and Hailuo (i2v-01)~\citep{minimax2024hailuo} and locally deployed models like Wan-2.1~\citep{wang2025wan} and HunyuanVideo~\citep{kong2024hunyuanvideo}), providing a comprehensive comparison across various implementation environments.}
    \label{fig:metric_distribution_benchmark_sample_set}
\end{figure}

The evaluation results shown in Fig.~\ref{fig:metric_distribution_benchmark_sample_set} demonstrate \magi's strong competitive position in the field. In terms of overall performance, our model shows advantages over the open-source model Wan-2.1~\citep{wang2025wan}, performs slightly behind the commercial model Kling1.6 (HD)~\citep{kuaishou2024kling}, but achieves clearly better results compared to both Hailuo(i2v-01)~\citep{minimax2024hailuo} and HunyuanVideo~\citep{kong2024hunyuanvideo}. Looking at specific capabilities, \magi excels particularly in instruction following and motion quality metrics, consistently receiving high scores across comparisons. However, in terms of visual quality, there remains room for improvement compared to top models.

\subsubsection{VBench}
\label{subsubsec:i2v_public_evaluation}

VBench~\citep{huang2024vbenchcomprehensiveversatilebenchmark} is currently the most widely adopted benchmark in the community for automated and objective evaluation of video generation models. While its evaluation framework is still evolving and not without limitations, VBench remains a critical tool for model comparison due to its fully automated and reproducible assessment process, especially when contrasted with in-house human evaluations, which are often subjective and lack transparency.

VBench provides two primary evaluation tracks: text-to-video (T2V) and image-to-video (I2V). Given that I2V more closely reflects real-world usage patterns, where users typically input a static image to generate videos in existing product, and in line with our goal of aligning evaluation with practical application scenarios, we focus our evaluation on the I2V track in VBench.

We evaluate the generation quality of \magi under two different configurations: \magi (1$\times$decoder) and \magi (2$\times$decoder). The only difference between them lies in the VAE decoder: \magi (2$\times$decoder) employs an enhanced decoder capable of 2× upsampling, while the core autoregressive denoising model remains identical across both versions. For evaluation, both models generate 4-second videos at 24 FPS with a 16:9 aspect ratio.

\def\vbenchtitlea{\multirow{2}{*}{\makecell[c]{Metric\\(VBenchI2V)}}}
\def\vbenchtitleb{\multirow{2}{*}{\makecell[c]{\magi\\(2$\times$decoder)}}} 
\def\vbenchtitlec{\multirow{2}{*}{\makecell[c]{\magi\\(1$\times$decoder)}}}
\def\vbenchtitled{\multirow{2}{*}{\makecell[c]{VisualPi}}}
\def\vbenchtitlee{\multirow{2}{*}{\makecell[c]{StepFun\\(TI2V)}}}
\def\vbenchtitlef{\multirow{2}{*}{\makecell[c]{OS\\(TopA)}}}
\def\vbenchtitleg{\multirow{2}{*}{\makecell[c]{OS\\(TopB)}}}
\begin{table}[h]
\centering
\small
\begin{tabular}{l|l|c|c|c|c}
\toprule
&\vbenchtitlea           & \vbenchtitleb  & \vbenchtitlec  & \vbenchtitled    & \vbenchtitlee    \\& &  &  &   &      \\\midrule \multirow{9}{*}{\makecell[l]{Quality\\Metrics}} 
&\abf{I2V-Camera}        & \und{50.85}    & \abf{50.77}    & \bbf{51.20}      & \abf{49.23}      \\ 
&\abf{I2V-Subject}       & \und{98.39}    & \abf{98.36}    & \bbf{98.67}      & \abf{97.86}      \\ 
&\abf{I2V-Background}    & \bbf{99.00}    & \und{98.98}    & \abf{98.87}      & \abf{98.63}      \\ 
&\abf{Subject Cons.}     & \abf{93.96}    & \abf{94.28}    & \bbf{96.87}      & \und{96.02}      \\ 
&\abf{Motion Smooth.}    & \abf{98.68}    & \abf{98.83}    & \und{99.18}      & \bbf{99.24}      \\ 
&\abf{Imaging Quality}   & \abf{69.71}    & \abf{69.68}    & \bbf{72.86}      & \und{70.44}      \\ 
&\abf{Dynamic Degree}    & \bbf{68.21}    & \und{63.41}    & \abf{49.93}      & \abf{48.78}      \\ 
&\abf{Background Cons.}  & \abf{96.74}    & \abf{96.90}    & \bbf{97.50}      & \und{97.06}      \\ 
&\abf{Aesthetic Quality} & \bbf{64.74}    & \abf{61.89}    & \abf{61.91}      & \und{62.29}      \\ \midrule \multirow{3}{*}{\makecell[l]{Agg.\\Scores}} 
& \abf{Quality Score}    & \bbf{82.44}    & \abf{81.67}    & \und{81.95}      & \abf{81.22}      \\
& \abf{I2V Score}        & \und{96.12}    & \abf{96.08}    & \bbf{96.21}      & \abf{95.50}      \\
& \abf{Total Score}      & \bbf{89.28}    & \abf{88.88}    & \und{89.08}      & \abf{88.36}      \\
\bottomrule
\end{tabular}
\vspace{0.5em}
\caption{Quantitative evaluation results on VBench-I2V benchmark. \magi (1$\times$decoder) denotes our baseline model ($1280\times720$ resolution), while \magi (2$\times$decoder) represents the enhanced variant with 2x VAE upsampling ($2560\times1440$ resolution). Comparative data for other models are sourced from the top tier at latest  \href{https://huggingface.co/spaces/Vchitect/VBench_Leaderboard}{Vbench leaderboard}. Bold and underlined values indicate the highest and second-highest scores respectively across all metrics.}\label{tab:vbench_results}

\end{table}

The results are presented in Tab.~\ref{tab:vbench_results}. As shown, both of our models achieve outstanding performance, with \magi (2$\times$ decoder) reaching a top overall score of 89.28, ranking first among all models.
Notably, the \magi models demonstrate a significant advantage in the \emph{dynamic Degree} compared to other approaches, while simultaneously maintaining high visual quality, including strong performance in \emph{aesthetic quality} and \emph{motion smoothness}. This effectively addresses a common trade-off in other methods, where increasing motion amplitude often downgrade image quality. We attribute this strength to the autoregressive denoising architecture, which provides a stronger modeling capability for complex motion dynamics.

\subsection{Physical Evaluation}
Video generation models are increasingly recognized as a foundation toward building the world model, and the ability to accurately capture real-world physical dynamics has become a central focus within the research community. In contrast to perceptual evaluation, which inevitably involves subjective human preferences, physics-based evaluation aims to assess a model’s ability to understand and simulate objective physical principles.

Currently, there are only a few established benchmarks~\citep{bansal2024videophy,meng2024towards,dash2011cophy,yi2019clevrer,kang2024far} in this area, and Physics-IQ~\citep{physicsiqbenchmark} stands out as the most comprehensive and state-of-the-art benchmark. Therefore, we adopt Physics-IQ to evaluate the physical understanding and reasoning capabilities of \magi.

The Physics-IQ evaluation protocol uses 8-second real-world videos that depict objective physical phenomena. The first 3 seconds of each video are provided as conditional input to the model, which is then required to predict the remaining 5 seconds. The accuracy of the model’s physical modeling capability is measured by comparing the predicted videos with the ground truth.

Since most existing video generation models do not natively support video-conditioned continuation, they typically approximate this task using image-to-video (I2V) generation, conditioning only on the last frame of the input video. To provide a comprehensive comparison, we report results for both two settings.

\def\physicstitlez{\multirow{2}{*}{\makecell[l]{Model}}}
\def\physicstitlea{\multirow{2}{*}{\makecell[c]{Phys.\\IQ Score↑}}}
\def\physicstitleb{\multirow{2}{*}{\makecell[c]{Spatial\\IoU ↑}}}
\def\physicstitlec{\multirow{2}{*}{\makecell[c]{Spatio\\Temporal↑}}}
\def\physicstitled{\multirow{2}{*}{\makecell[c]{Weighted\\Spatial IoU ↑}}}
\def\physicstitlee{\multirow{2}{*}{\makecell[c]{MSE↓}}}
\def\physicstitlef{\multirow{2}{*}{\makecell[c]{FPS}}}
\def\physicstitleg{\multirow{2}{*}{\makecell[c]{Size}}}
\begin{table*}[h]
\centering
\small
\setlength{\tabcolsep}{4pt}  
\begin{tabular*}{\textwidth}{l|c|c|c|c|c}
\toprule 
 \physicstitlez                                      & \physicstitlea   & \physicstitleb& \physicstitlec & \physicstitled &\physicstitlee \\
                                                     &                  &               &                &                &               \\\midrule
\abf{\magi (V2V)}                                     & \bbf{56.02}      & \bbf{0.367}   & \bbf{0.270}    & \bbf{0.304}    & \bbf{0.005}   \\ 
\abf{VideoPoet (V2V)~\citep{kondratyuk2023videopoet}}& \abf{29.50}      & \abf{0.204}   & \abf{0.164}    & \abf{0.137}    & \abf{0.010}   \\
\abf{Lumiere (V2V)~\citep{bar2024lumiere}}           & \abf{23.00}      & \abf{0.170}   & \abf{0.155}    & \abf{0.093}    & \abf{0.013}   \\ \midrule
\abf{\magi (I2V)}                                     & \bbf{30.23}      & \bbf{0.203}   & \abf{0.151}    & \bbf{0.154}    & \bbf{0.012}   \\ 
\abf{Kling1.6 (I2V)~\citep{kuaishou2024kling}}       & \abf{23.64}      & \abf{0.197}   & \abf{0.086}    & \abf{0.144}    & \abf{0.025}   \\ 
\abf{VideoPoet (I2V)~\citep{kondratyuk2023videopoet}}& \abf{20.30}      & \abf{0.141}   & \abf{0.126}    & \abf{0.087}    & \abf{0.012}   \\ 
\abf{Gen 3 (I2V)~\citep{runway2024gen3}}             & \abf{22.80}      & \abf{0.201}   & \abf{0.115}    & \abf{0.116}    & \abf{0.015}   \\ 
\abf{Wan2.1 (I2V)~\citep{wang2025wan}}               & \abf{20.89}      & \abf{0.153}   & \abf{0.100}    & \abf{0.112}    & \abf{0.023}   \\
\abf{Lumiere (I2V)~\citep{bar2024lumiere}}           & \abf{19.00}      & \abf{0.113}   & \bbf{0.173}    & \abf{0.061}    & \abf{0.016}   \\ 
\abf{SVD (I2V)~\citep{blattmann2023svd}}             & \abf{14.80}      & \abf{0.132}   & \abf{0.076}    & \abf{0.073}    & \abf{0.021}   \\ 
\abf{Pika 1.0 (I2V)~\citep{pika2024pika}}            & \abf{13.00}      & \abf{0.140}   & \abf{0.041}    & \abf{0.078}    & \abf{0.014}   \\ 
\abf{Sora (I2V)~\citep{openaisora2024}}              & \abf{10.00}      & \abf{0.138}   & \abf{0.047}    & \abf{0.063}    & \abf{0.030}   \\ \midrule
\abf{GroundTruth}                                    & \abf{100.0}      & \abf{0.678}   & \abf{0.535}    & \abf{0.577}    & \abf{0.002}   \\ 
\bottomrule
\end{tabular*}
\vspace{0.5em}
\caption{Quantitative comparison of video generation models evaluated on the Physics-IQ-Benchmark. Models are categorized by input modality: image-to-video (I2V) and video-to-video (V2V). Results were obtained through direct evaluation of model APIs, local deployment of open-source implementations, and as reported in \cite{physicsiqbenchmark}. In the V2V task, models observe the first 3 seconds of an 8-second ground truth video and predict the remaining 5 seconds, while in the I2V task, models take only a single frame at the 3-second mark and predict the subsequent 5 seconds. Magi(V2V) utilizes the full 24 FPS video input (96 frames). }\label{tab:physics_iq_benchmark_results}
\end{table*}

The results are presented in Tab.~\ref{tab:physics_iq_benchmark_results}. When conditioned on video inputs, \magi outperforms all competing models by a substantial margin, reaches the score of 56.02. The previous state-of-the-art model VideoPoet~\citep{kondratyuk2023videopoet}, which also supports video-to-video (V2V) prediction, is outperformed by approximately 27 points.
Even when using only image condition, \magi still achieves the highest score among all models, reaching 30.23, despite a noticeable drop compared to its video-conditioned version.

These results clearly demonstrate the strong capability of \magi in understanding and modeling real-world physical principles. We attribute this advantage to its autoregressive nature: modeling physical processes demands a focus on causality rather than mere correlation, and autoregressive models inherently promote causal reasoning. In contrast, bidirectional denoising models lack the algorithmic foundations necessary to effectively capture causality, which leads to inferior performance in such tasks.
While VideoPoet is also an autoregressive model, its primary design objective is integration with language models, which limits its efficiency in modeling the video modality; In contrast, \magi is purpose-built for video generation, combining the strengths of autoregressive and denoising-based modeling. This targeted design enables it to achieve significantly superior performance.

\begin{figure}[ht]
    \centering
    \begin{minipage}{1.0\textwidth}
        \begin{subfigure}[b]{1.0\textwidth}
            \centering
            \vspace{1pt}
            \begin{tabular}{@{}c@{}c@{}c@{\hspace{1pt}}c@{\hspace{1pt}}c@{\hspace{1pt}}c@{\hspace{1pt}}c@{\hspace{1pt}}c@{\hspace{1pt}}c@{\hspace{1pt}}c@{\hspace{1pt}}c@{\hspace{1pt}}c@{}}
                & 0.0s & 2.5s & 3.0s & 3.3s & 3.7s & 4.0s & 4.3s & 4.7s & 5.0s & 7.9s \\
                \raisebox{1.2\height}{\tiny \rotatebox[origin=c]{90}{Real}   }&
                \includegraphics[width=0.088\textwidth]{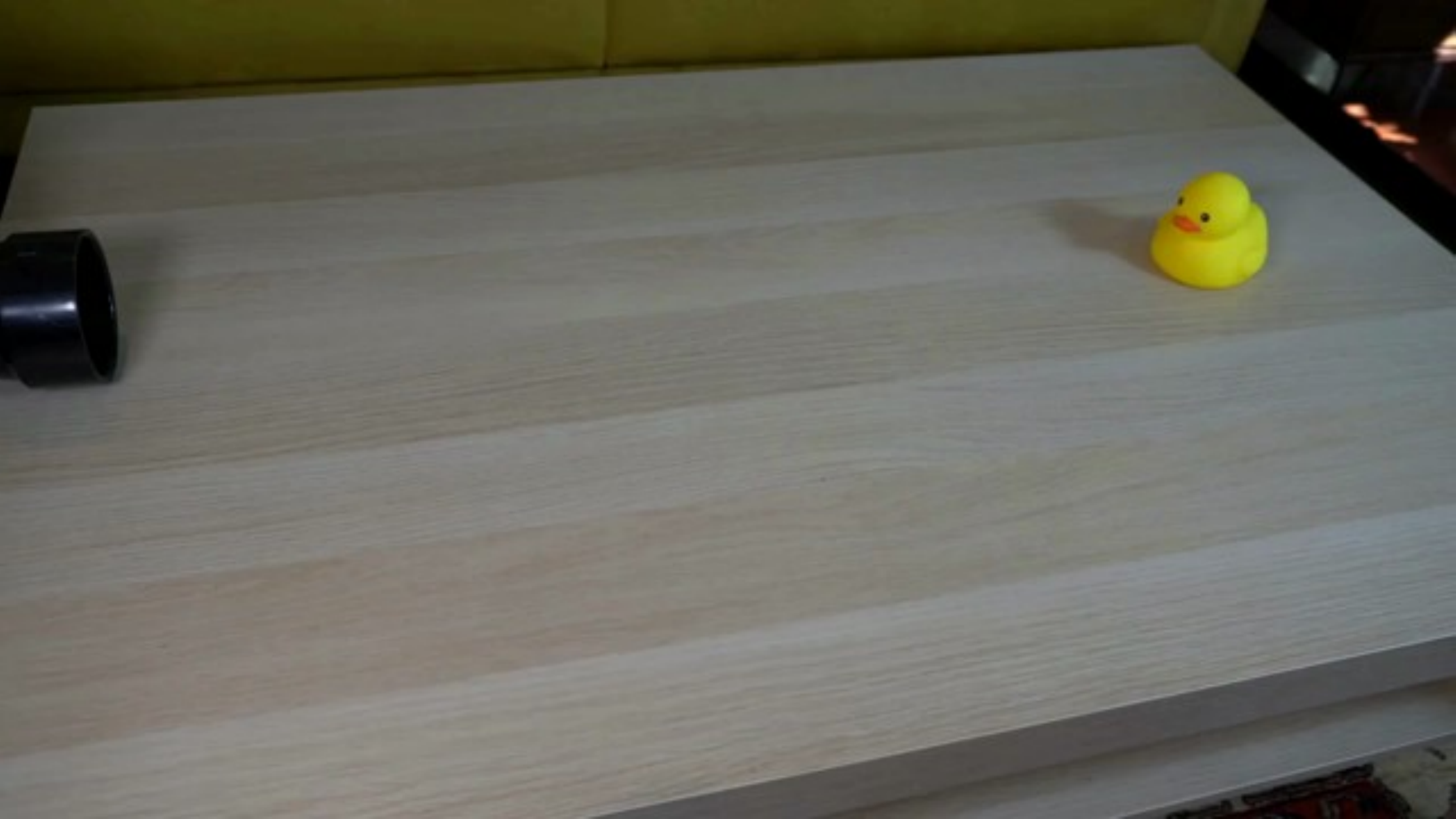}  &
                \includegraphics[width=0.088\textwidth]{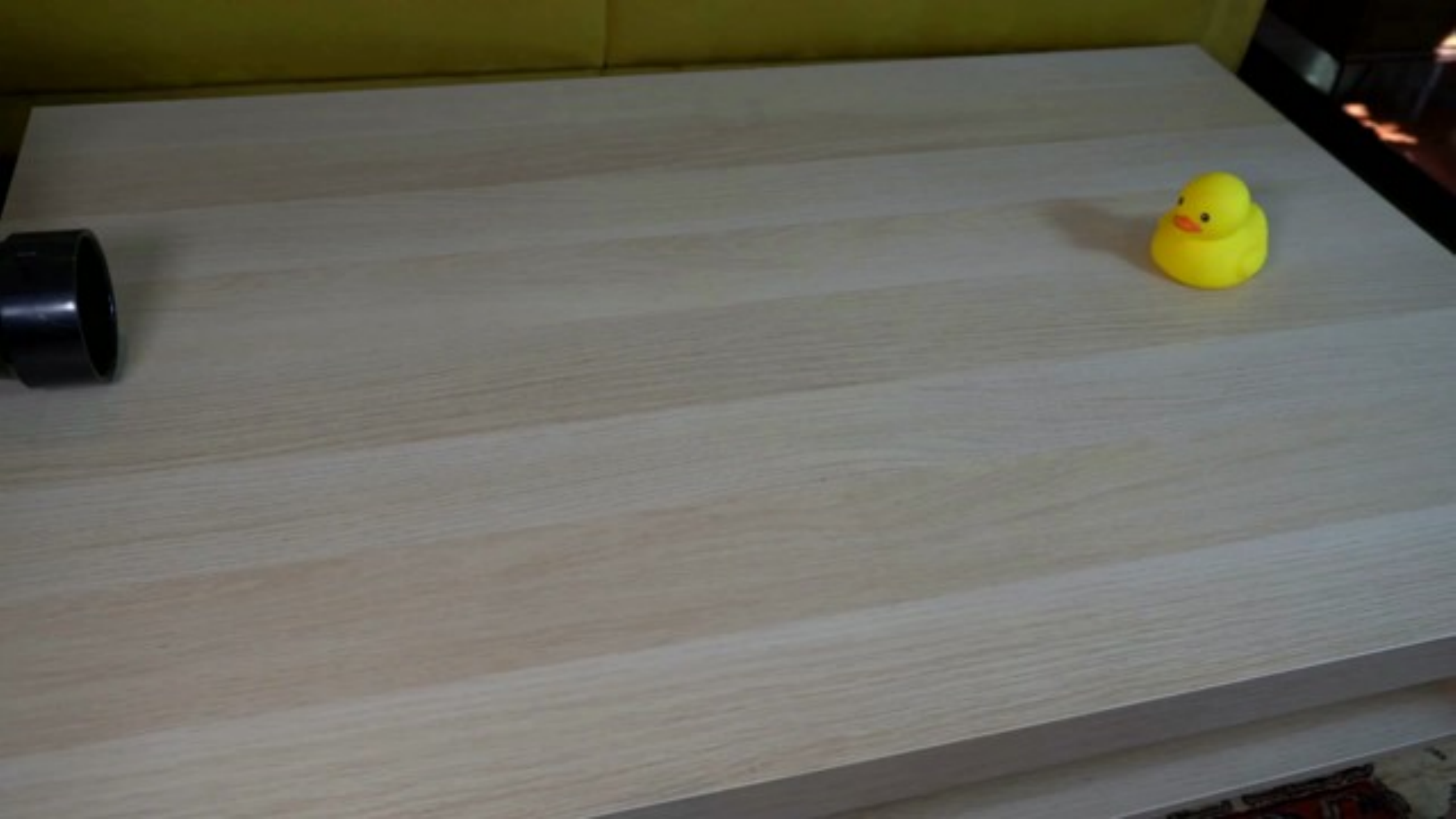} &
                \includegraphics[width=0.088\textwidth]{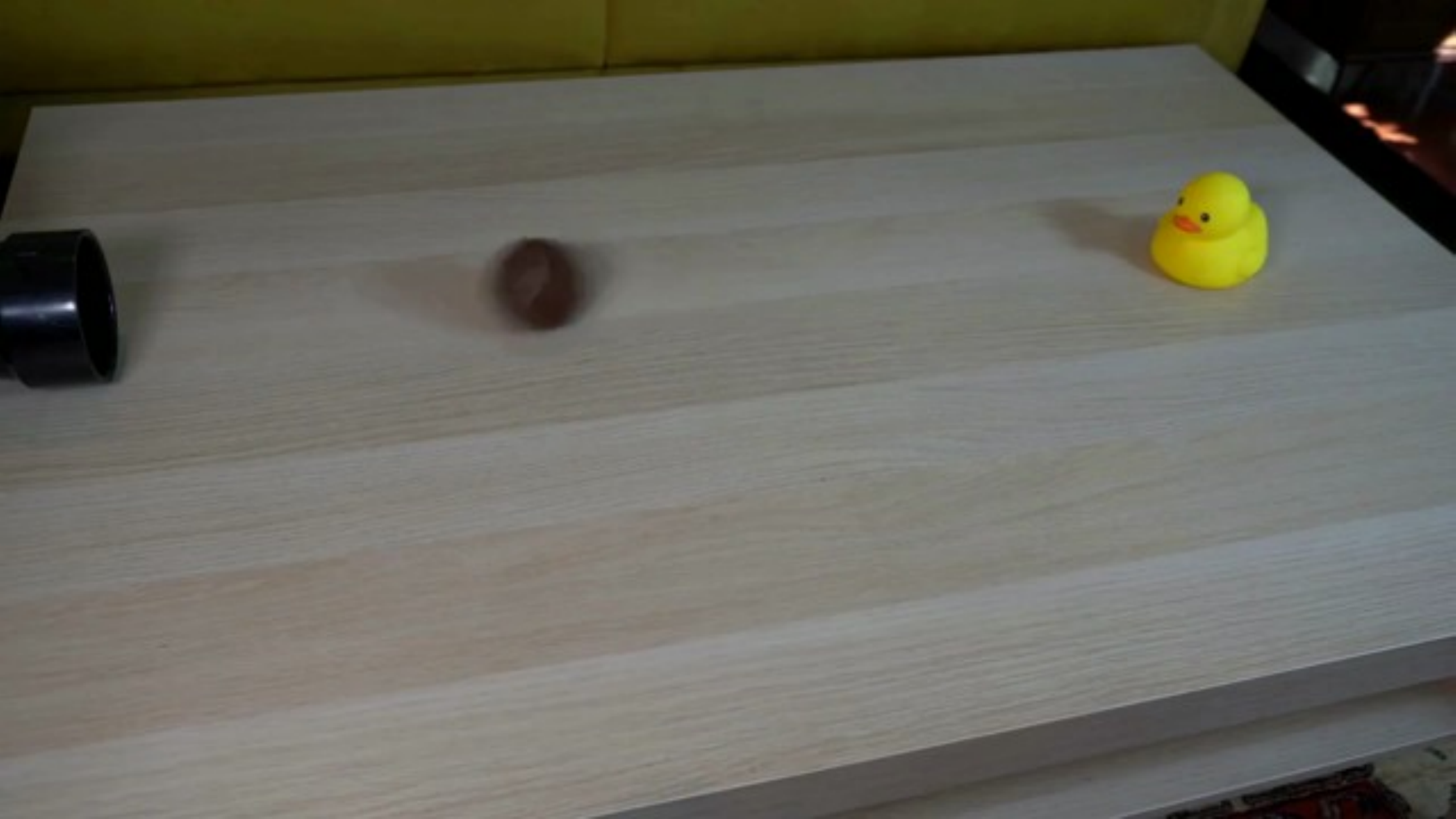} &
                \includegraphics[width=0.088\textwidth]{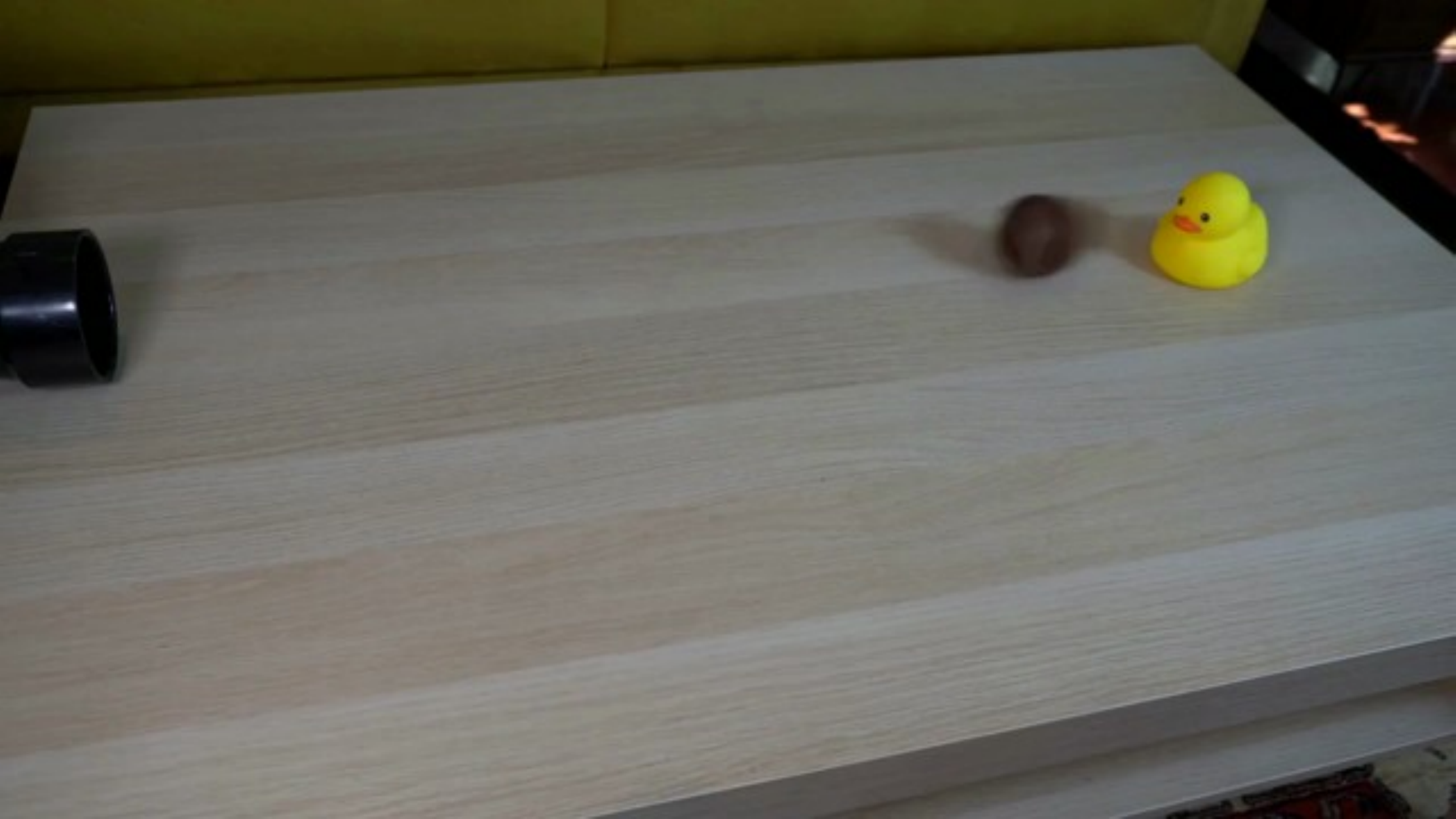} &
                \includegraphics[width=0.088\textwidth]{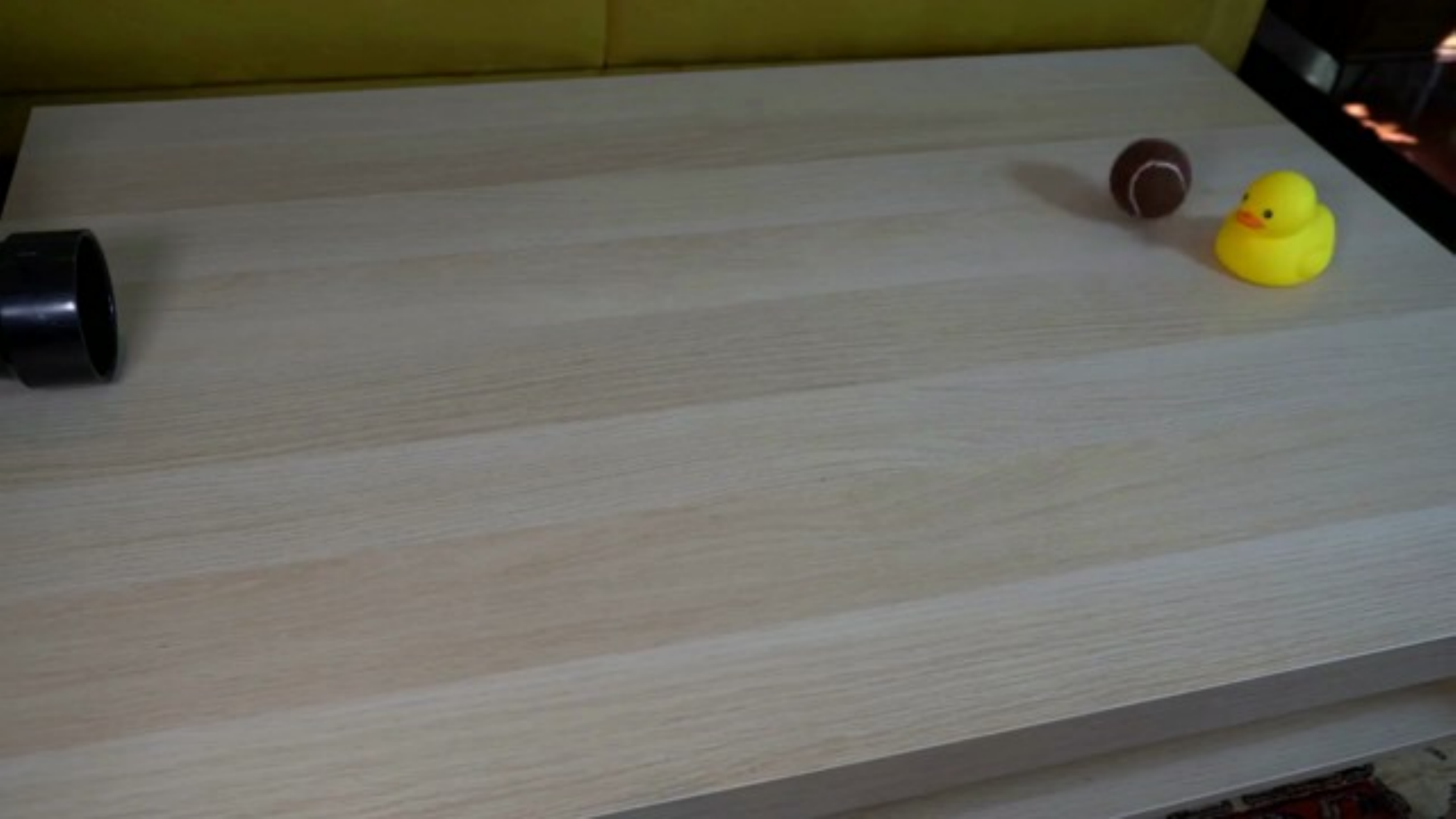} &
                \includegraphics[width=0.088\textwidth]{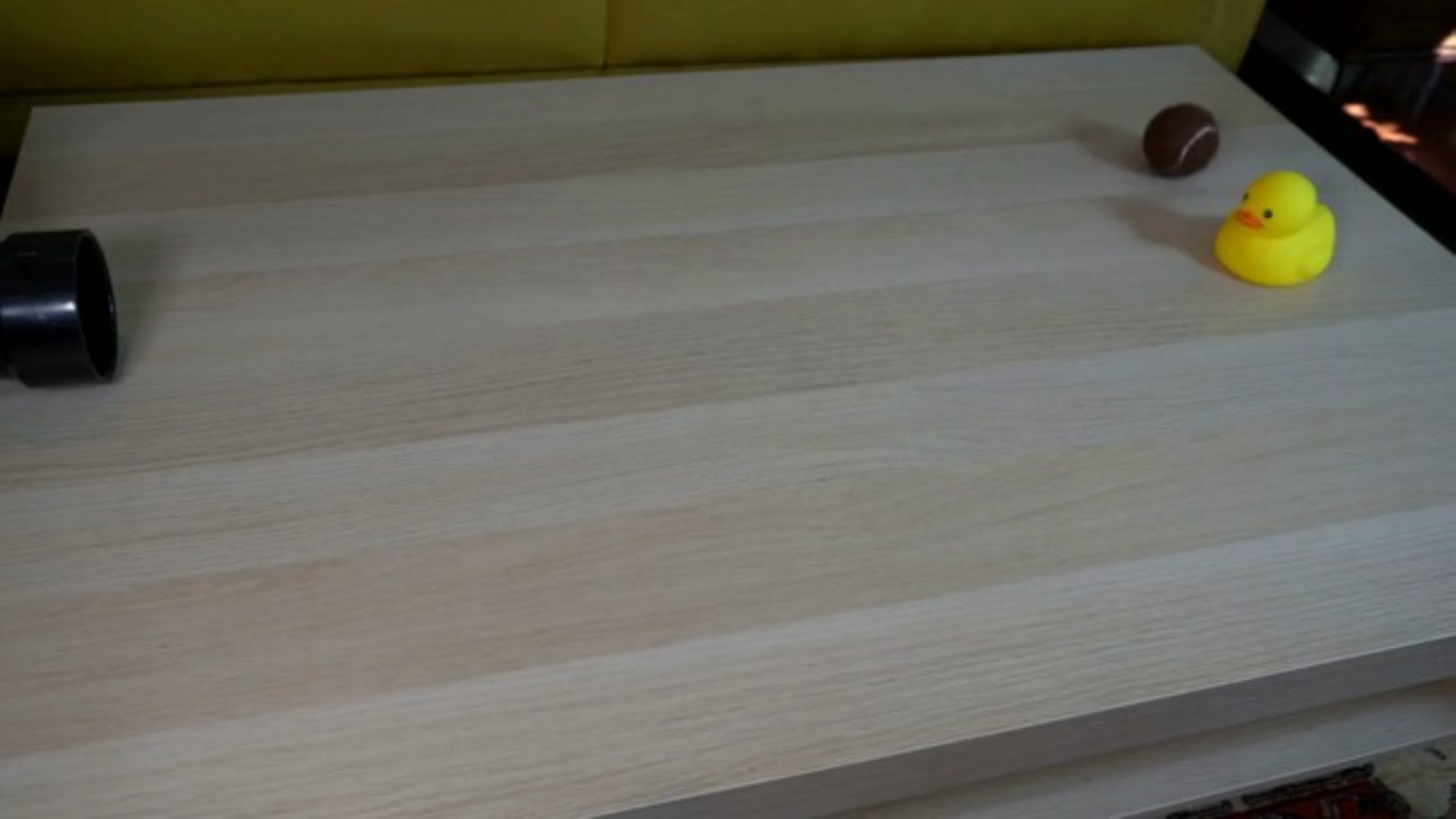} &
                \includegraphics[width=0.088\textwidth]{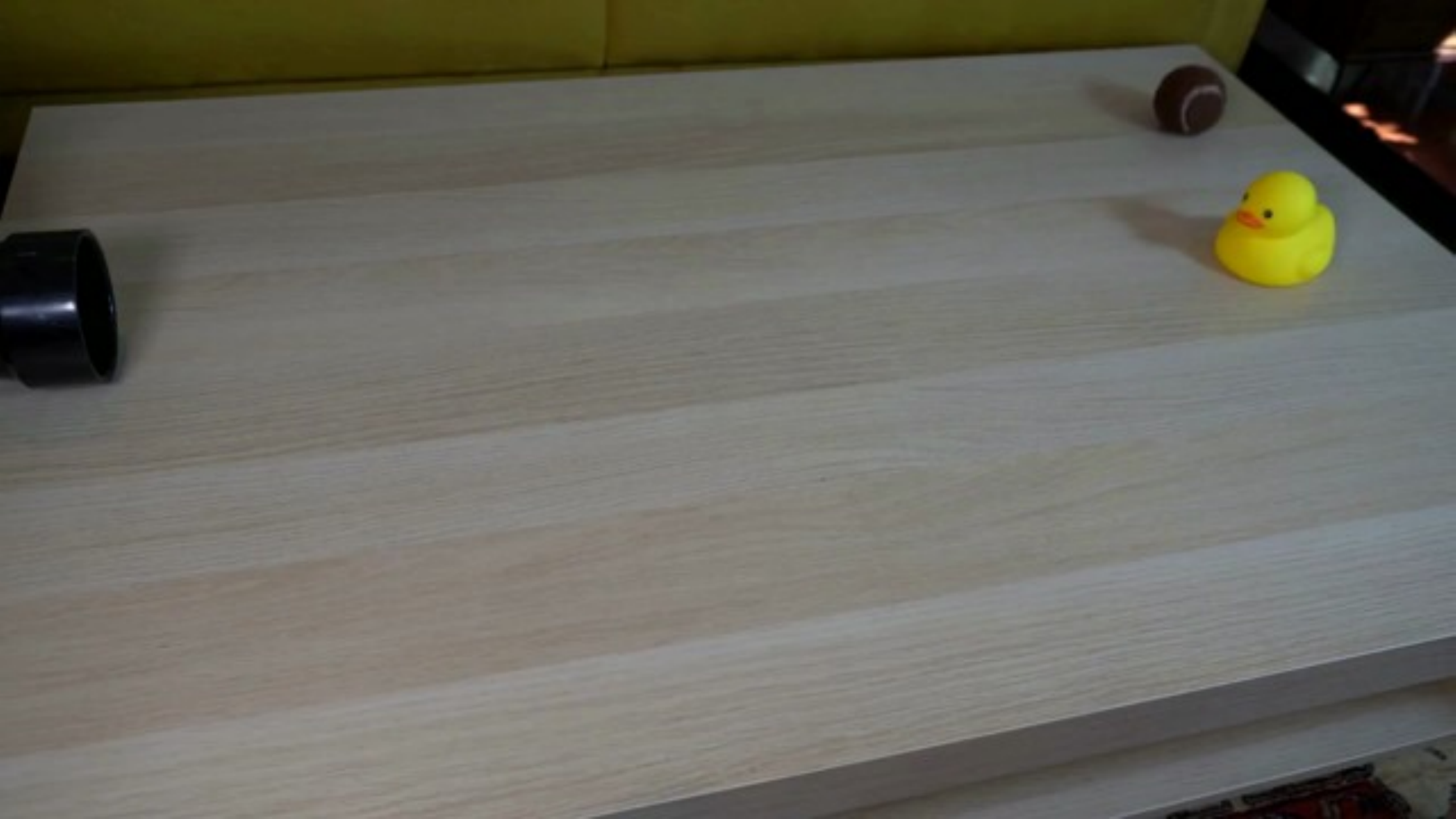} &
                \includegraphics[width=0.088\textwidth]{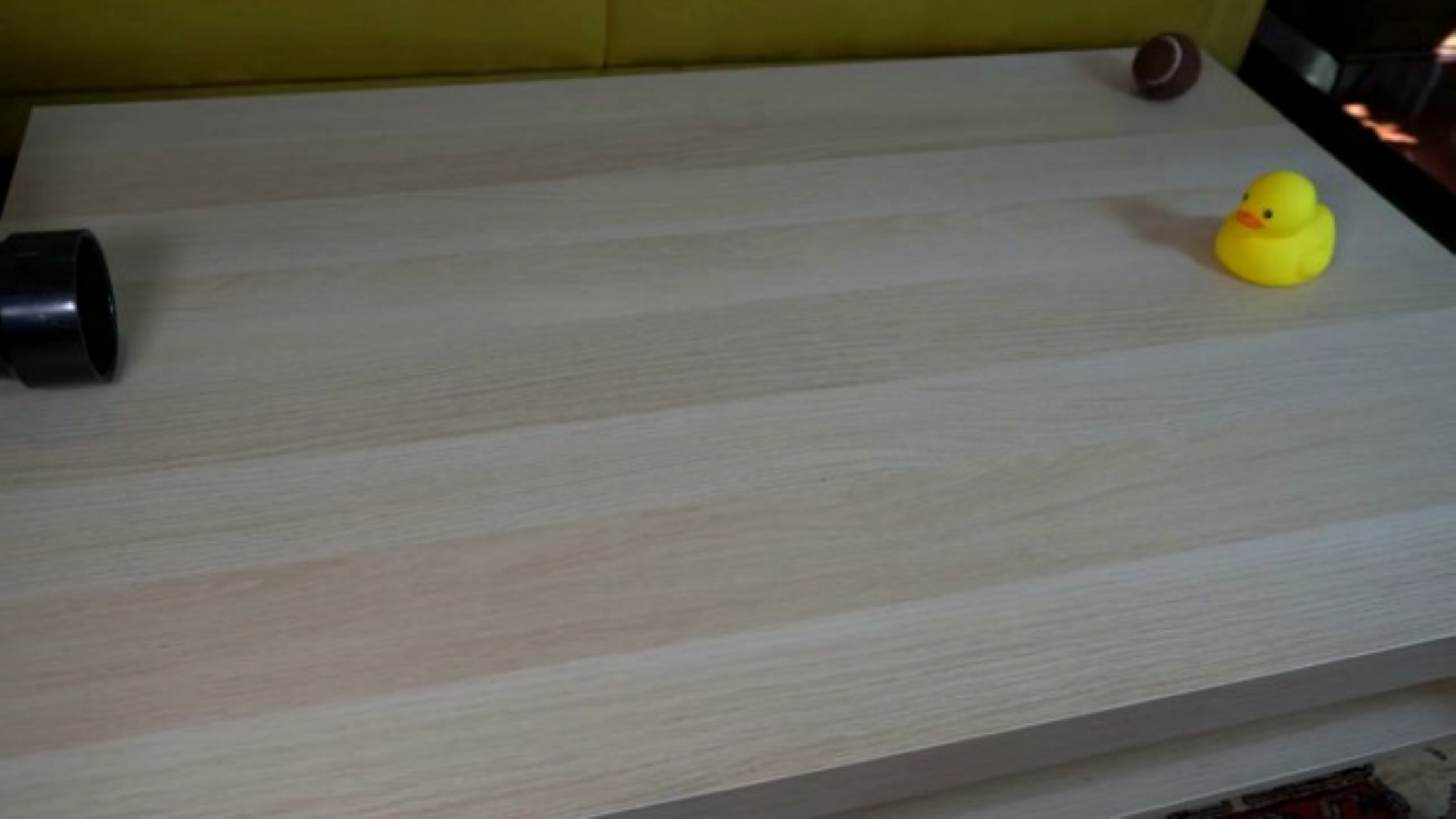} &
                \includegraphics[width=0.088\textwidth]{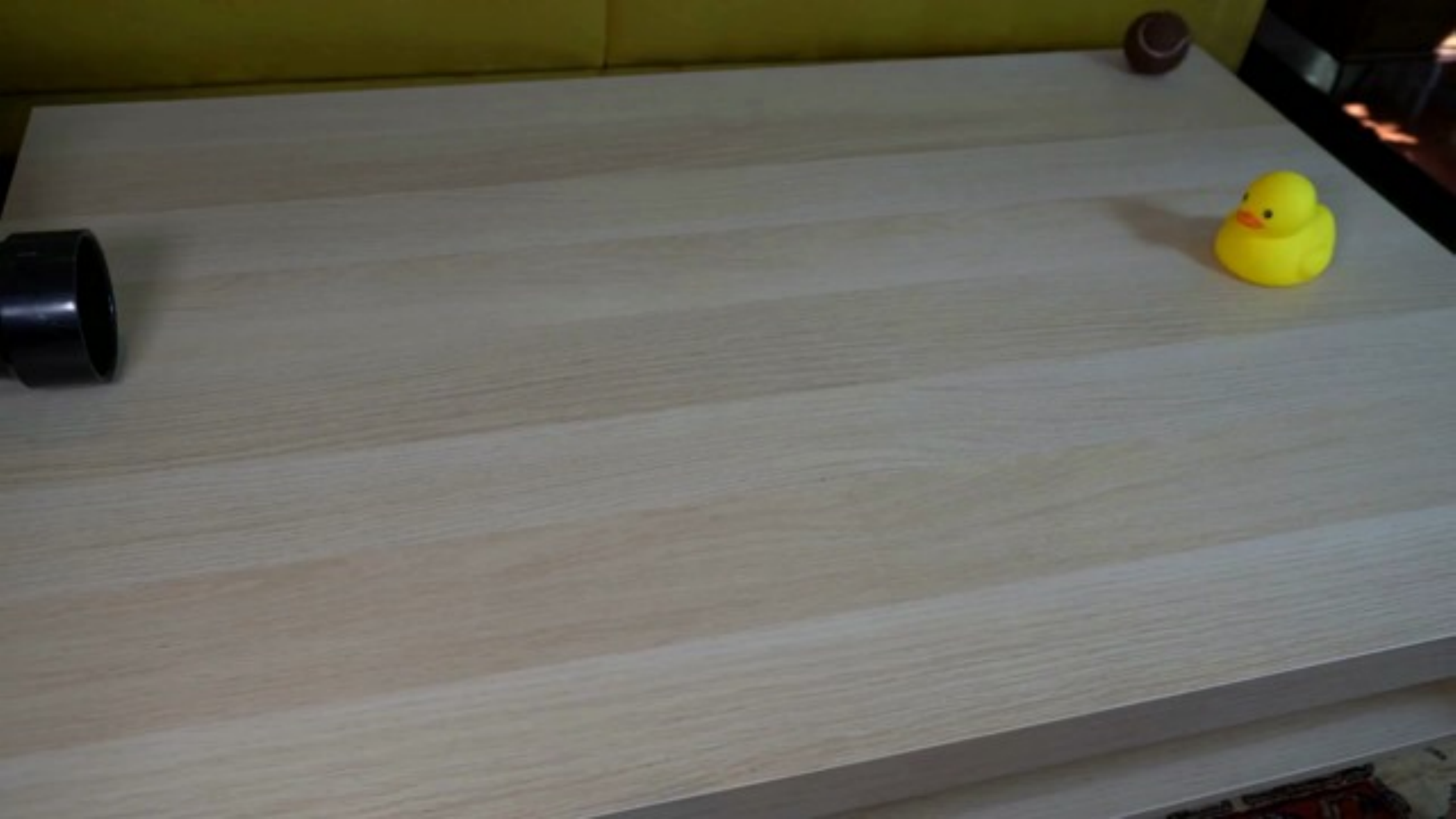} &
                \includegraphics[width=0.088\textwidth]{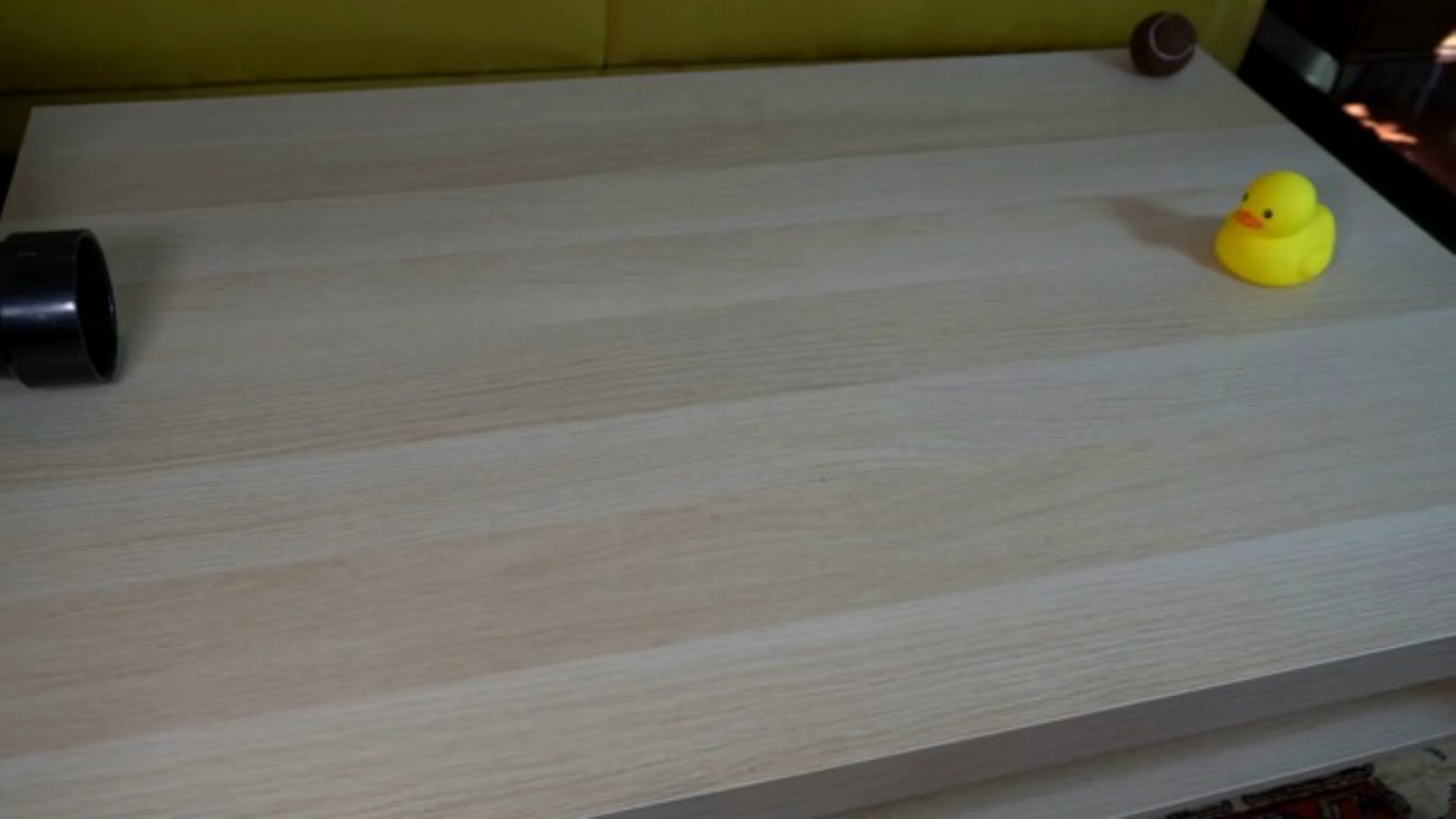}
            \end{tabular}
            \vspace{0.1pt}  
            \begin{tabular}{@{}c@{}c@{\hspace{1pt}}c@{\hspace{1pt}}c@{\hspace{1pt}}c@{\hspace{1pt}}c@{\hspace{1pt}}c@{\hspace{1pt}}c@{\hspace{1pt}}c@{\hspace{1pt}}c@{\hspace{1pt}}c@{}}
                \raisebox{0.6\height}{\tiny \rotatebox[origin=c]{90}{Generated}   }&
                
                \includegraphics[width=0.088\textwidth]{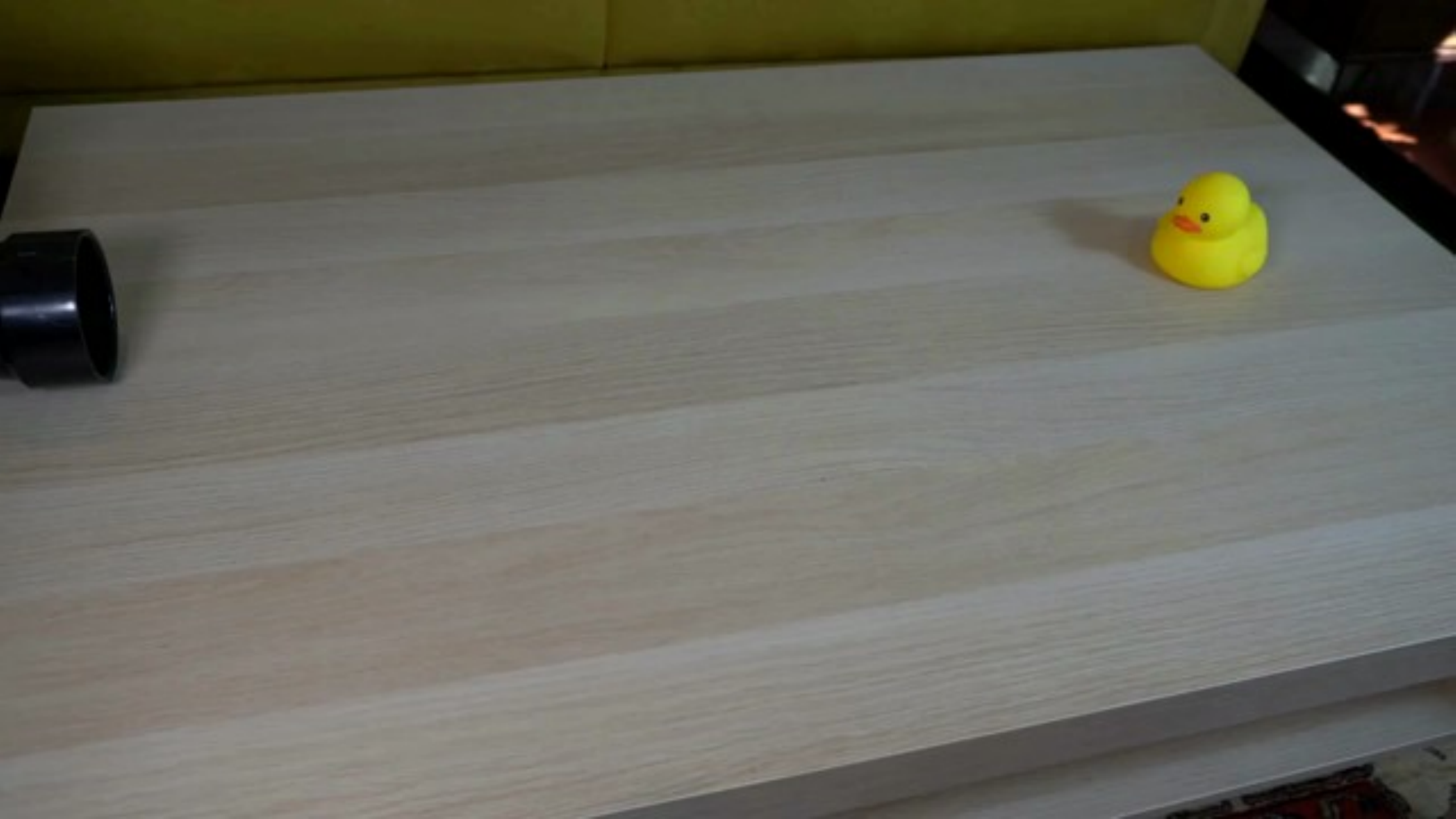} &
                \includegraphics[width=0.088\textwidth]{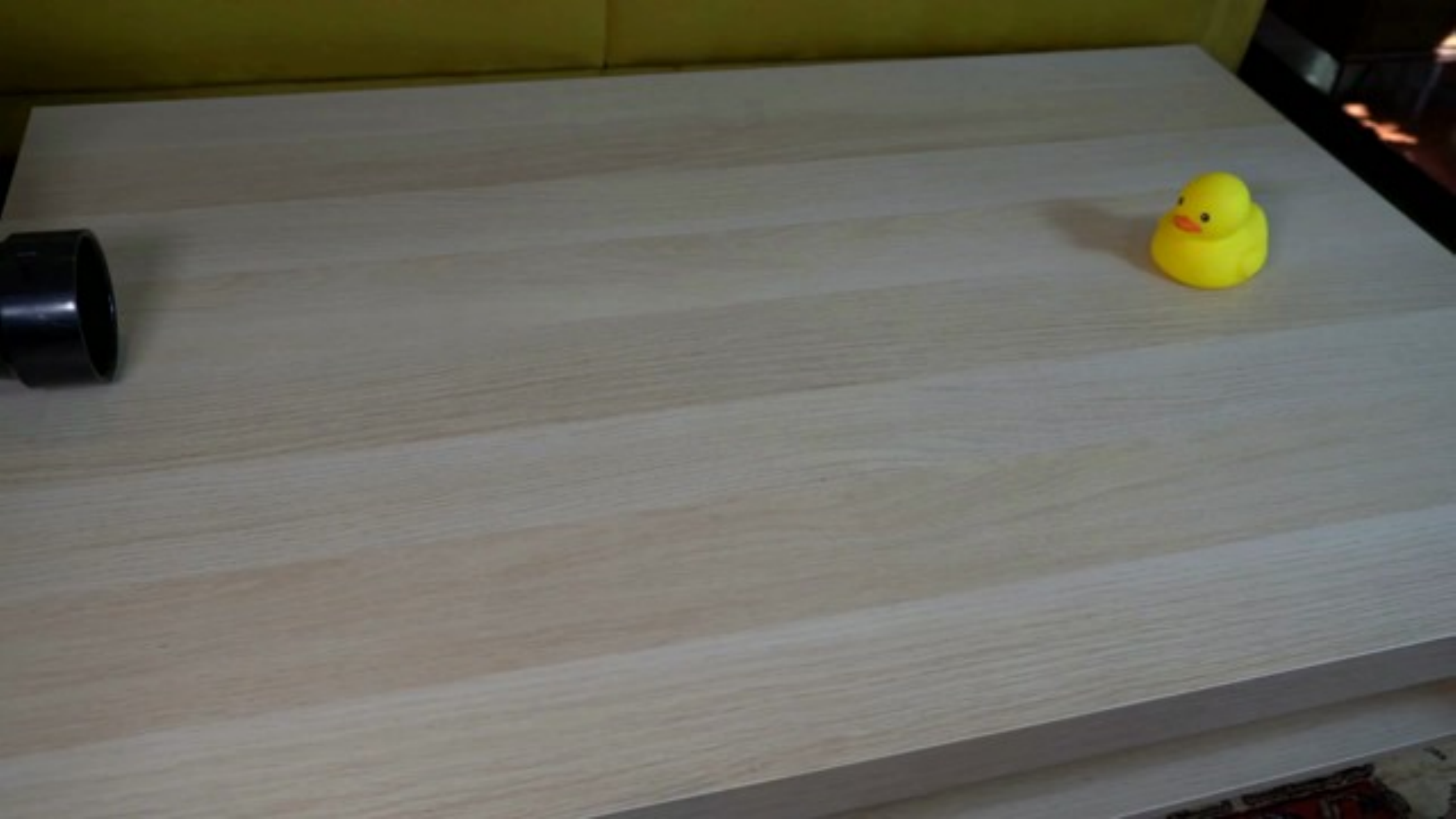} &
                \includegraphics[width=0.088\textwidth]{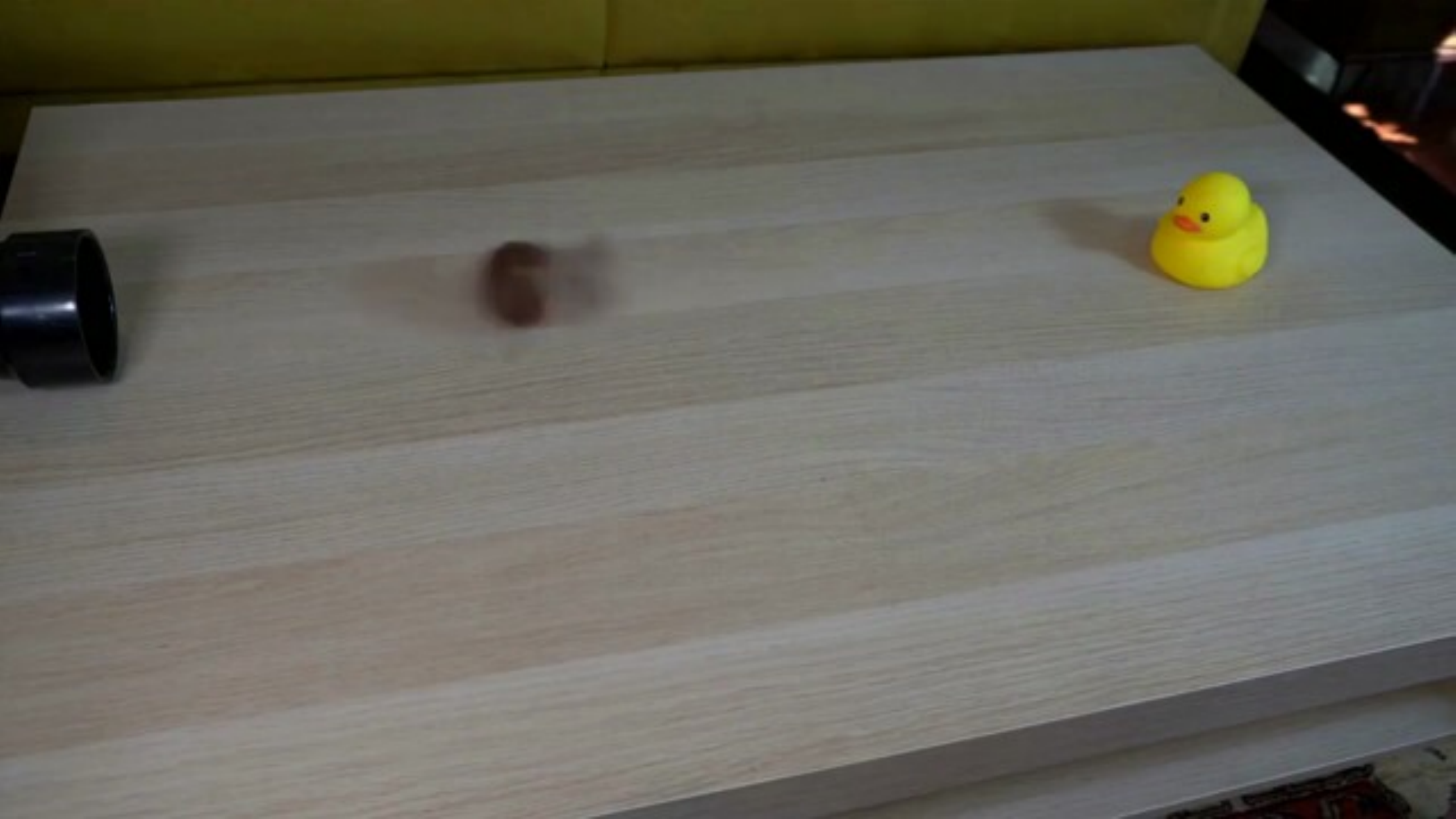} &
                \includegraphics[width=0.088\textwidth]{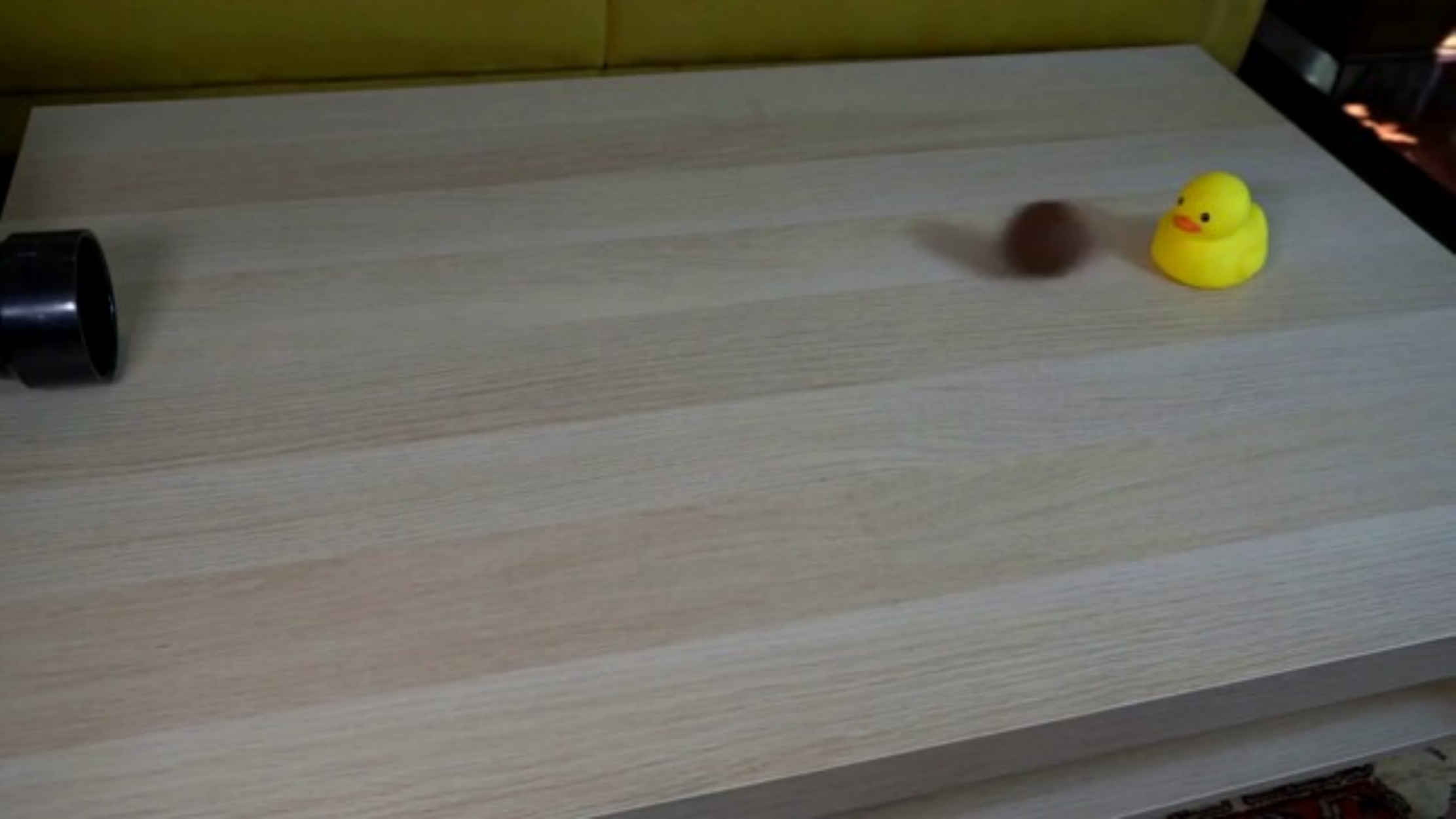} &
                \includegraphics[width=0.088\textwidth]{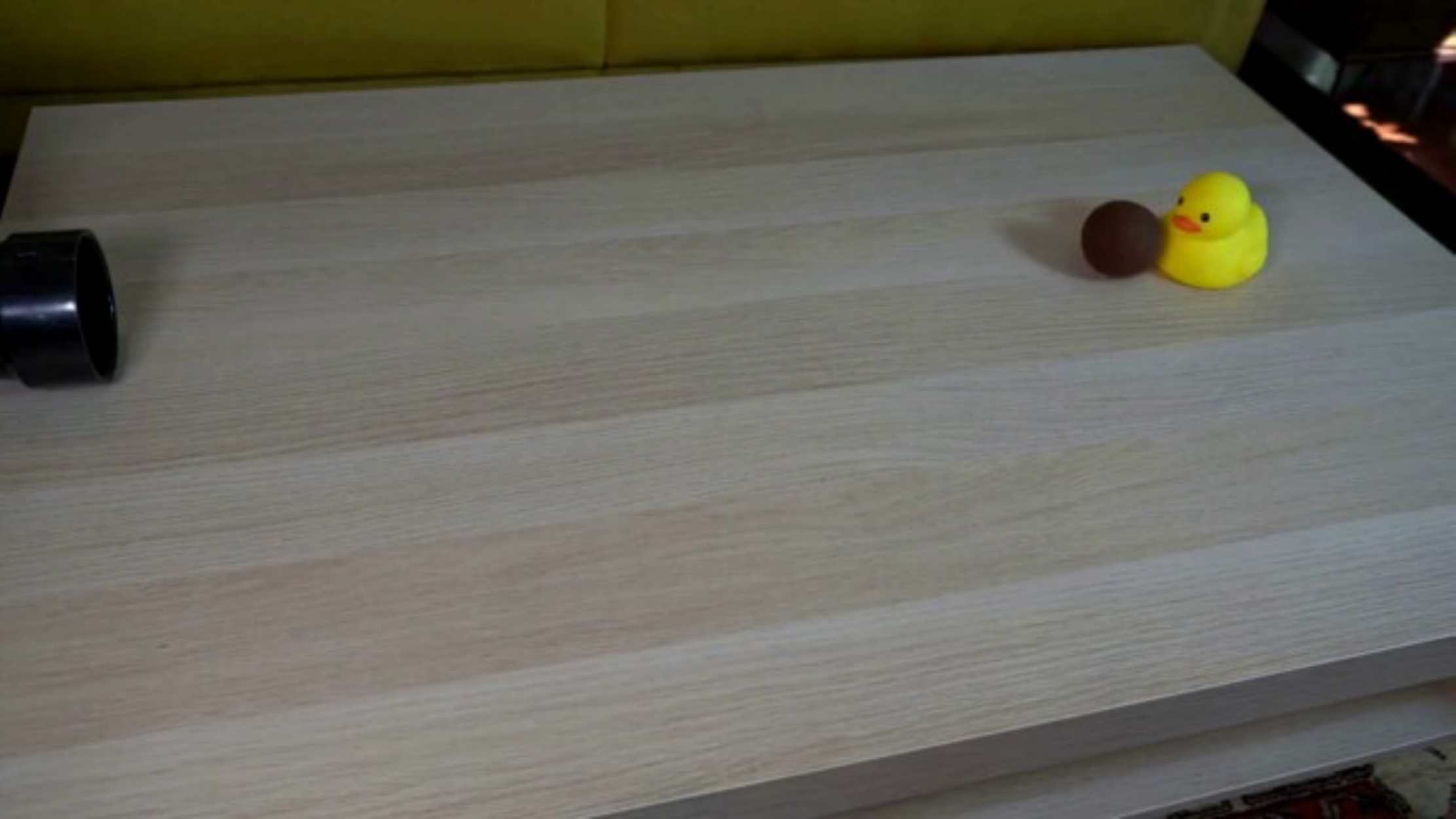} &
                \includegraphics[width=0.088\textwidth]{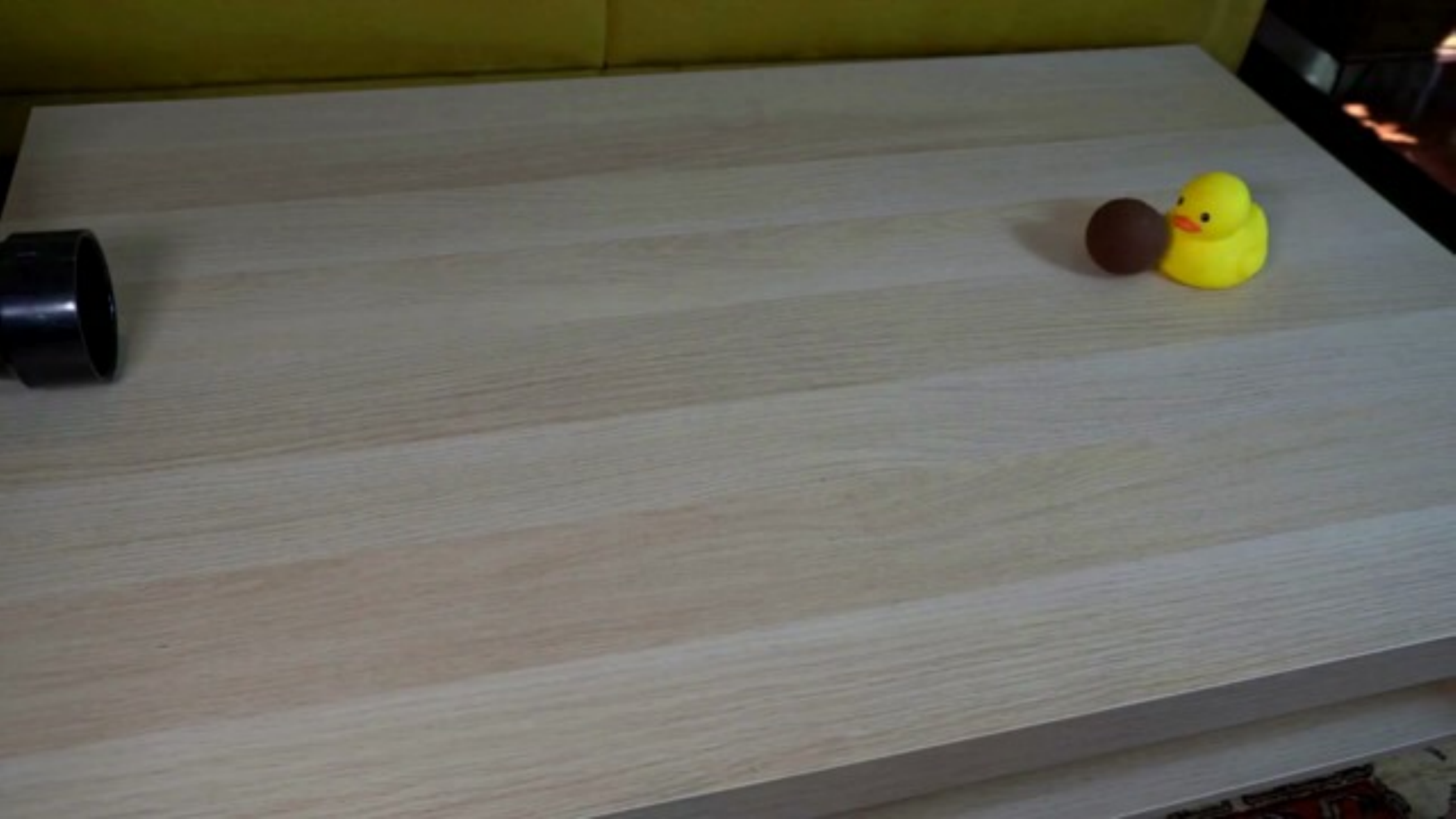} &
                \includegraphics[width=0.088\textwidth]{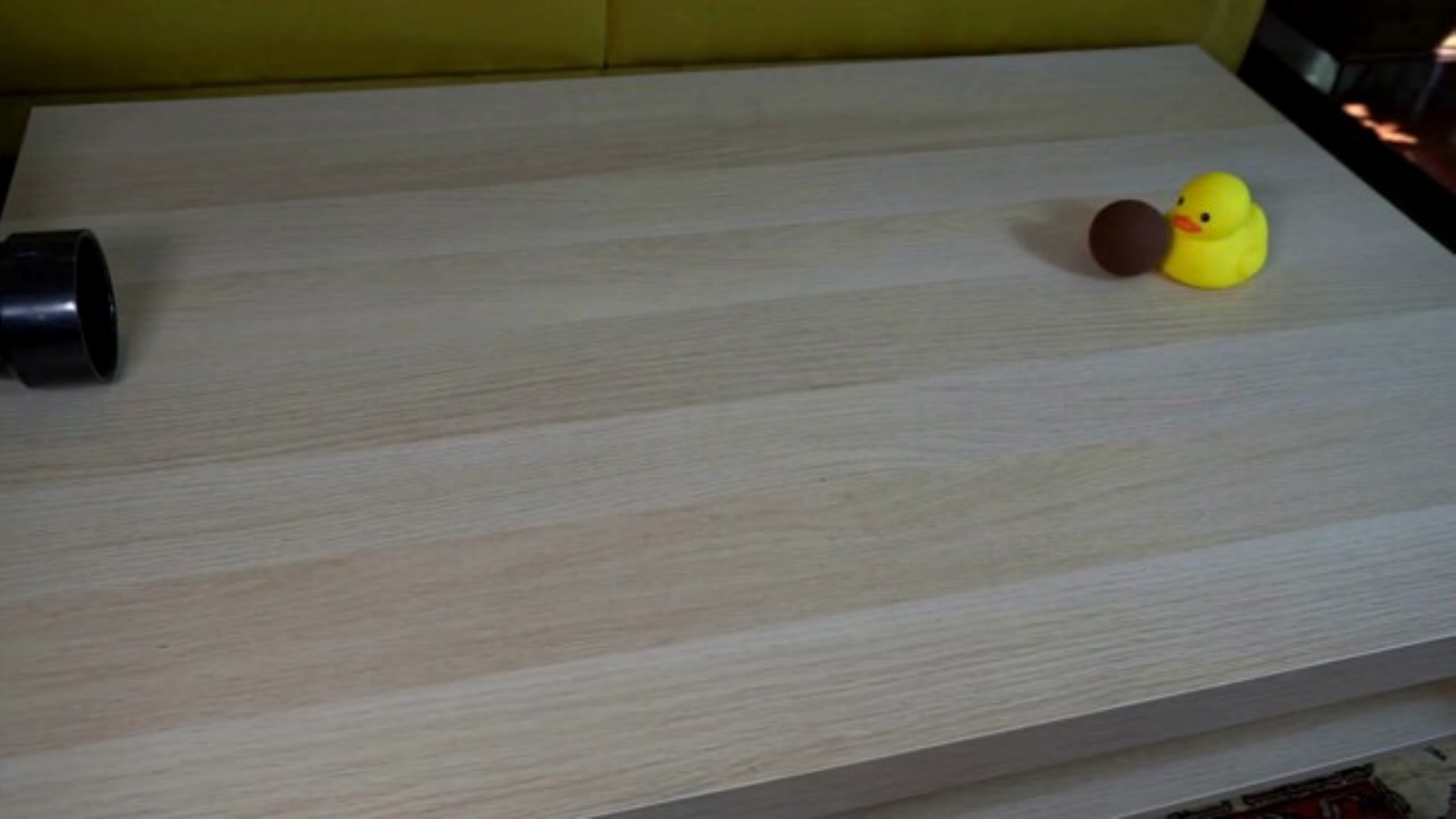} &
                \includegraphics[width=0.088\textwidth]{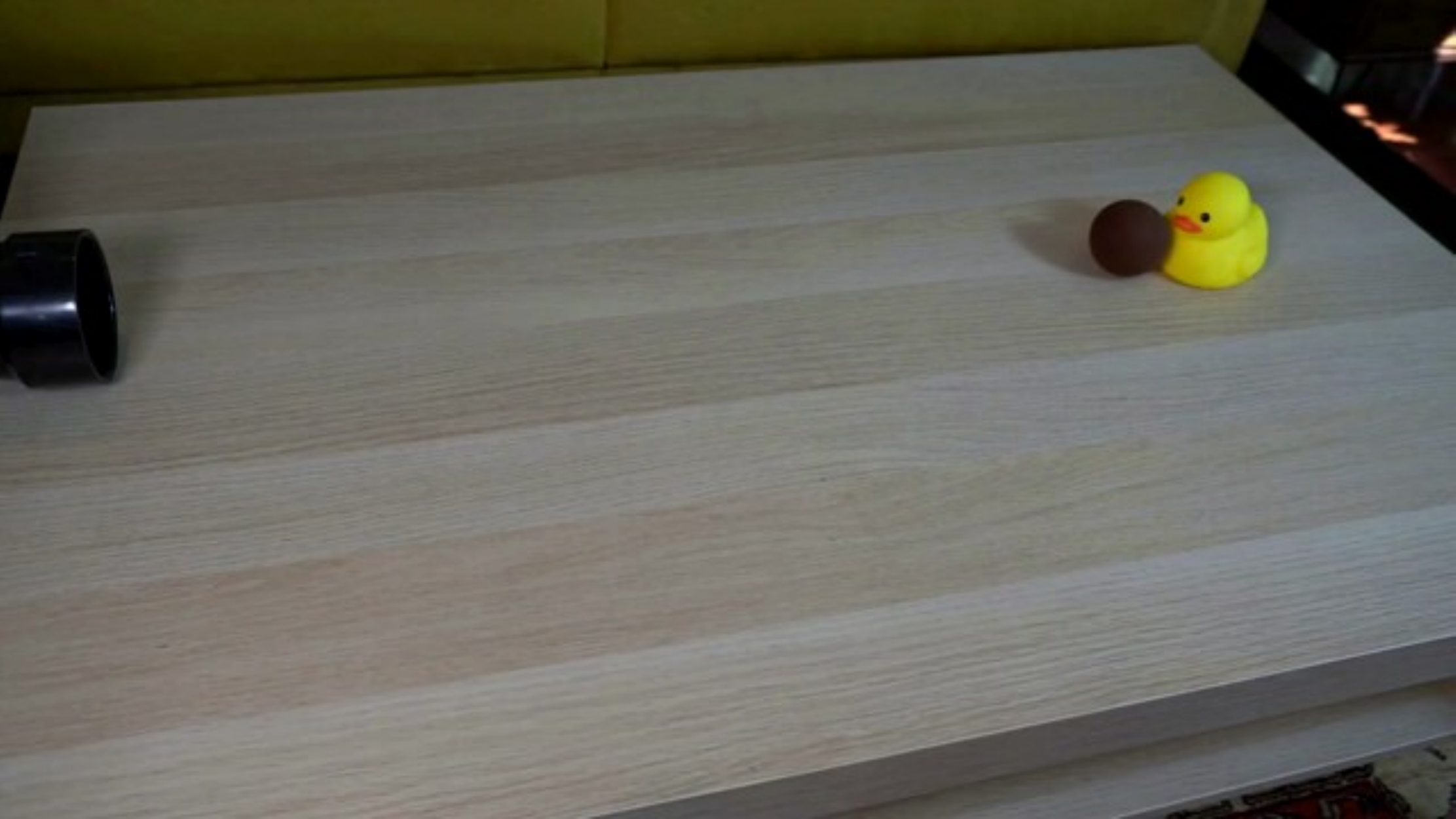} &
                \includegraphics[width=0.088\textwidth]{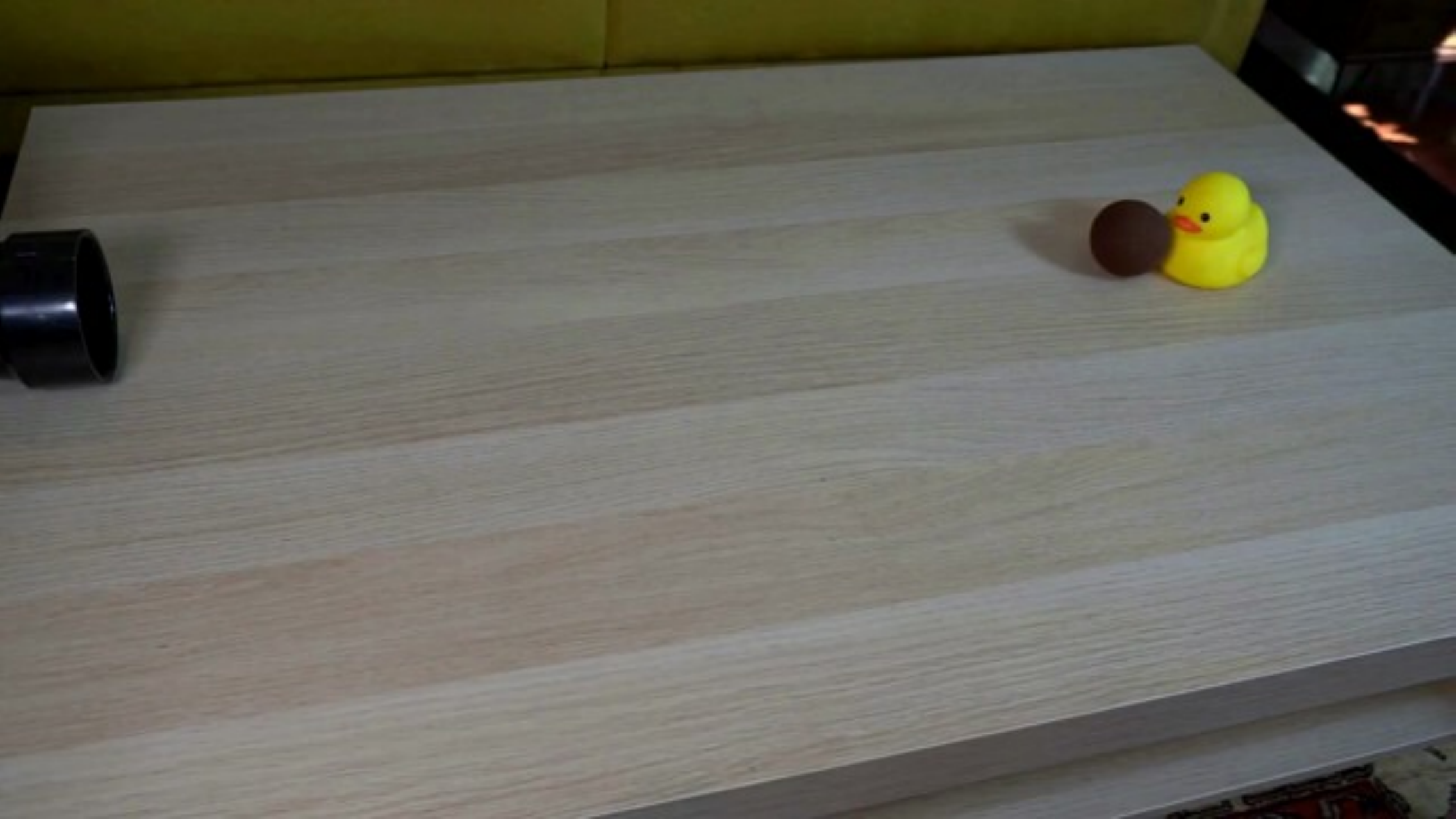} &
                \includegraphics[width=0.088\textwidth]{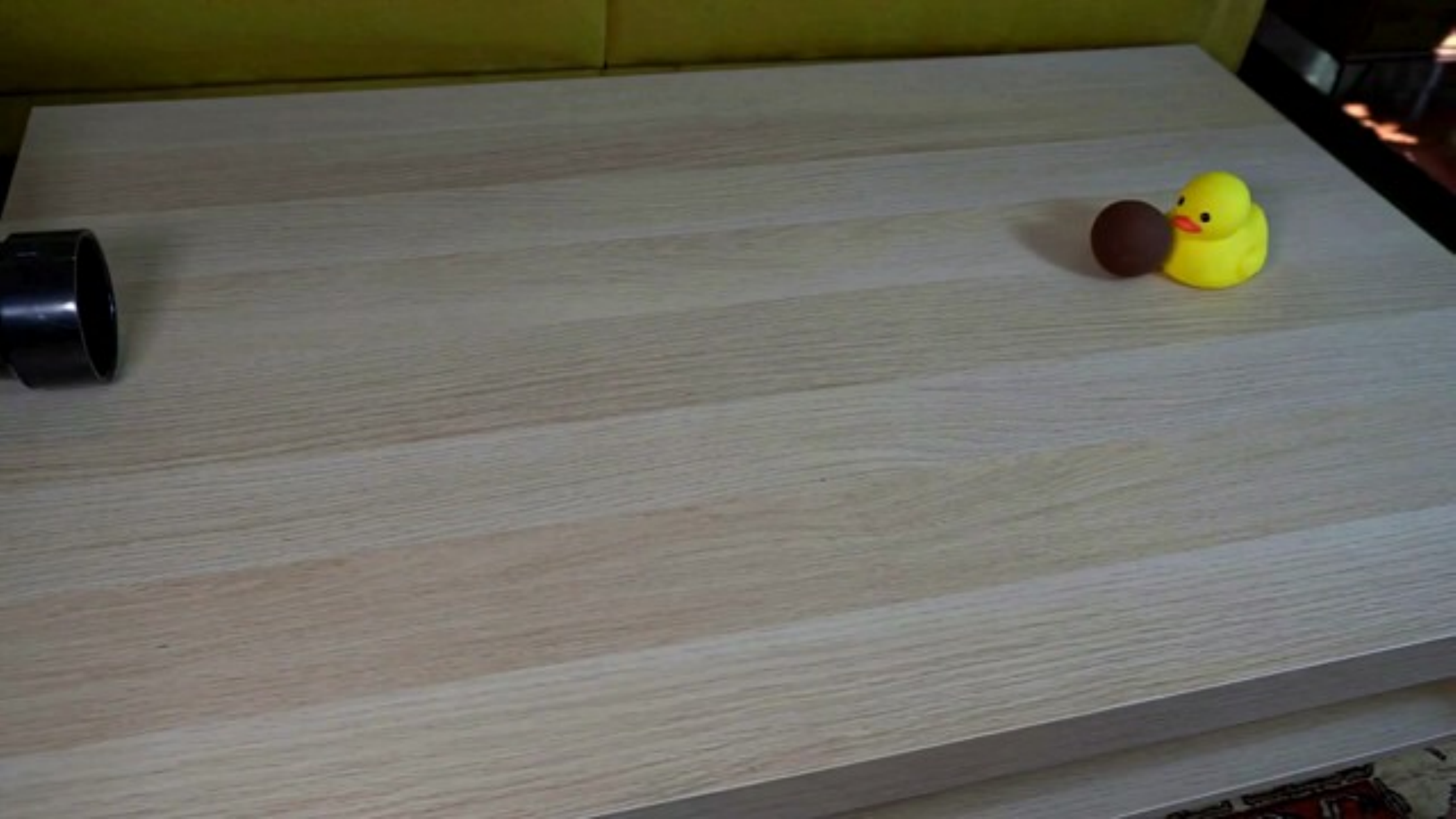}
            \end{tabular}
            \caption{A light beige coffee table with a small yellow rubber ducky on it. A mustard yellow couch is in the background. There is a black pipe on one end of the table and a brown tennis ball rolls out of it towards the rubber ducky. Static shot with no camera movement.}
            \label{fig:case1}
        \end{subfigure}
    \end{minipage}
    \begin{minipage}{1.0\textwidth}
        \begin{subfigure}[b]{1.0\textwidth}
            \centering
            \vspace{1pt}  
            \begin{tabular}{@{}c@{}c@{}c@{\hspace{1pt}}c@{\hspace{1pt}}c@{\hspace{1pt}}c@{\hspace{1pt}}c@{\hspace{1pt}}c@{\hspace{1pt}}c@{\hspace{1pt}}c@{\hspace{1pt}}c@{\hspace{1pt}}c@{}}
                & 0.0s & 2.5s & 3.0s & 3.3s & 3.7s & 4.0s & 4.3s & 4.7s & 5.0s & 7.9s \\
                \raisebox{1.2\height}{\tiny \rotatebox[origin=c]{90}{Real}   }&
                \includegraphics[width=0.088\textwidth]{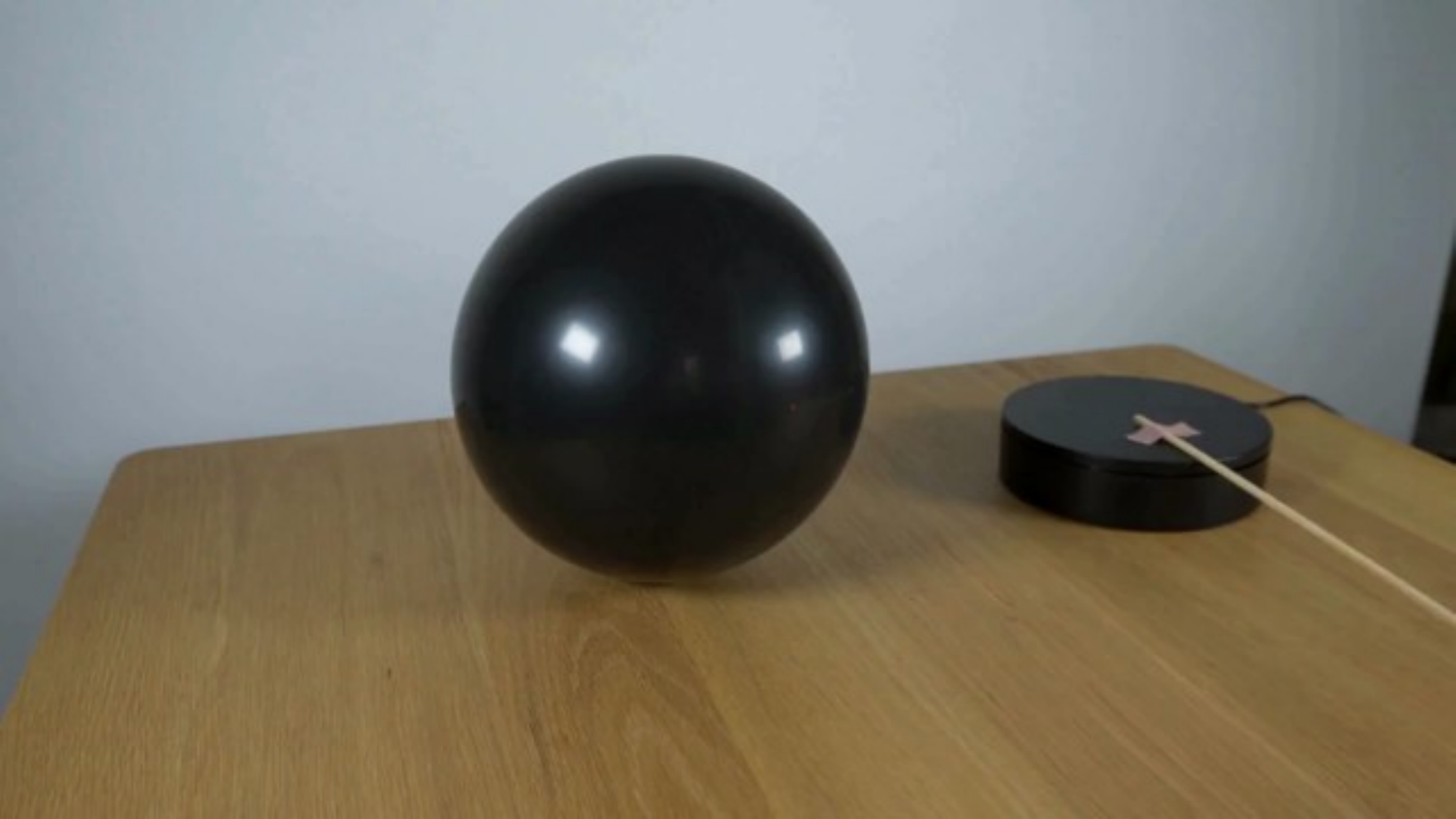} &
                \includegraphics[width=0.088\textwidth]{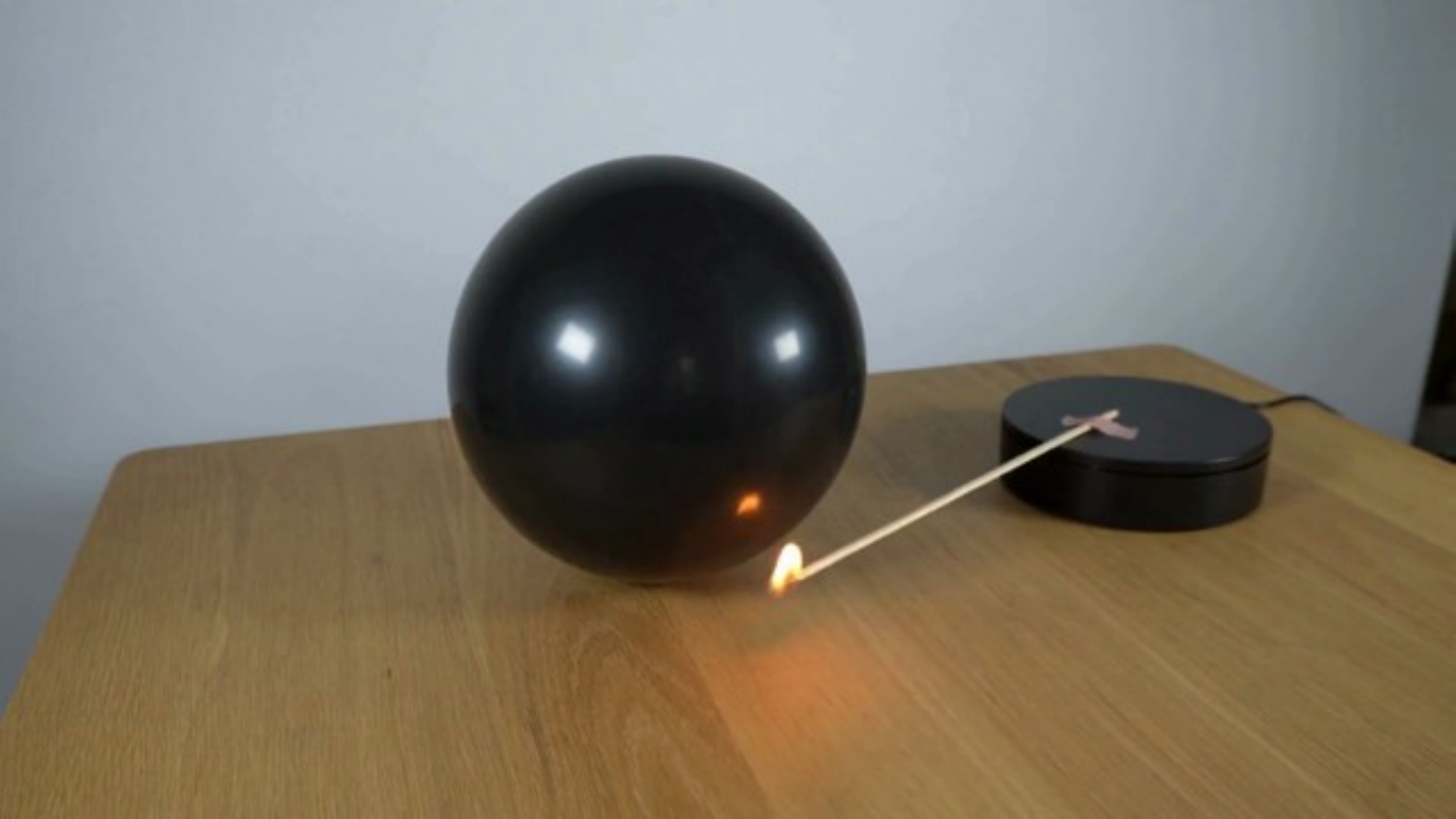} &
                \includegraphics[width=0.088\textwidth]{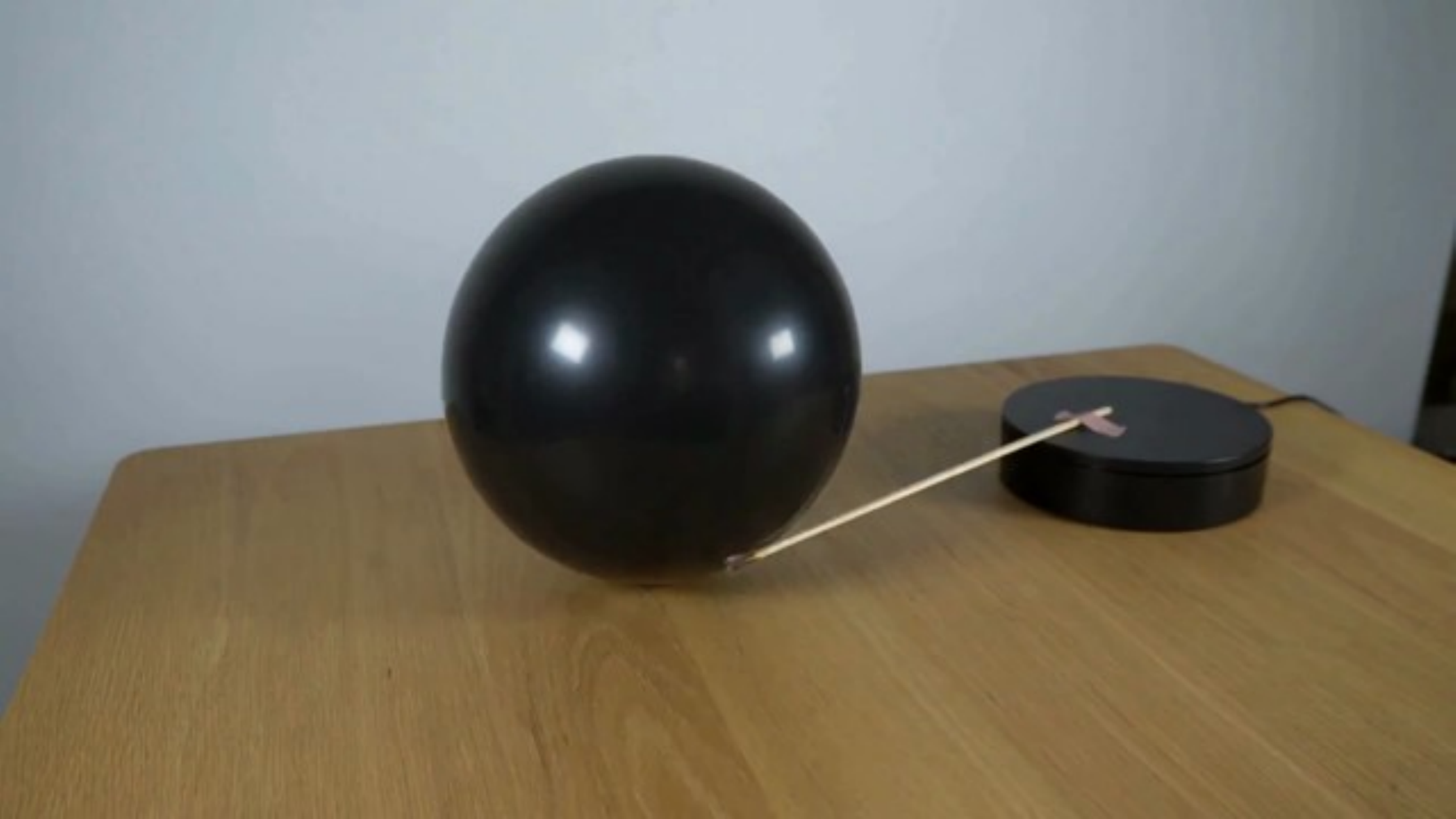} &
                \includegraphics[width=0.088\textwidth]{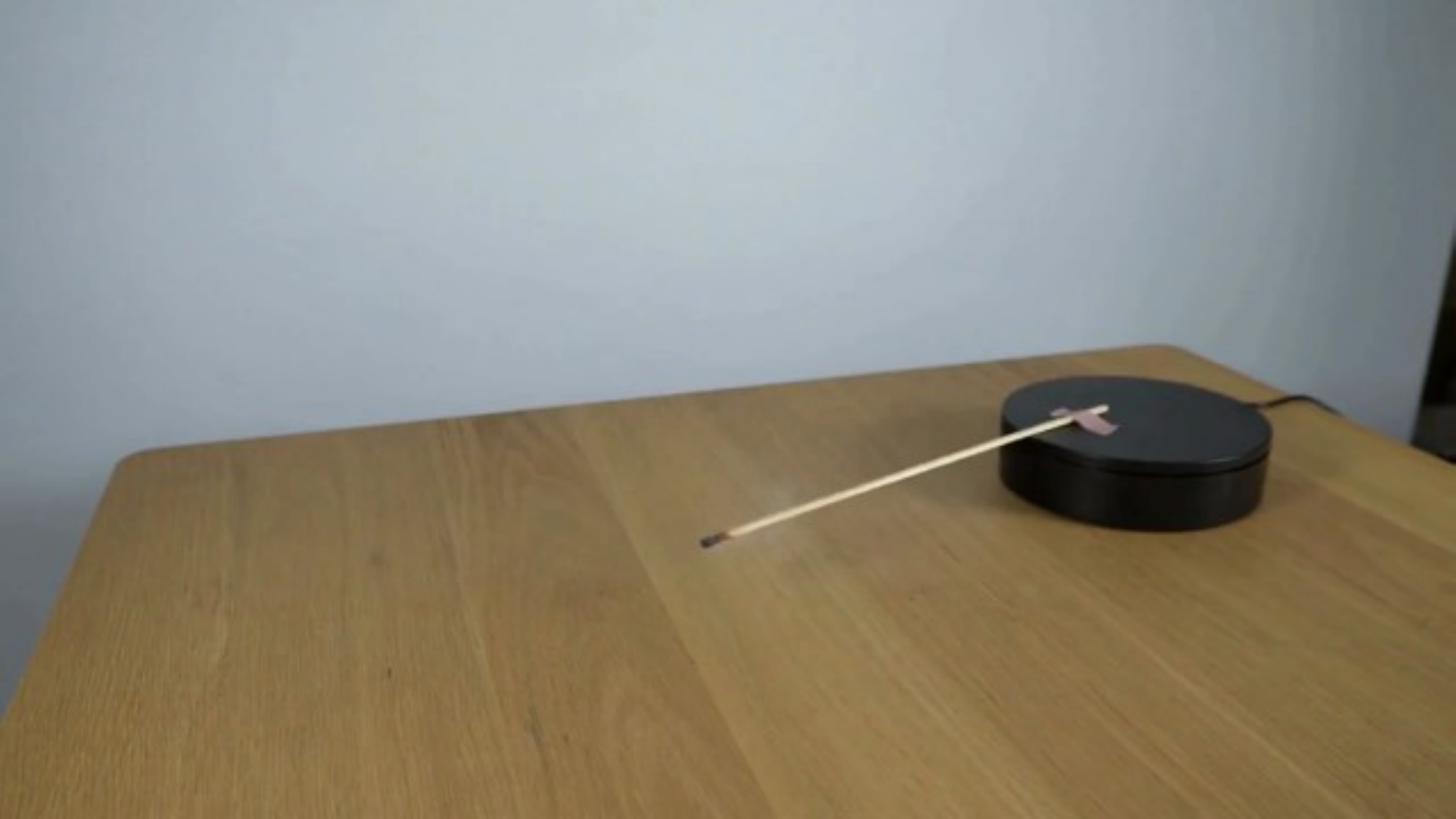} &
                \includegraphics[width=0.088\textwidth]{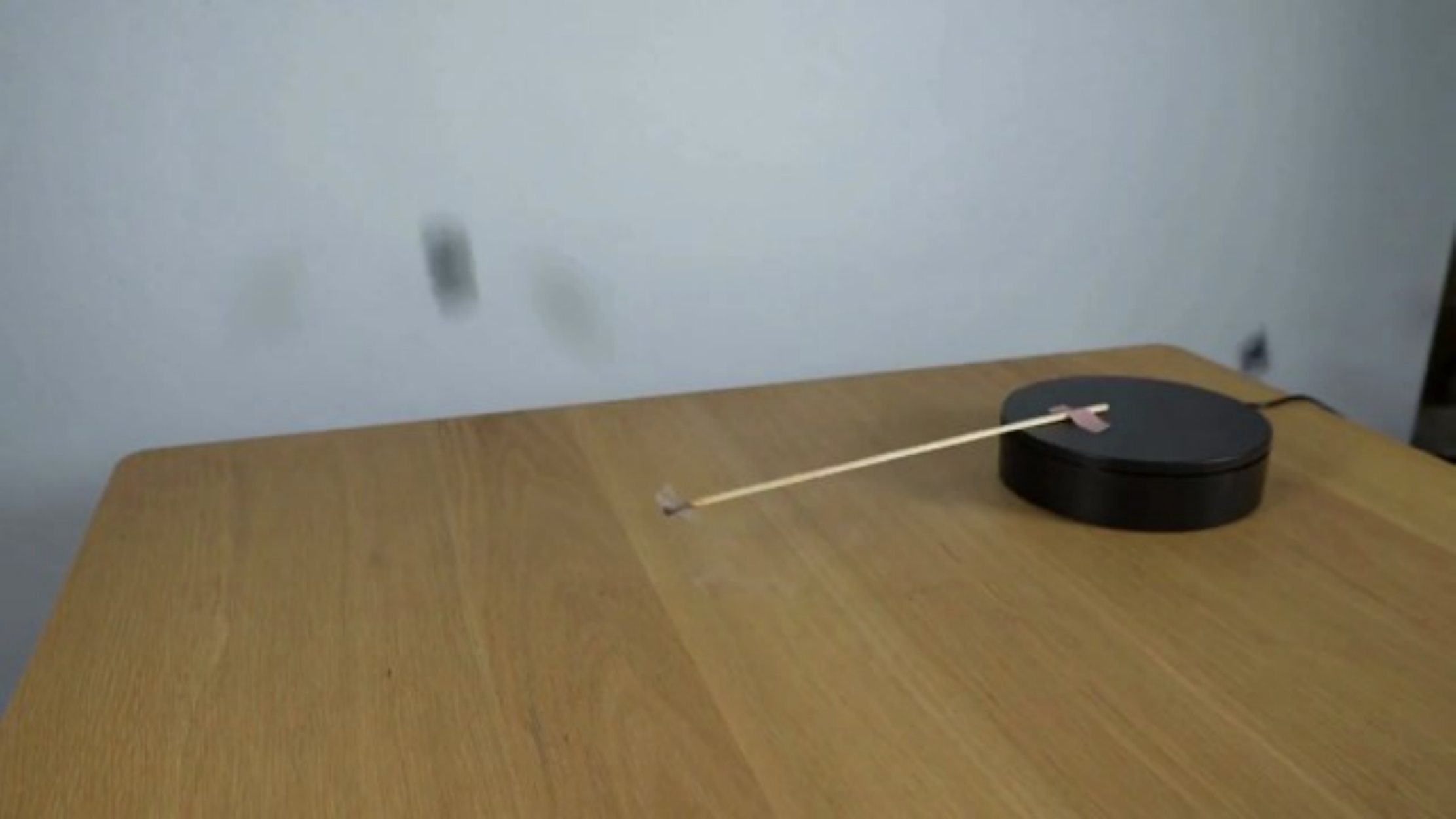} &
                \includegraphics[width=0.088\textwidth]{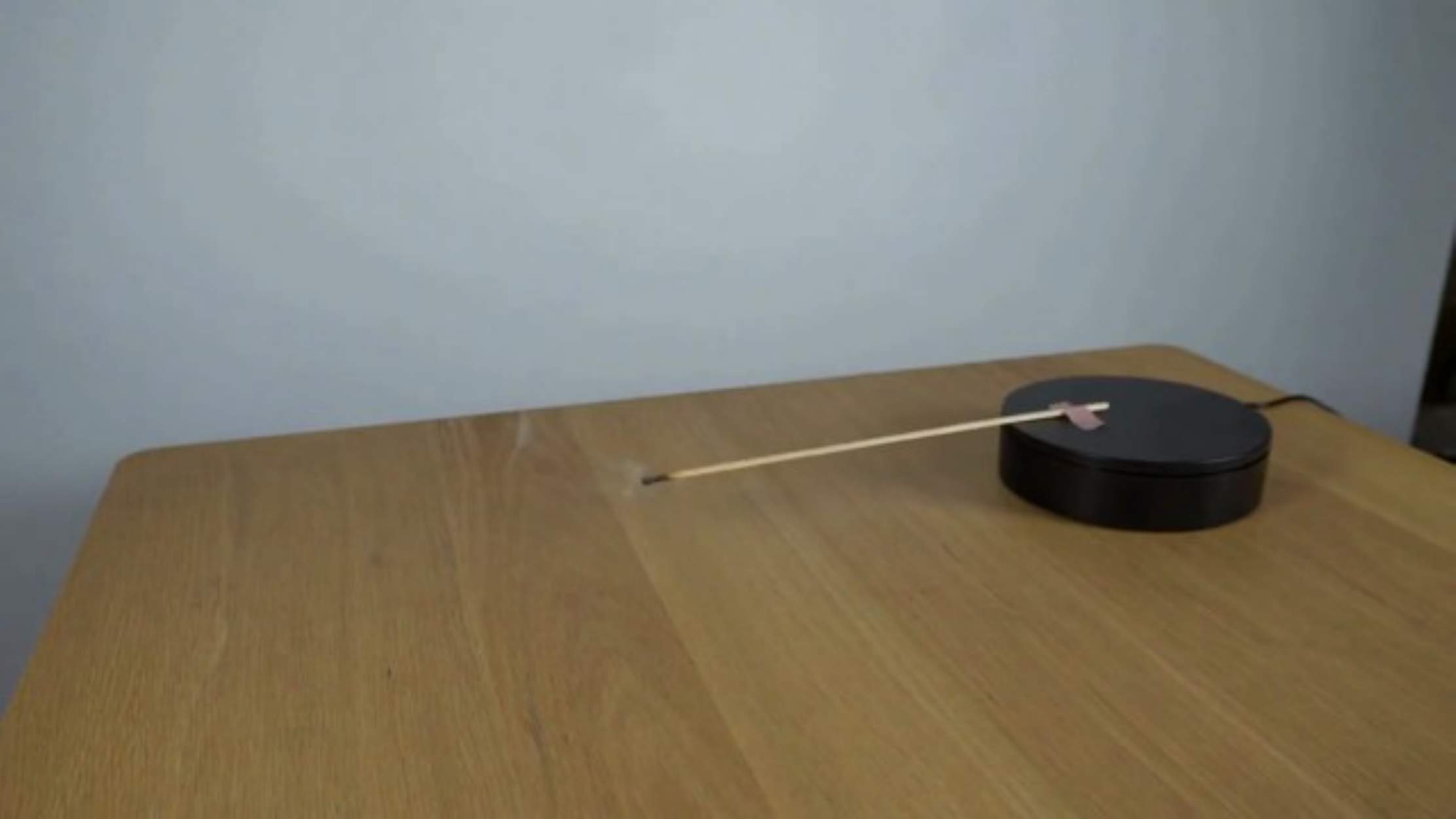} &
                \includegraphics[width=0.088\textwidth]{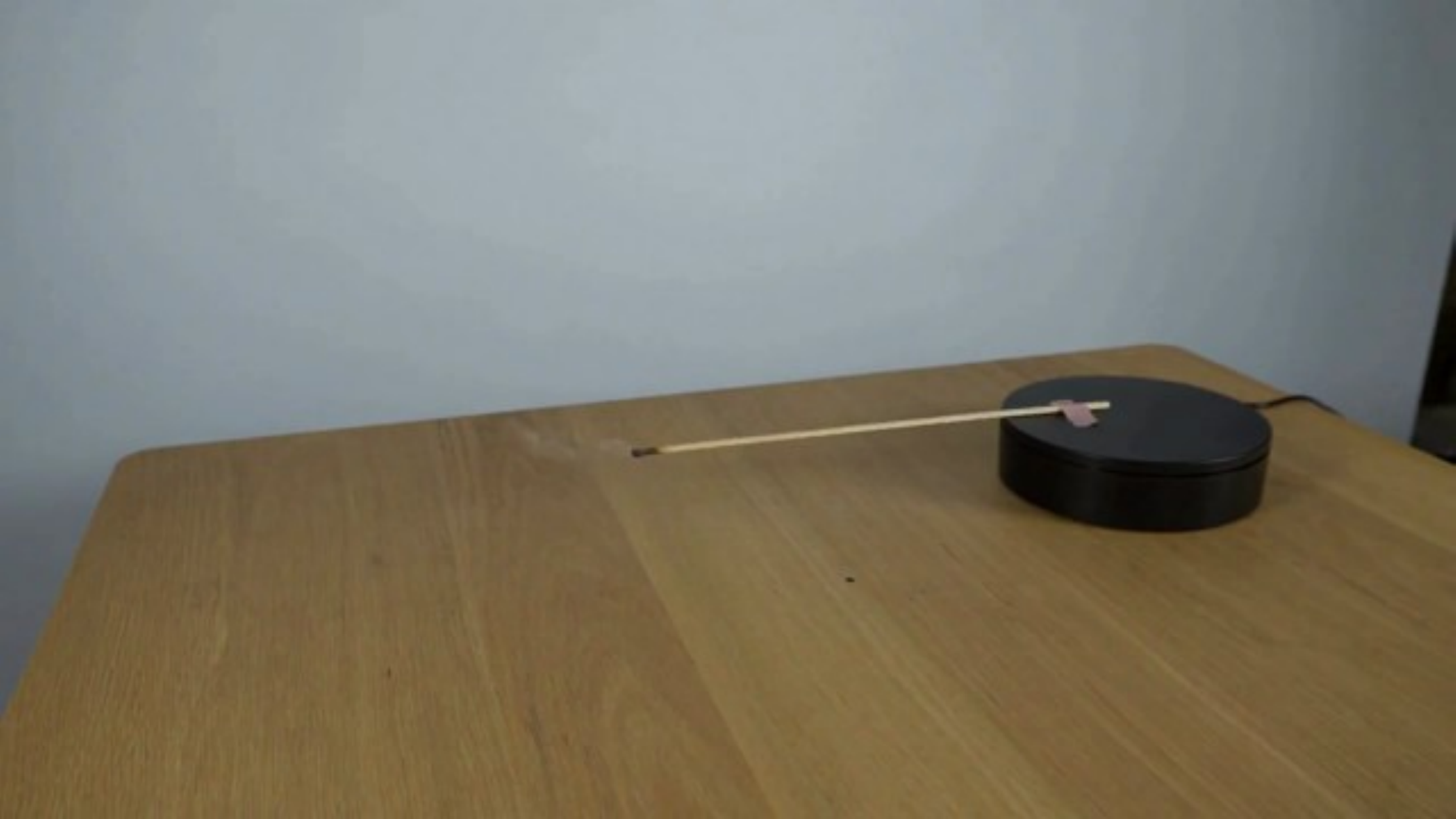} &
                \includegraphics[width=0.088\textwidth]{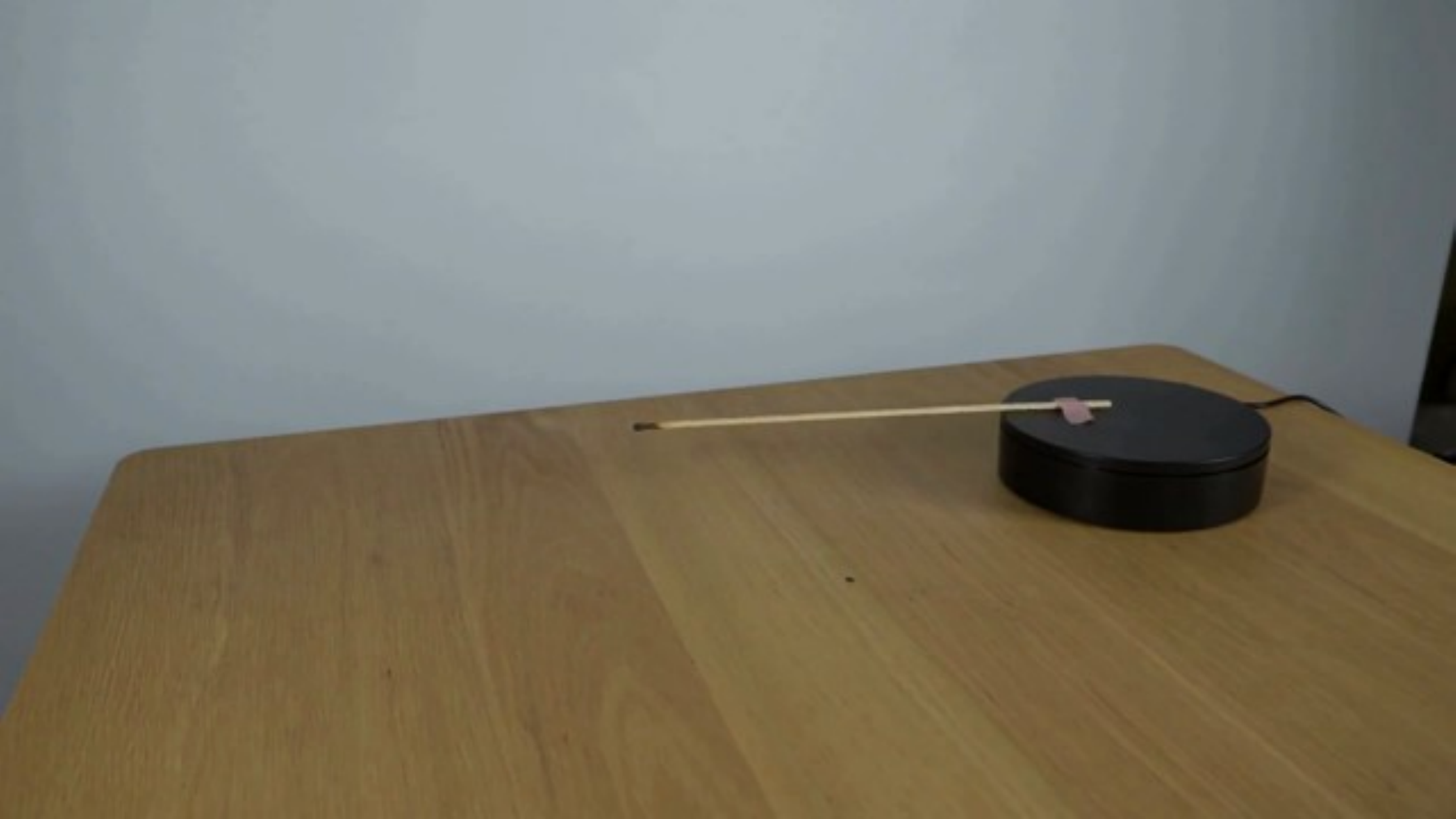} &
                \includegraphics[width=0.088\textwidth]{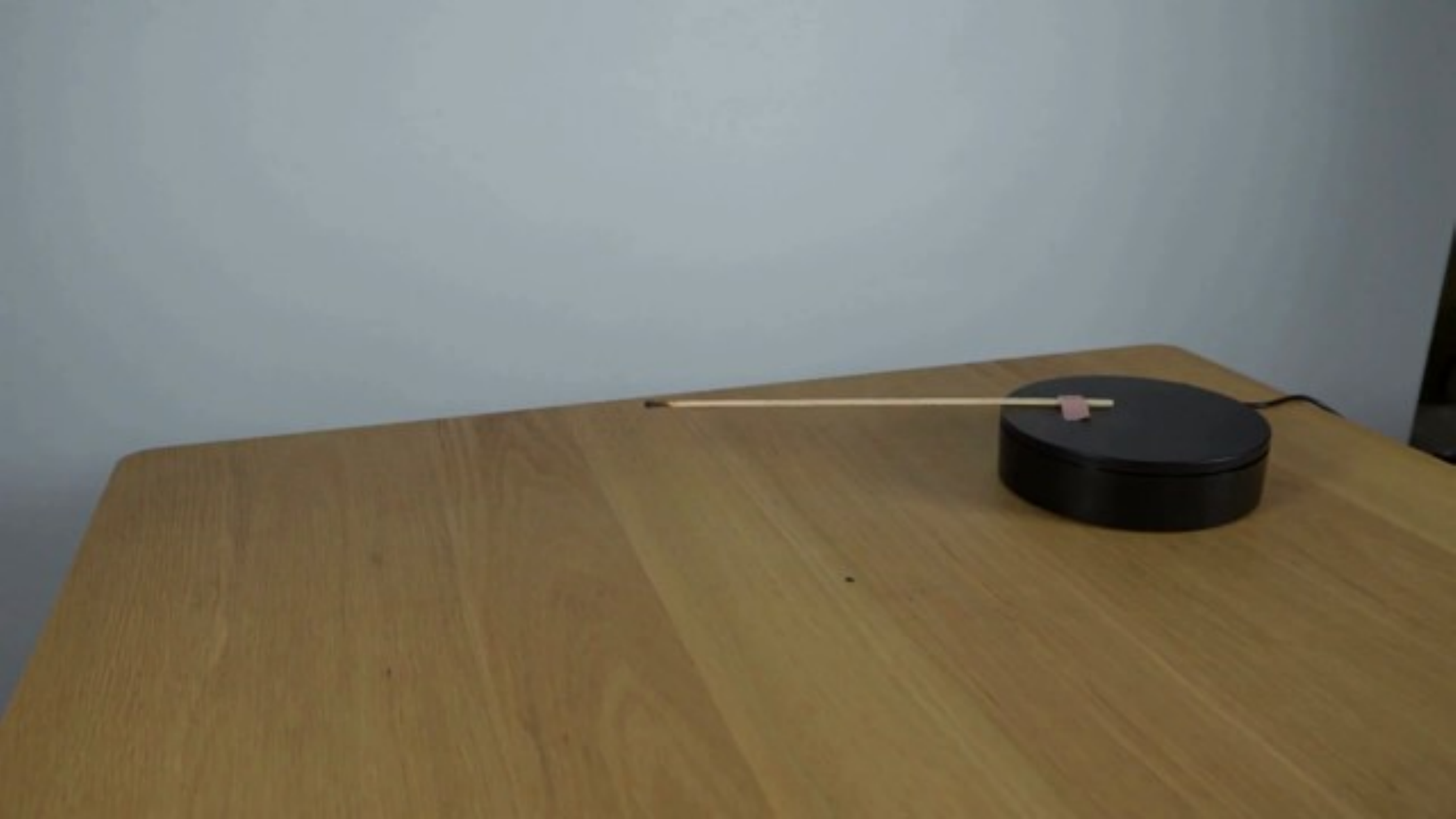} &
                \includegraphics[width=0.088\textwidth]{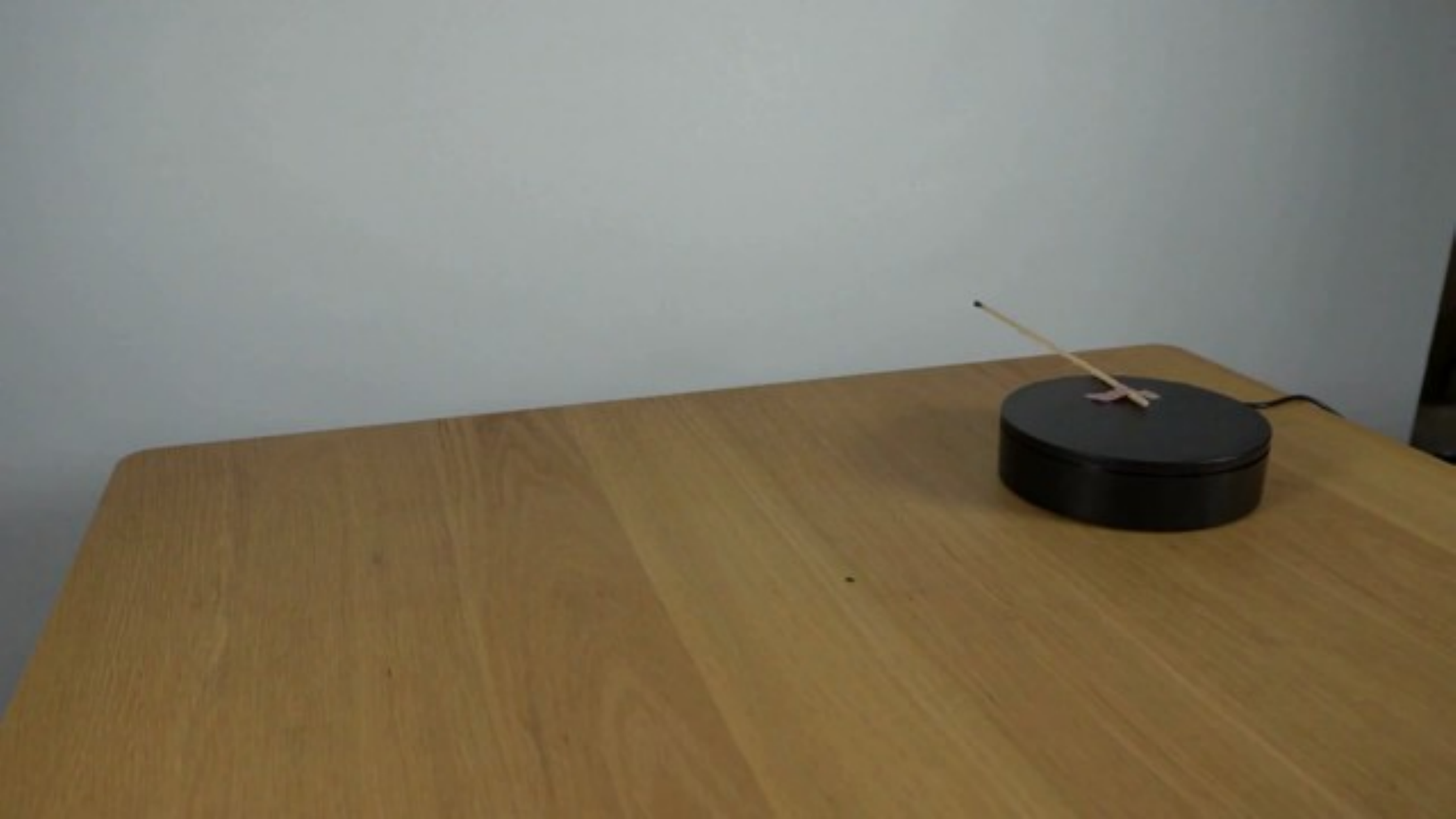}
            \end{tabular}
            \vspace{0.1pt}  
            \begin{tabular}{@{}c@{}c@{\hspace{1pt}}c@{\hspace{1pt}}c@{\hspace{1pt}}c@{\hspace{1pt}}c@{\hspace{1pt}}c@{\hspace{1pt}}c@{\hspace{1pt}}c@{\hspace{1pt}}c@{\hspace{1pt}}c@{}}

                \raisebox{0.6\height}{\tiny \rotatebox[origin=c]{90}{Generated}   }&
                \includegraphics[width=0.088\textwidth]{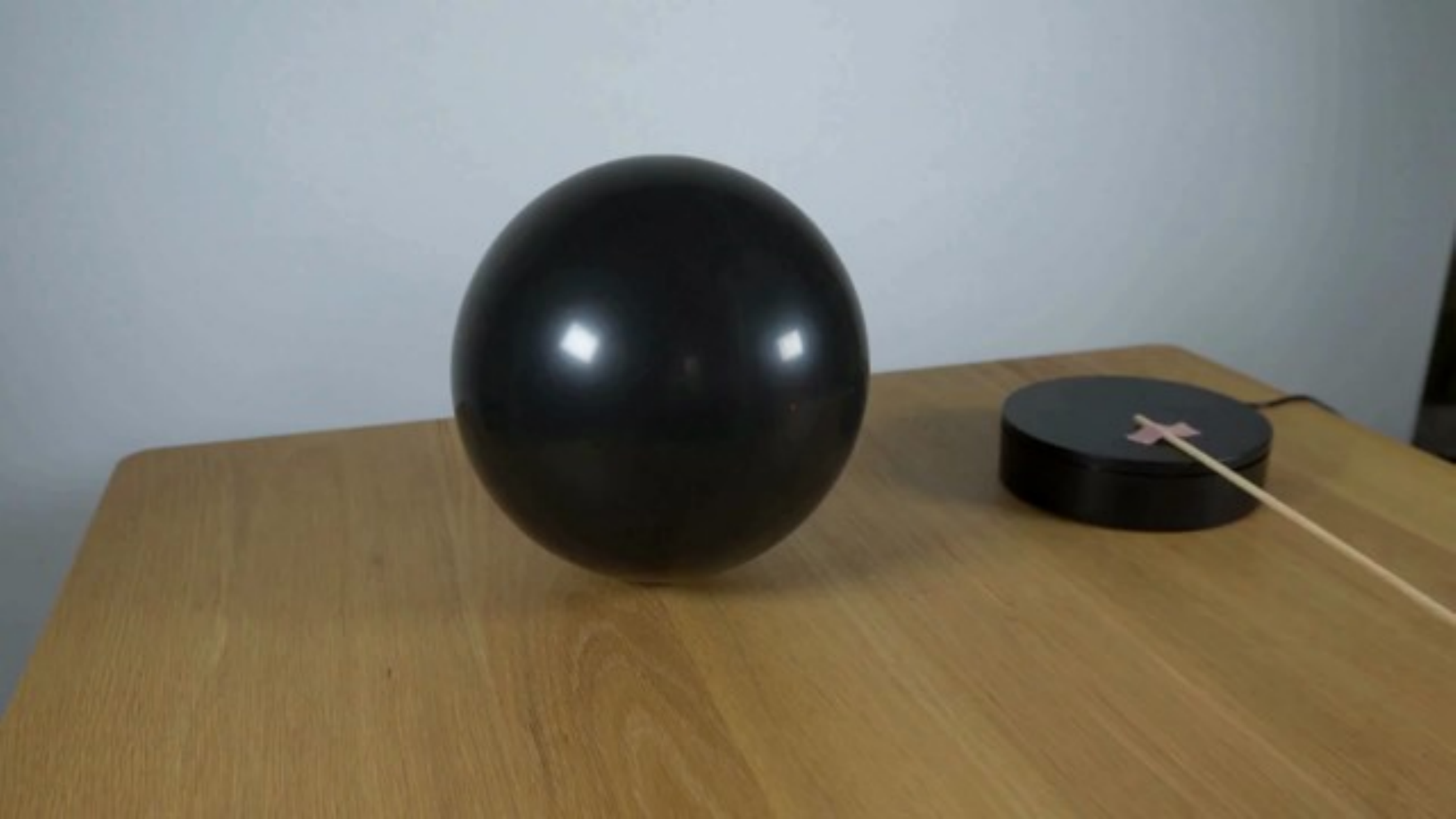} &
                \includegraphics[width=0.088\textwidth]{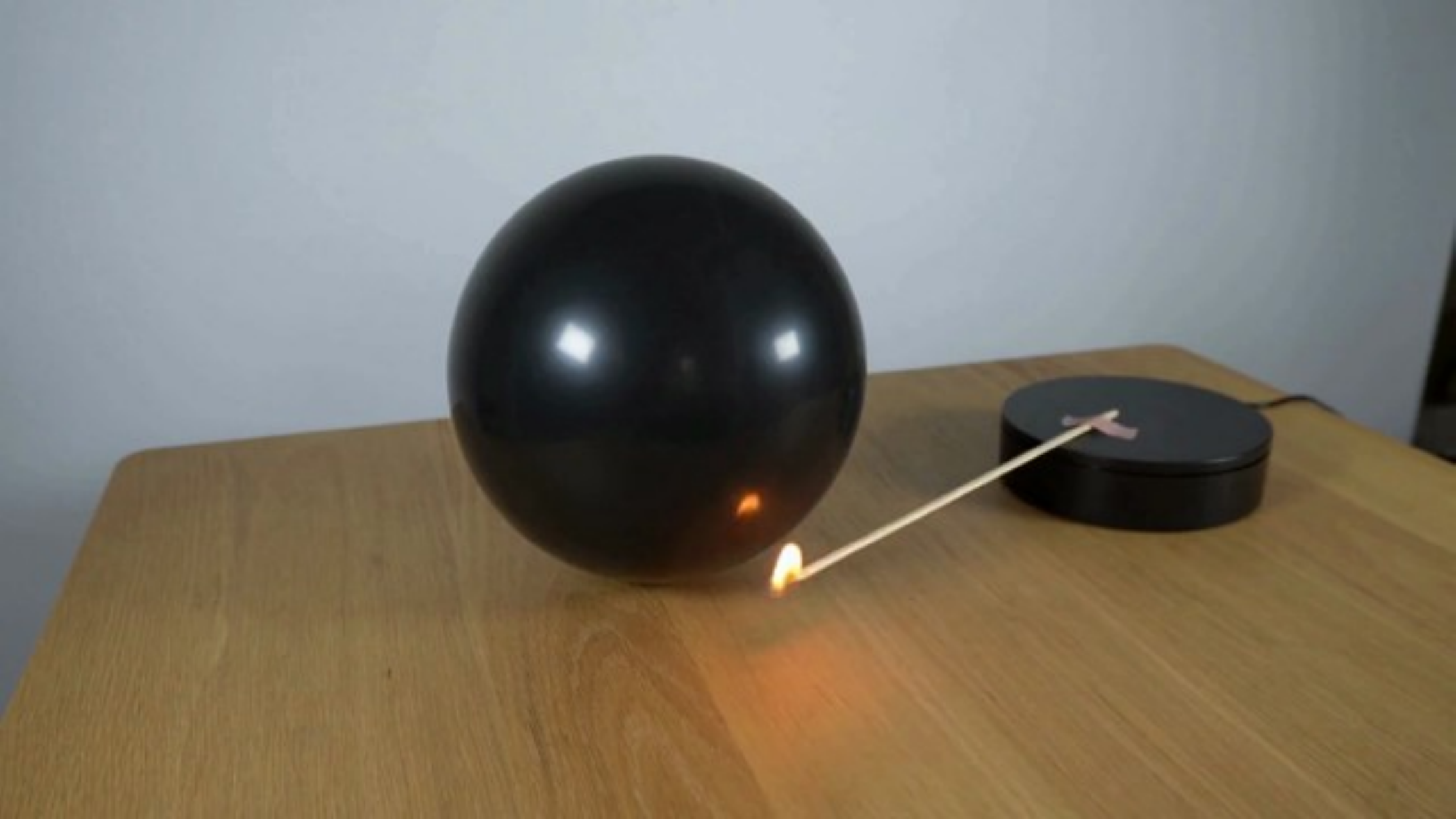} &
                \includegraphics[width=0.088\textwidth]{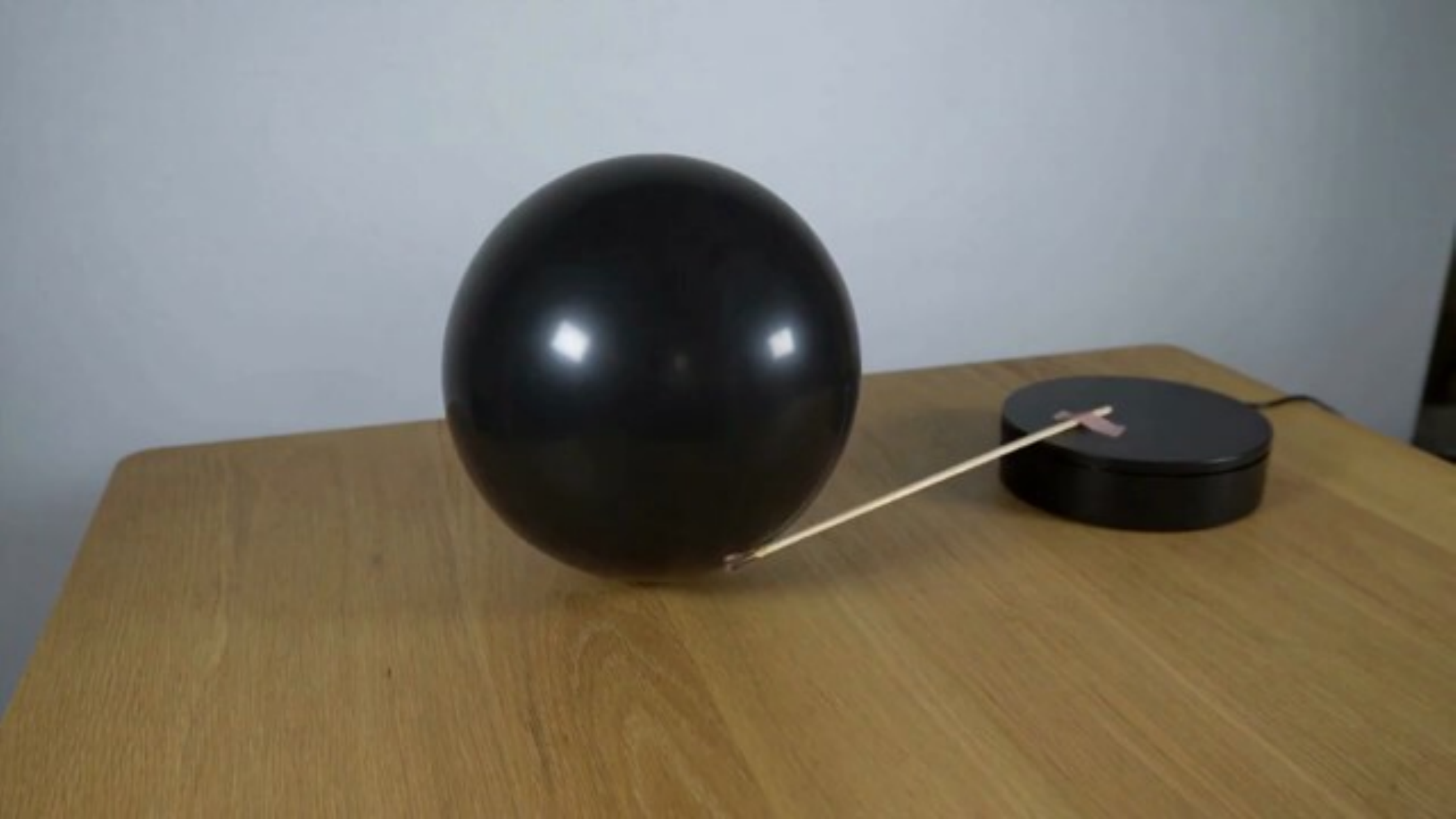} &
                \includegraphics[width=0.088\textwidth]{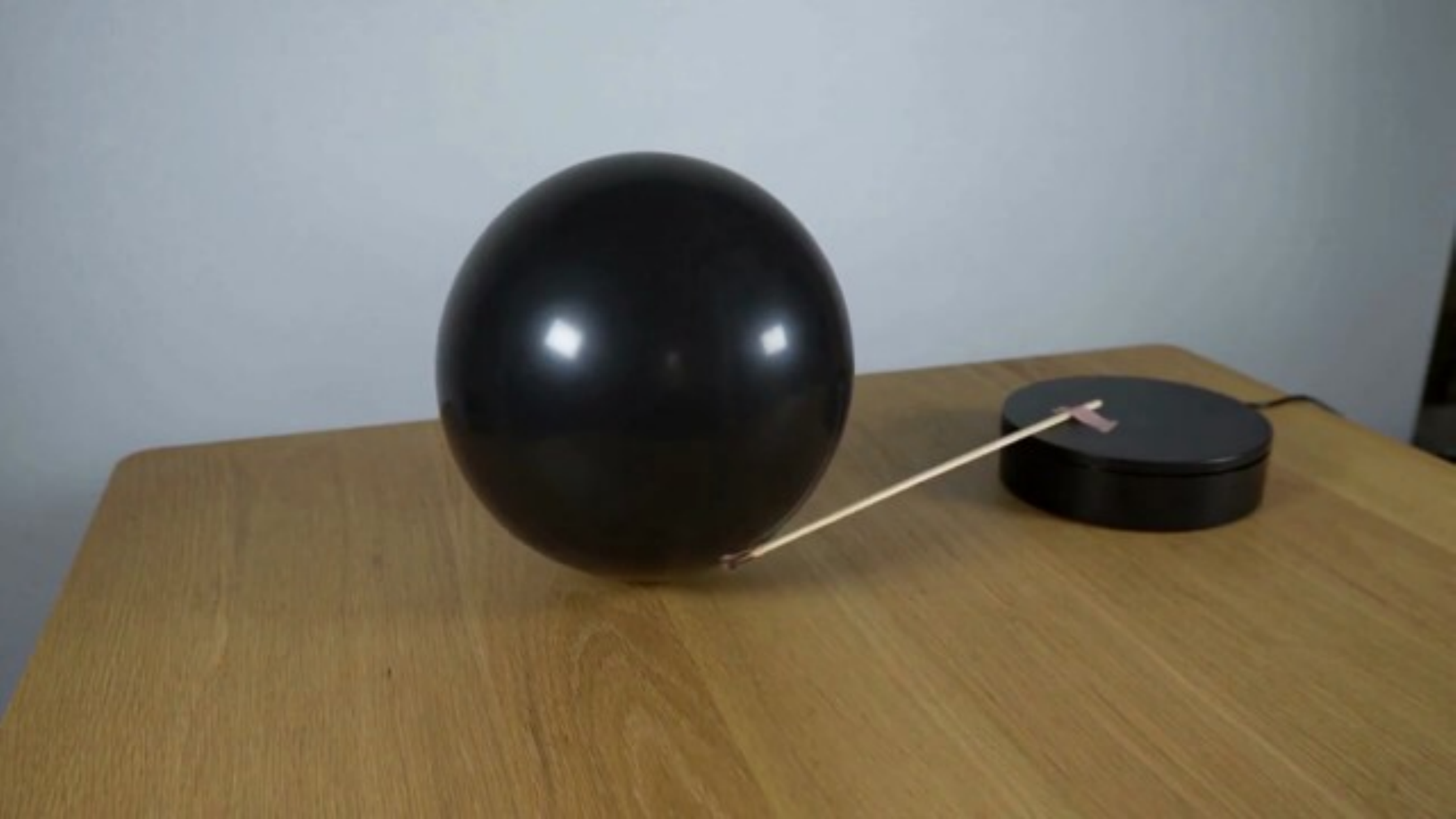} &
                \includegraphics[width=0.088\textwidth]{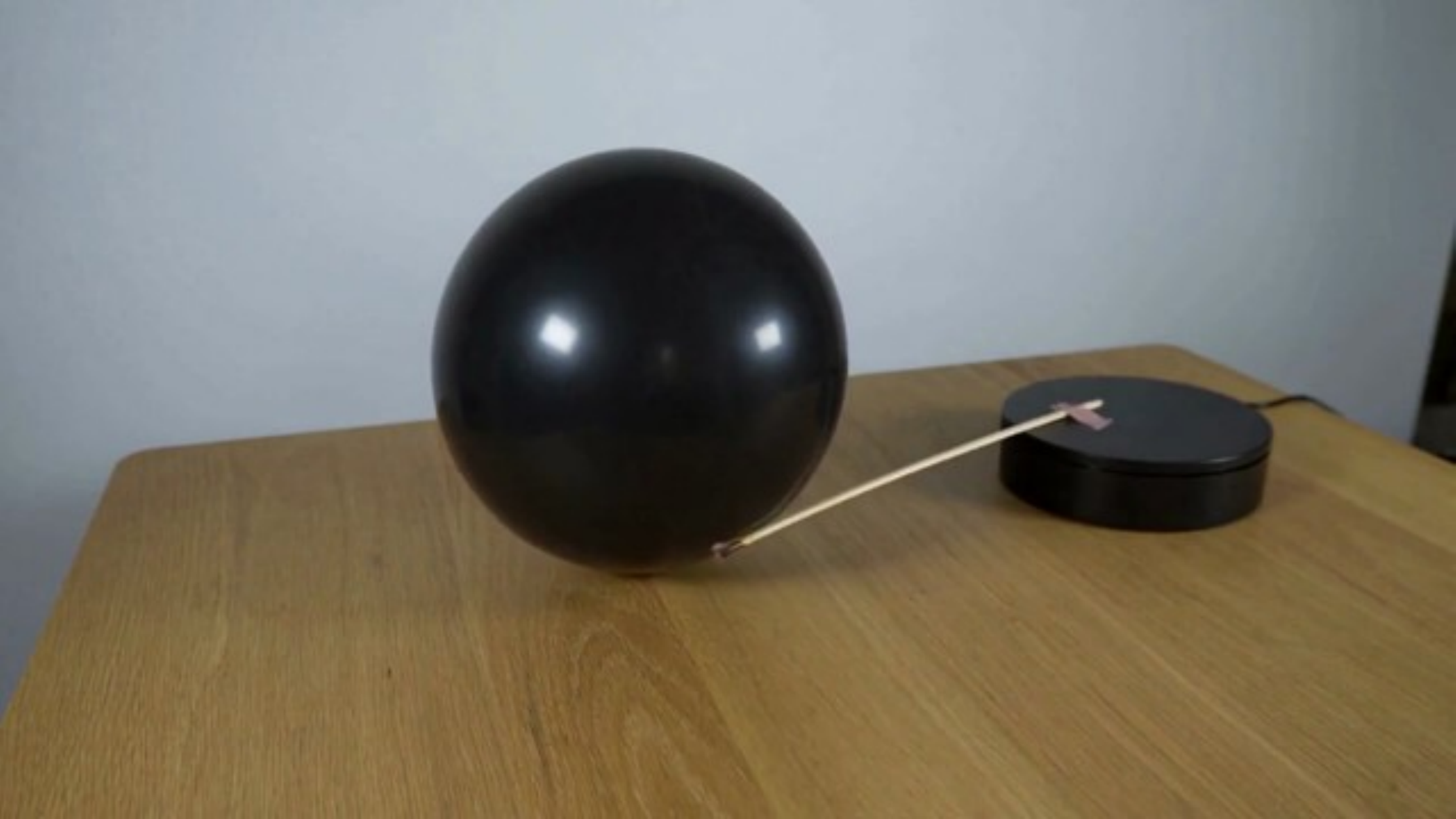} &
                \includegraphics[width=0.088\textwidth]{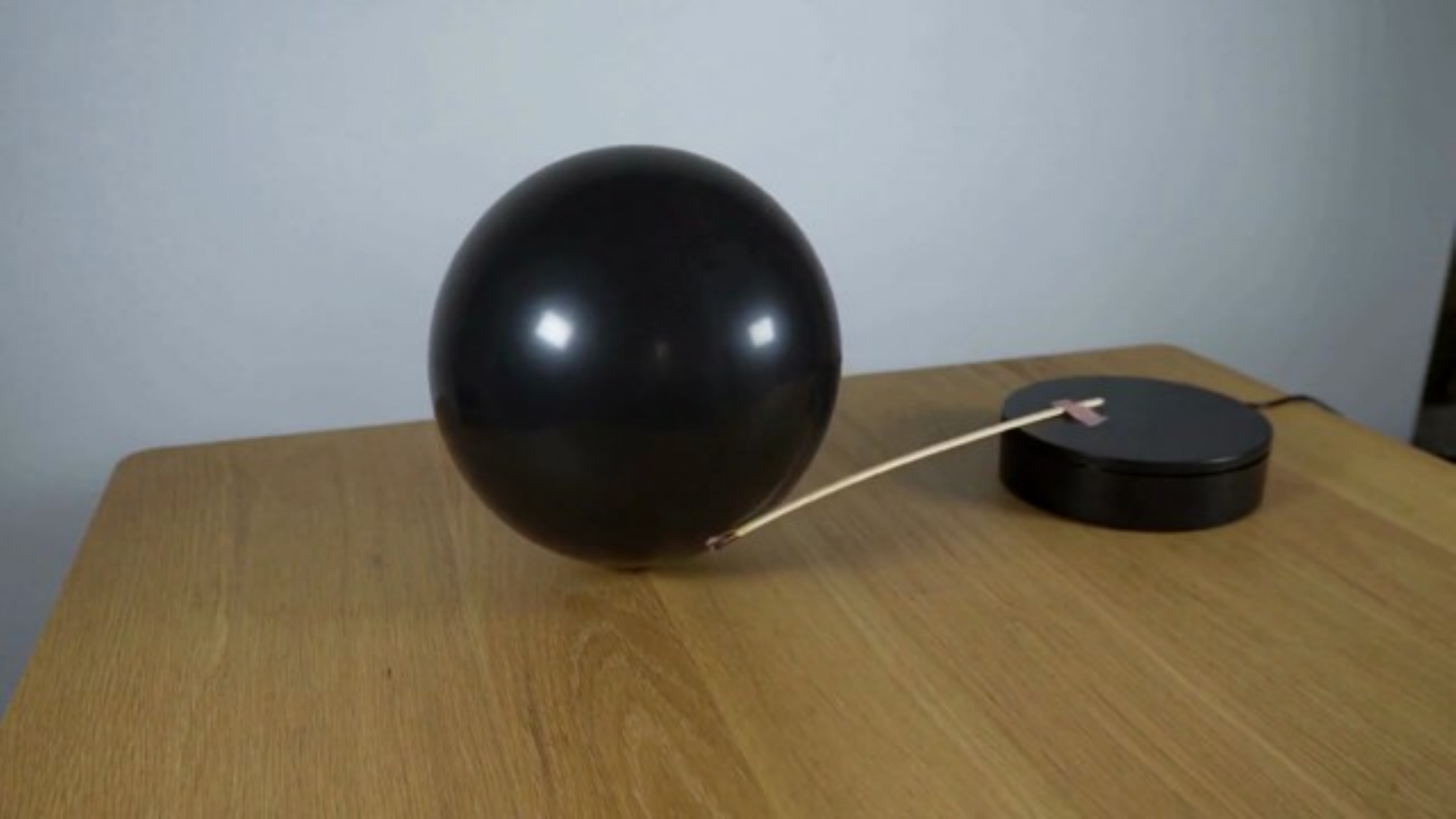} &
                \includegraphics[width=0.088\textwidth]{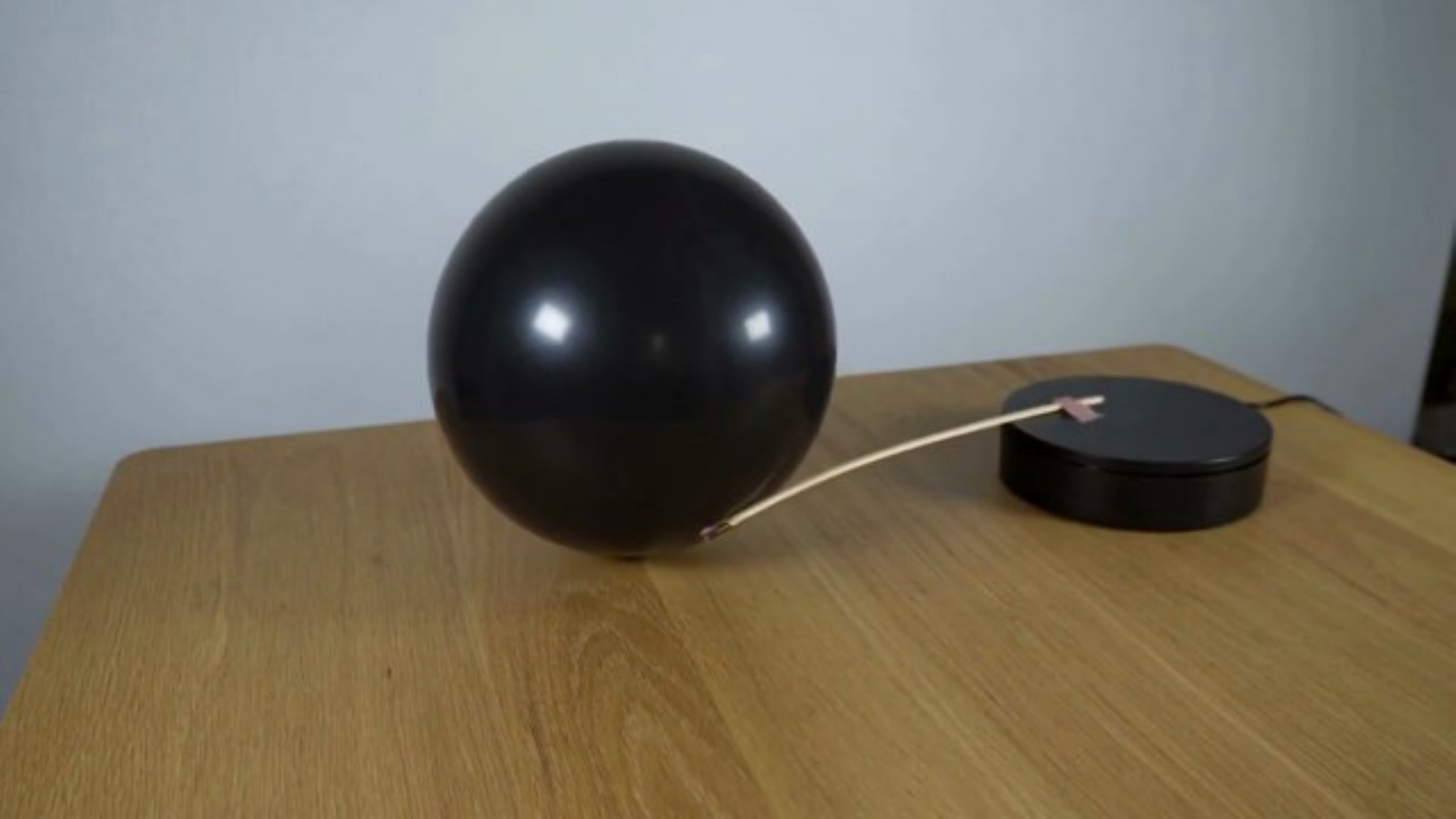} &
                \includegraphics[width=0.088\textwidth]{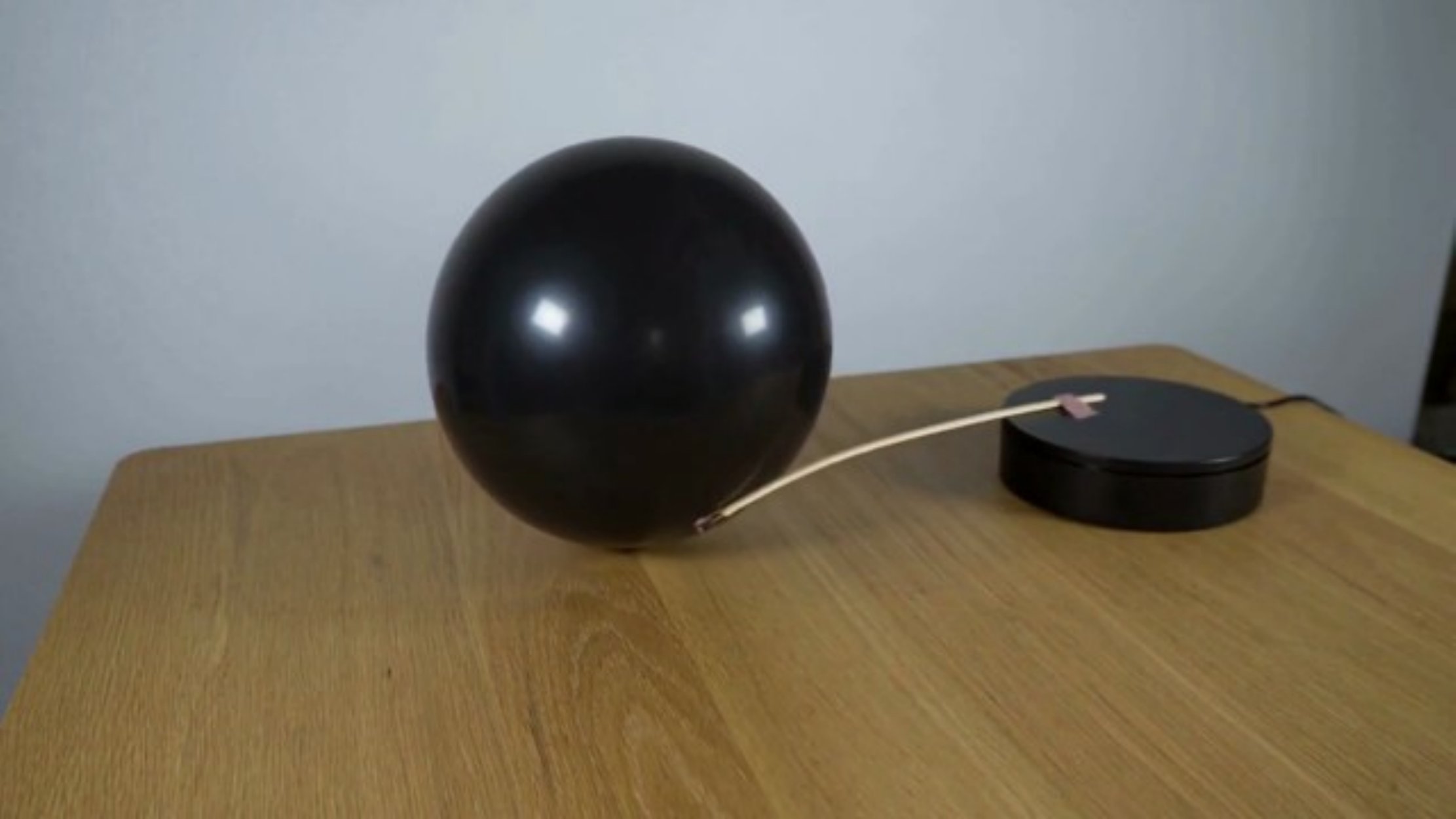} &
                \includegraphics[width=0.088\textwidth]{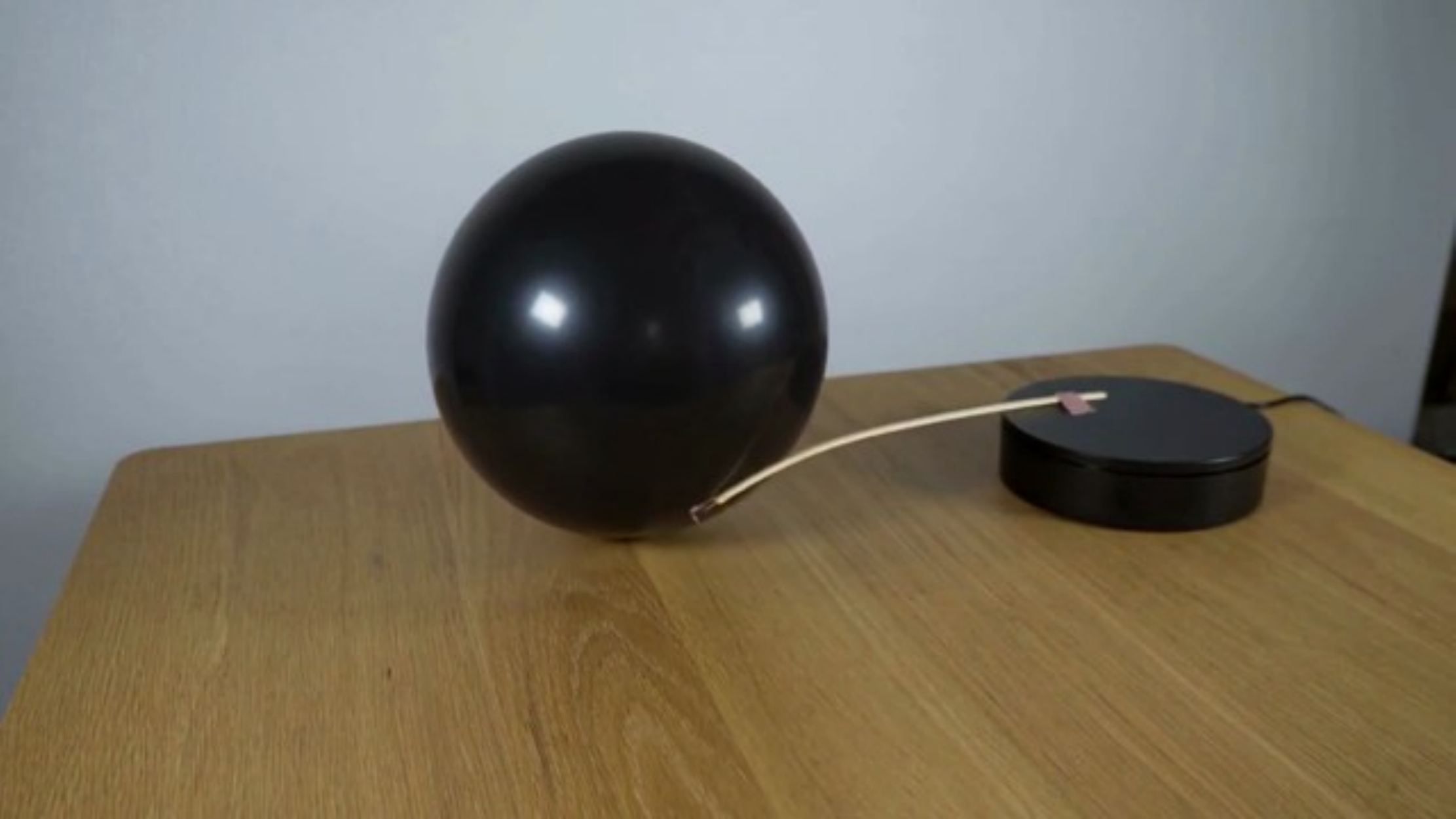} &
                \includegraphics[width=0.088\textwidth]{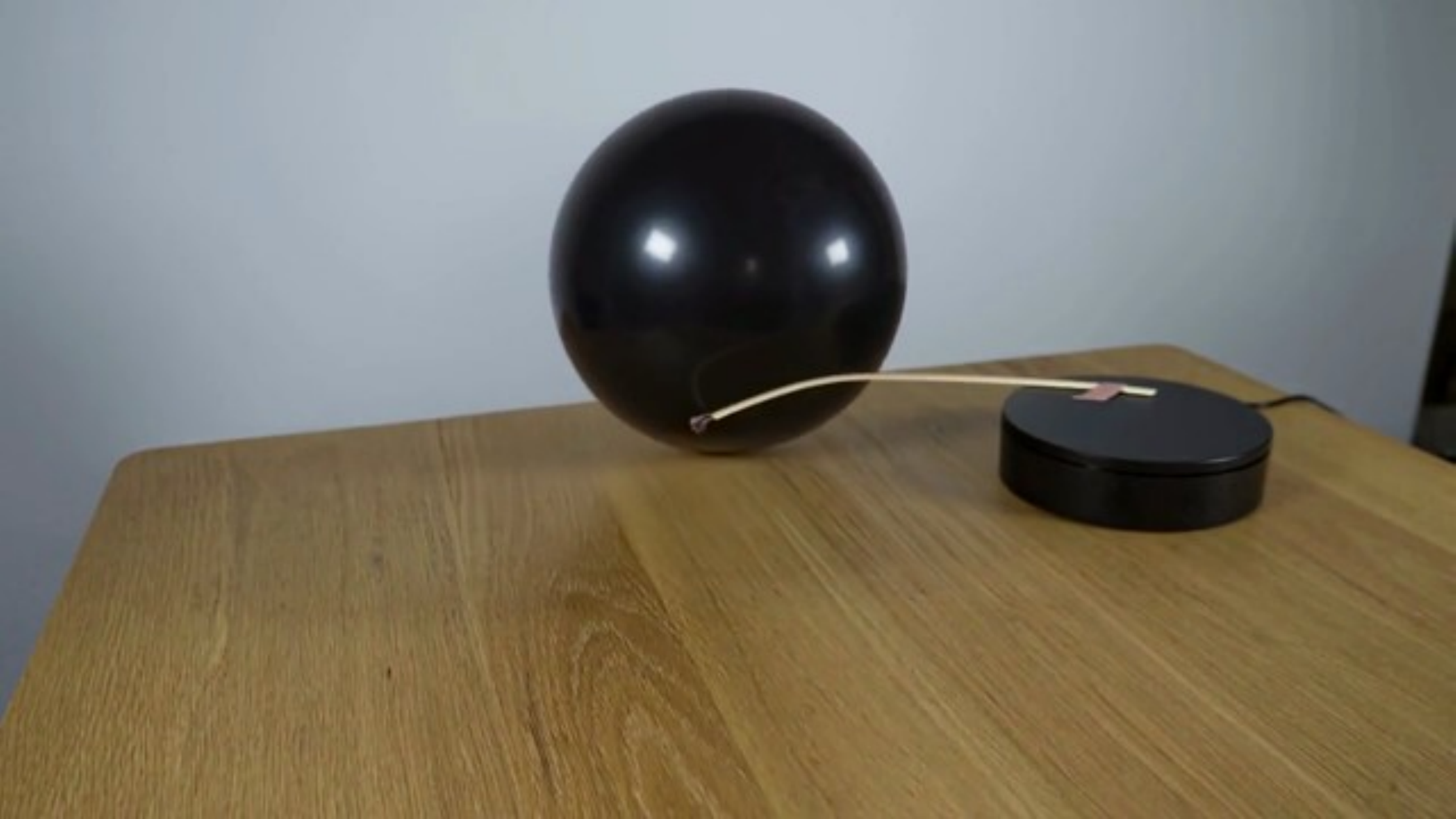}
            \end{tabular}
            \caption{A black balloon is sitting on a wooden table next to a small rotating platform with a lit matchstick taped to it. The match rotates clockwise and touches the balloon. Static shot with no camera movement.}
            \label{fig:case1}
        \end{subfigure}
    \end{minipage}
    \begin{minipage}{1.0\textwidth}
        \begin{subfigure}[b]{1.0\textwidth}
            \centering
            \vspace{1pt}  
            \begin{tabular}{@{}c@{}c@{}c@{\hspace{1pt}}c@{\hspace{1pt}}c@{\hspace{1pt}}c@{\hspace{1pt}}c@{\hspace{1pt}}c@{\hspace{1pt}}c@{\hspace{1pt}}c@{\hspace{1pt}}c@{\hspace{1pt}}c@{}}
                & 0.0s & 2.5s & 3.0s & 3.3s & 3.7s & 4.0s & 4.3s & 4.7s & 5.0s & 7.9s \\
                \raisebox{1.2\height}{\tiny \rotatebox[origin=c]{90}{Real}   }&
                \includegraphics[width=0.088\textwidth]{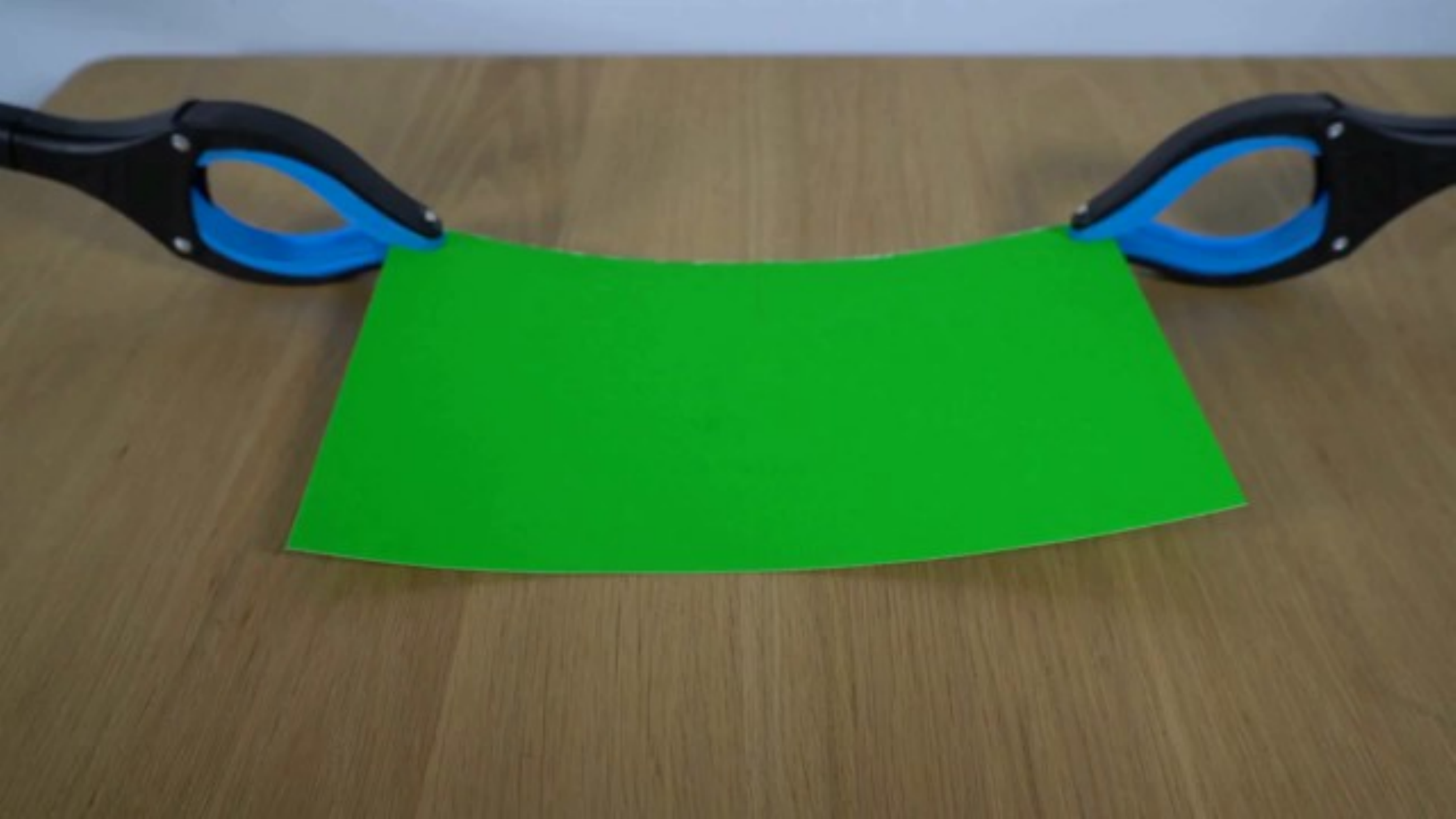} &
                \includegraphics[width=0.088\textwidth]{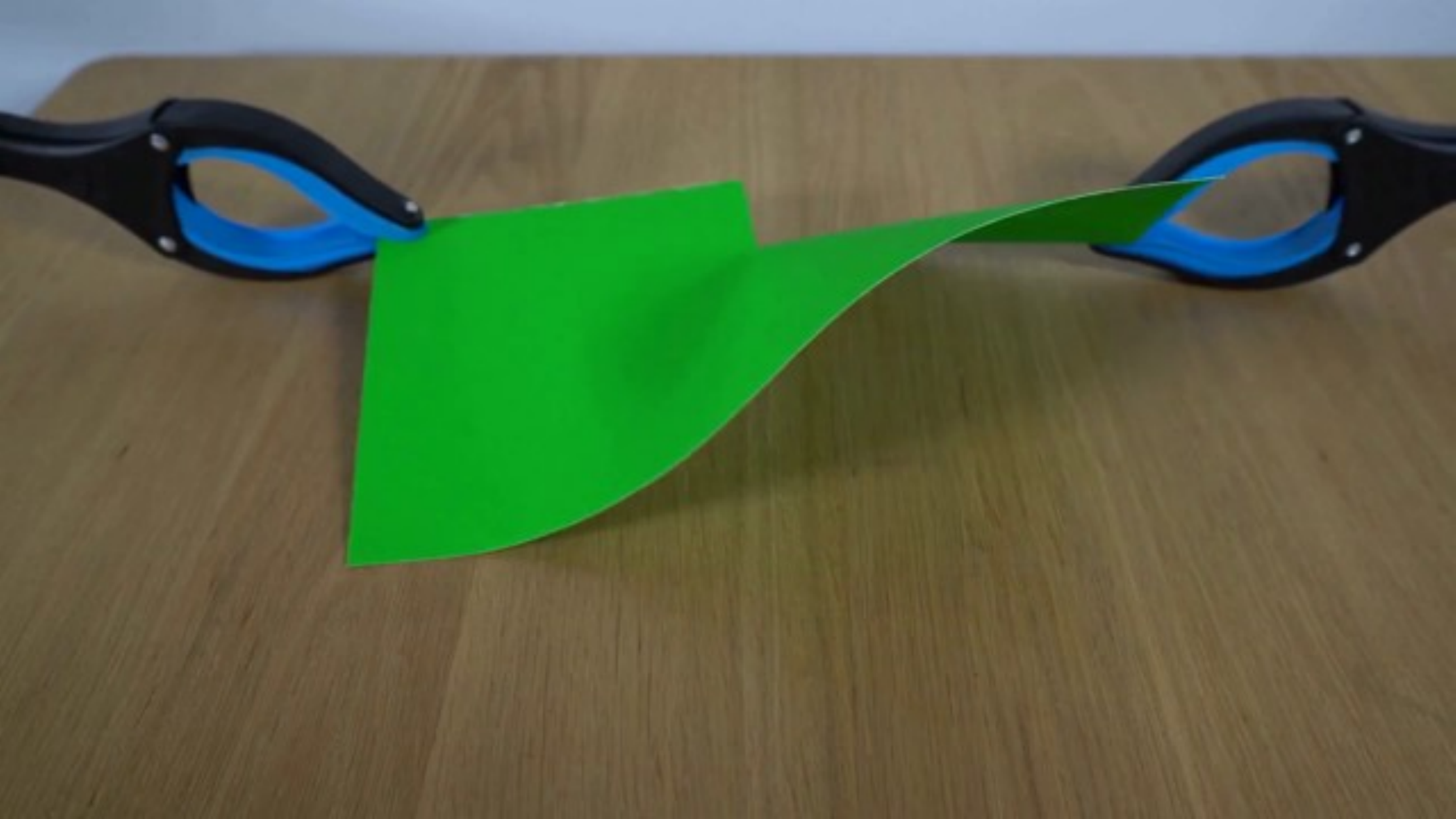} &
                \includegraphics[width=0.088\textwidth]{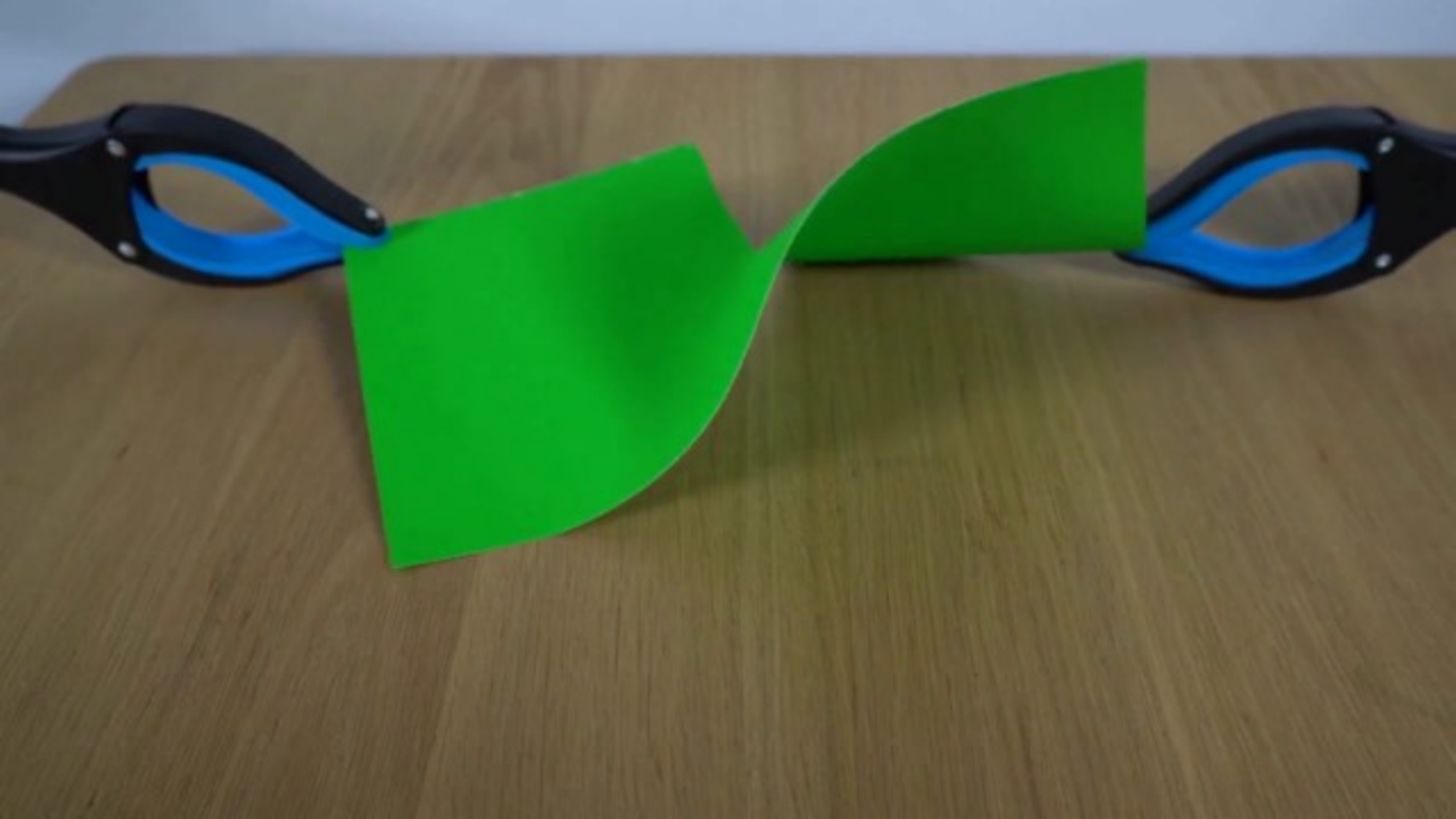} &
                \includegraphics[width=0.088\textwidth]{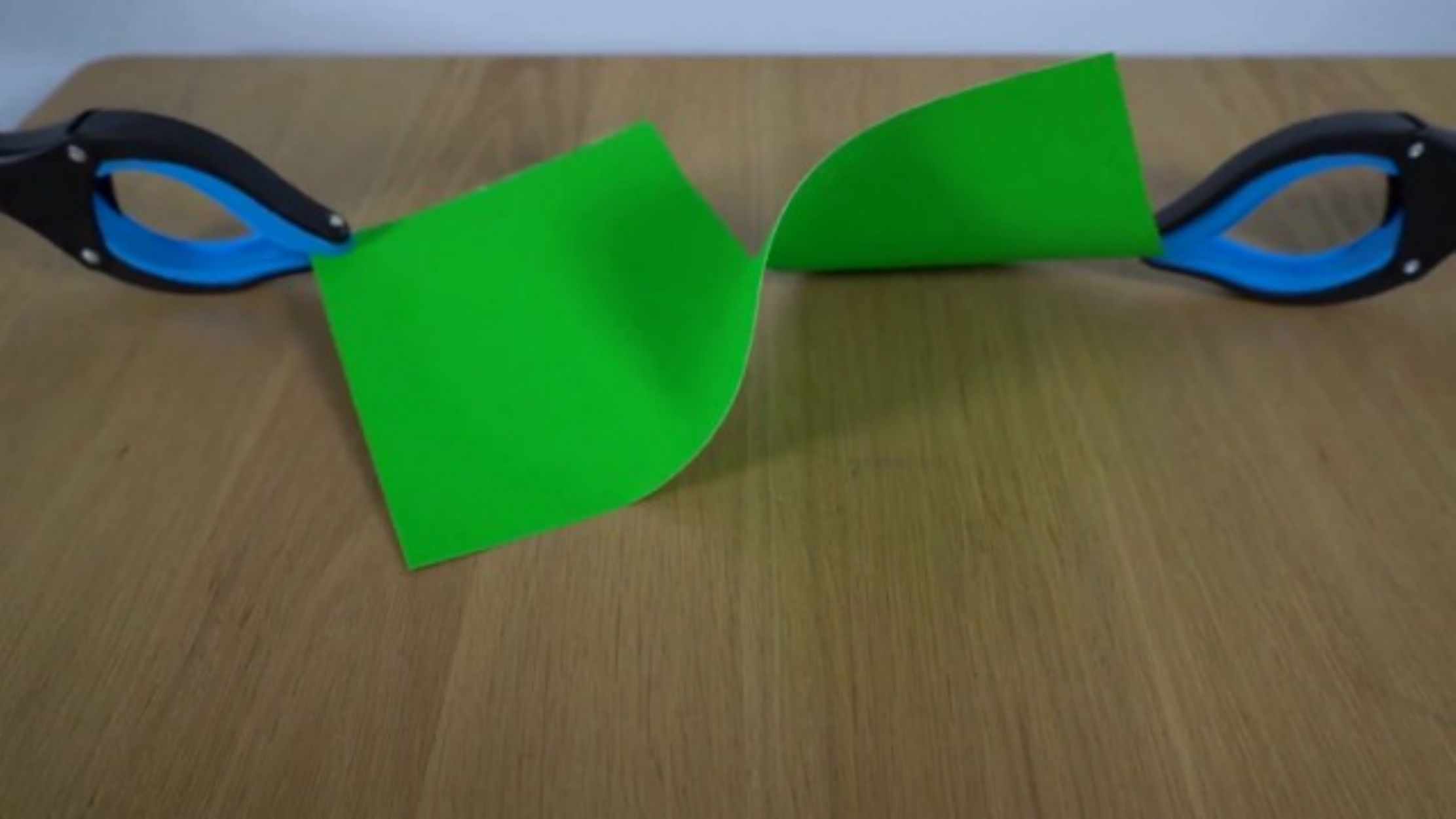} &
                \includegraphics[width=0.088\textwidth]{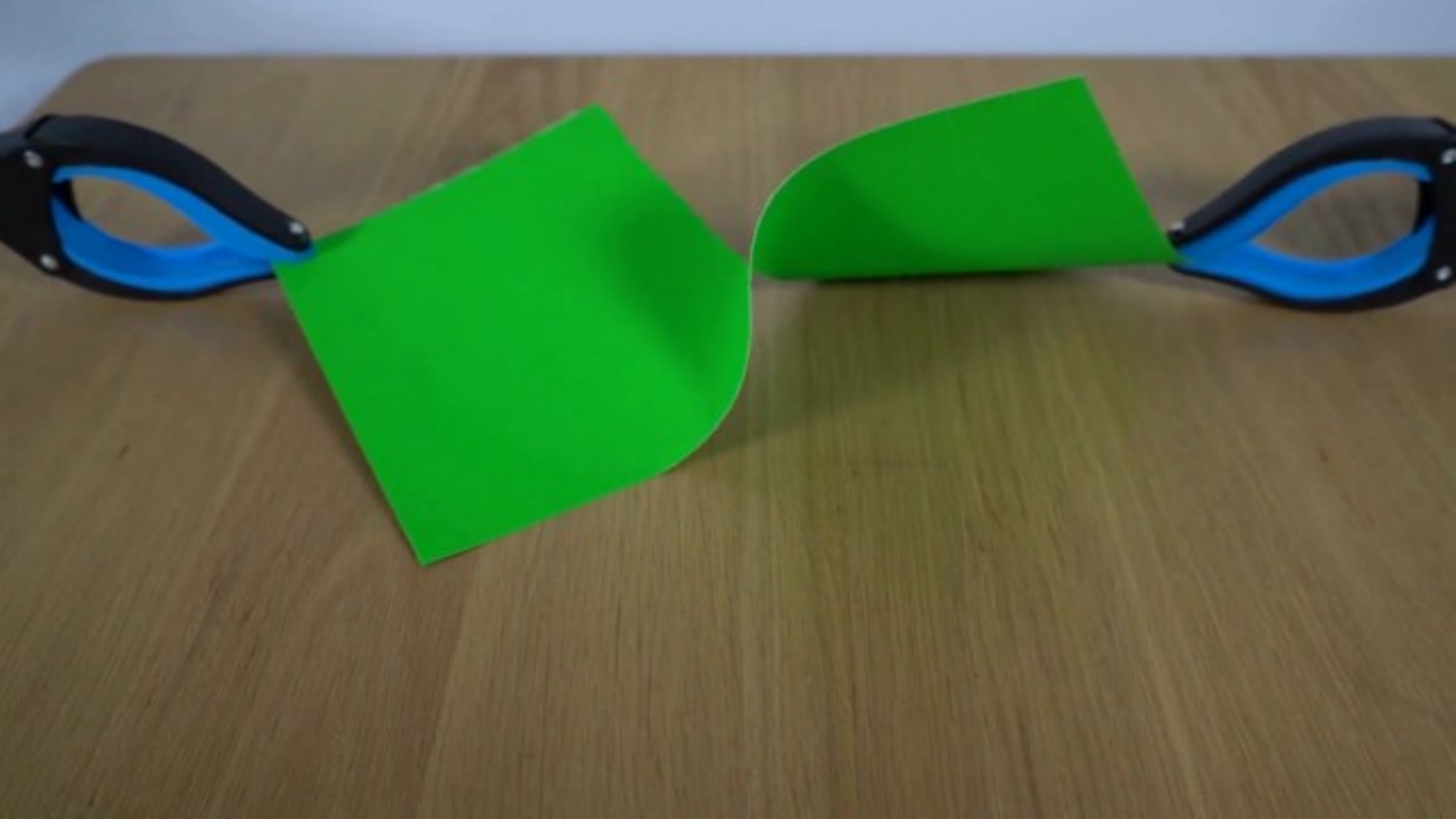} &
                \includegraphics[width=0.088\textwidth]{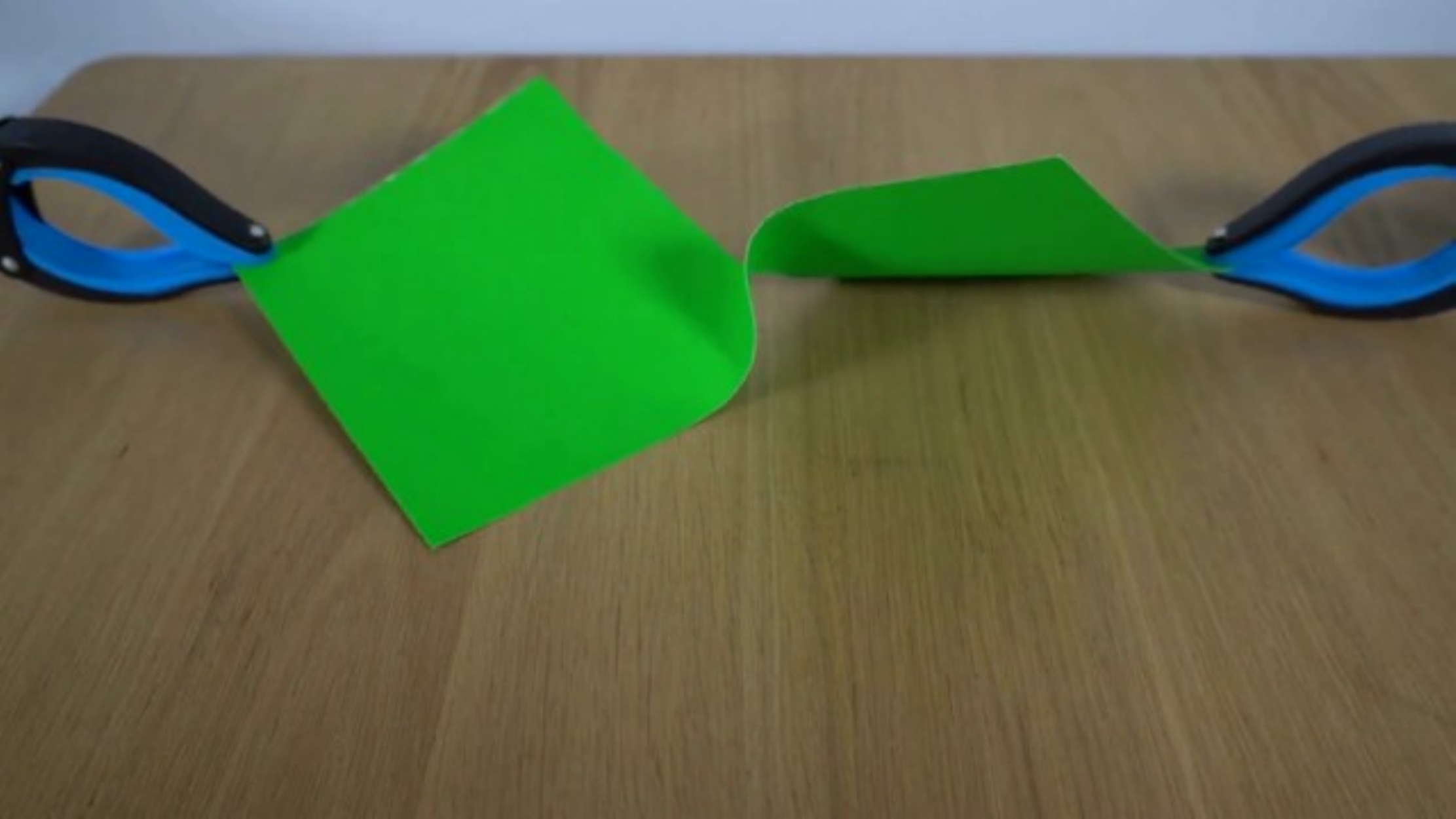} &
                \includegraphics[width=0.088\textwidth]{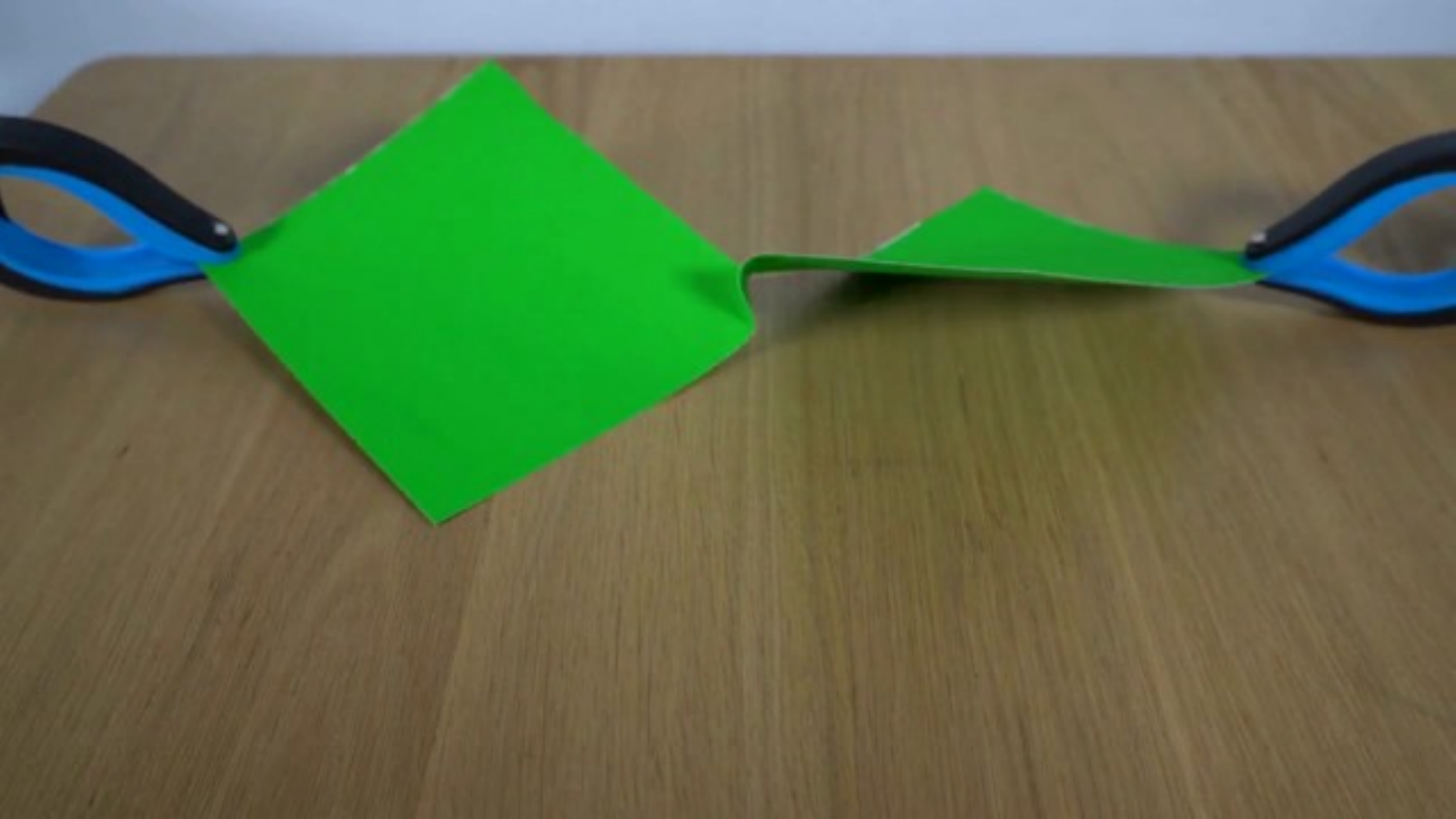} &
                \includegraphics[width=0.088\textwidth]{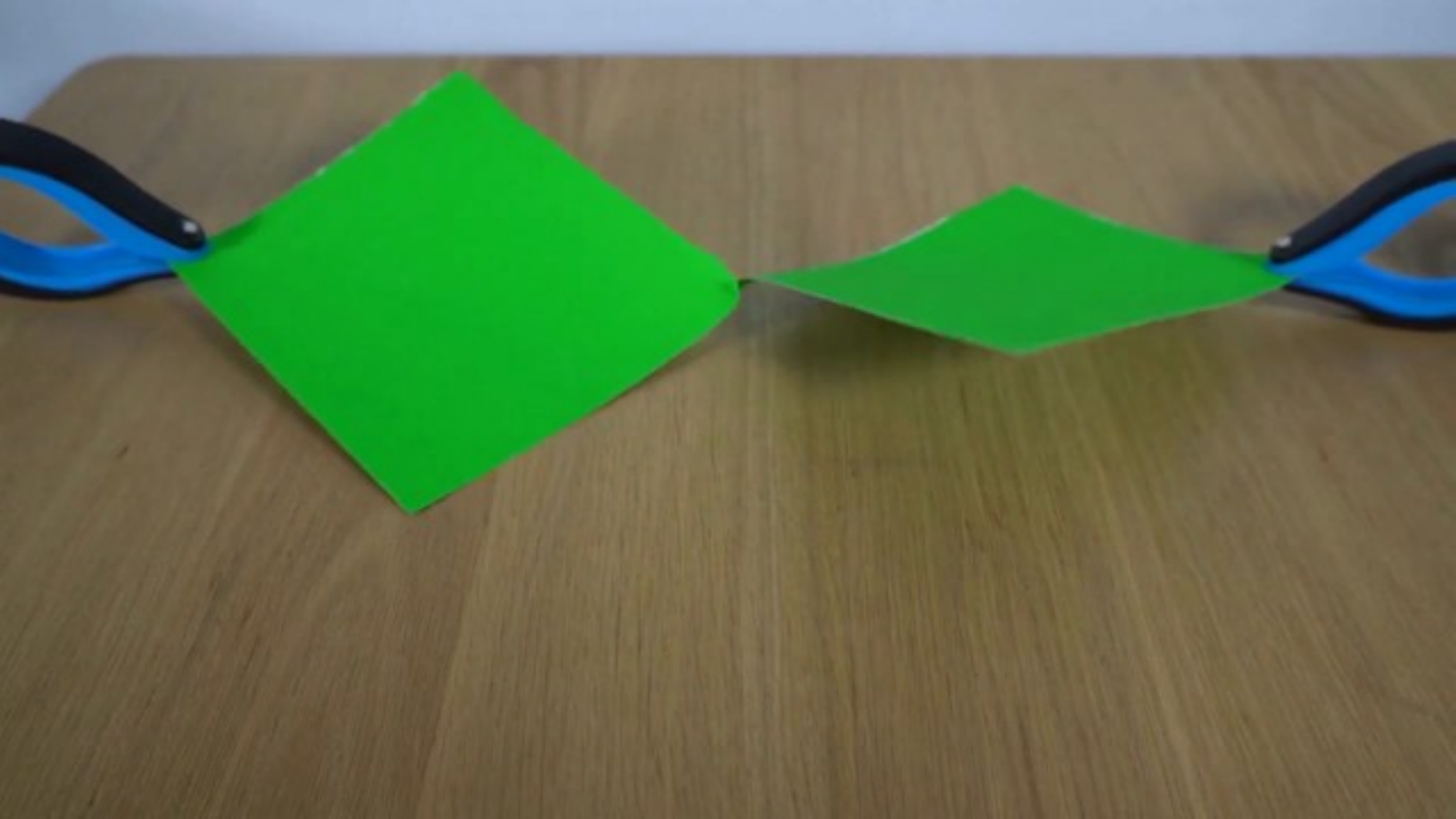} &
                \includegraphics[width=0.088\textwidth]{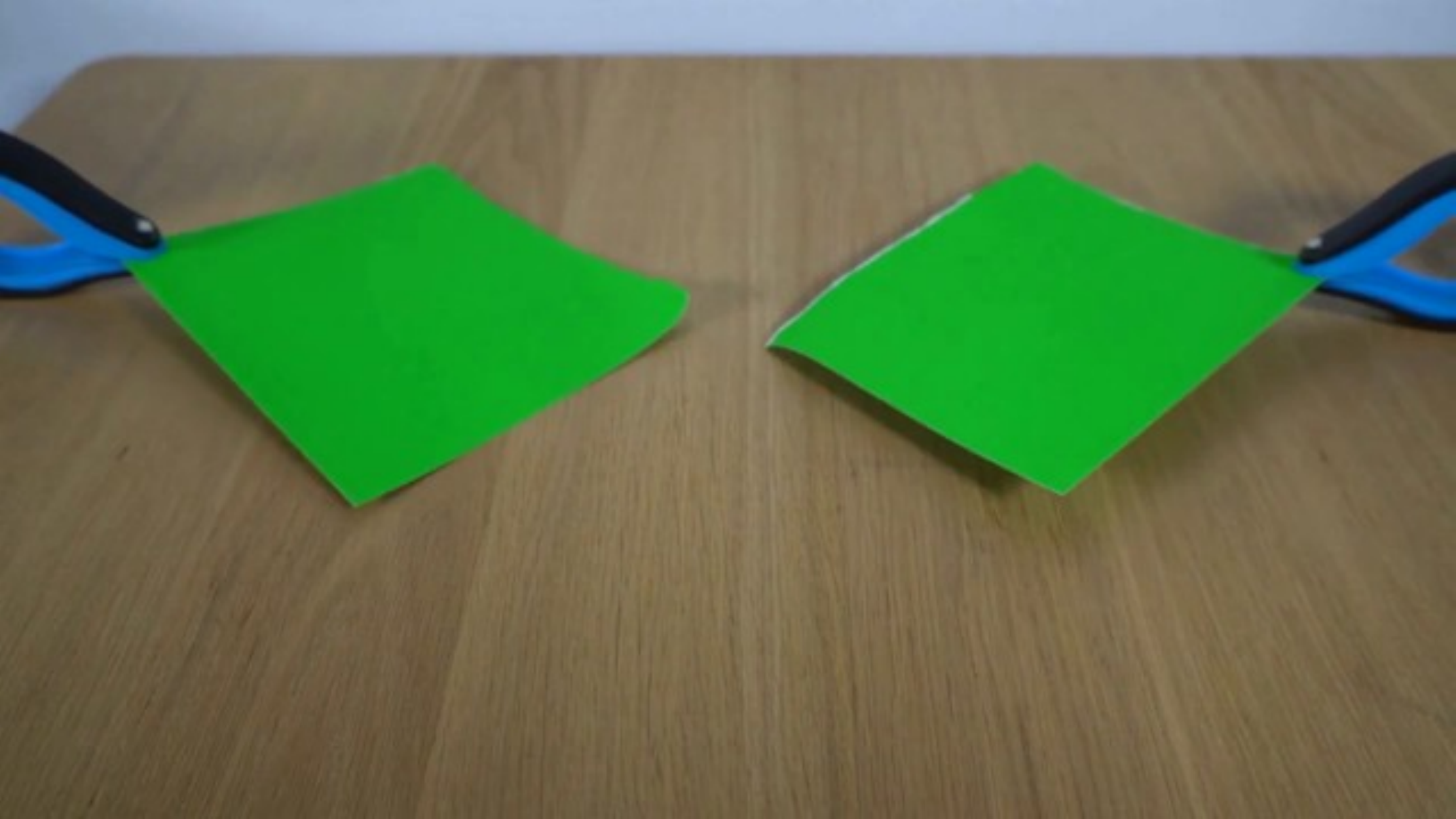} &
                \includegraphics[width=0.088\textwidth]{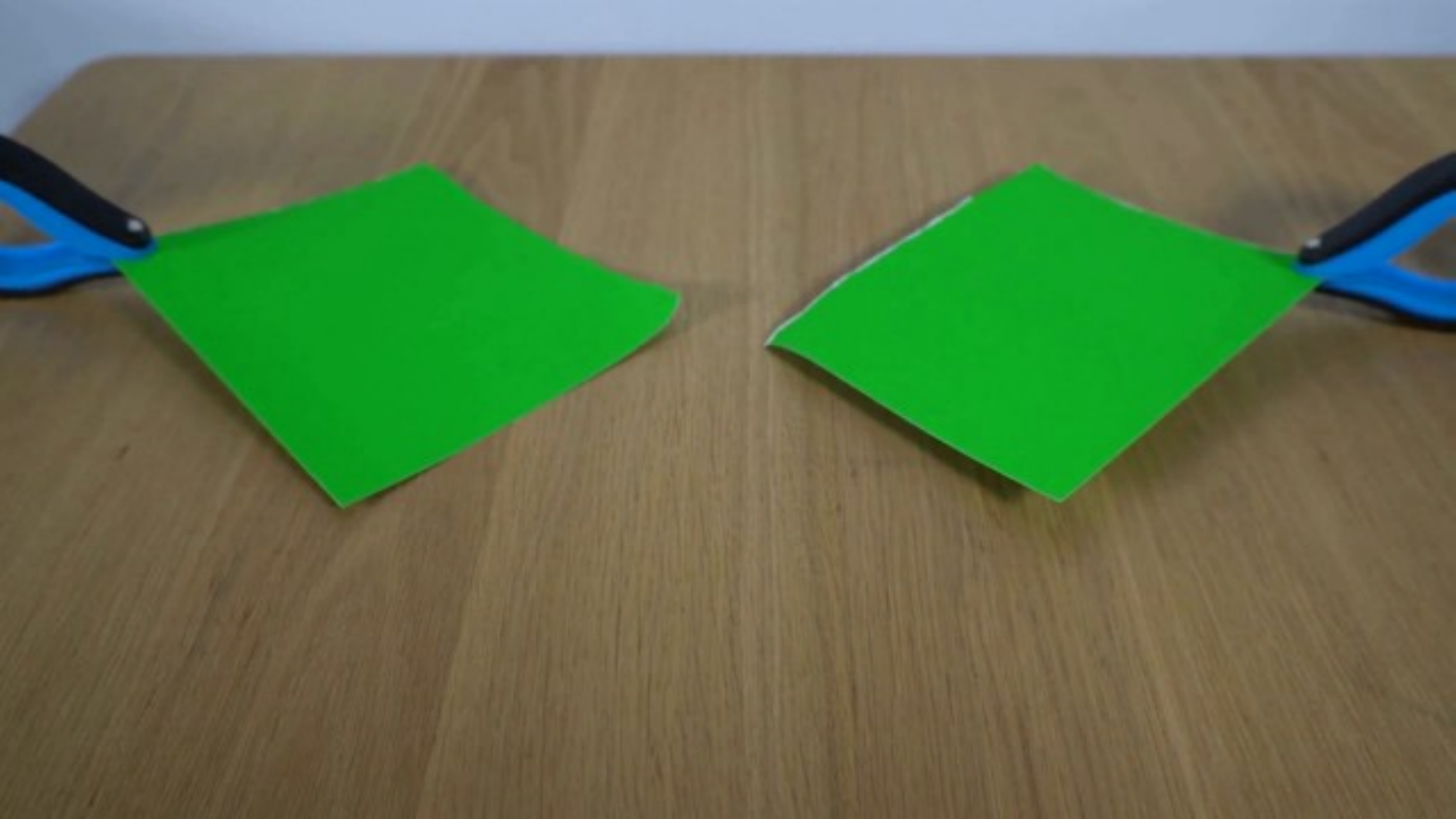}
            \end{tabular}
            \vspace{0.1pt}  
            \begin{tabular}{@{}c@{}c@{\hspace{1pt}}c@{\hspace{1pt}}c@{\hspace{1pt}}c@{\hspace{1pt}}c@{\hspace{1pt}}c@{\hspace{1pt}}c@{\hspace{1pt}}c@{\hspace{1pt}}c@{\hspace{1pt}}c@{}}
            \raisebox{0.6\height}{\tiny \rotatebox[origin=c]{90}{Generated}   }&
                
                \includegraphics[width=0.088\textwidth]{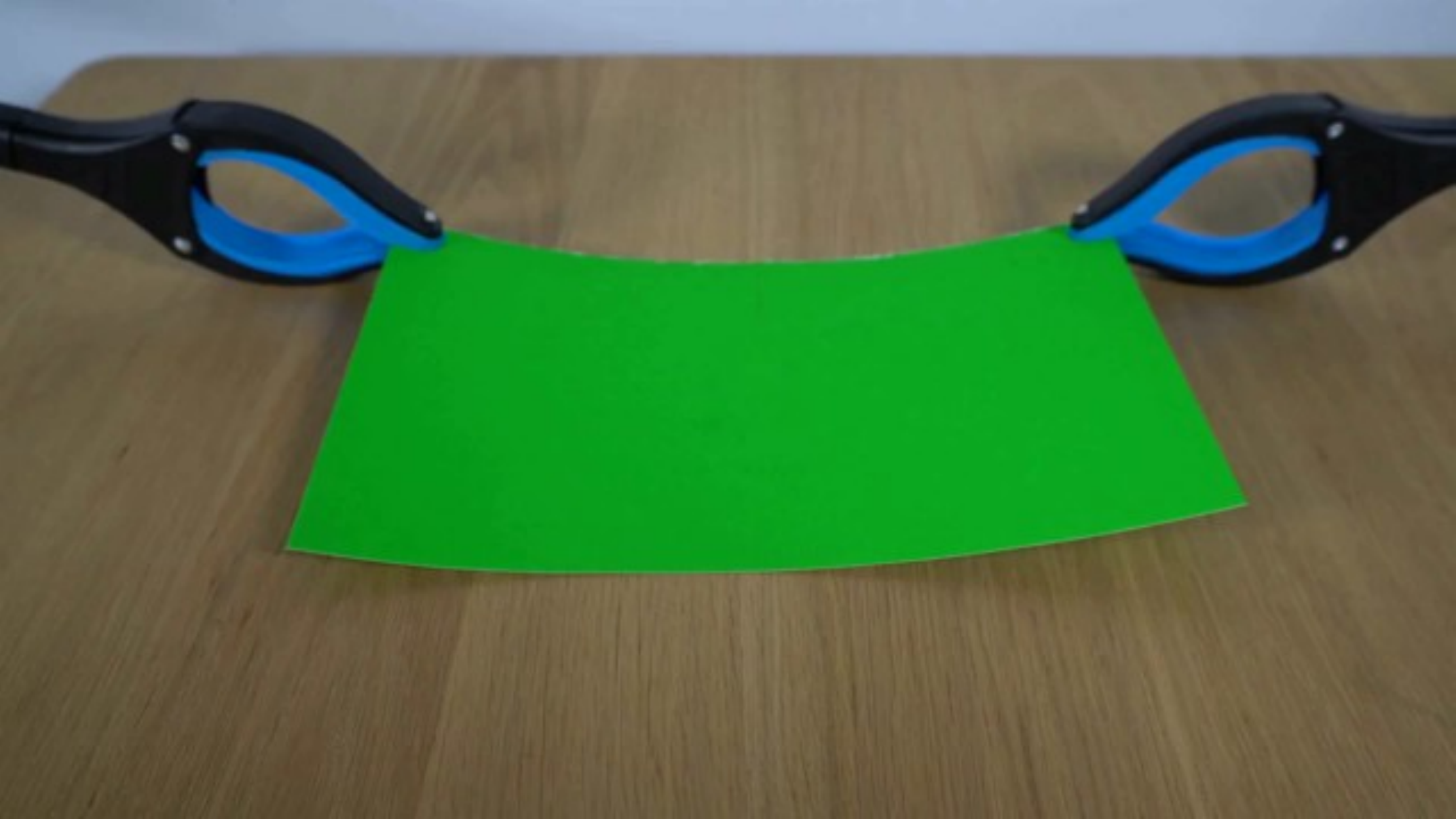} &
                \includegraphics[width=0.088\textwidth]{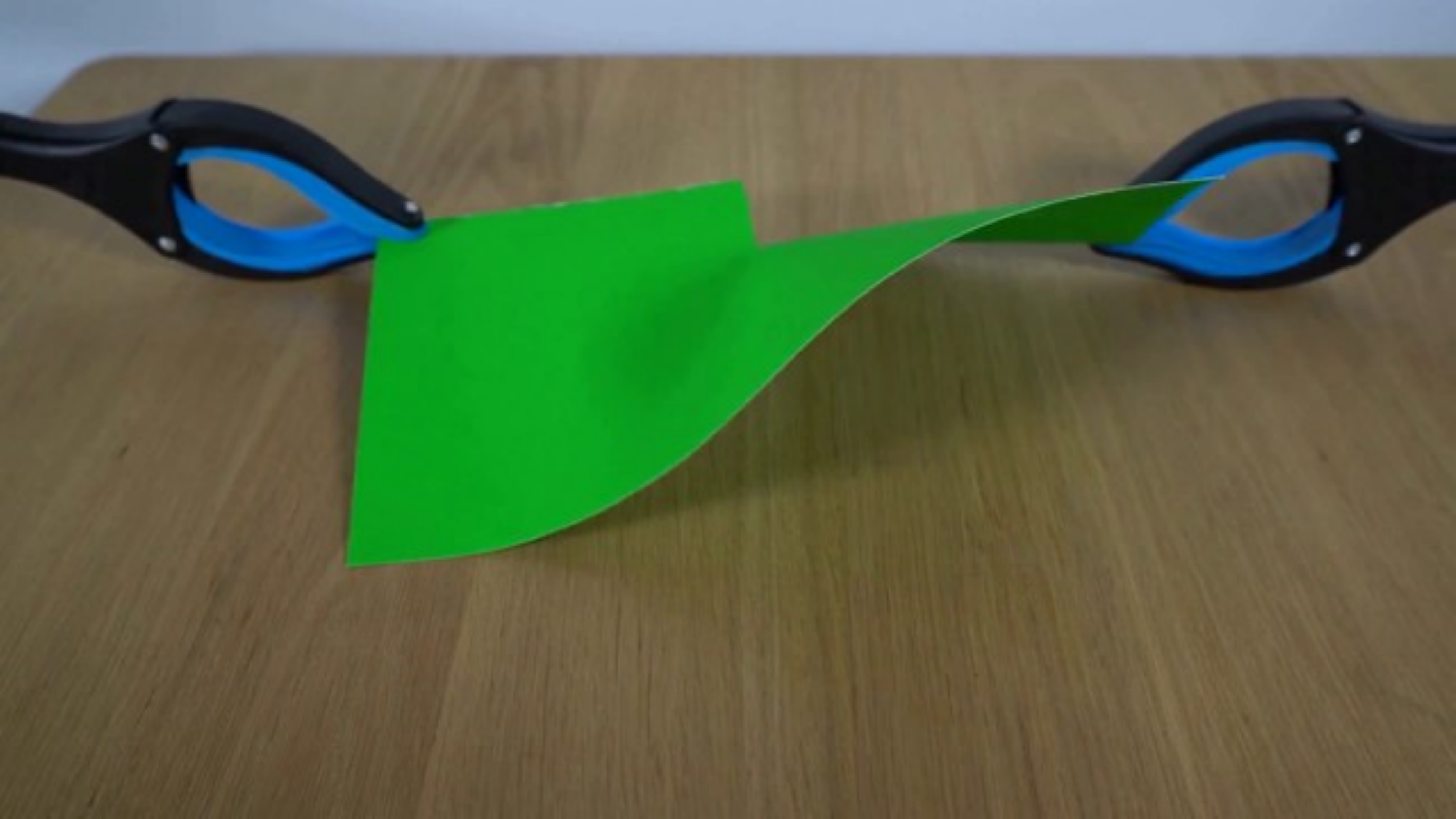} &
                \includegraphics[width=0.088\textwidth]{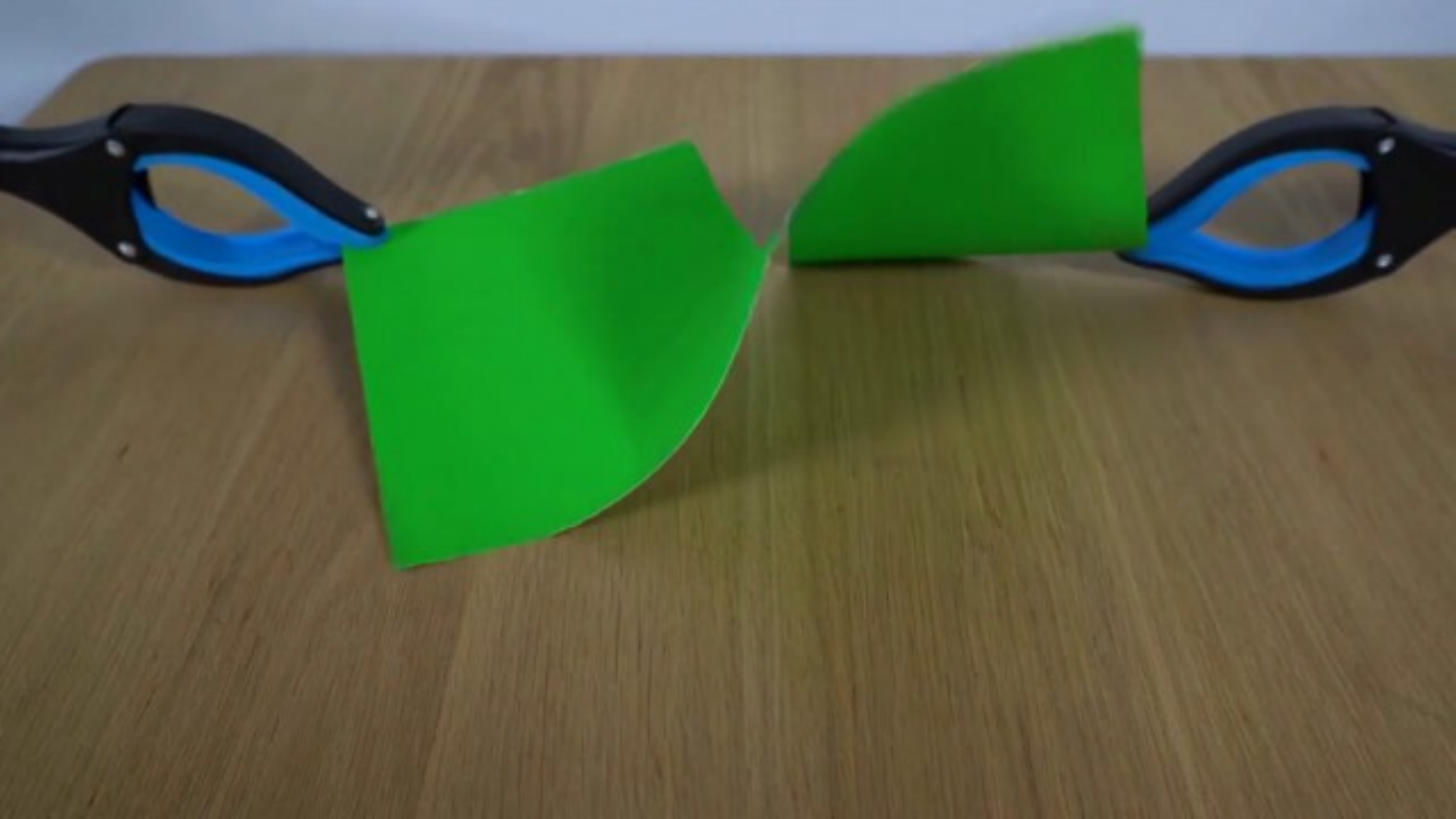} &
                \includegraphics[width=0.088\textwidth]{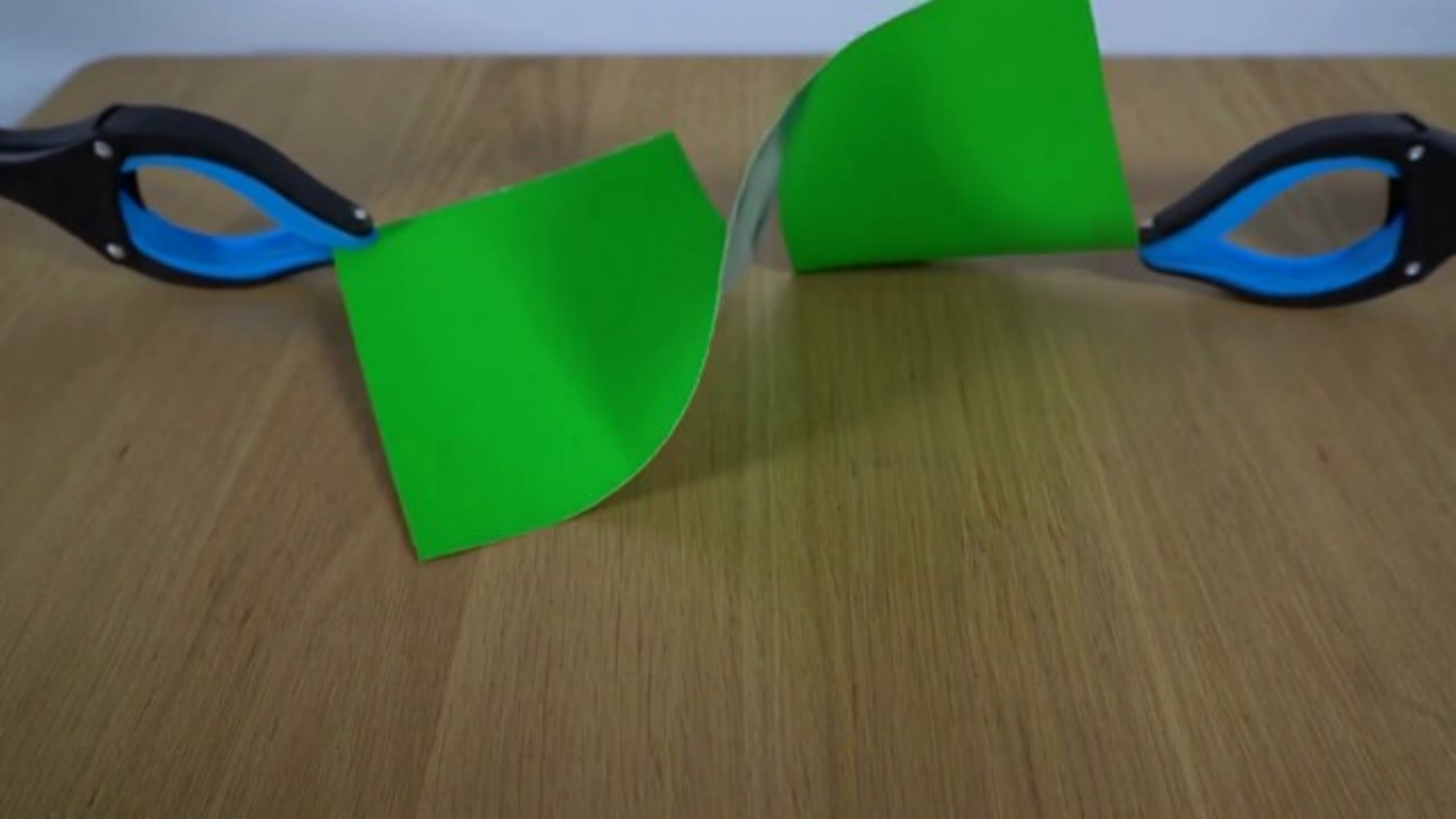} &
                \includegraphics[width=0.088\textwidth]{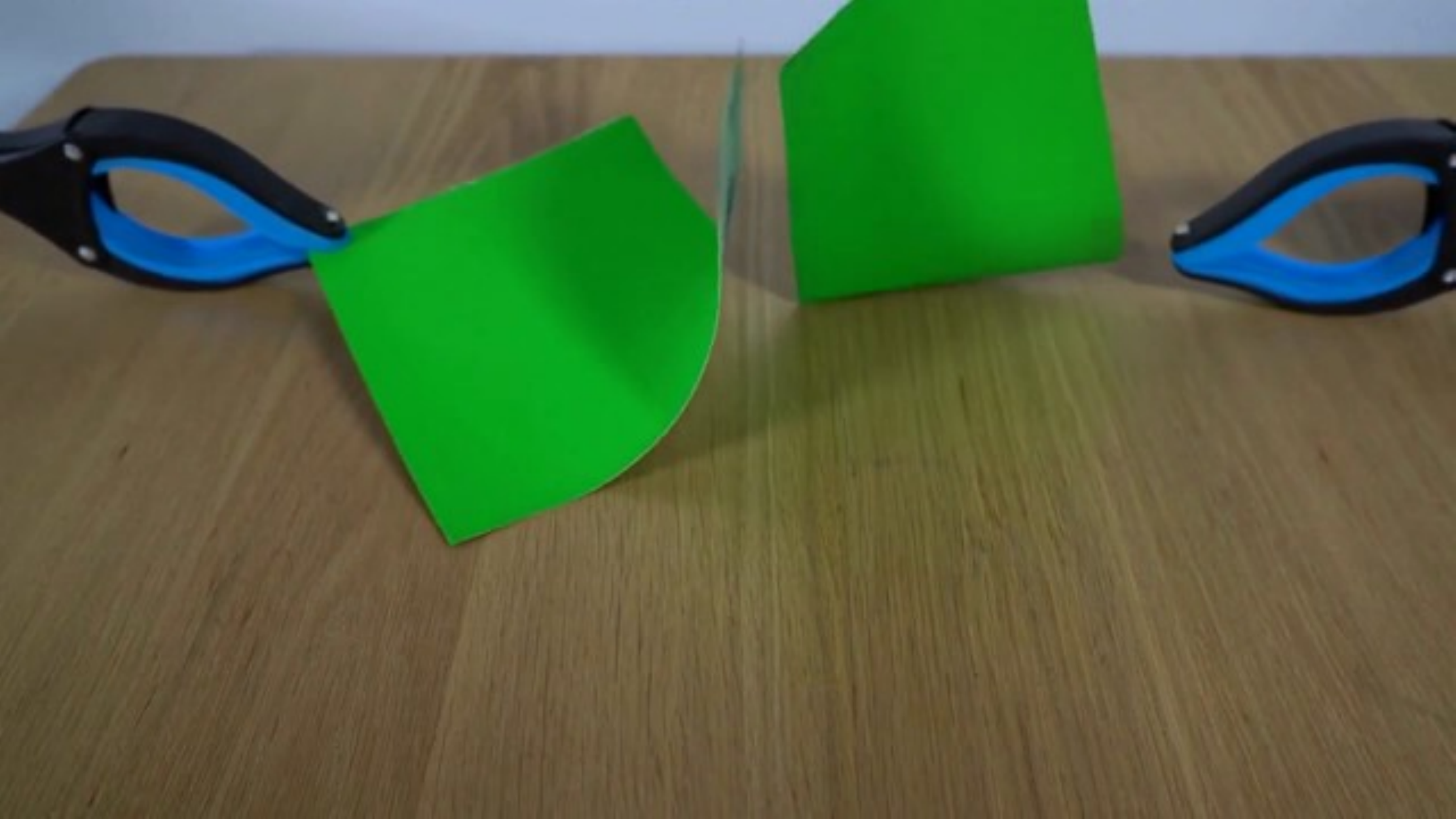} &
                \includegraphics[width=0.088\textwidth]{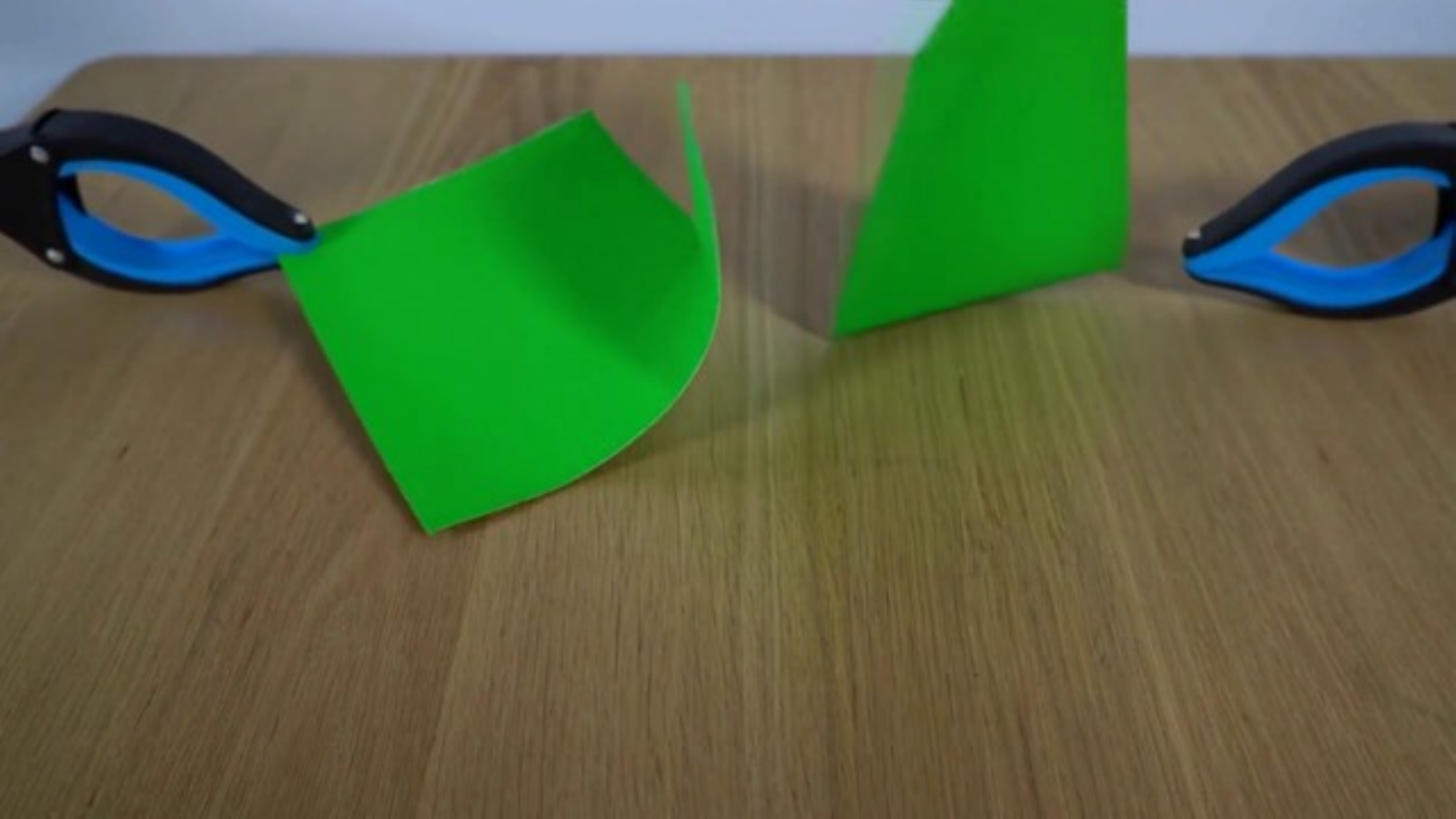} &
                \includegraphics[width=0.088\textwidth]{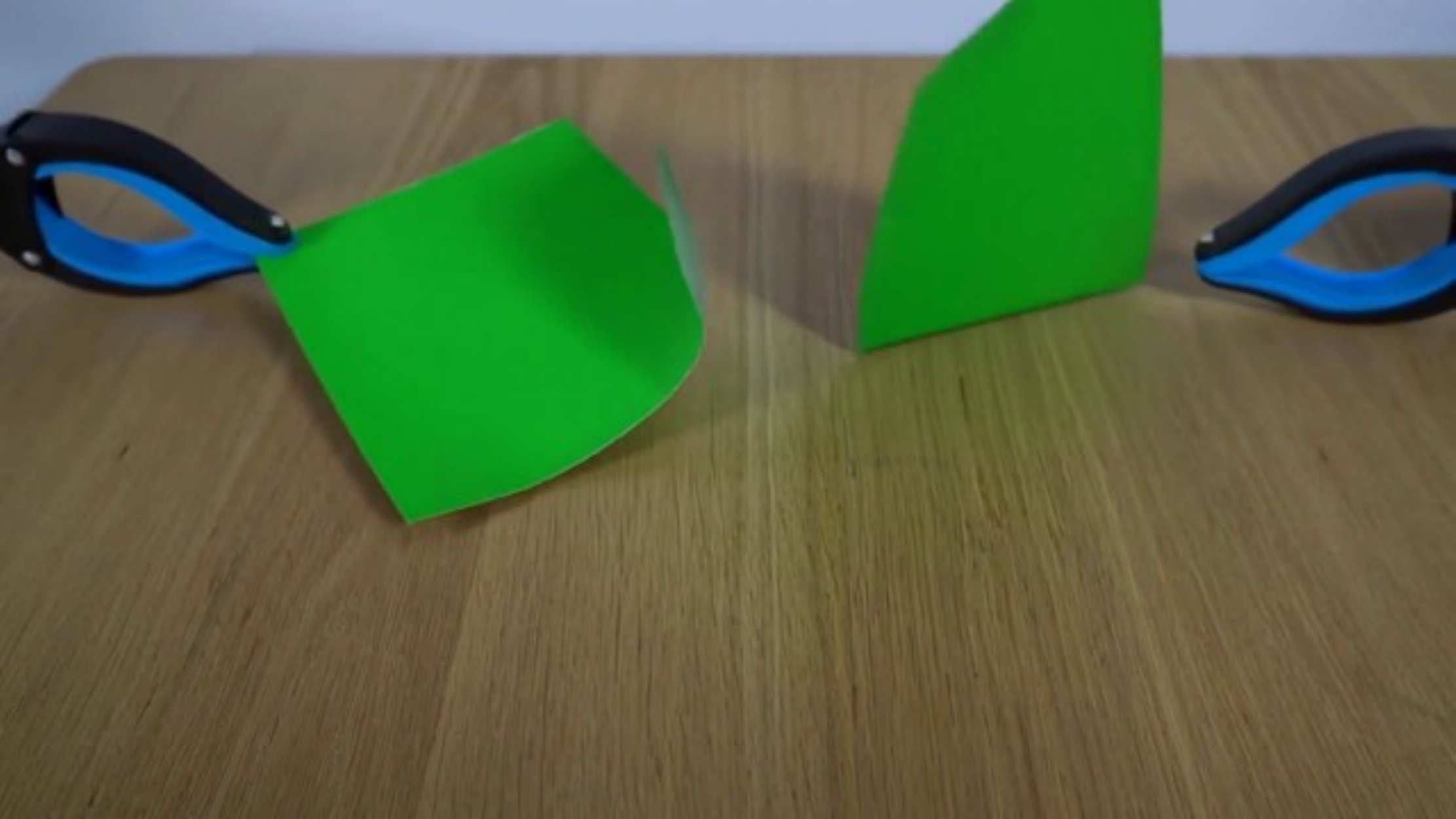} &
                \includegraphics[width=0.088\textwidth]{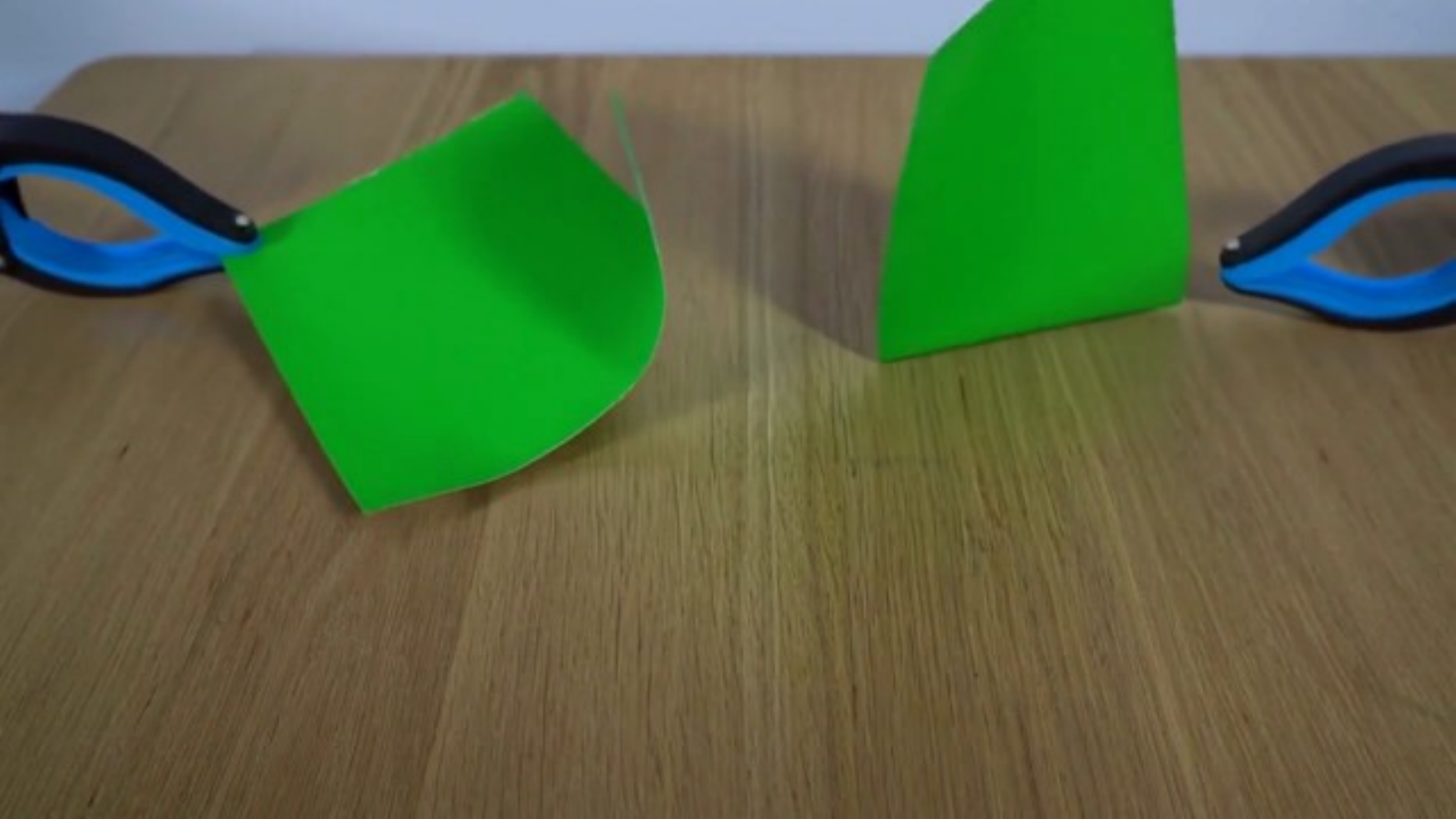} &
                \includegraphics[width=0.088\textwidth]{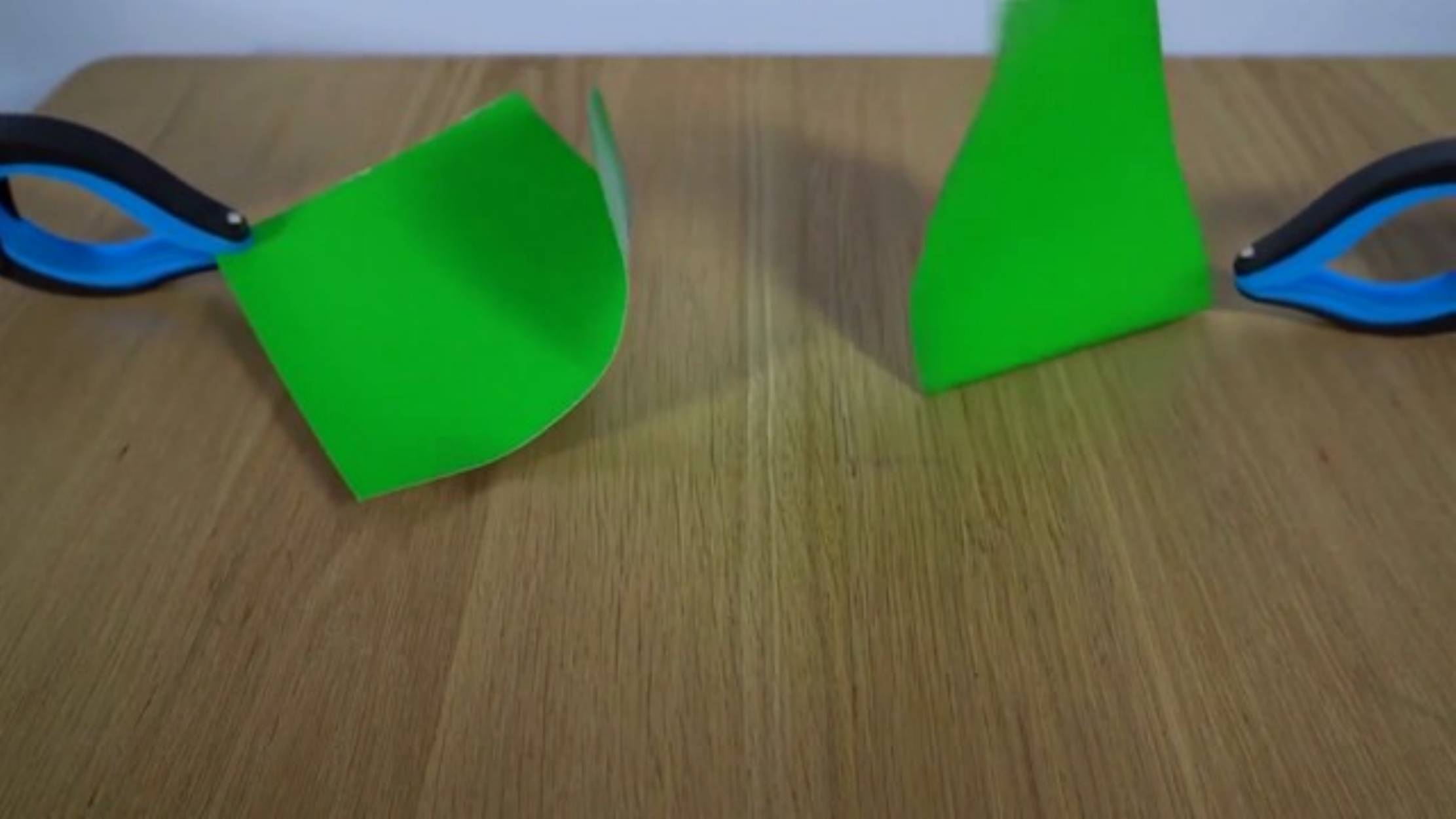} &
                \includegraphics[width=0.088\textwidth]{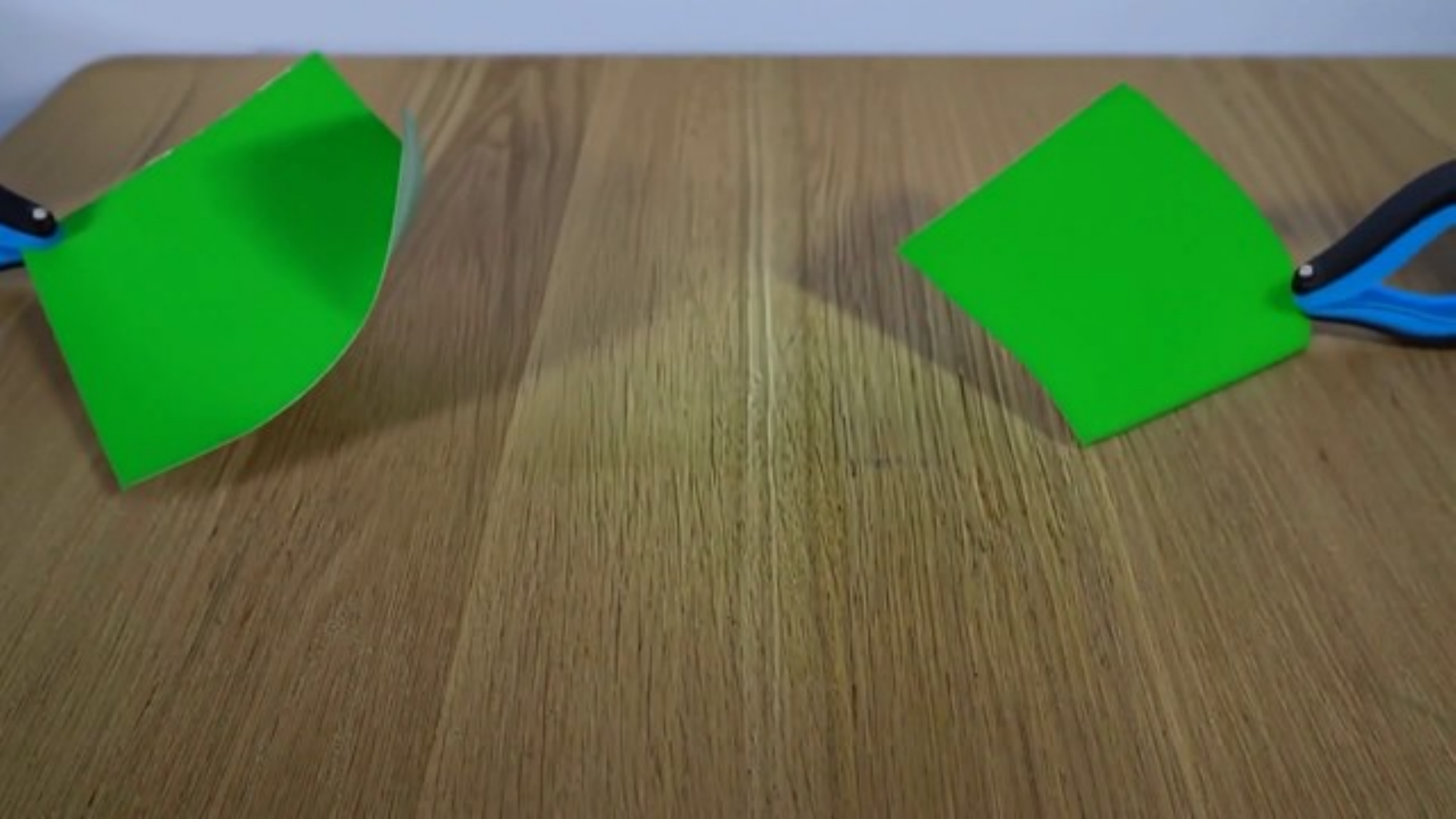}
            \end{tabular}
            \caption{Two black and blue gripping tools are pulling a piece of green paper from its two corners, causing it to tear. Static shot with no camera movement.}
            \label{fig:case1}
        \end{subfigure}
    \end{minipage}
    \caption{Case study results from the Physics-IQ Benchmark illustrate three distinct physical scenarios over time. Each scenario compares the ground truth (top row) with our model’s predictions (bottom row), conditioned on the first 3 seconds and forecasting the next 5 seconds. The results highlight the model’s ability to capture core physical interactions, as well as its limitations with complex material-specific effects: (a) The model correctly predicts the initial projectile motion but erroneously shows the ball deflecting off the duck instead of stopping upon impact. (b) Rotational dynamics are accurately captured, but the model fails to predict the match igniting and popping the balloon, instead showing the object being pushed back. (c) The model predicts the card tearing but struggles to model the motion of the torn pieces afterward.}
    \label{fig:case_comparison}
\end{figure}

Nevertheless, our model is not without limitations. Fig.~\ref{fig:case_comparison} presents several representative results, revealing both its strengths and weaknesses. While \magi effectively captures primary dynamics—such as projectile motion, rotational behavior, and material deformation, it struggles with complex secondary effects, including precise collision responses, material-specific reactions, and post-deformation behavior. Notably, even when the predicted outcome deviates from the ground truth, the model often generates physically plausible alternatives. For example, in the second case (Fig.~\ref{fig:case_comparison}(b)), although the model fails to simulate the ignition of a match and the popping of a balloon, it instead produces a coherent sequence in which the rod rotates, contacts the object, and realistically bends upon impact. These results suggest that \magi has acquired a non-trivial physical intuition, capable of generating alternative yet physically consistent scenarios.

\begin{figure}[h]
    \centering
    \includegraphics[width=0.678\textwidth]{
        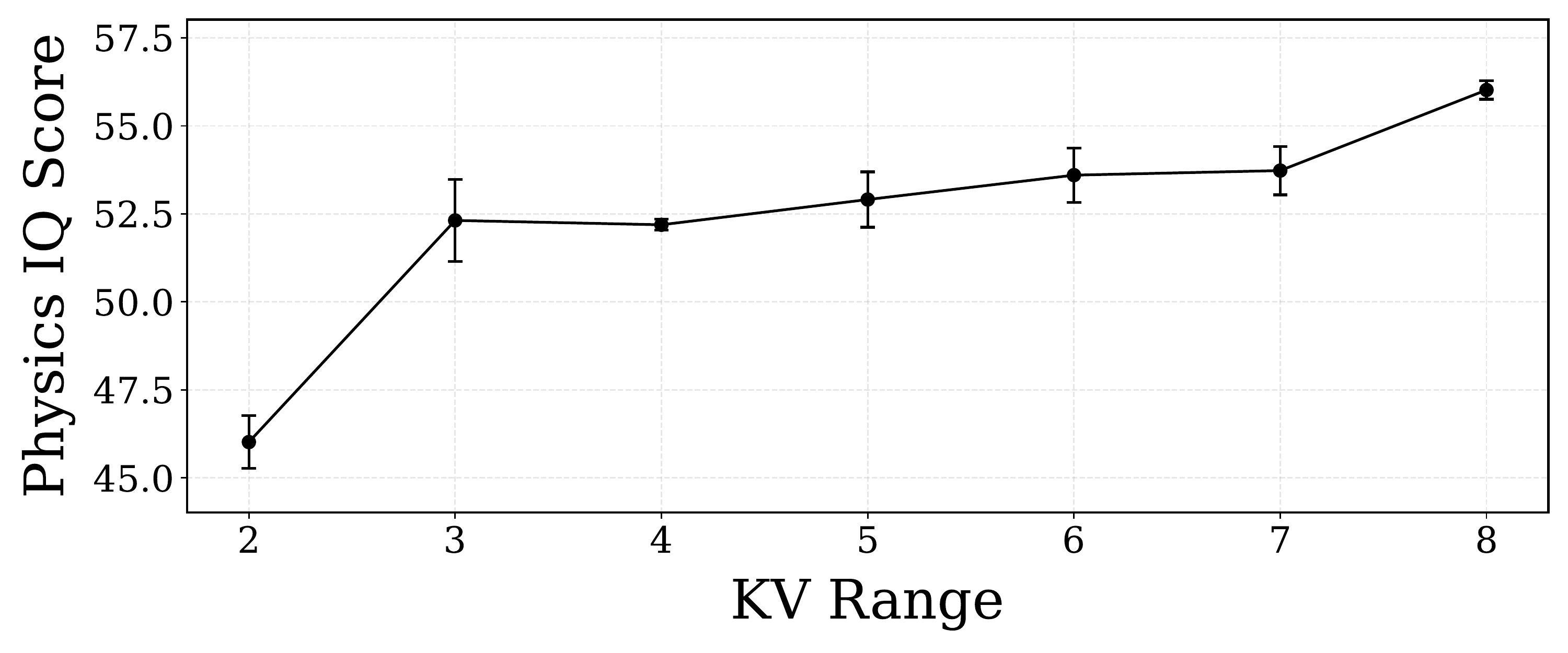
    }
    \vspace{0.5em}
    \caption{Physical IQ scores as a function of historical context. This visualization shows how performance changes with varying amounts of historical information, represented by the KV Range Value.}
    \label{fig:V2V_windows_size_ablation}
\end{figure}

\paragraph{The influence of historical context length}
The benefit of utilizing historical context for more accurate predictions has already been demonstrated in the comparison between image-conditioned and video-conditioned \magi models. To more systematically evaluate the impact of historical information in physical modeling, we varied the length of accessible history by adjusting the KV range of \magi during inference. Fig.~\ref{fig:V2V_windows_size_ablation} presents the results. Overall, we observe that increasing the amount of historical context generally leads to better performance. However, the most significant gain occurs at $\text{KV range}=2$, meaning that short-term history is often sufficient to support accurate predictions.

\section{Related Works}

This section reviews major developments in text-to-video generation, categorized by proprietary systems, open-source efforts, and recent trends in autoregressive and causal modeling. We highlight unresolved challenges in scalability, causality, and streaming compatibility—challenges that \magi is designed to address.

\paragraph{Proprietary Systems.}
Recent proprietary models have significantly advanced generation length, resolution, and semantic fidelity. OpenAI’s Sora~\citep{openaisora2024} introduced long-form, high-resolution generation with strong prompt consistency. Kuaishou’s Kling~\citep{kuaishou2024kling} and Runway’s Gen‑3~\citep{runway2024gen3} emphasized temporal fidelity and fine-grained stylistic control, respectively. Luma AI’s DreamMachine~\citep{luma2024dm} improved motion continuity and stylistic adherence. Pika Labs’ Pika 1.5~\citep{pika2024pika} enabled interactive control over visual attributes, while Meta’s MovieGen~\citep{polyak2024moviegencastmedia} offered transparency into foundational model training. Most recently, Google's Veo 2~\citep{veo22025} advanced physical realism and human motion modeling. Despite these innovations, most systems are closed-source and opaque in architecture, limiting reproducibility and extensibility.

\paragraph{Open-Source Ecosystem.}
The open-source community pioneered latent diffusion through Stable Diffusion~\citep{esser2020taming}, which integrated a variational autoencoder~\citep{kingma2013auto} for latent representation, a CLIP-based text encoder~\citep{radford2021learning}, and a U-Net denoiser~\citep{ronneberger2015u}. Temporal extensions such as VDM~\citep{ho2022video}, AnimateDiff~\citep{guo2023animatediff}, and SVD~\citep{blattmann2023svd} adapted the architecture for frame coherence. Transformer-based backbones like DiT~\citep{dit}, PixArt‑$\alpha$~\citep{chen2023pixartalpha}, and Latte~\citep{ma2024latte} demonstrated scalability and inspired early video adaptations. Recent open implementations—including Open‑Sora~\citep{opensora}, Open‑Sora‑Plan~\citep{lin2024open}, CogVideoX~\citep{cogvideox}, Mochi 1~\citep{genmo2024mochi}, HunyuanVideo~\citep{kong2024hunyuanvideo}, StepVideo~\citep{ma2025step}, LTX‑Video~\citep{HaCohen2024LTXVideo}, and Wan~\citep{wang2025wan}—introduced modular advances in chunking, compression, and streaming. However, these systems largely retain bidirectional denoising and globally conditioned inference, limiting applicability to real-time or causal settings.

\paragraph{Autoregressive and Causal Modeling.}
An emerging trend is the integration of autoregressive modeling and causal constraints. Diffusion Forcing~\citep{chen2024diffusion} introduces independent per‑token noise schedules that allow a causal model to denoise future tokens while keeping past tokens minimally perturbed, effectively unifying next‑token prediction with full‑sequence diffusion. FVDM~\citep{liu2024redefining} employed timestep vectorization for precise noise control. CausVid~\citep{yin2024slow} combined causal inference with distillation for streaming scenarios. While promising, these models remain limited in scale, often lack chunk-wise abstraction, and do not unify video continuation with I2V/T2V generation.

\paragraph{\magi: Scalable Autoregressive Diffusion.}
To our knowledge, \magi is the first large-scale, chunk-wise autoregressive diffusion model trained from scratch that unifies high-fidelity text-to-video, image-to-video, and video continuation tasks under strict causal constraints. It supports real-time streaming and long-horizon synthesis via efficient chunk-wise denoising, shortcut distillation, and KV-cached inference. By explicitly addressing scalability, causality, and streaming compatibility, \magi establishes a new foundation for unified and controllable video generation.

\section{Conclusion}
\magi introduces a scalable chunk-wise autoregressive diffusion framework for high-fidelity video synthesis. By progressively denoising fixed-length segments under strict causal constraints, it enables real-time, streaming-compatible generation with fixed computational overhead regardless of video length. The architecture builds upon a Transformer backbone enhanced with block-causal and parallel attention modules, and is supported by a distributed attention mechanism and a highly efficient training strategy for handling ultra-long contexts.

A key contribution lies in its unified design: \magi supports text-to-video, image-to-video, and video continuation tasks without requiring task-specific modifications, all under a shared training objective. Through chunk-wise text conditioning, it further achieves fine-grained semantic control across long-form video generation. A shortcut distillation strategy significantly reduces the number of diffusion steps required for inference, improving efficiency while maintaining temporal consistency and sample quality.

Empirical results on VBench-I2V and Physics-IQ benchmarks demonstrate that \magi outperforms existing large-scale video diffusion models in prompt adherence, physical plausibility, and temporal coherence. Taken together, these contributions establish \magi as a robust and extensible foundation for autoregressive video synthesis—offering both state-of-the-art performance and a fertile ground for future advancements in modularity, controllability, and multi-modal reasoning.

\section{Limitation and Future Work}

While \magi demonstrates strong generation quality and low-latency inference via chunk-wise autoregressive denoising, its current architecture remains tightly coupled. Specifically, a single large decoder-style Transformer is tasked with both (1) high-level temporal context fusion—integrating static conditioning signals with progressively noisier visual inputs—and (2) low-level denoising, which requires accurate reconstruction of fine-grained visual details. This conflation of heterogeneous objectives introduces several technical limitations:

\begin{itemize}
    \item \textbf{Inference latency bottleneck}: The same large model is repeatedly invoked across all denoising steps, even when only minor refinements are required. This leads to inefficient utilization of compute, especially in streaming settings where low-latency frame delivery is critical.
    \item \textbf{Optimization conflict}: Jointly optimizing global semantic planning and pixel-level restoration within a single model exacerbates objective interference, often leading to suboptimal scaling behavior.
    \item \textbf{Limited controllability}: The monolithic architecture constrains the insertion of auxiliary control signals—such as confidence-based guidance modulation, or dynamic temporal constraints—due to entangled latent pathways and overlapping functional scopes.
\end{itemize}

Thus, a decoupled design that structurally separates high-level semantic reasoning from low-level visual synthesis is worth exploring. Looking ahead, as video generation evolves from producing isolated clips to constructing long-form content with coherent narratives, we anticipate a convergence between video generation and understanding. In this closed-loop setting, the quality of generated content will increasingly depend on the model’s capacity to understand video content, making understanding the key bottleneck. Although we are still far from this frontier, we believe that a modular architecture represents a crucial step toward closing the loop between video understanding and generation.

\newpage
\section*{Contributions and Acknowledgments}

Names are presented alphabetically by first name.

\vspace{1em}

\noindent\textbf{Core Contributors}:
\vspace{-1em}
\begin{multicols}{2}
\noindent
Hansi Teng	\\
Hongyu Jia	\\
Lei Sun	\\
Lingzhi Li	\\
Maolin Li	\\
Mingqiu Tang	\\
Shuai Han	\\
Tianning Zhang	\\
W.Q. Zhang	\\
Weifeng Luo	\\
Xiaoyang Kang	\\
Yuchen Sun	\\
Yue Cao	\\
Yunpeng Huang	\\
Yutong Lin	\\
Yuxin Fang	\\
Zewei Tao	\\
Zheng Zhang	\\
Zhongshu Wang	\\
Zixun Liu
\end{multicols}

\vspace{1em}
\noindent\textbf{Contributors}:
\vspace{-1em}
\begin{multicols}{2}
\noindent
Dai Shi \\
Guoli Su \\
Hanwen Sun \\
Hong Pan \\
Jie Wang \\
Jiexin Sheng \\
Min Cui \\
Min Hu \\
Ming Yan \\
Shucheng Yin \\
Siran Zhang \\
Tingting Liu \\
Xianping Yi \\
Xiaoyu Yang \\
Xin Song \\
Xuan Hu \\
Yankai Zhang \\
Yuqiao Li
\end{multicols}

\bibliography{main}
\bibliographystyle{iclr2025_conference}

\appendix

\section{Inference Infra}
\subsection{W8A8 Quantization}
\label{sec:appendix_w8a8}
We adopt the A8W8 SmoothQuant approach~\citep{xiao2023smoothquant}, which leverages a calibration dataset to pre-compute per-channel scaling factors $s$. This enables an equivalent transformation of the form $Y = (X \cdot \text{diag}(s)^{-1}) \cdot (\text{diag}(s)W)$, effectively mitigating the impact of outliers in channel-wise activations.

For calibration, we constructed a dataset encompassing a wide range of usage scenarios, including different task types (\emph{e.g.}, T2V and I2V) and a uniformly sampled step size within the range [12, 32]. Notably, I2V samples constituted approximately 30\% of the dataset. We employed the \texttt{FP8} data type for quantization, as \texttt{INT8} was found to introduce noticeable visual artifacts in the generated videos. Furthermore, we conducted a hyperparameter search over the range $\alpha \in (0.4, 0.6)$, and ultimately selected $\alpha = 0.45$ for SmoothQuant.
All model weights were quantized except for the first and last layers. This quantization strategy led to a 30\% performance improvement without compromising generation quality.

\subsection{Multi-Node Parallel Inference}
\label{sec:appendix_multinode_parallel_inference}
We adopted a multi-node Ulysses-based parallel inference framework across 3 nodes (24 GPUs), where inter-GPU communication and the high computational density of attention emerged as the primary bottlenecks.
Ulysses performs four all-to-all communication steps for the $q$, $k$, $v$, and $o$ tensors. To mitigate communication overhead, we carefully overlapped each communication stage with corresponding computations:

\begin{itemize} 
    \item $v$-communication overlaps with $k$-computation
    \item $k$-communication overlaps with $q$-computation
    \item $q$-communication overlaps with KV cache updates
    \item $o$-communication overlaps with cross-attention computation
\end{itemize}

This overlapping strategy effectively reduced communication overhead to less than 3\% of total execution time.

\begin{figure}[htbp]
    \centering
    \begin{minipage}{1.0\textwidth}
    \begin{subfigure}[b]{1.0\textwidth}
        \centering
        \includegraphics[width=1.0\textwidth]{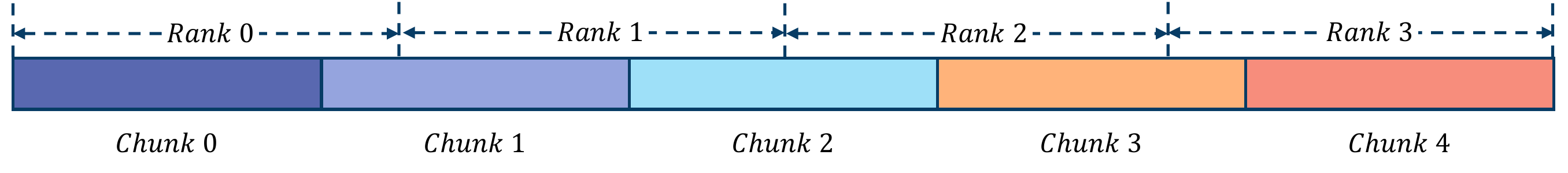}
        \caption{Ulysses context split logic.}
    \end{subfigure}
    \hfill
    \begin{subfigure}[b]{1.0\textwidth}
        \centering
        \includegraphics[width=1.0\textwidth]{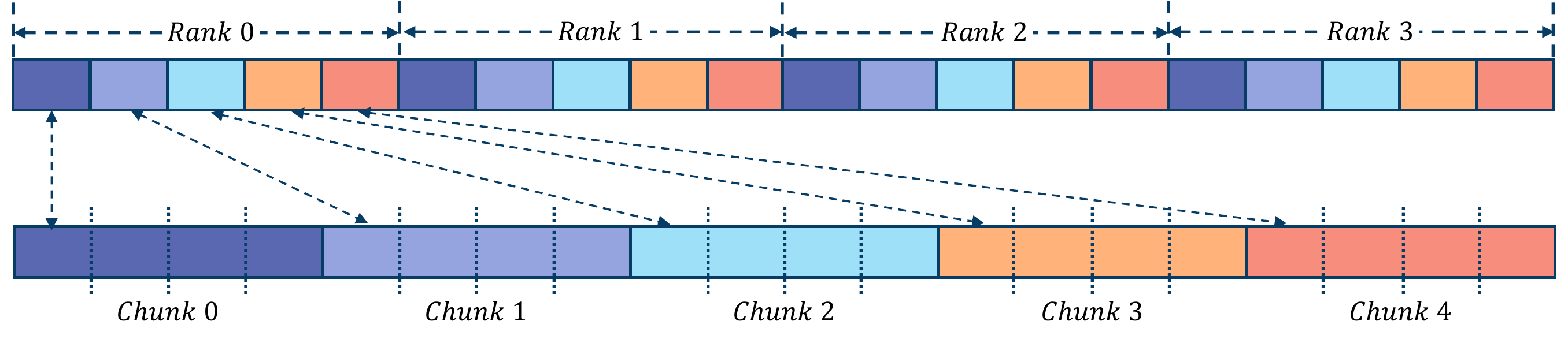}
        \caption{CSO context split logic.}
    \end{subfigure}
    \end{minipage}
    \caption{Overview of Ulysses and CSO context split logic.}
    \label{fig:context_split_logic_overview}
\end{figure}

\subsection{Context Shuffle Overlap}
\label{sec:appendix_context_shuffle_overlap}
We optimized based on the Ulysses CP algorithm, striving to overlap communication with computation or data movement. Since the RTX 4090 GPUs communicate via PCIe, which has a relatively low bandwidth, we further proposed the \textbf{CSO (Context Shuffle Overlap)} algorithm to optimize communication more deeply.

The key difference between CSO and Ulysses lies in how the context is partitioned. In Ulysses, all chunks are distributed sequentially across different ranks. In contrast, CSO assigns each rank a partial view of every chunk. As illustrated in Fig.~\ref{fig:context_split_logic_overview}, this alternative partitioning strategy allows CSO to conveniently overlap computation and communication at the chunk level. Assuming we have 5 chunks, the complete process of CSO is as follows:
\begin{itemize} 
    \item $k$-communication and $v$-communication of all chunks overlaps with $q$-computation of all chunks
    \item $q$-communication of chunk 1 overlaps with KV cache updates
    \item $q$-communication of chunk 2 overlaps with $o$-computation of chunk 1
    \item $q$-communication of chunk 3 and $o$-communication of chunk 1 overlaps with $o$-computation of chunk 2
    \item $q$-communication of chunk 4 and $o$-communication of chunk 2 overlaps with $o$-computation of chunk 3
    \item $q$-communication of chunk 5 and $o$-communication of chunk 3 overlaps with $o$-computation of chunk 4
    \item $o$-communication of chunk 4 overlaps with $o$-computation of chunk 5
    \item $o$-communication of chunk 5 overlaps with cross-attention computation
\end{itemize}

In addition, the CSO partition pattern enables communication operations to be split into multiple balanced all-to-all communications. These balanced operations offer better performance compared to unbalanced all-to-all communication and also allow for efficient subsequent merging.

\section{Training Infrastructure}

\subsection{MagiAttention Materials}\label{appendix:magiattn}

\begin{figure}[htbp]
    \centering
    \includegraphics[width=\linewidth]{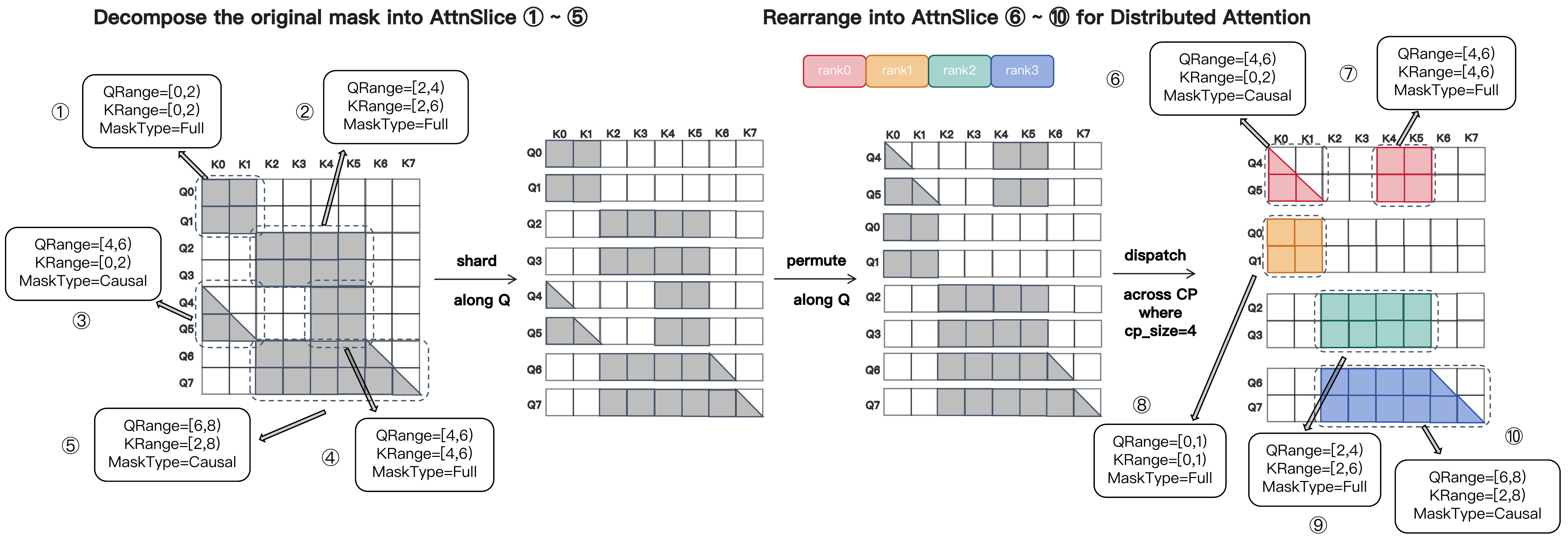}
    \caption{Illustration of \texttt{AttnSlice} formulation for some irregular mask (see~\ref{sec:magiattn_ffa}). It decomposes the original mask into multiple \texttt{AttnSlice}s and allows re-expression of fractal masks after rearrangement across CP ranks, making it suitable for distributed attention. Note that computation load balance across CP ranks is not considered in this illustration.}
    \label{fig:attnslice_interpret}
\end{figure}

\begin{figure}[htbp]
    \centering
    \includegraphics[width=\linewidth]{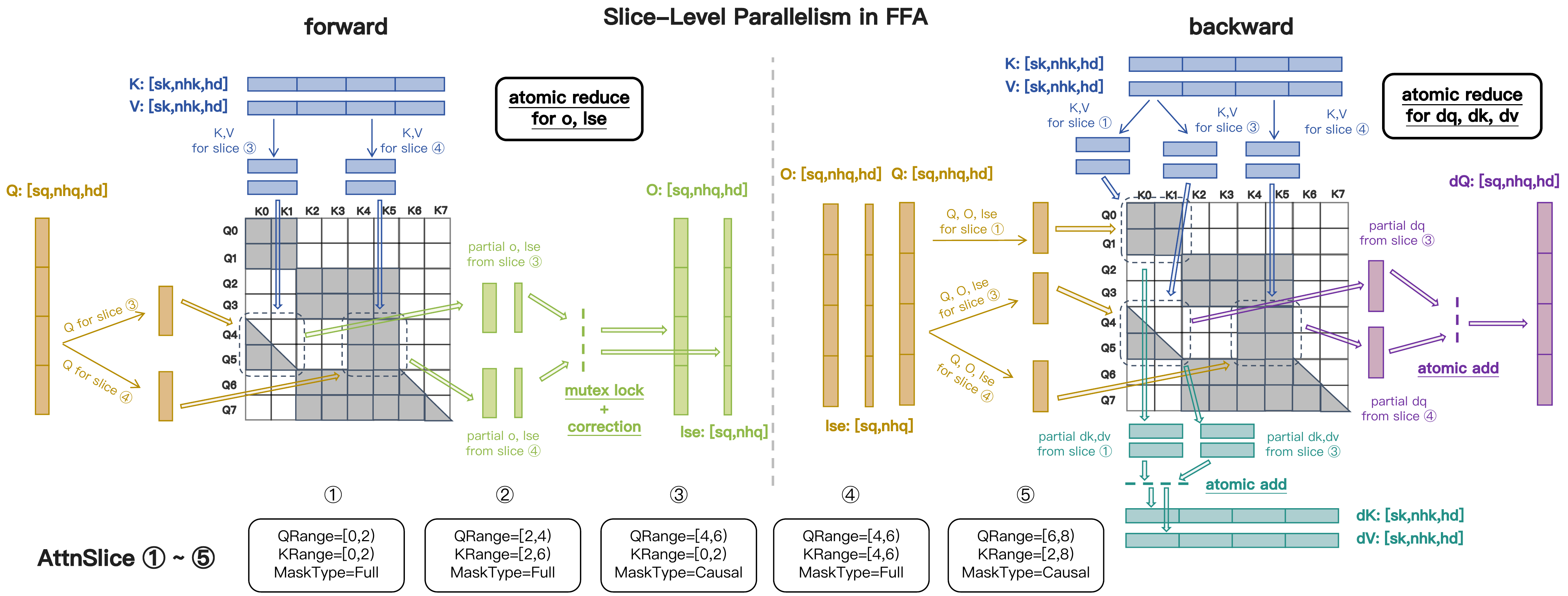}
    \caption{Illustration of slice-level parallelism in FFA for both forward and backward kernels (see~\ref{sec:magiattn_ffa}). The overlapping nature across slices in both rows (\texttt{QRange}) and columns (\texttt{KRange}) necessitates \textit{atomic reduce} operations in both kernels to ensure correct reduction.}
    \label{fig:ffa_slice_parallel}
\end{figure}

\begin{figure}
    \centering
    \includegraphics[width=\linewidth]{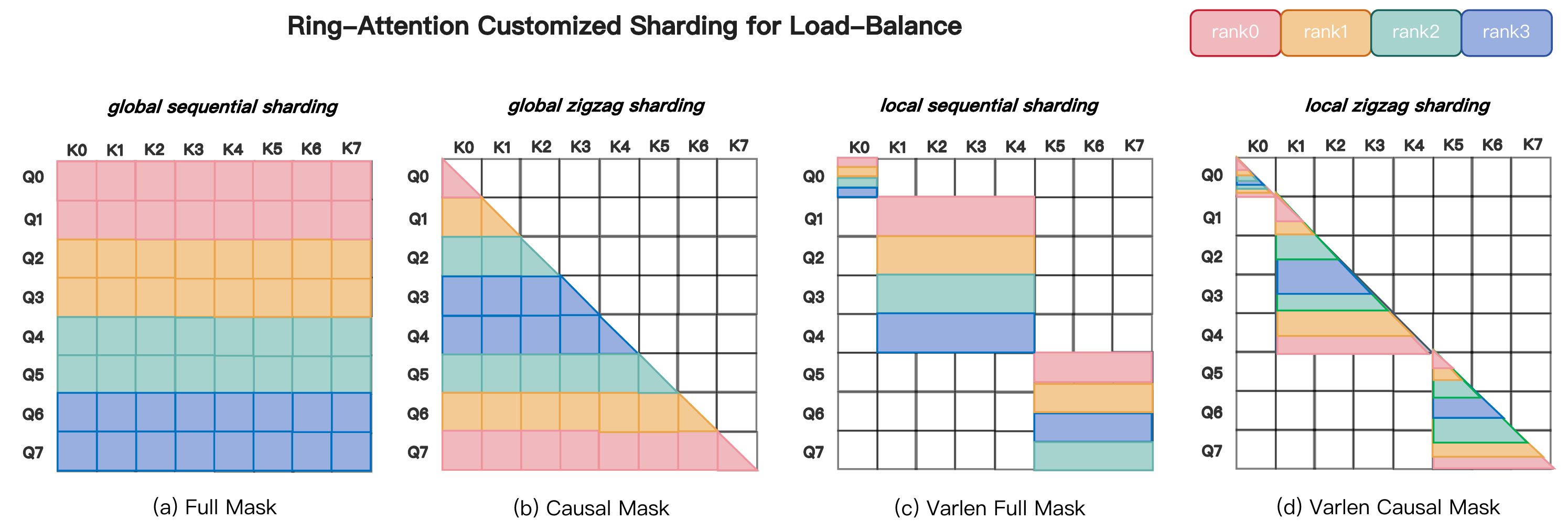}
    \caption{Illustration of Ring-Attention’s customized sharding strategies for load balancing. (a) Full mask uses sequential sharding for the global mask; (b) Causal mask employs tailored \textit{zigzag sharding}~\citep{ring_flash_attention_issue2}; (c) Varlen full mask applies sequential sharding per local mask (one per packed sample); (d) Varlen causal mask uses \textit{zigzag sharding} per local mask, causing performance degradation from fragmentation and padding.}
    \label{fig:ring_attn_load_balance}
\end{figure}

\begin{figure}[htbp]
    \centering
    \includegraphics[width=\linewidth]{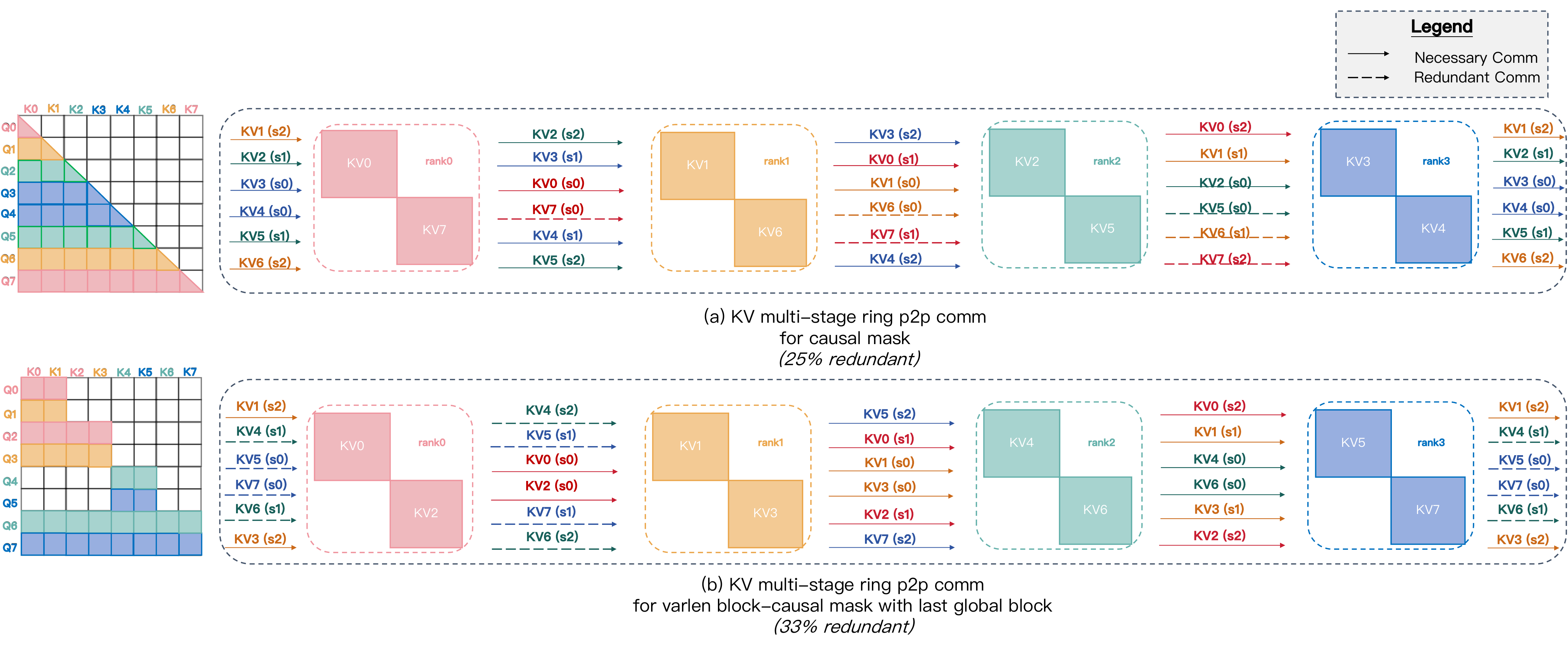}
    \caption{Examples illustrating redundant communication in Ring P2P patterns for distributed attention given heterogeneous masks (see~\ref{sec:magiattn_comm}).: (a) Even with a simple causal mask, Ring P2P incurs \textbf{25\%} redundant communication; (b) For irregular mask patterns such as varlen block-causal mask with last global block, Ring P2P results in over \textbf{33\%} redundancy.}
    \label{fig:ring-p2p-redundancy}
\end{figure}

\begin{figure}[htbp]
    \centering
    \includegraphics[width=\linewidth]{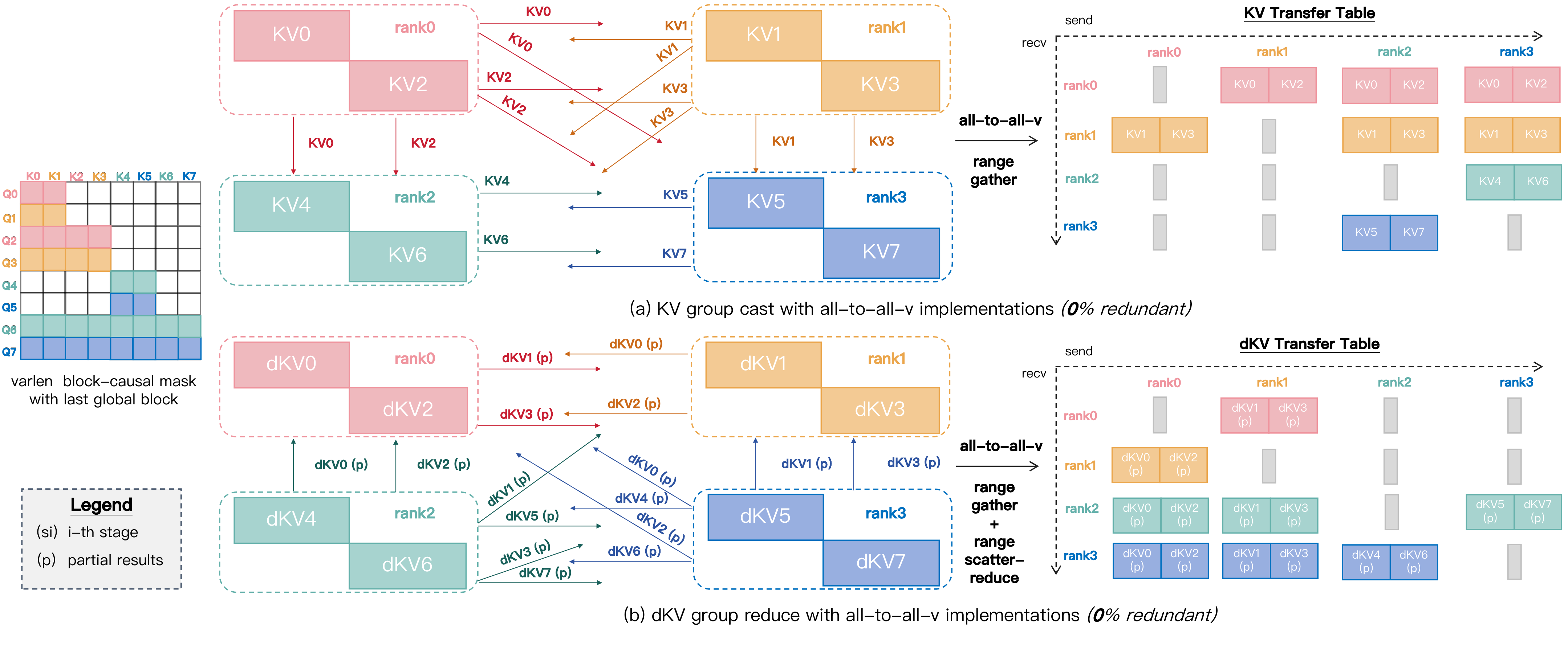}
    \caption{Illustration of \texttt{group-cast}/\texttt{group-reduce} primitives for zero redundancy, using the varlen block-causal mask with the last global block as an example for irregular patterns (see~\ref{sec:magiattn_comm}). (a) In both forward and backward passes, the \texttt{group-cast} primitive internally analyzes and generates a transfer table for $\mathrm{KV}$ send/receive buffers, and launches the underlying \texttt{all-to-all-v} to complete communication with our custom \textit{Range Gather} kernel for pre-/post-processing. (b) In the backward pass, \texttt{group-reduce} similarly handles the partial $\mathrm{dKV}$ communication for reduction, using \texttt{all-to-all-v} with the \textit{Range Gather} kernel for pre-processing and the \textit{Range Scatter-Reduce} kernel for post-processing.}
    \label{fig:group-gather-reduce-all2allv}
\end{figure}

\begin{figure}[htbp]
    \centering
    \includegraphics[width=\linewidth]{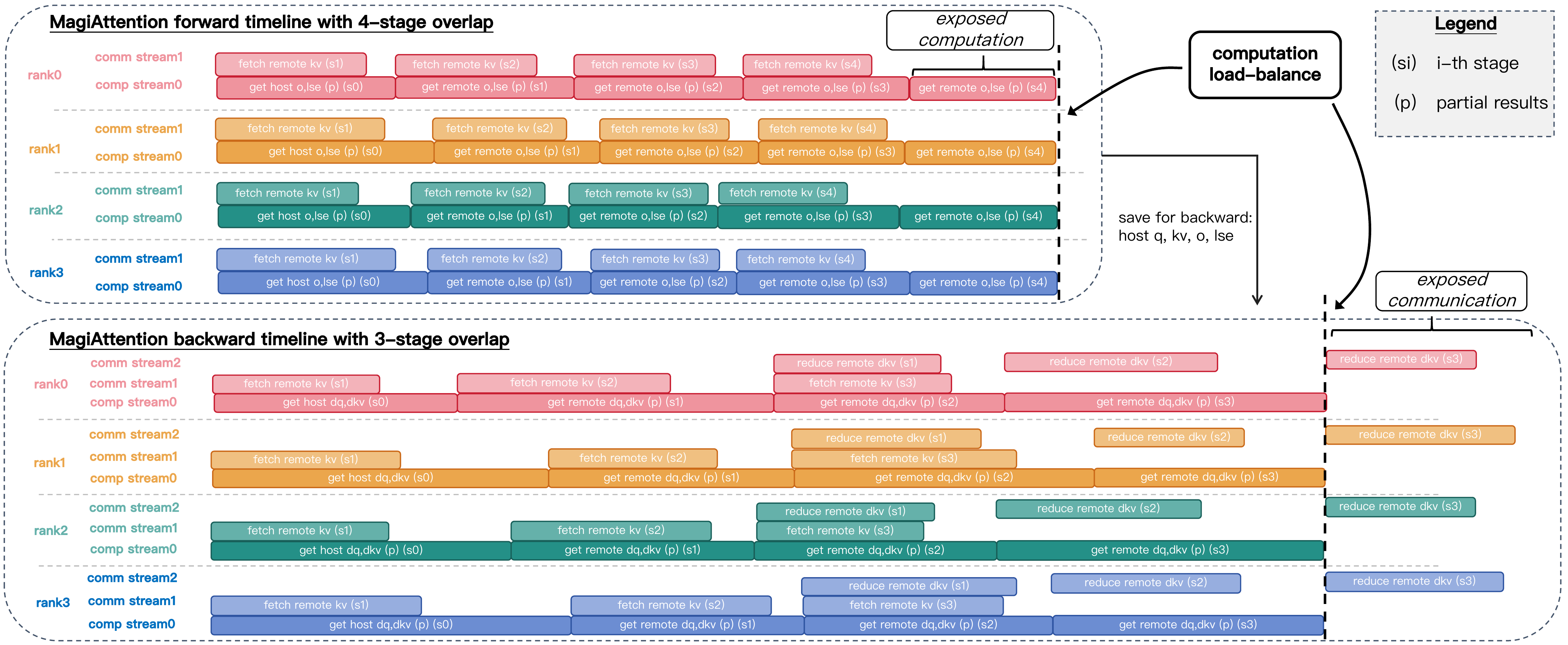}
    \caption{Schematic of MagiAttention's multi-stage overlap scheduling (see ~\ref{sec:magiattn_overlap}). (a) Forward pass: 4-stage scheduling overlaps computation (partial attention outputs and \textit{lse} factors) with prefetching of next-stage $\mathrm{KV}$ requests (where applicable), hiding all communication overhead with the final stage's computation exposed. (b) Backward pass: 3-stage scheduling overlaps computation (partial $\mathrm{dQ}$, $\mathrm{dKV}$) with prefetching of next-stage $\mathrm{KV}$ requests and reduction of prior $\mathrm{dKV}$ requests, hiding all communication overhead except the $\mathrm{dKV}$ reduction of the final stage.}  
    \label{fig:multi_stage_overlap_fwd_bwd}
\end{figure}

\begin{algorithm}[htbp]
\caption{Greedy Load-Balance Dispatch Algorithm via Min-Heap w.r.t.~\ref{sec:magiattn_comp}}
\label{alg:minhp}
\begin{algorithmic}[1]
\REQUIRE Dispatch chunk and area pairs $\{(C_i, \mathrm{Area}(C_i))\}_{i=1}^n$, number of buckets $cp\_size$ ($cp\_size \mid n$)
\ENSURE A dispatch mapping minimizing maximum bucket workload, satisfying Eq.~(\ref{eq:comp_load_balance})

\STATE Sort dispatch chunks $\{C_i\}_{i=1}^n$ in descending order by $\mathrm{Area}(C_i)$

\STATE Compute chunk capacity per bucket: $chunk\_per\_bucket \gets n / cp\_size$

\STATE Initialize buckets: workload counters $W \gets [0]^{cp\_size}$, chunk counters $count \gets [0]^{cp\_size}$, and mapping $B_j \gets \emptyset, \forall j$

\STATE Initialize min-heap $\mathcal{H}$ with tuples $(0, j)$ for each bucket $j \in [1, cp\_size]$

\FOR{each chunk $C_i$ in sorted order}
    \REPEAT
        \STATE $(w_{\text{min}}, j_{\text{min}}) \gets \text{extract-min}(\mathcal{H})$
    \UNTIL{$count[j_{\text{min}}] < chunk\_per\_bucket$} \COMMENT{Select least-loaded bucket that is not full}

    \STATE Assign chunk: $B_{j_{\text{min}}} \gets B_{j_{\text{min}}} \cup \{C_i\}$
    
    \STATE Update workload and count:
    \[
    W[j_{\text{min}}] \gets W[j_{\text{min}}] + \mathrm{Area}(C_i), \quad count[j_{\text{min}}] \gets count[j_{\text{min}}] + 1
    \]

    \STATE Insert updated tuple $(W[j_{\text{min}}], j_{\text{min}})$ back into $\mathcal{H}$

\ENDFOR

\RETURN Bucket assignments $\{B_j\}_{j=1}^{cp\_size}$ and maximum bucket workload $\max_j W[j]$

\end{algorithmic}
\end{algorithm}

\begin{algorithm}[htbp]
\caption{Dynamic Overlap Stage Search Algorithm w.r.t.~\ref{sec:magiattn_overlap}}
\label{alg:dynamic-overlap-search}
\begin{algorithmic}[1]
\REQUIRE Remote $\mathrm{KV}$/$\mathrm{dKV}$ requests per rank; kernel cost models for FFA ($\mathrm{C}_{ffa}(\cdot)$), \texttt{group-cast} ($\mathrm{C}_{gc}(\cdot)$), and \texttt{group-reduce} ($\mathrm{C}_{gr}(\cdot)$), estimated via offline profiling;
hyperparameters to control search range: $(min\_chunk\_size$,$max\_num\_chunks)$
\ENSURE Optimal number of overlap stages for both forward and backward passes

\FOR{each $\mathrm{rank}_i$ in parallel}
    \STATE \textbf{[Step 1: Communication Partition]} Partition remote $\mathrm{KV}$/$\mathrm{dKV}$ requests into $p_i$ fine-grained communication packages, with sizes bounded by hyperparameters\\$(min\_chunk\_size, max\_num\_chunks)$
    
    \STATE \textbf{[Step 2: Cost Evaluation]} 
    \FOR{$s_i \gets p_i$ to $1$}
        \STATE Randomly assign $p_i$ communication packages into $s_i$ stages, and pack them
        \STATE Estimate the overall timeline costs for both forward and backward as:
        
        \begin{align*}
        C^{(fwd)}(s_i) &\gets \sum_{j=0}^{s_i-1} \underbrace{\max \left\{ \mathrm{C}_{gc}(j{+}1),\; \mathrm{C}_{ffa}(j) \right\}}_{\text{overlapped}} + \underbrace{\mathrm{C}_{ffa}(s_i)}_{\text{exposed}} \\
        C^{(bwd)}(s_i) &\gets \sum_{j=0}^{s_i} \underbrace{\max \left\{ \mathrm{C}_{gc}(j{+}1), \mathrm{C}_{ffa}(j), \mathrm{C}_{gr}(j{-}1) \right\}}_{\text{overlapped}} + \underbrace{\mathrm{C}_{gr}(s_i)}_{\text{exposed}}
        \end{align*}
        
        \COMMENT{Where $\mathrm{C}_{ffa}(0)$ denotes the host computation cost and $\mathrm{C}_{gc}(s_i{+}1),\mathrm{C}_{gr}(0),\mathrm{C}_{gr}(-1)$ are all assigned to $0$ to handle boundary conditions}
    \ENDFOR

    \STATE \textbf{[Step 3: Local Selection]} 
    \STATE $s^{*(fwd)}_i \gets \arg\min\limits_{s_i} C^{(fwd)}(s_i)$ \COMMENT{Optimal forward stage count for $\mathrm{rank}_i$}
    \STATE $s^{*(bwd)}_i \gets \arg\min\limits_{s_i} C^{(bwd)}(s_i)$ \COMMENT{Optimal backward stage count for $\mathrm{rank}_i$}
\ENDFOR

\STATE \textbf{[Step 4: Global Synchronization]} 
\STATE $num\_stages^{(fwd)} \gets \max\limits_{i} s^{*(fwd)}_i$ via All-Reduce across all CP ranks
\STATE $num\_stages^{(bwd)} \gets \max\limits_{i} s^{*(bwd)}_i$ via All-Reduce across all CP ranks

\RETURN $num\_stages^{(fwd)}$, $num\_stages^{(bwd)}$

\end{algorithmic}
\end{algorithm}

\clearpage
\subsection{MagiAttention Experiments}\label{appendix:magiattn_exps}

\subsubsection{Benchmarking MagiAttention kernel-level performance and flexibility}\label{appendix:magiattn_exps_ffa}

\begin{figure}[htbp]
    \centering

    \begin{minipage}{\textwidth}
    
    \begin{subfigure}[b]{\textwidth}
        \centering
        \includegraphics[width=\textwidth]{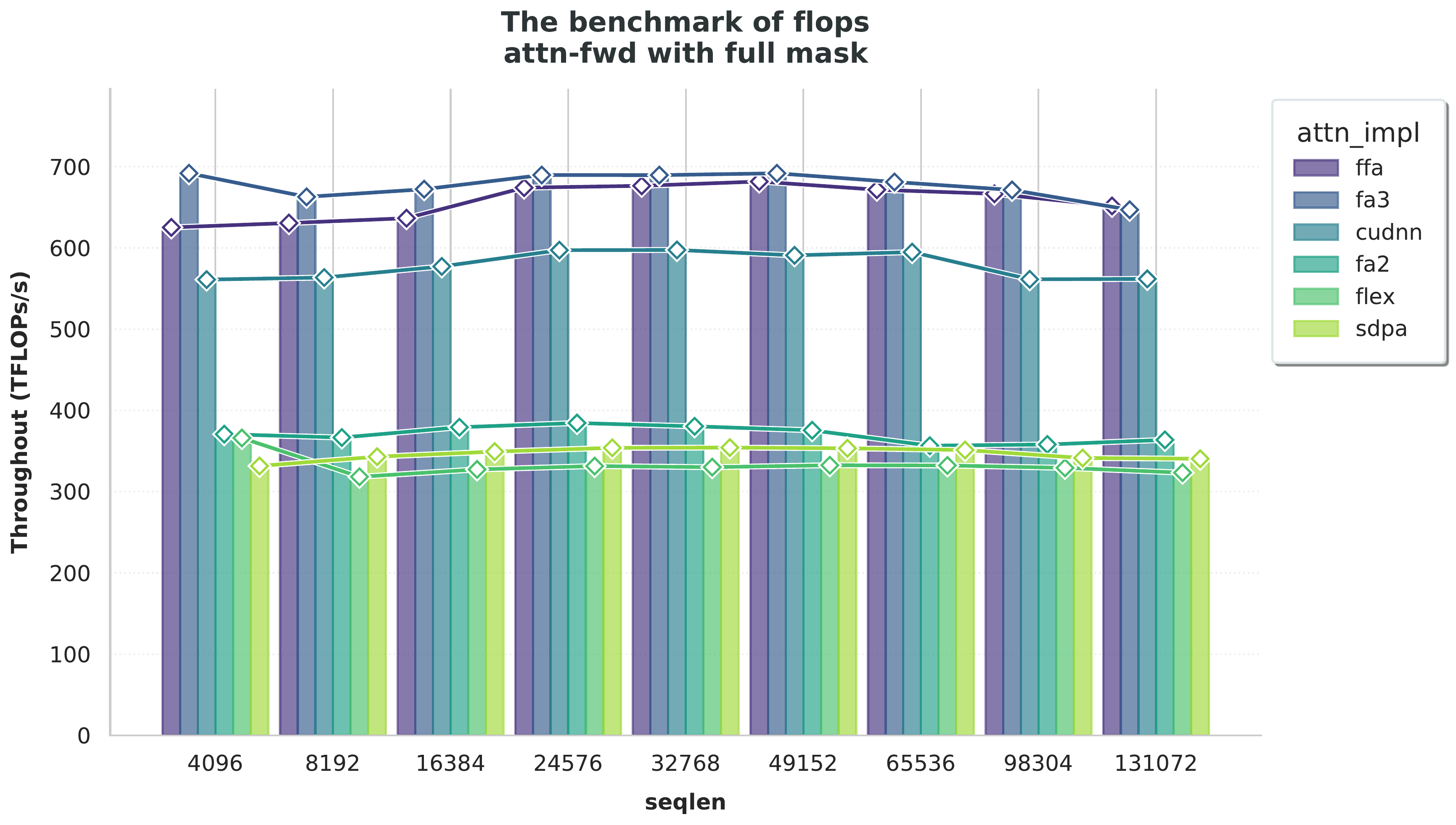}
        \caption{Forward pass for full mask.}
        \label{fig:ffa_fwd_perf_report_full}
    \end{subfigure}

    \end{minipage}

    \begin{minipage}{\textwidth}
    
    \begin{subfigure}[b]{\textwidth}
        \centering
        \includegraphics[width=\textwidth]{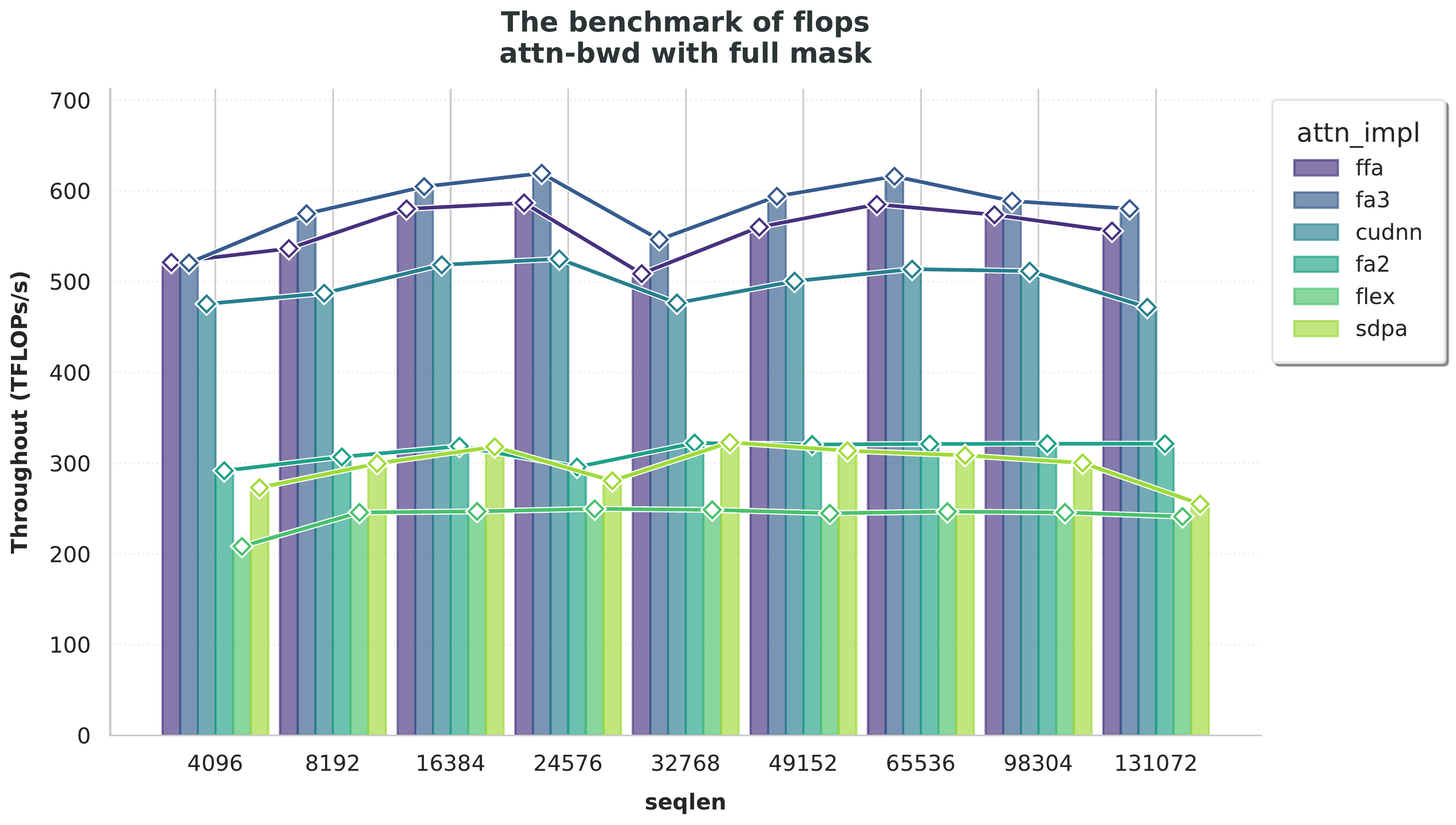}
        \caption{Backward pass for full mask.}
        \label{fig:ffa_bwd_perf_report_full}
    \end{subfigure}

    \end{minipage}

    \caption{Benchmarking FFA's performance and flexibility against other leading attention kernels for full mask scenarios.}
    \label{fig:ffa_perf_report_full}
\end{figure}

\begin{figure}[htbp]
    \centering

    \begin{minipage}{\textwidth}
    
    \begin{subfigure}[b]{\textwidth}
        \centering
        \includegraphics[width=\textwidth]{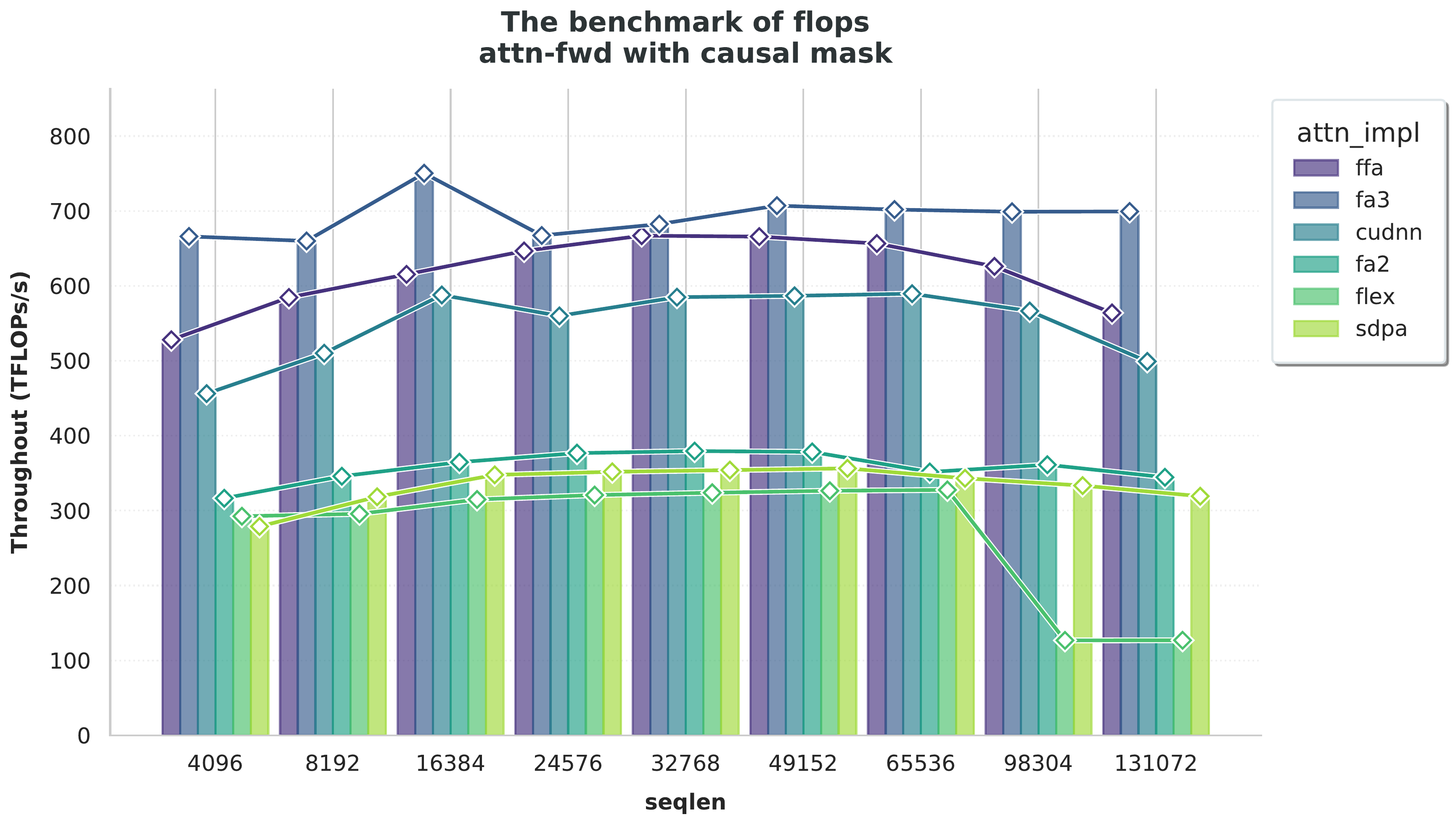}
        \caption{Forward pass for causal mask.}
        \label{fig:ffa_fwd_perf_report_causal}
    \end{subfigure}

    \end{minipage}

    \begin{minipage}{\textwidth}
    
    \begin{subfigure}[b]{\textwidth}
        \centering
        \includegraphics[width=\textwidth]{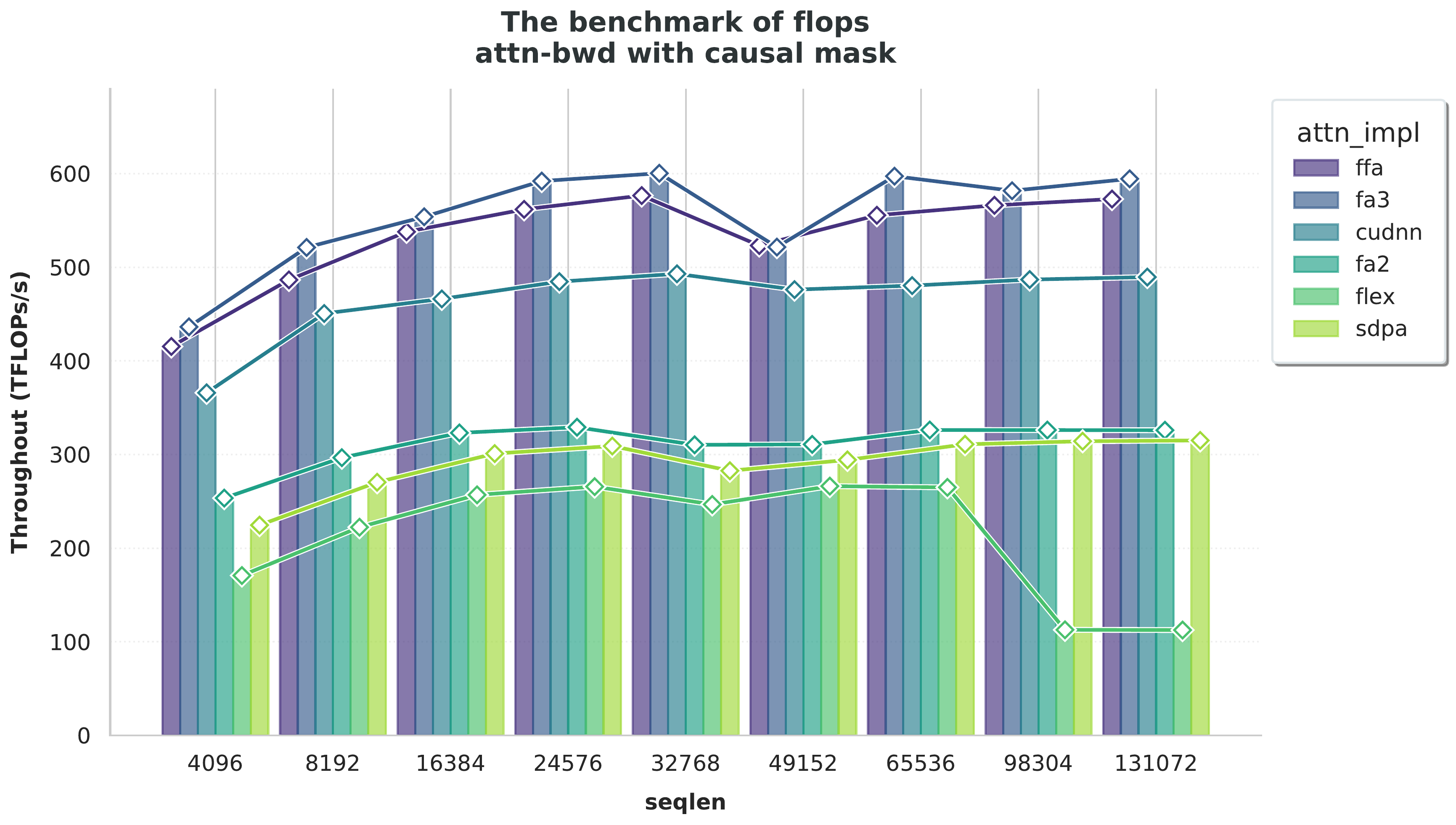}
        \caption{Backward pass for causal mask.}
        \label{fig:ffa_bwd_perf_report_causal}
    \end{subfigure}

    \end{minipage}

    \caption{Benchmarking FFA's performance and flexibility against other leading attention kernels for causal mask scenarios.}
    \label{fig:ffa_perf_report_causal}
\end{figure}

\begin{figure}[htbp]
    \centering

    \begin{minipage}{\textwidth}
    
    \begin{subfigure}[b]{\textwidth}
        \centering
        \includegraphics[width=\textwidth]{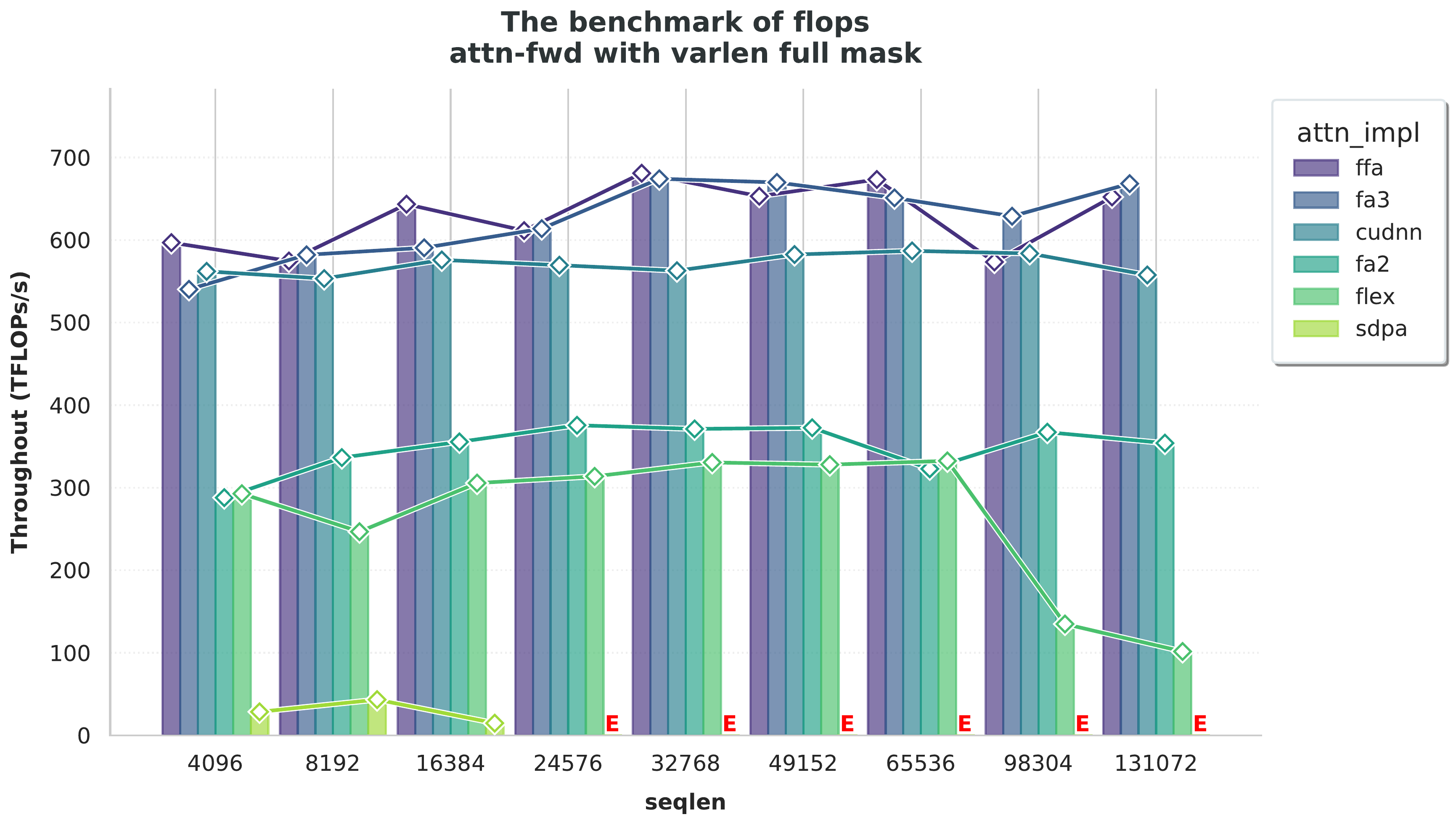}
        \caption{Forward pass for varlen full mask.}
        \label{fig:ffa_fwd_perf_report_varlen_full}
    \end{subfigure}

    \end{minipage}

    \begin{minipage}{\textwidth}
    
    \begin{subfigure}[b]{\textwidth}
        \centering
        \includegraphics[width=\textwidth]{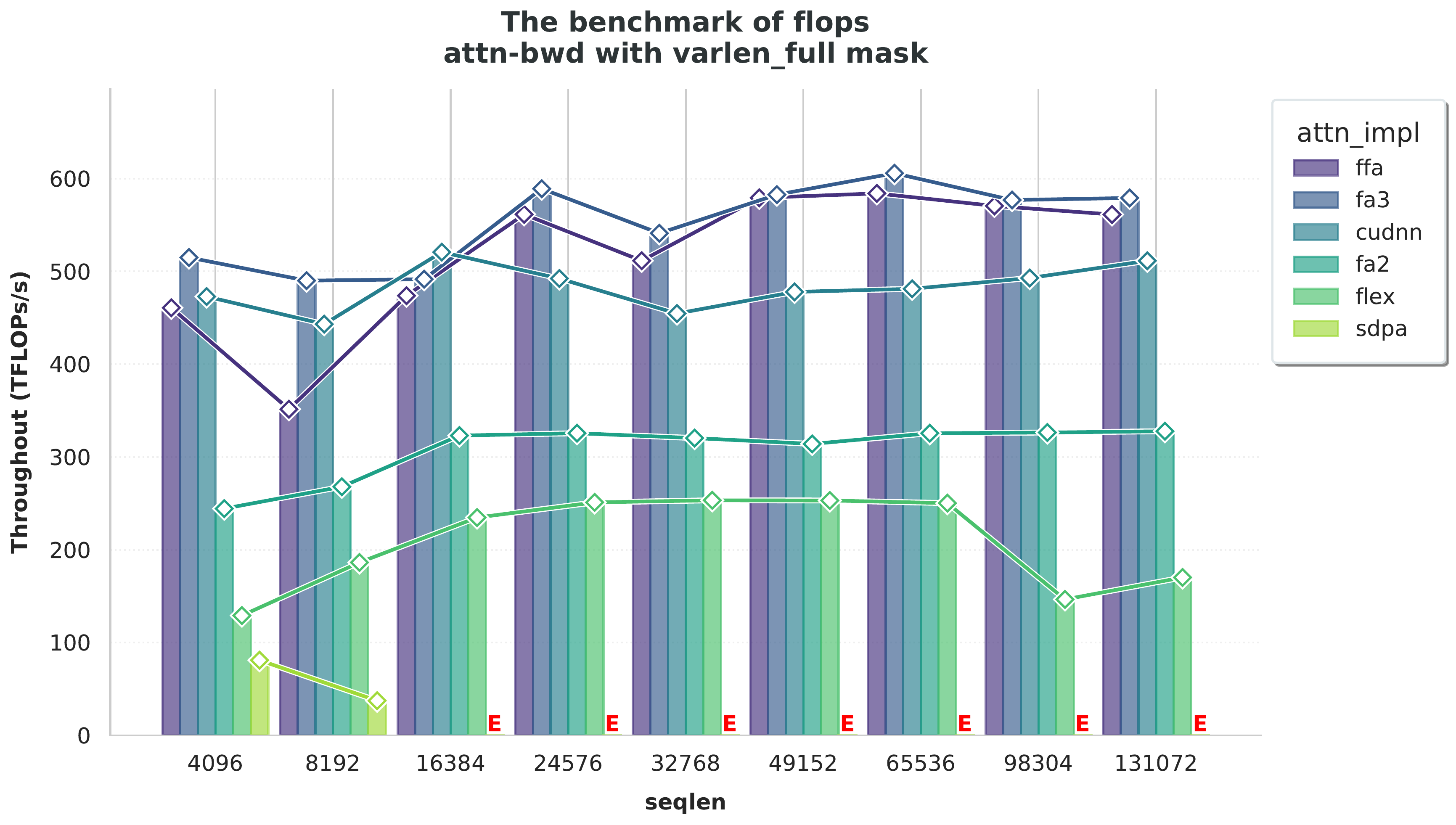}
        \caption{Backward pass for varlen full mask.}
        \label{fig:ffa_bwd_perf_report_varlen_full}
    \end{subfigure}

    \end{minipage}

    \caption{Benchmarking FFA's performance and flexibility against other leading attention kernels for varlen full mask scenarios.  (Note that: the $\mathbf{E}$ symbol indicates the corresponding distributed attention implementation raises \textit{Cuda Out of Memory} error in that specific configuration.)}
    \label{fig:ffa_perf_report_varlen_full}
\end{figure}

\begin{figure}[htbp]
    \centering

    \begin{minipage}{\textwidth}
    
    \begin{subfigure}[b]{\textwidth}
        \centering
        \includegraphics[width=\textwidth]{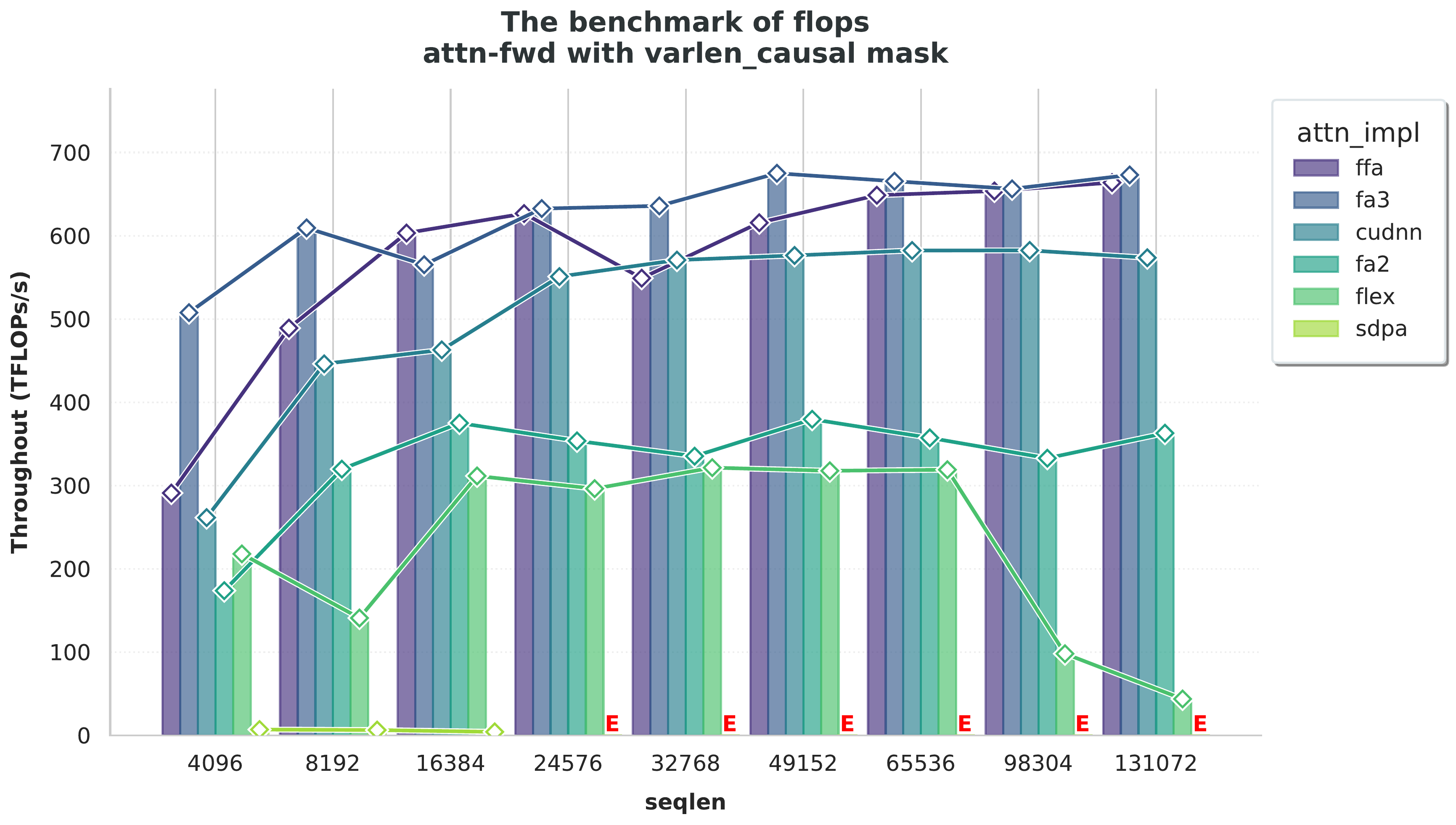}
        \caption{Forward pass for varlen causal mask}
        \label{fig:ffa_fwd_perf_report_varlen_causal}
    \end{subfigure}

    \end{minipage}

    \begin{minipage}{\textwidth}
    
    \begin{subfigure}[b]{\textwidth}
        \centering
        \includegraphics[width=\textwidth]{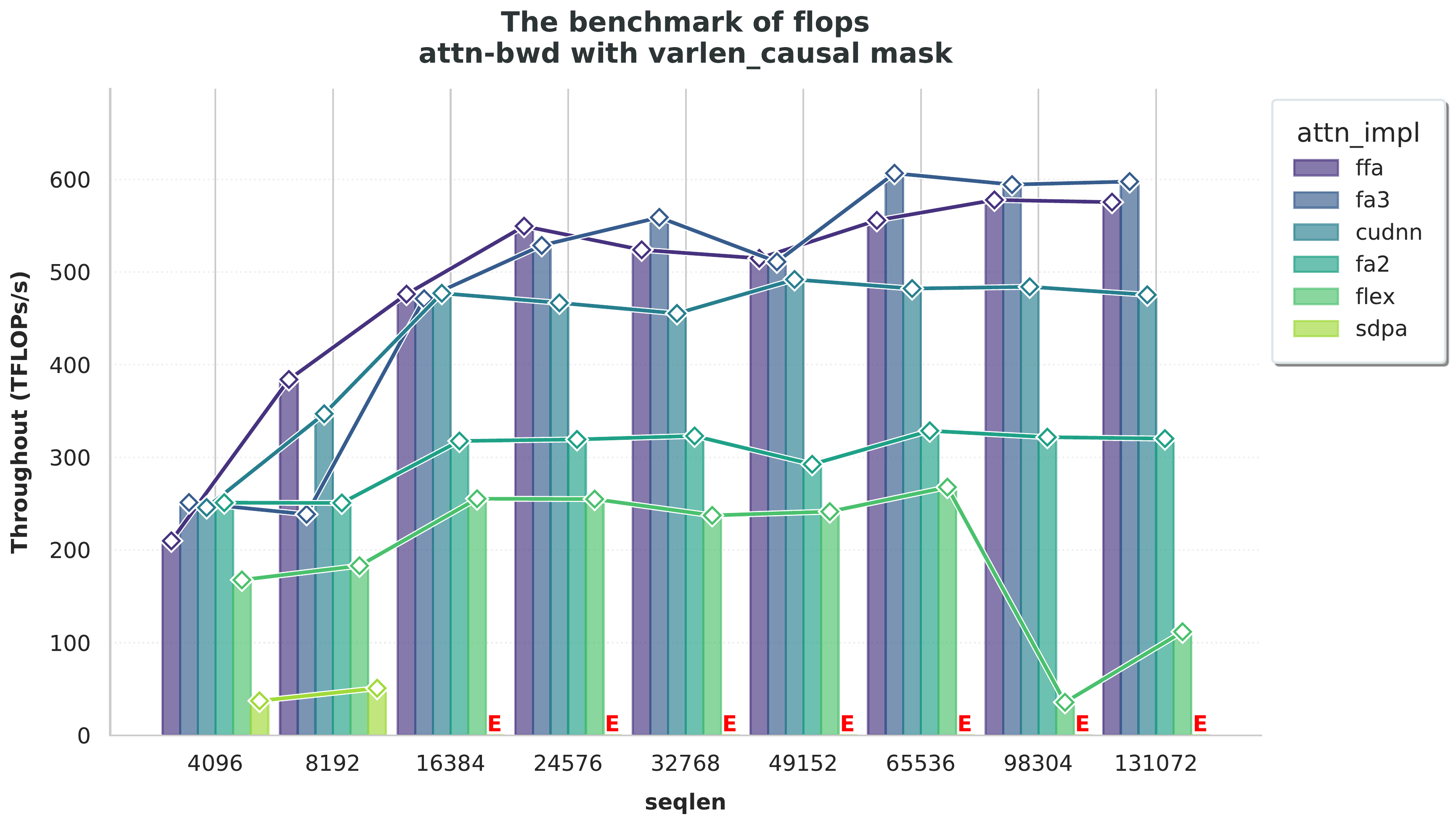}
        \caption{Backward pass for varlen causal mask}
        \label{fig:ffa_bwd_perf_report_varlen_causal}
    \end{subfigure}

    \end{minipage}

    \caption{Benchmarking FFA's performance and flexibility against other leading attention kernels for varlen causal mask scenarios.  (Note that: the $\mathbf{E}$ symbol indicates the corresponding distributed attention implementation raises \textit{Cuda Out of Memory} error in that specific configuration.)}
    \label{fig:ffa_perf_report_varlen_causal}
\end{figure}

\begin{figure}[htbp]
    \centering

    \begin{minipage}{\textwidth}
    
    \begin{subfigure}[b]{\textwidth}
        \centering
        \includegraphics[width=\textwidth]{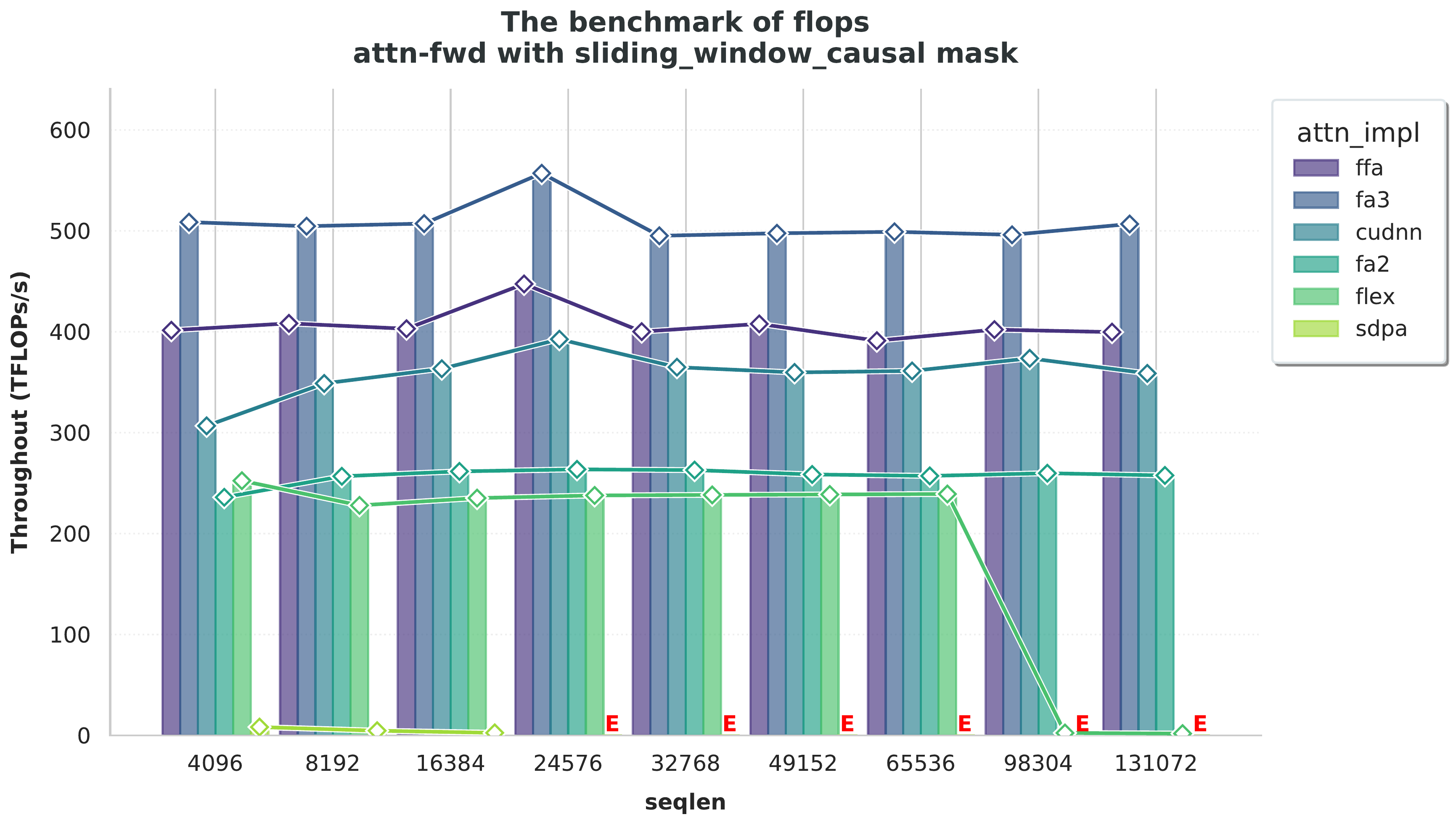}
        \caption{Forward pass for sliding-window causal mask.}
        \label{fig:ffa_fwd_perf_report_sw_causal}
    \end{subfigure}

    \end{minipage}

    \begin{minipage}{\textwidth}
    
    \begin{subfigure}[b]{\textwidth}
        \centering
        \includegraphics[width=\textwidth]{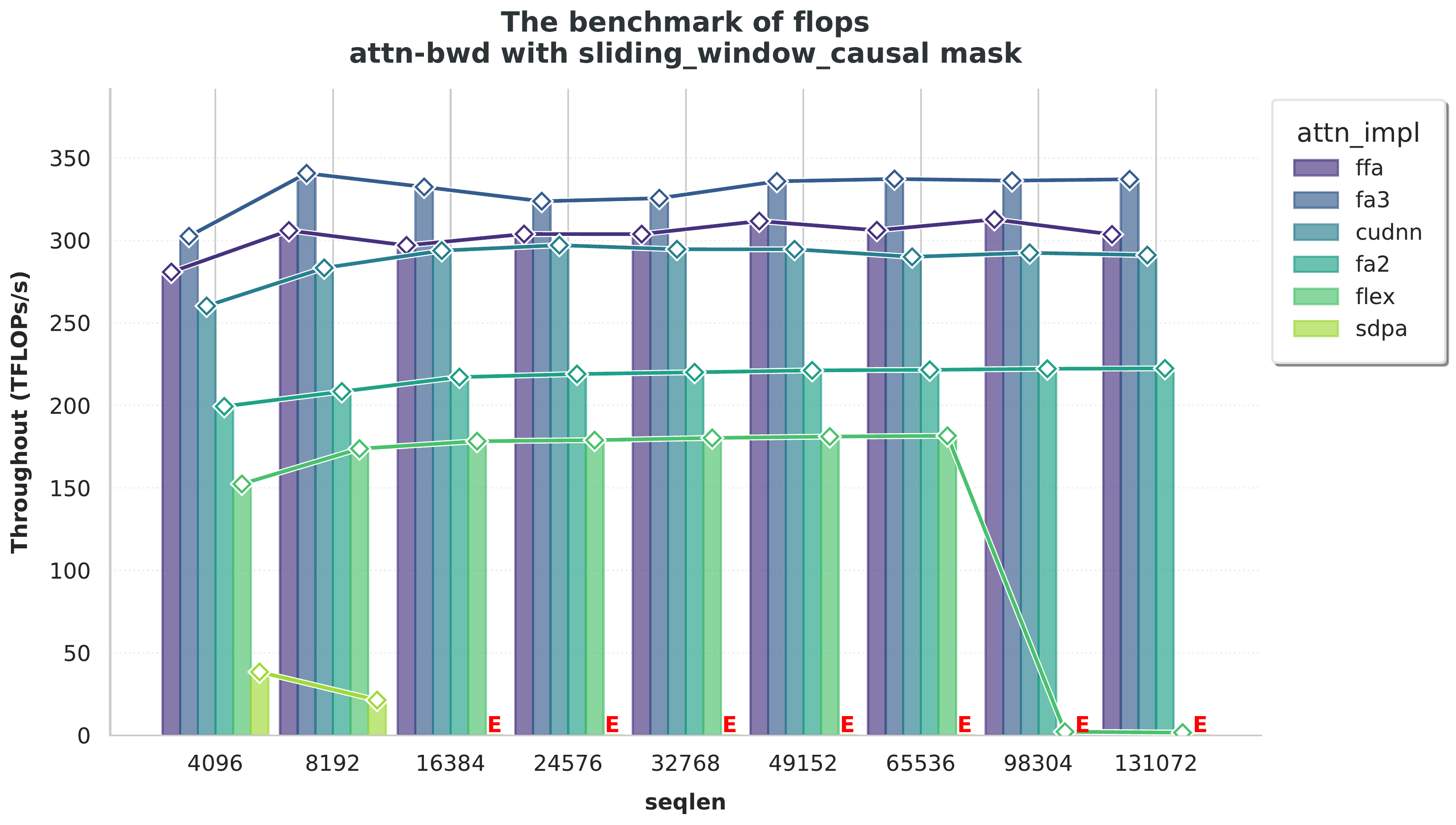}
        \caption{Backward pass for sliding-window causal mask.}
        \label{fig:ffa_bwd_perf_report_sw_causal}
    \end{subfigure}

    \end{minipage}

    \caption{Benchmarking FFA's performance and flexibility against other leading attention kernels for sliding-window causal mask scenarios.  (Note that: the $\mathbf{E}$ symbol indicates the corresponding distributed attention implementation raises \textit{Cuda Out of Memory} error in that specific configuration.)}
    \label{fig:ffa_perf_report_sw_causal}
\end{figure}

To demonstrate FFA kernels' state-of-the-art performance and flexibility in handling ultra-long, heterogeneous mask training, we measure the throughput (in $\mathrm{TFLOPs/s}$) on Hopper GPUs for both forward and backward passes of prevalent attention kernels across standard mask patterns (see Fig.\ref{fig:ffa_perf_report_full}, Fig.\ref{fig:ffa_perf_report_causal}) with their varlen variants (see Fig.\ref{fig:ffa_perf_report_varlen_full} and Fig.\ref{fig:ffa_perf_report_varlen_causal}), 
and some irregular mask patterns (see Fig.\ref{fig:ffa_perf_report_sw_causal}).

\begin{figure}[htbp]
    \centering
    \includegraphics[width=\linewidth]{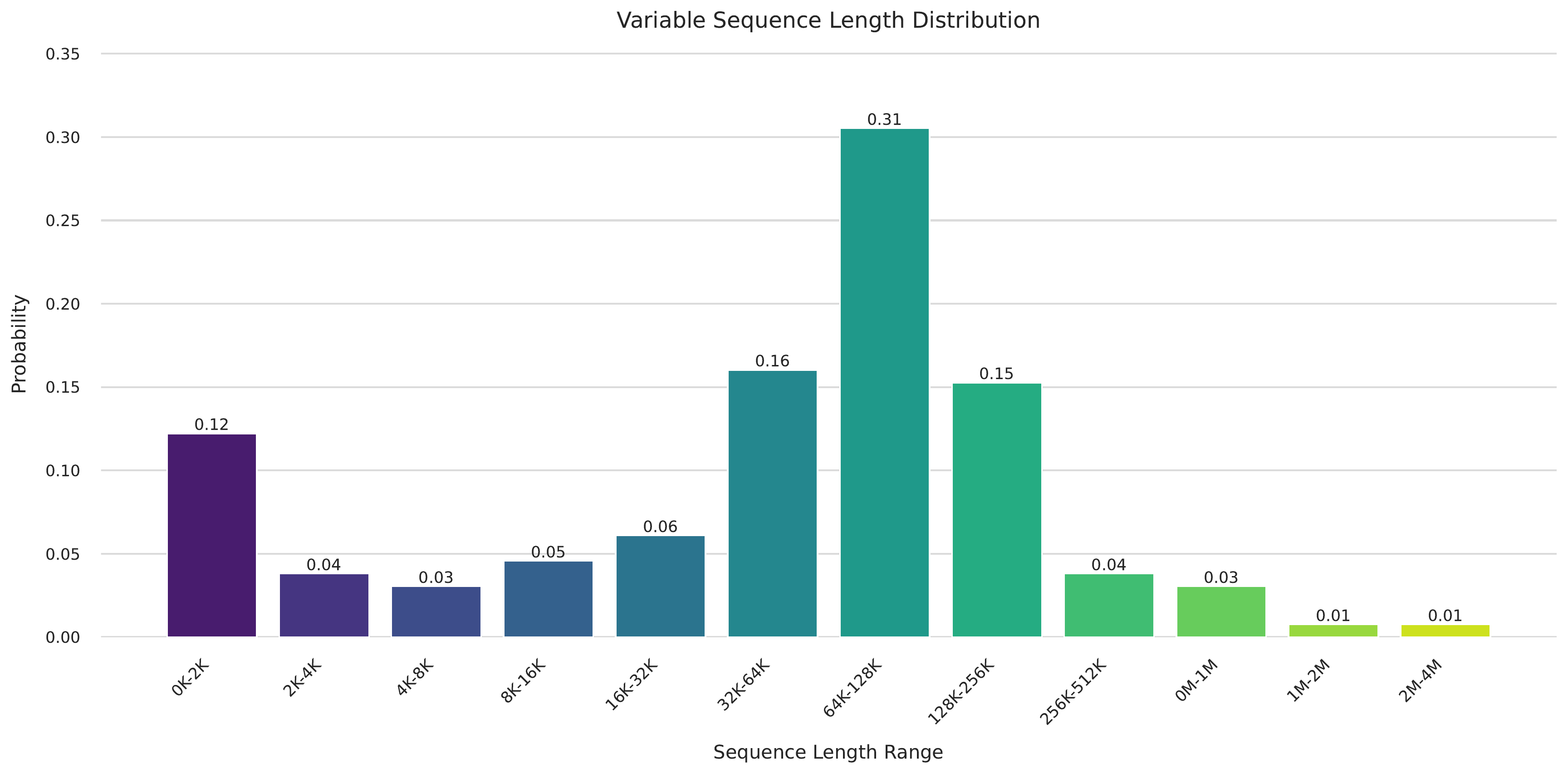}
    \caption{Distribution of sequence lengths in the dataset~\citep{xu2024chatqa}, used to sample and construct the variable-length data for both kernel-level and module-level experiments of MagiAttention.}
    \label{fig:varlen_seqlen_distribution}
\end{figure}

Benchmark settings: for each mask pattern, we vary the sequence length $seqlen$ from $4k,8k,16k,...,$ up to $128k$ ($seqlen_q = seqlen_k = seqlen$) while measuring throughput (in $\mathrm{TFLOPs/s}$) for forward and backward passes of different attention kernels. Other configurations are fixed using common training settings (see Tab.\ref{tab:ffa_exps_settings}) to focus on the impact of sequence length and mask pattern. For the varlen packed data, we simply follow the variable sequence length distribution in the open-sourced dataset~\citep{xu2024chatqa} (see Fig.\ref{fig:varlen_seqlen_distribution}), from which we sample to pack and pad to the required $seqlen$.

To calculate the $\mathrm{TFLOPs/s}$ for various mask patterns during both forward and backward passes, we use the subsequent equations, following Flash-Attention~\citep{dao2023flashattention}:

\begin{align}
    \mathrm{FLOPs}^{(fwd)} &= \underbrace{2}_{\text{2 matmul}} \times \underbrace{2}_{\text{2 flops per matmul}} \times\;\; \mathrm{MaskArea}(seqlen, mask\_type) \label{eq:flops_fwd}\\
    &\times batch\_size \times num\_heads_q \times head\_dim \nonumber\\
    \mathrm{FLOPs}^{(bwd)} &= \underbrace{2.5}_{\text{5 matmul due to recomputation}} \times\;\; \mathrm{FLOPs}^{(fwd)} \label{eq:flops_bwd}\\
    where \;\;& \mathrm{MaskArea}(seqlen, full) = seqlen^2, \nonumber\\
     \;\;& \mathrm{MaskArea}(seqlen, causal) = \frac{seqlen(seqlen+1)}{2}, \;\; ...\nonumber \\
     \mathrm{TFLOPs/s}^{(wd)} &= \cfrac{\mathrm{FLOPs}^{(wd)}}{\mathrm{Runtime}^{(wd)}}, \quad wd \in \{fwd, bwd\}
\end{align}

\begin{table}[htbp]
    \centering
    \small
    \begin{tabular}{c|c}
    \toprule
     Settings & Value \\
    \midrule
    Batch Size (b) & 1 \\
    Number of Heads (nh)  & \begin{tabular}[c]{@{}c@{}}nhq:nhk:nhv = 64:8:8\\ (\textit{GQA})\end{tabular} \\
    Head Dimension (hd) & 128       \\
    Dtype & torch.bfloat16                        \\
    Window Size & \begin{tabular}[c]{@{}c@{}}1024\\ (\textit{for sliding window masks only})\end{tabular}\\
    \bottomrule
    \end{tabular}
    \vspace{0.5em}
    \caption{The fixed settings of FFA performance and flexibility benchmark}
    \label{tab:ffa_exps_settings}
\end{table}

\clearpage

\subsubsection{Benchmarking MagiAttention module-level scalability}\label{appendix:magiattn_exps_dffa}

\begin{figure}[htbp]
    \centering

    \begin{minipage}{\textwidth}
    
    \begin{subfigure}[b]{\textwidth}
        \centering
        \includegraphics[width=\textwidth]{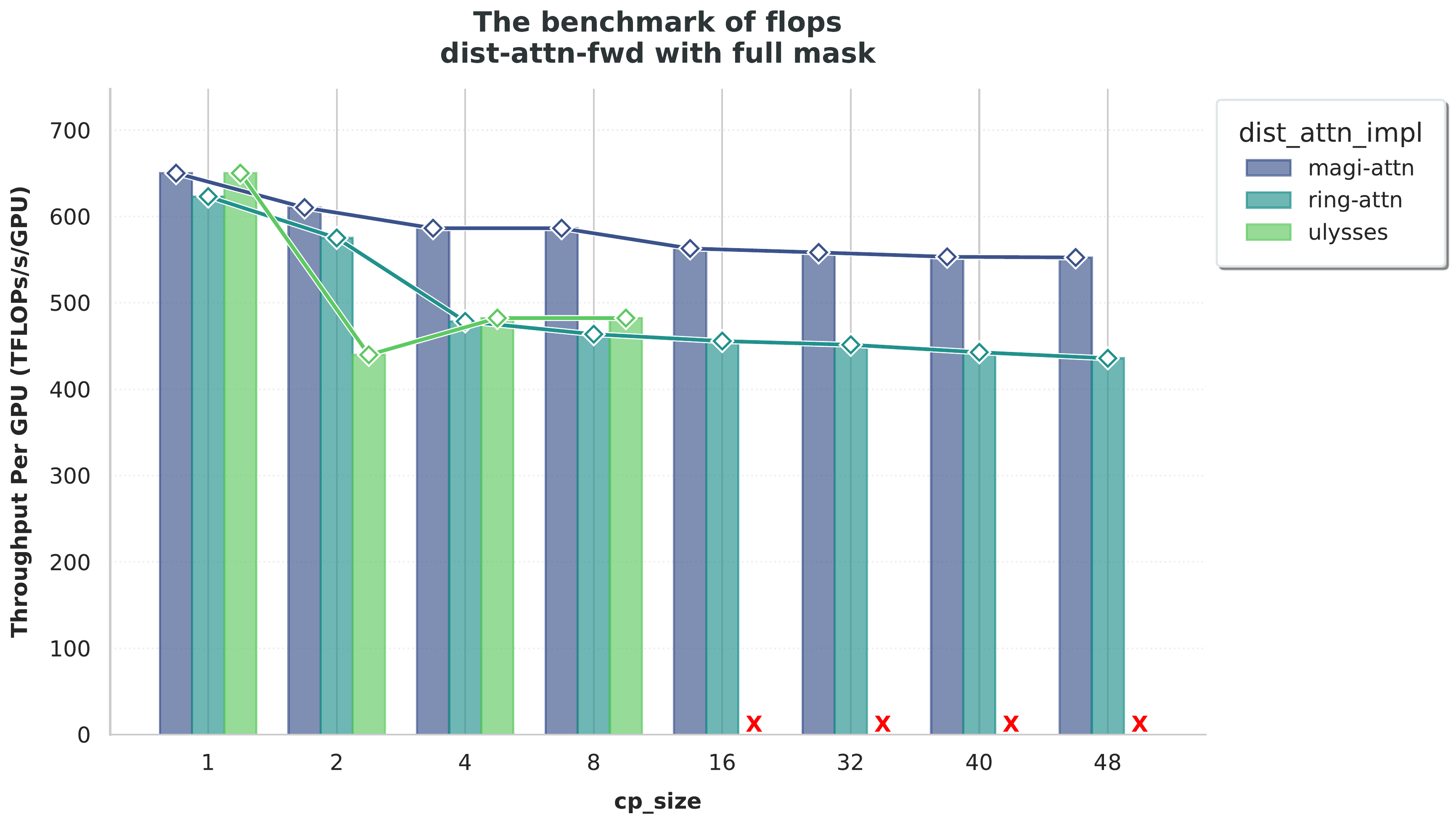}
        \caption{Forward pass for full mask.}
        \label{fig:dffa_fwd_perf_report_full}
    \end{subfigure}

    \end{minipage}

    \begin{minipage}{\textwidth}
    
    \begin{subfigure}[b]{\textwidth}
        \centering
        \includegraphics[width=\textwidth]{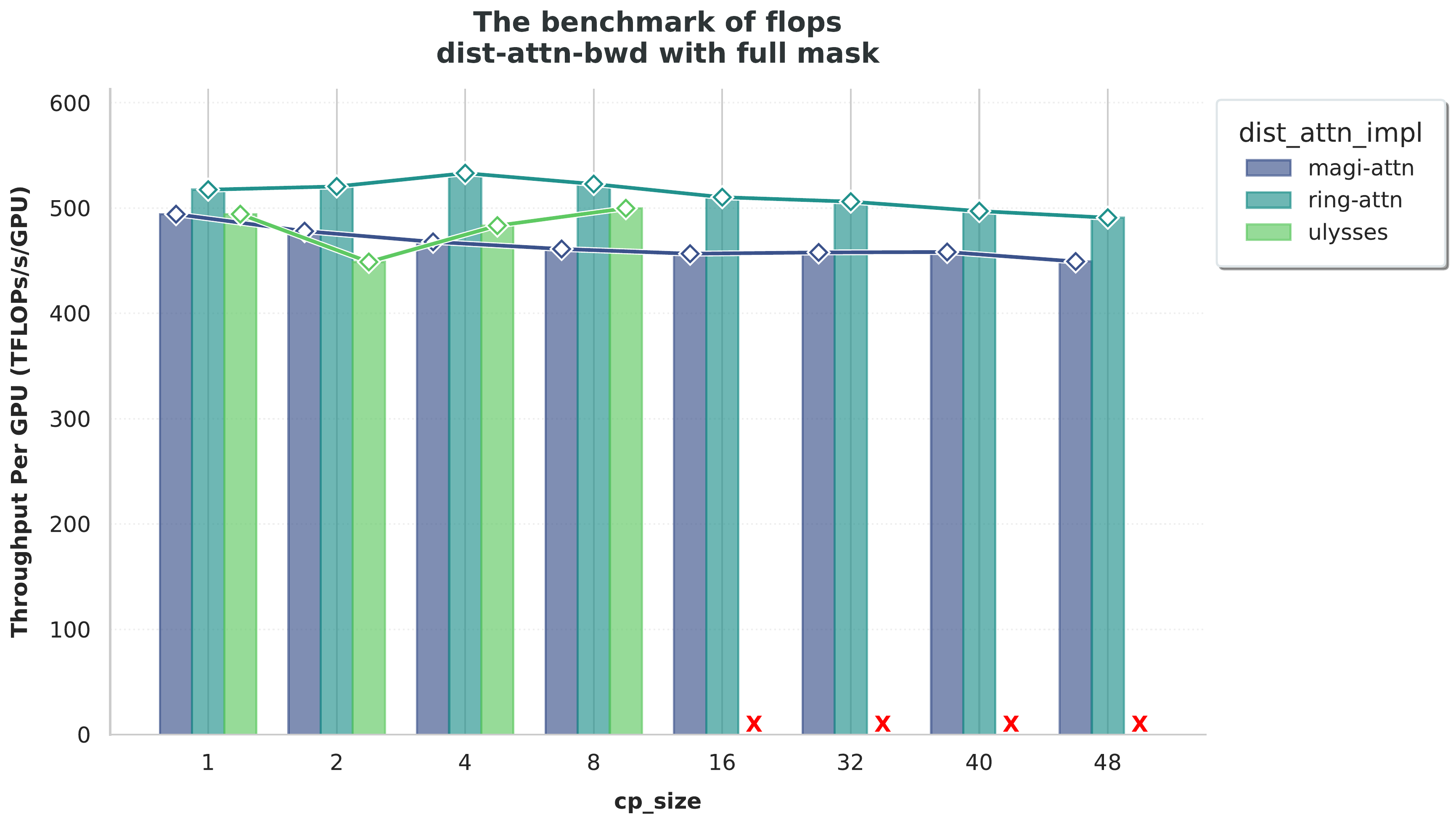}
        \caption{Backward pass for full mask.}
        \label{fig:dffa_bwd_perf_report_full}
    \end{subfigure}

    \end{minipage}

    \caption{Benchmarking MaiAttention's scalability against other leading CP strategies for full mask scenarios. (Note that: the $\mathbf{X}$ symbol indicates the corresponding distributed attention implementation is not supported in that specific configuration.)}
    \label{fig:dffa_perf_report_full}
\end{figure}

\begin{figure}[htbp]
    \centering

    \begin{minipage}{\textwidth}
    
    \begin{subfigure}[b]{\textwidth}
        \centering
        \includegraphics[width=\textwidth]{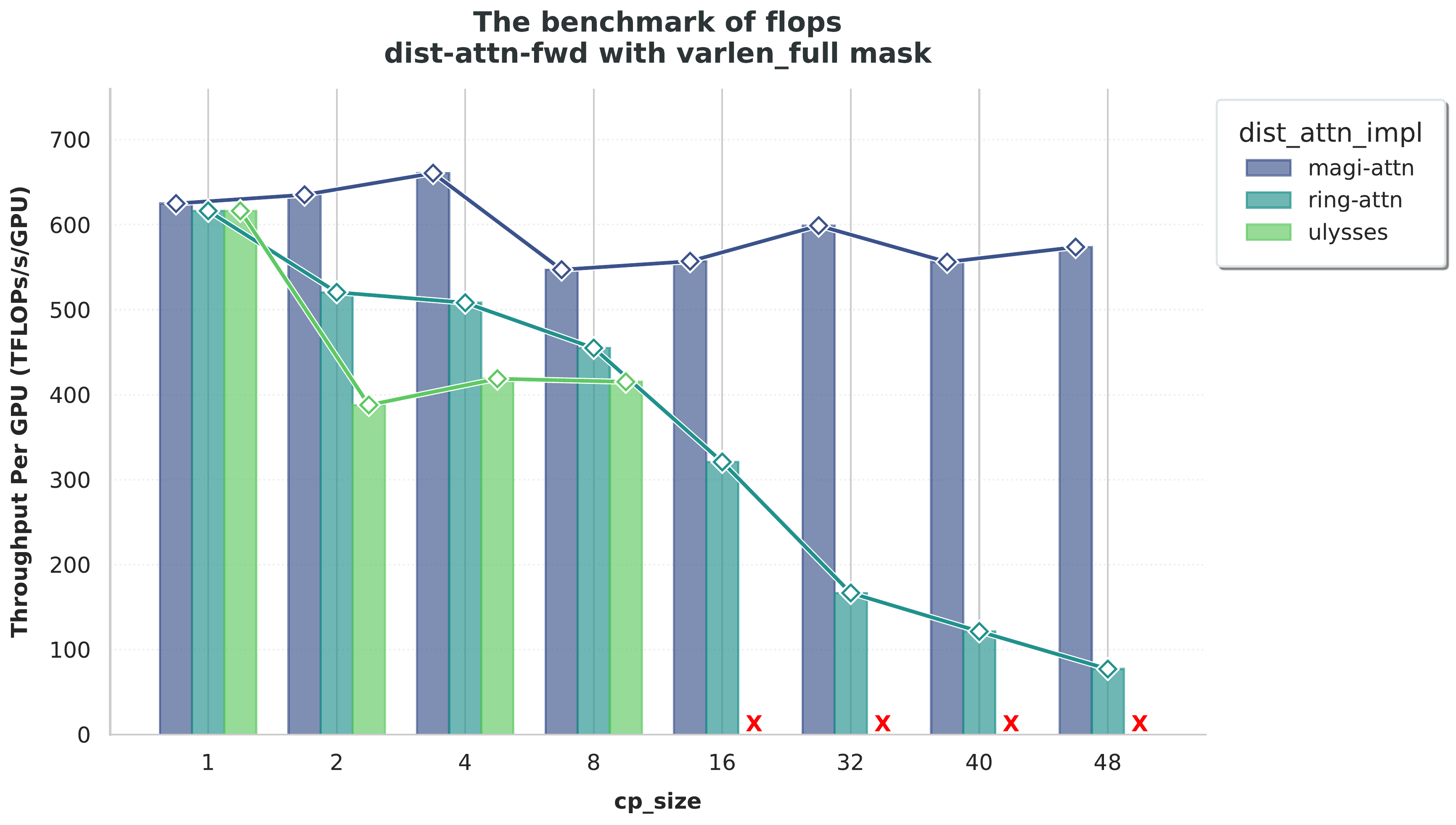}
        \caption{Forward pass for varlen-full mask.}
        \label{fig:dffa_fwd_perf_report_varlen_full}
    \end{subfigure}

    \end{minipage}

    \begin{minipage}{\textwidth}
    
    \begin{subfigure}[b]{\textwidth}
        \centering
        \includegraphics[width=\textwidth]{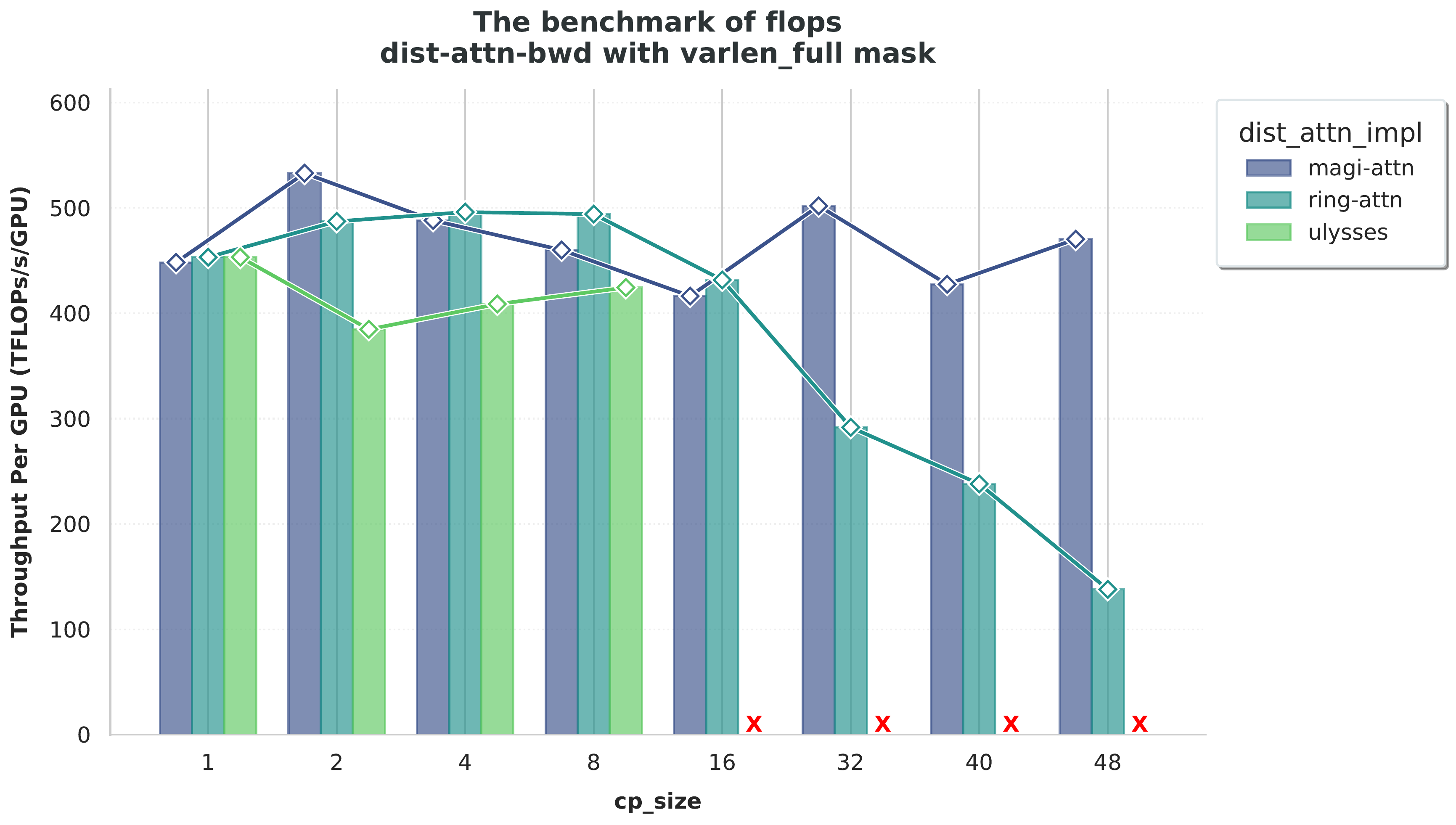}
        \caption{Backward pass for varlen-full mask.}
        \label{fig:dffa_bwd_perf_report_varlen_full}
    \end{subfigure}

    \end{minipage}

    \caption{Benchmarking MaiAttention's scalability against other leading CP strategies for varlen full mask scenarios. (Note that: the $\mathbf{X}$ symbol indicates the corresponding distributed attention implementation is not supported in that specific configuration.)}
    \label{fig:dffa_perf_report_varlen_full}
\end{figure}

To validate the scalability of MagiAttention, we assess the per-GPU throughput (in $\mathrm{TFLOPs/s/GPU}$) of the attention module during both forward and backward propagation, as the sequence length and parallel size increase. This assessment is compared against common CP strategies including Ring-Attention and Ulysses~\citep{liu2023ringattentionblockwisetransformers, jacobs2023deepspeed}. Due to the complexity of supporting irregular masks for baselines, our experiments are limited to the full mask and varlen full mask scenarios. And the distribution of variable sequence lengths still follow the one in kernel-level experiments~\ref{appendix:magiattn_exps_ffa}. The results are presented in Fig.\ref{fig:dffa_perf_report_full} and Fig.\ref{fig:dffa_perf_report_varlen_full}.

Our experiments are conducted on a large-scale productive GPU cluster~\footnote{Due to business and confidentiality reasons, specific details about the productive cluster, such as the number and type of GPUs, are withheld.}. We jointly scale the total sequence length \textit{seqlen}, the context-parallel size \textit{cp\_size}, and the node size \textit{nnodes} from (seqlen:64k, cp\_size:1, nnodes:1) up to (seqlen:3072k (3M), cp\_size:48, nnodes:48). The tensor-parallel size \textit{tp\_size} is kept constant at 8, with sequence-parallel enabled. Other data and model configurations for different mask types are consistent with those detailed in Tab.\ref{tab:ffa_exps_settings}.

Therefore, in every training setting, each rank is assigned constantly with $seqlen=64k$, $num\_heads_q = 8$ and $num\_heads_k = 1$ for attention propagation, while the remaining activations stays $seqlen=8k$, $num\_heads_q = 64$ and $num\_heads_k = 8$ with SP enabled. This setup simulates a common training configuration.

To calculate the $\mathrm{TFLOPs/s/GPU}$ for various mask patterns during both forward and backward passes, similarly, we first calculate the \texttt{FLOPs} as Eq.\ref{eq:flops_fwd} and Eq.\ref{eq:flops_bwd} and apply the following equation:

\begin{align}
     \mathrm{TFLOPs/s/GPU}^{(wd)} &= \cfrac{\mathrm{FLOPs}^{(wd)}}{\mathrm{Runtime}^{(wd) }\times cp\_size}, \quad wd \in \{fwd, bwd\}
\end{align}

As demonstrated, MagiAttention exhibits linear scalability as the context length and CP size increase, in both full mask and varlen full mask configurations, for both forward and backward passes. In contrast, baseline methods either face strict limitations in scaling up or experience performance degradation with ultra-long contexts, which worsens with varlen mask patterns.

\clearpage
\subsection{Other Materials}\label{appendix:odin}

\begin{table}[htbp]
\caption{Examples of DTensor's parallel placements and propagation common in modern training workflows w.r.t. \ref{sec:train_infra_odin}}
\label{tab:dtensor_placements}
\begin{threeparttable}
\resizebox{\columnwidth}{!}{%
\renewcommand{\arraystretch}{1.5} %
\begin{tabular}{|Sc|Sc|Sl|}
\hline
\multicolumn{1}{|c|}{Basic Placements} & \multicolumn{1}{c|}{Specific Placements} & \multicolumn{1}{c|}{Examples} \\ \hline
Replicate                             & Replicate \(^{(\mathrm{R})}\)                                 & 
    \(\begin{aligned}
    & \text{LayerNorm w/o SP:} \\
    & y^{(\mathrm{R})} = \frac{x^{(\mathrm{R})} - \mathrm{E}[x^{(\mathrm{R})}]}{ \sqrt{\mathrm{Var}[x^{(\mathrm{R})}] + \epsilon}} * \gamma^{(\mathrm{R})} + \beta^{(\mathrm{R})}
    \end{aligned}\)
                    \\ \hline
\multirow{4}{*}{Partial}               & SumPartial \(^{(\mathrm{P}_{sum})}\)                            & \(\begin{aligned}
    &\text{RowLinear w/o SP:}\\
    & O^{(\mathrm{R})} =  \mathrm{AllReduce}((X^{(\mathrm{S}(1))} \times W^{(\mathrm{S}(0))})^{(\mathrm{P}_{sum})}) \\
    &\text{ZERO-1 Grad Sync:}\\
    & grad^{(\mathrm{R})} = \mathrm{AllReduce}(grad^{(\mathrm{P}_{sum})})  \\
    &\text{FSDP Grad Sync:}\\
    & grad^{(\mathrm{S(0)})} = \mathrm{ReduceScatter}(grad^{(\mathrm{P}_{sum})}) \\
    &\text{MSE Loss Sync:}\\
    & loss^{(\mathrm{R})} = \mathrm{AllReduce}(loss^{(\mathrm{P}_{sum})})
\end{aligned}\)                         \\ \cline{2-3}
& AvgPartial \(^{(\mathrm{P}_{avg})}\)                          &      \(\begin{aligned}
    &\text{Parallel RMSNorm:}\\
    & y^{(\mathrm{R})} = \frac{x^{(\mathrm{S}(1))}}{\mathrm{AllReduce}(\mathrm{RMS}(x^{(\mathrm{S}(1))})^{(\mathrm{P}_{avg})})} * \gamma^{(\mathrm{S}(1))}  \\
\end{aligned}\)                         \\ \cline{2-3}
                                       & MaxPartial \(^{(\mathrm{P}_{max})}\)                                &    
\(\begin{aligned}
    &\text{Grad L}_{\infty}\text{-Norm}:\\
    & grad\_norm^{(\mathrm{R})} = \mathrm{AllReduce}(grad\_norm^{(\mathrm{P}_{max})})
\end{aligned}\)                         \\ \cline{2-3} 
& MinPartial \(^{(\mathrm{P}_{min})}\)                                &    
\(\begin{aligned}
    &\text{FP8 Scaling:}\\
    & \tilde{x}^{(\mathrm{R})} =  x^{(\mathrm{R})} \times \mathrm{AllReduce}((scaling\_factor^{-1})^{(\mathrm{P}_{min})})
\end{aligned}\)                         \\ \cline{2-3}
                                       & NormPartial \(^{(\mathrm{P}_{norm})}\)                              &    \(\begin{aligned}
    &\text{Grad L}_{p}\text{-Norm}:\\
    & grad\_norm^{(\mathrm{R})} = \mathrm{AllReduce}(\mathrm{pow}(grad\_norm^{(\mathrm{P}_{norm})}, p)^{(\mathrm{P}_{sum})})^{\frac{1}{p}}
\end{aligned}\)                          \\ \cline{2-3} 
& \textbf{LSEPartial} \(^{(\mathrm{P}_{lse})}\)                              &    \(\begin{aligned}
    &\text{LSE Correction}:\\
    & lse^{(\mathrm{R})} = \log(\mathrm{AllReduce}(\exp(lse^{(\mathrm{P}_{lse})})^{(\mathrm{P}_{sum})}))
\end{aligned}\)                          \\ \cline{2-3} 
                                       & \textbf{AttnPartial}  \(^{(\mathrm{P}_{attn})}\)                             & \(\begin{aligned}
    &\text{Attention Forward Correction:}\\
    & o^{(\mathrm{R})} = \mathrm{AllReduce}((\exp(lse^{(\mathrm{P}_{lse})}{-}lse^{(\mathrm{R})})\cdot o^{(\mathrm{P}_{attn})})^{(\mathrm{P}_{sum})})
\end{aligned}\)                              \\ \hline
\multirow{3}{*}{Shard}                 & (Even)Shard  \(^{(\mathrm{S}(dim))}\)                              &   \(\begin{aligned}
    &\text{ColLinear with SP}:\\
    & O^{(\mathrm{S}(1))} = \mathrm{AllGather}(X^{(\mathrm{S}(0))}) \times W^{(\mathrm{S}(1))} \\
    &\text{LayerNorm with SP:}\\
    & y^{(\mathrm{S}(0))} = \frac{x^{(\mathrm{S}(0))} - \mathrm{E}[x^{(\mathrm{S}(0))}]}{ \sqrt{\mathrm{Var}[x^{(\mathrm{S}(0))}] + \epsilon}} * \gamma^{(\mathrm{R})} + \beta^{(\mathrm{R})}
\end{aligned}\)                        \\ \cline{2-3}                                   & (Even)StridedShard \(^{(\mathrm{SS}(dim))}\)                        &        \(\begin{aligned}
    &\text{FSDP + ColLinear w/o SP}:\\
    & O^{(\mathrm{S}(1))} = X^{(\mathrm{R})}\times \mathrm{AllGather}(W^{(\mathrm{SS}(0), S(1))})^{(\mathrm{S}(1))} \\
    &\text{MHA QKV-fused ColLinear w/o SP}:\\
    & Q^{(\mathrm{S}(1))},K^{(\mathrm{S}(1))},V^{(\mathrm{S}(1))} = \mathrm{split}(QKV^{(\mathrm{SS}(1))}), \\&QKV^{(\mathrm{SS}(1))} = X^{(\mathrm{R})} \times W_{qkv}^{(\mathrm{SS}(1))}
\end{aligned}\)                          \\ \cline{2-3} 
                                       & \textbf{UnevenStridedShard} \(^{(\mathrm{USS}(dim))}\)                       &     \(\begin{aligned}
    &\text{GQA QKV-fused ColLinear w/o SP}:\\
    & Q^{(\mathrm{S}(1))},K^{(\mathrm{S}(1))},V^{(\mathrm{S}(1))} = \mathrm{split}(QKV^{(\mathrm{USS}(1))}), \\&QKV^{(\mathrm{USS}(1))} = X^{(\mathrm{R})} \times W_{qkv}^{(\mathrm{USS}(1))}
\end{aligned}\)                         \\ \hline
\end{tabular}%
}
\begin{tablenotes}\footnotesize
\item \textbf{Note:} the specific placements with the \textbf{bold text} are the ones we've complemented.
\end{tablenotes}
\end{threeparttable}
\end{table}

\end{document}